\documentstyle[psfig,twoside]{report}

\newcommand{\word}[1]{{\em #1}}
\newcommand{\pageline}{\underline{\hspace{\textwidth}}}

\newcounter{treecount}
\newcounter{branchcount}
\newsavebox{\parentbox}
\newsavebox{\treebox}
\newsavebox{\treeboxone}
\newsavebox{\treeboxtwo}
\newsavebox{\treeboxthree}
\newsavebox{\treeboxfour}
\newsavebox{\treeboxfive}
\newsavebox{\treeboxsix}
\newsavebox{\treeboxseven}
\newsavebox{\treeboxeight}
\newsavebox{\treeboxnine}
\newsavebox{\treeboxten}
\newsavebox{\treeboxeleven}
\newsavebox{\treeboxtwelve}
\newsavebox{\treeboxthirteen}
\newsavebox{\treeboxfourteen}
\newsavebox{\treeboxfifteen}
\newsavebox{\treeboxsixteen}
\newsavebox{\treeboxseventeen}
\newsavebox{\treeboxeighteen}
\newsavebox{\treeboxnineteen}
\newsavebox{\treeboxtwenty}
\newlength{\treeoffsetone}
\newlength{\treeoffsettwo}
\newlength{\treeoffsetthree}
\newlength{\treeoffsetfour}
\newlength{\treeoffsetfive}
\newlength{\treeoffsetsix}
\newlength{\treeoffsetseven}
\newlength{\treeoffseteight}
\newlength{\treeoffsetnine}
\newlength{\treeoffsetten}
\newlength{\treeoffseteleven}
\newlength{\treeoffsettwelve}
\newlength{\treeoffsetthirteen}
\newlength{\treeoffsetfourteen}
\newlength{\treeoffsetfifteen}
\newlength{\treeoffsetsixteen}
\newlength{\treeoffsetseventeen}
\newlength{\treeoffseteighteen}
\newlength{\treeoffsetnineteen}
\newlength{\treeoffsettwenty}

\newlength{\treeshiftone}
\newlength{\treeshifttwo}
\newlength{\treeshiftthree}
\newlength{\treeshiftfour}
\newlength{\treeshiftfive}
\newlength{\treeshiftsix}
\newlength{\treeshiftseven}
\newlength{\treeshifteight}
\newlength{\treeshiftnine}
\newlength{\treeshiftten}
\newlength{\treeshifteleven}
\newlength{\treeshifttwelve}
\newlength{\treeshiftthirteen}
\newlength{\treeshiftfourteen}
\newlength{\treeshiftfifteen}
\newlength{\treeshiftsixteen}
\newlength{\treeshiftseventeen}
\newlength{\treeshifteighteen}
\newlength{\treeshiftnineteen}
\newlength{\treeshifttwenty}
\newlength{\treewidthone}
\newlength{\treewidthtwo}
\newlength{\treewidththree}
\newlength{\treewidthfour}
\newlength{\treewidthfive}
\newlength{\treewidthsix}
\newlength{\treewidthseven}
\newlength{\treewidtheight}
\newlength{\treewidthnine}
\newlength{\treewidthten}
\newlength{\treewidtheleven}
\newlength{\treewidthtwelve}
\newlength{\treewidththirteen}
\newlength{\treewidthfourteen}
\newlength{\treewidthfifteen}
\newlength{\treewidthsixteen}
\newlength{\treewidthseventeen}
\newlength{\treewidtheighteen}
\newlength{\treewidthnineteen}
\newlength{\treewidthtwenty}
\newlength{\daughteroffsetone}
\newlength{\daughteroffsettwo}
\newlength{\daughteroffsetthree}
\newlength{\daughteroffsetfour}
\newlength{\branchwidthone}
\newlength{\branchwidthtwo}
\newlength{\branchwidththree}
\newlength{\branchwidthfour}
\newlength{\parentoffset}
\newlength{\treeoffset}
\newlength{\daughteroffset}
\newlength{\branchwidth}
\newlength{\parentwidth}
\newlength{\treewidth}
\newcommand{\ontop}[1]{\begin{tabular}{c}#1\end{tabular}}
\newcommand{\poptree}{%
\ifnum\value{treecount}=0\typeout{QobiTeX warning---Tree stack underflow}\fi%
\addtocounter{treecount}{-1}%
\setlength{\treeoffsettwo}{\treeoffsetthree}%
\setlength{\treeoffsetthree}{\treeoffsetfour}%
\setlength{\treeoffsetfour}{\treeoffsetfive}%
\setlength{\treeoffsetfive}{\treeoffsetsix}%
\setlength{\treeoffsetsix}{\treeoffsetseven}%
\setlength{\treeoffsetseven}{\treeoffseteight}%
\setlength{\treeoffseteight}{\treeoffsetnine}%
\setlength{\treeoffsetnine}{\treeoffsetten}%
\setlength{\treeoffsetten}{\treeoffseteleven}%
\setlength{\treeoffseteleven}{\treeoffsettwelve}%
\setlength{\treeoffsettwelve}{\treeoffsetthirteen}%
\setlength{\treeoffsetthirteen}{\treeoffsetfourteen}%
\setlength{\treeoffsetfourteen}{\treeoffsetfifteen}%
\setlength{\treeoffsetfifteen}{\treeoffsetsixteen}%
\setlength{\treeoffsetsixteen}{\treeoffsetseventeen}%
\setlength{\treeoffsetseventeen}{\treeoffseteighteen}%
\setlength{\treeoffseteighteen}{\treeoffsetnineteen}%
\setlength{\treeoffsetnineteen}{\treeoffsettwenty}%
\setlength{\treeshifttwo}{\treeshiftthree}%
\setlength{\treeshiftthree}{\treeshiftfour}%
\setlength{\treeshiftfour}{\treeshiftfive}%
\setlength{\treeshiftfive}{\treeshiftsix}%
\setlength{\treeshiftsix}{\treeshiftseven}%
\setlength{\treeshiftseven}{\treeshifteight}%
\setlength{\treeshifteight}{\treeshiftnine}%
\setlength{\treeshiftnine}{\treeshiftten}%
\setlength{\treeshiftten}{\treeshifteleven}%
\setlength{\treeshifteleven}{\treeshifttwelve}%
\setlength{\treeshifttwelve}{\treeshiftthirteen}%
\setlength{\treeshiftthirteen}{\treeshiftfourteen}%
\setlength{\treeshiftfourteen}{\treeshiftfifteen}%
\setlength{\treeshiftfifteen}{\treeshiftsixteen}%
\setlength{\treeshiftsixteen}{\treeshiftseventeen}%
\setlength{\treeshiftseventeen}{\treeshifteighteen}%
\setlength{\treeshifteighteen}{\treeshiftnineteen}%
\setlength{\treeshiftnineteen}{\treeshifttwenty}%
\setlength{\treewidthtwo}{\treewidththree}%
\setlength{\treewidththree}{\treewidthfour}%
\setlength{\treewidthfour}{\treewidthfive}%
\setlength{\treewidthfive}{\treewidthsix}%
\setlength{\treewidthsix}{\treewidthseven}%
\setlength{\treewidthseven}{\treewidtheight}%
\setlength{\treewidtheight}{\treewidthnine}%
\setlength{\treewidthnine}{\treewidthten}%
\setlength{\treewidthten}{\treewidtheleven}%
\setlength{\treewidtheleven}{\treewidthtwelve}%
\setlength{\treewidthtwelve}{\treewidththirteen}%
\setlength{\treewidththirteen}{\treewidthfourteen}%
\setlength{\treewidthfourteen}{\treewidthfifteen}%
\setlength{\treewidthfifteen}{\treewidthsixteen}%
\setlength{\treewidthsixteen}{\treewidthseventeen}%
\setlength{\treewidthseventeen}{\treewidtheighteen}%
\setlength{\treewidtheighteen}{\treewidthnineteen}%
\setlength{\treewidthnineteen}{\treewidthtwenty}%
\sbox{\treeboxtwo}{\usebox{\treeboxthree}}%
\sbox{\treeboxthree}{\usebox{\treeboxfour}}%
\sbox{\treeboxfour}{\usebox{\treeboxfive}}%
\sbox{\treeboxfive}{\usebox{\treeboxsix}}%
\sbox{\treeboxsix}{\usebox{\treeboxseven}}%
\sbox{\treeboxseven}{\usebox{\treeboxeight}}%
\sbox{\treeboxeight}{\usebox{\treeboxnine}}%
\sbox{\treeboxnine}{\usebox{\treeboxten}}%
\sbox{\treeboxten}{\usebox{\treeboxeleven}}%
\sbox{\treeboxeleven}{\usebox{\treeboxtwelve}}%
\sbox{\treeboxtwelve}{\usebox{\treeboxthirteen}}%
\sbox{\treeboxthirteen}{\usebox{\treeboxfourteen}}%
\sbox{\treeboxfourteen}{\usebox{\treeboxfifteen}}%
\sbox{\treeboxfifteen}{\usebox{\treeboxsixteen}}%
\sbox{\treeboxsixteen}{\usebox{\treeboxseventeen}}%
\sbox{\treeboxseventeen}{\usebox{\treeboxeighteen}}%
\sbox{\treeboxeighteen}{\usebox{\treeboxnineteen}}%
\sbox{\treeboxnineteen}{\usebox{\treeboxtwenty}}}
\newcommand{\leaf}[1]{%
\ifnum\value{treecount}=20\typeout{QobiTeX warning---Tree stack overflow}\fi%
\addtocounter{treecount}{1}%
\sbox{\treeboxtwenty}{\usebox{\treeboxnineteen}}%
\sbox{\treeboxnineteen}{\usebox{\treeboxeighteen}}%
\sbox{\treeboxeighteen}{\usebox{\treeboxseventeen}}%
\sbox{\treeboxseventeen}{\usebox{\treeboxsixteen}}%
\sbox{\treeboxsixteen}{\usebox{\treeboxfifteen}}%
\sbox{\treeboxfifteen}{\usebox{\treeboxfourteen}}%
\sbox{\treeboxfourteen}{\usebox{\treeboxthirteen}}%
\sbox{\treeboxthirteen}{\usebox{\treeboxtwelve}}%
\sbox{\treeboxtwelve}{\usebox{\treeboxeleven}}%
\sbox{\treeboxeleven}{\usebox{\treeboxten}}%
\sbox{\treeboxten}{\usebox{\treeboxnine}}%
\sbox{\treeboxnine}{\usebox{\treeboxeight}}%
\sbox{\treeboxeight}{\usebox{\treeboxseven}}%
\sbox{\treeboxseven}{\usebox{\treeboxsix}}%
\sbox{\treeboxsix}{\usebox{\treeboxfive}}%
\sbox{\treeboxfive}{\usebox{\treeboxfour}}%
\sbox{\treeboxfour}{\usebox{\treeboxthree}}%
\sbox{\treeboxthree}{\usebox{\treeboxtwo}}%
\sbox{\treeboxtwo}{\usebox{\treeboxone}}%
\sbox{\treeboxone}{\ontop{#1}}%
\sbox{\treeboxone}{\raisebox{-\ht\treeboxone}{\usebox{\treeboxone}}}%
\setlength{\treeoffsettwenty}{\treeoffsetnineteen}%
\setlength{\treeoffsetnineteen}{\treeoffseteighteen}%
\setlength{\treeoffseteighteen}{\treeoffsetseventeen}%
\setlength{\treeoffsetseventeen}{\treeoffsetsixteen}%
\setlength{\treeoffsetsixteen}{\treeoffsetfifteen}%
\setlength{\treeoffsetfifteen}{\treeoffsetfourteen}%
\setlength{\treeoffsetfourteen}{\treeoffsetthirteen}%
\setlength{\treeoffsetthirteen}{\treeoffsettwelve}%
\setlength{\treeoffsettwelve}{\treeoffseteleven}%
\setlength{\treeoffseteleven}{\treeoffsetten}%
\setlength{\treeoffsetten}{\treeoffsetnine}%
\setlength{\treeoffsetnine}{\treeoffseteight}%
\setlength{\treeoffseteight}{\treeoffsetseven}%
\setlength{\treeoffsetseven}{\treeoffsetsix}%
\setlength{\treeoffsetsix}{\treeoffsetfive}%
\setlength{\treeoffsetfive}{\treeoffsetfour}%
\setlength{\treeoffsetfour}{\treeoffsetthree}%
\setlength{\treeoffsetthree}{\treeoffsettwo}%
\setlength{\treeoffsettwo}{\treeoffsetone}%
\setlength{\treeoffsetone}{0.5\wd\treeboxone}%
\setlength{\treeshifttwenty}{\treeshiftnineteen}%
\setlength{\treeshiftnineteen}{\treeshifteighteen}%
\setlength{\treeshifteighteen}{\treeshiftseventeen}%
\setlength{\treeshiftseventeen}{\treeshiftsixteen}%
\setlength{\treeshiftsixteen}{\treeshiftfifteen}%
\setlength{\treeshiftfifteen}{\treeshiftfourteen}%
\setlength{\treeshiftfourteen}{\treeshiftthirteen}%
\setlength{\treeshiftthirteen}{\treeshifttwelve}%
\setlength{\treeshifttwelve}{\treeshifteleven}%
\setlength{\treeshifteleven}{\treeshiftten}%
\setlength{\treeshiftten}{\treeshiftnine}%
\setlength{\treeshiftnine}{\treeshifteight}%
\setlength{\treeshifteight}{\treeshiftseven}%
\setlength{\treeshiftseven}{\treeshiftsix}%
\setlength{\treeshiftsix}{\treeshiftfive}%
\setlength{\treeshiftfive}{\treeshiftfour}%
\setlength{\treeshiftfour}{\treeshiftthree}%
\setlength{\treeshiftthree}{\treeshifttwo}%
\setlength{\treeshifttwo}{\treeshiftone}%
\setlength{\treeshiftone}{0pt}%
\setlength{\treewidthtwenty}{\treewidthnineteen}%
\setlength{\treewidthnineteen}{\treewidtheighteen}%
\setlength{\treewidtheighteen}{\treewidthseventeen}%
\setlength{\treewidthseventeen}{\treewidthsixteen}%
\setlength{\treewidthsixteen}{\treewidthfifteen}%
\setlength{\treewidthfifteen}{\treewidthfourteen}%
\setlength{\treewidthfourteen}{\treewidththirteen}%
\setlength{\treewidththirteen}{\treewidthtwelve}%
\setlength{\treewidthtwelve}{\treewidtheleven}%
\setlength{\treewidtheleven}{\treewidthten}%
\setlength{\treewidthten}{\treewidthnine}%
\setlength{\treewidthnine}{\treewidtheight}%
\setlength{\treewidtheight}{\treewidthseven}%
\setlength{\treewidthseven}{\treewidthsix}%
\setlength{\treewidthsix}{\treewidthfive}%
\setlength{\treewidthfive}{\treewidthfour}%
\setlength{\treewidthfour}{\treewidththree}%
\setlength{\treewidththree}{\treewidthtwo}%
\setlength{\treewidthtwo}{\treewidthone}%
\setlength{\treewidthone}{\wd\treeboxone}}
\newcommand{\branch}[2]{%
\setcounter{branchcount}{#1}%
\ifnum\value{branchcount}=1\sbox{\parentbox}{\ontop{#2}}%
\setlength{\parentoffset}{\treeoffsetone}%
\addtolength{\parentoffset}{-0.5\wd\parentbox}%
\setlength{\daughteroffset}{0in}%
\ifdim\parentoffset<0in%
\setlength{\daughteroffset}{-\parentoffset}%
\setlength{\parentoffset}{0in}\fi%
\setlength{\parentwidth}{\parentoffset}%
\addtolength{\parentwidth}{\wd\parentbox}%
\setlength{\treeoffset}{\daughteroffset}%
\addtolength{\treeoffset}{\treeoffsetone}%
\setlength{\treewidth}{\wd\treeboxone}%
\addtolength{\treewidth}{\daughteroffset}%
\ifdim\treewidth<\parentwidth\setlength{\treewidth}{\parentwidth}\fi%
\sbox{\treebox}{\begin{minipage}{\treewidth}%
\begin{flushleft}%
\hspace*{\parentoffset}\usebox{\parentbox}\\
{\setlength{\unitlength}{2ex}%
\hspace*{\treeoffset}\begin{picture}(0,1)%
\put(0,0){\line(0,1){1}}%
\end{picture}}\\
\vspace{-\baselineskip}
\hspace*{\daughteroffset}%
\raisebox{-\ht\treeboxone}{\usebox{\treeboxone}}%
\end{flushleft}%
\end{minipage}}%
\setlength{\treeoffsetone}{\parentoffset}%
\addtolength{\treeoffsetone}{0.5\wd\parentbox}%
\setlength{\treeshiftone}{0pt}%
\setlength{\treewidthone}{\treewidth}%
\sbox{\treeboxone}{\usebox{\treebox}}%
\else\ifnum\value{branchcount}=2\sbox{\parentbox}{\ontop{#2}}%
\setlength{\branchwidthone}{\treewidthtwo}%
\addtolength{\branchwidthone}{\treeoffsetone}%
\addtolength{\branchwidthone}{-\treeshiftone}%
\addtolength{\branchwidthone}{-\treeoffsettwo}%
\setlength{\branchwidth}{\branchwidthone}%
\setlength{\daughteroffsetone}{\branchwidth}%
\addtolength{\daughteroffsetone}{-\branchwidthone}%
\addtolength{\daughteroffsetone}{-\treeshiftone}%
\setlength{\parentoffset}{-0.5\wd\parentbox}%
\addtolength{\parentoffset}{\treeoffsettwo}%
\addtolength{\parentoffset}{0.5\branchwidth}%
\setlength{\daughteroffset}{0in}%
\ifdim\parentoffset<0in%
\setlength{\daughteroffset}{-\parentoffset}%
\setlength{\parentoffset}{0in}\fi%
\setlength{\parentwidth}{\parentoffset}%
\addtolength{\parentwidth}{\wd\parentbox}%
\setlength{\treeoffset}{\daughteroffset}%
\addtolength{\treeoffset}{\treeoffsettwo}%
\setlength{\treewidth}{\wd\treeboxone}%
\addtolength{\treewidth}{\daughteroffsetone}%
\addtolength{\treewidth}{\treewidthtwo}%
\addtolength{\treewidth}{\daughteroffset}%
\ifdim\treewidth<\parentwidth\setlength{\treewidth}{\parentwidth}\fi%
\sbox{\treebox}{\begin{minipage}{\treewidth}%
\begin{flushleft}%
\hspace*{\parentoffset}\usebox{\parentbox}\\
{\setlength{\unitlength}{0.5\branchwidth}%
\hspace*{\treeoffset}\begin{picture}(2,0.5)%
\put(0,0){\line(2,1){1}}%
\put(2,0){\line(-2,1){1}}%
\end{picture}}\\
\vspace{-\baselineskip}
\hspace*{\daughteroffset}%
\makebox[\treewidthtwo][l]%
{\raisebox{-\ht\treeboxtwo}{\usebox{\treeboxtwo}}}%
\hspace*{\daughteroffsetone}%
\raisebox{-\ht\treeboxone}{\usebox{\treeboxone}}%
\end{flushleft}%
\end{minipage}}%
\setlength{\treeoffsetone}{\parentoffset}%
\addtolength{\treeoffsetone}{0.5\wd\parentbox}%
\setlength{\treeshiftone}{0pt}%
\setlength{\treewidthone}{\treewidth}%
\sbox{\treeboxone}{\usebox{\treebox}}\poptree%
\else\ifnum\value{branchcount}=3\sbox{\parentbox}{\ontop{#2}}%
\setlength{\branchwidthone}{\treewidthtwo}%
\addtolength{\branchwidthone}{\treeoffsetone}%
\addtolength{\branchwidthone}{-\treeshiftone}%
\addtolength{\branchwidthone}{-\treeoffsettwo}%
\setlength{\branchwidthtwo}{\treewidththree}%
\addtolength{\branchwidthtwo}{\treeoffsettwo}%
\addtolength{\branchwidthtwo}{-\treeshifttwo}%
\addtolength{\branchwidthtwo}{-\treeoffsetthree}%
\setlength{\branchwidth}{\branchwidthone}%
\ifdim\branchwidthtwo>\branchwidth%
\setlength{\branchwidth}{\branchwidthtwo}\fi%
\setlength{\daughteroffsetone}{\branchwidth}%
\addtolength{\daughteroffsetone}{-\branchwidthone}%
\addtolength{\daughteroffsetone}{-\treeshiftone}%
\setlength{\daughteroffsettwo}{\branchwidth}%
\addtolength{\daughteroffsettwo}{-\branchwidthtwo}%
\addtolength{\daughteroffsettwo}{-\treeshifttwo}%
\setlength{\parentoffset}{-0.5\wd\parentbox}%
\addtolength{\parentoffset}{\treeoffsetthree}%
\addtolength{\parentoffset}{\branchwidth}%
\setlength{\daughteroffset}{0in}%
\ifdim\parentoffset<0in%
\setlength{\daughteroffset}{-\parentoffset}%
\setlength{\parentoffset}{0in}\fi%
\setlength{\parentwidth}{\parentoffset}%
\addtolength{\parentwidth}{\wd\parentbox}%
\setlength{\treeoffset}{\daughteroffset}%
\addtolength{\treeoffset}{\treeoffsetthree}%
\setlength{\treewidth}{\wd\treeboxone}%
\addtolength{\treewidth}{\daughteroffsetone}%
\addtolength{\treewidth}{\treewidthtwo}%
\addtolength{\treewidth}{\daughteroffsettwo}%
\addtolength{\treewidth}{\treewidththree}%
\addtolength{\treewidth}{\daughteroffset}%
\ifdim\treewidth<\parentwidth\setlength{\treewidth}{\parentwidth}\fi%
\sbox{\treebox}{\begin{minipage}{\treewidth}%
\begin{flushleft}%
\hspace*{\parentoffset}\usebox{\parentbox}\\
{\setlength{\unitlength}{0.5\branchwidth}%
\hspace*{\treeoffset}\begin{picture}(4,1)%
\put(0,0){\line(2,1){2}}%
\put(2,0){\line(0,1){1}}%
\put(4,0){\line(-2,1){2}}%
\end{picture}}\\
\vspace{-\baselineskip}
\hspace*{\daughteroffset}%
\makebox[\treewidththree][l]%
{\raisebox{-\ht\treeboxthree}{\usebox{\treeboxthree}}}%
\hspace*{\daughteroffsettwo}%
\makebox[\treewidthtwo][l]%
{\raisebox{-\ht\treeboxtwo}{\usebox{\treeboxtwo}}}%
\hspace*{\daughteroffsetone}%
\raisebox{-\ht\treeboxone}{\usebox{\treeboxone}}%
\end{flushleft}%
\end{minipage}}%
\setlength{\treeoffsetone}{\parentoffset}%
\addtolength{\treeoffsetone}{0.5\wd\parentbox}%
\setlength{\treeshiftone}{0pt}%
\setlength{\treewidthone}{\treewidth}%
\sbox{\treeboxone}{\usebox{\treebox}}\poptree\poptree%
\else\ifnum\value{branchcount}=4\sbox{\parentbox}{\ontop{#2}}%
\setlength{\branchwidthone}{\treewidthtwo}%
\addtolength{\branchwidthone}{\treeoffsetone}%
\addtolength{\branchwidthone}{-\treeshiftone}%
\addtolength{\branchwidthone}{-\treeoffsettwo}%
\setlength{\branchwidthtwo}{\treewidththree}%
\addtolength{\branchwidthtwo}{\treeoffsettwo}%
\addtolength{\branchwidthtwo}{-\treeshifttwo}%
\addtolength{\branchwidthtwo}{-\treeoffsetthree}%
\setlength{\branchwidththree}{\treewidthfour}%
\addtolength{\branchwidththree}{\treeoffsetthree}%
\addtolength{\branchwidththree}{-\treeshiftthree}%
\addtolength{\branchwidththree}{-\treeoffsetfour}%
\setlength{\branchwidth}{\branchwidthone}%
\ifdim\branchwidthtwo>\branchwidth%
\setlength{\branchwidth}{\branchwidthtwo}\fi%
\ifdim\branchwidththree>\branchwidth%
\setlength{\branchwidth}{\branchwidththree}\fi%
\setlength{\daughteroffsetone}{\branchwidth}%
\addtolength{\daughteroffsetone}{-\branchwidthone}%
\addtolength{\daughteroffsetone}{-\treeshiftone}%
\setlength{\daughteroffsettwo}{\branchwidth}%
\addtolength{\daughteroffsettwo}{-\branchwidthtwo}%
\addtolength{\daughteroffsettwo}{-\treeshifttwo}%
\setlength{\daughteroffsetthree}{\branchwidth}%
\addtolength{\daughteroffsetthree}{-\branchwidththree}%
\addtolength{\daughteroffsetthree}{-\treeshiftthree}%
\setlength{\parentoffset}{-0.5\wd\parentbox}%
\addtolength{\parentoffset}{\treeoffsetfour}%
\addtolength{\parentoffset}{1.5\branchwidth}%
\setlength{\daughteroffset}{0in}%
\ifdim\parentoffset<0in%
\setlength{\daughteroffset}{-\parentoffset}%
\setlength{\parentoffset}{0in}\fi%
\setlength{\parentwidth}{\parentoffset}%
\addtolength{\parentwidth}{\wd\parentbox}%
\setlength{\treeoffset}{\daughteroffset}%
\addtolength{\treeoffset}{\treeoffsetfour}%
\setlength{\treewidth}{\wd\treeboxone}%
\addtolength{\treewidth}{\daughteroffsetone}%
\addtolength{\treewidth}{\treewidthtwo}%
\addtolength{\treewidth}{\daughteroffsettwo}%
\addtolength{\treewidth}{\treewidththree}%
\addtolength{\treewidth}{\daughteroffsetthree}%
\addtolength{\treewidth}{\treewidthfour}%
\addtolength{\treewidth}{\daughteroffset}%
\ifdim\treewidth<\parentwidth\setlength{\treewidth}{\parentwidth}\fi%
\sbox{\treebox}{\begin{minipage}{\treewidth}%
\begin{flushleft}%
\hspace*{\parentoffset}\usebox{\parentbox}\\
{\setlength{\unitlength}{0.5\branchwidth}%
\hspace*{\treeoffset}\begin{picture}(6,1)%
\put(0,0){\line(3,1){3}}%
\put(2,0){\line(1,1){1}}%
\put(4,0){\line(-1,1){1}}%
\put(6,0){\line(-3,1){3}}%
\end{picture}}\\
\vspace{-\baselineskip}
\hspace*{\daughteroffset}%
\makebox[\treewidthfour][l]%
{\raisebox{-\ht\treeboxfour}{\usebox{\treeboxfour}}}%
\hspace*{\daughteroffsetthree}%
\makebox[\treewidththree][l]%
{\raisebox{-\ht\treeboxthree}{\usebox{\treeboxthree}}}%
\hspace*{\daughteroffsettwo}%
\makebox[\treewidthtwo][l]%
{\raisebox{-\ht\treeboxtwo}{\usebox{\treeboxtwo}}}%
\hspace*{\daughteroffsetone}%
\raisebox{-\ht\treeboxone}{\usebox{\treeboxone}}%
\end{flushleft}%
\end{minipage}}%
\setlength{\treeoffsetone}{\parentoffset}%
\addtolength{\treeoffsetone}{0.5\wd\parentbox}%
\setlength{\treeshiftone}{0pt}%
\setlength{\treewidthone}{\treewidth}%
\sbox{\treeboxone}{\usebox{\treebox}}\poptree\poptree\poptree%
\else\ifnum\value{branchcount}=5\sbox{\parentbox}{\ontop{#2}}%
\setlength{\branchwidthone}{\treewidthtwo}%
\addtolength{\branchwidthone}{\treeoffsetone}%
\addtolength{\branchwidthone}{-\treeshiftone}%
\addtolength{\branchwidthone}{-\treeoffsettwo}%
\setlength{\branchwidthtwo}{\treewidththree}%
\addtolength{\branchwidthtwo}{\treeoffsettwo}%
\addtolength{\branchwidthtwo}{-\treeshifttwo}%
\addtolength{\branchwidthtwo}{-\treeoffsetthree}%
\setlength{\branchwidththree}{\treewidthfour}%
\addtolength{\branchwidththree}{\treeoffsetthree}%
\addtolength{\branchwidththree}{-\treeshiftthree}%
\addtolength{\branchwidththree}{-\treeoffsetfour}%
\setlength{\branchwidthfour}{\treewidthfive}%
\addtolength{\branchwidthfour}{\treeoffsetfour}%
\addtolength{\branchwidthfour}{-\treeshiftfour}%
\addtolength{\branchwidthfour}{-\treeoffsetfive}%
\setlength{\branchwidth}{\branchwidthone}%
\ifdim\branchwidthtwo>\branchwidth%
\setlength{\branchwidth}{\branchwidthtwo}\fi%
\ifdim\branchwidththree>\branchwidth%
\setlength{\branchwidth}{\branchwidththree}\fi%
\ifdim\branchwidthfour>\branchwidth%
\setlength{\branchwidth}{\branchwidthfour}\fi%
\setlength{\daughteroffsetone}{\branchwidth}%
\addtolength{\daughteroffsetone}{-\branchwidthone}%
\addtolength{\daughteroffsetone}{-\treeshiftone}%
\setlength{\daughteroffsettwo}{\branchwidth}%
\addtolength{\daughteroffsettwo}{-\branchwidthtwo}%
\addtolength{\daughteroffsettwo}{-\treeshifttwo}%
\setlength{\daughteroffsetthree}{\branchwidth}%
\addtolength{\daughteroffsetthree}{-\branchwidththree}%
\addtolength{\daughteroffsetthree}{-\treeshiftthree}%
\setlength{\daughteroffsetfour}{\branchwidth}%
\addtolength{\daughteroffsetfour}{-\branchwidthfour}%
\addtolength{\daughteroffsetfour}{-\treeshiftfour}%
\setlength{\parentoffset}{-0.5\wd\parentbox}%
\addtolength{\parentoffset}{\treeoffsetfive}%
\addtolength{\parentoffset}{2\branchwidth}%
\setlength{\daughteroffset}{0in}%
\ifdim\parentoffset<0in%
\setlength{\daughteroffset}{-\parentoffset}%
\setlength{\parentoffset}{0in}\fi%
\setlength{\parentwidth}{\parentoffset}%
\addtolength{\parentwidth}{\wd\parentbox}%
\setlength{\treeoffset}{\daughteroffset}%
\addtolength{\treeoffset}{\treeoffsetfive}%
\setlength{\treewidth}{\wd\treeboxone}%
\addtolength{\treewidth}{\daughteroffsetone}%
\addtolength{\treewidth}{\treewidthtwo}%
\addtolength{\treewidth}{\daughteroffsettwo}%
\addtolength{\treewidth}{\treewidththree}%
\addtolength{\treewidth}{\daughteroffsetthree}%
\addtolength{\treewidth}{\treewidthfour}%
\addtolength{\treewidth}{\daughteroffsetfour}%
\addtolength{\treewidth}{\treewidthfive}%
\addtolength{\treewidth}{\daughteroffset}%
\ifdim\treewidth<\parentwidth\setlength{\treewidth}{\parentwidth}\fi%
\sbox{\treebox}{\begin{minipage}{\treewidth}%
\begin{flushleft}%
\hspace*{\parentoffset}\usebox{\parentbox}\\
{\setlength{\unitlength}{0.5\branchwidth}%
\hspace*{\treeoffset}\begin{picture}(8,1)%
\put(0,0){\line(4,1){4}}%
\put(2,0){\line(2,1){2}}%
\put(4,0){\line(0,1){1}}%
\put(6,0){\line(-2,1){2}}%
\put(8,0){\line(-4,1){4}}%
\end{picture}}\\
\vspace{-\baselineskip}
\hspace*{\daughteroffset}%
\makebox[\treewidthfive][l]%
{\raisebox{-\ht\treeboxfour}{\usebox{\treeboxfive}}}%
\hspace*{\daughteroffsetfour}%
\makebox[\treewidthfour][l]%
{\raisebox{-\ht\treeboxfour}{\usebox{\treeboxfour}}}%
\hspace*{\daughteroffsetthree}%
\makebox[\treewidththree][l]%
{\raisebox{-\ht\treeboxthree}{\usebox{\treeboxthree}}}%
\hspace*{\daughteroffsettwo}%
\makebox[\treewidthtwo][l]%
{\raisebox{-\ht\treeboxtwo}{\usebox{\treeboxtwo}}}%
\hspace*{\daughteroffsetone}%
\raisebox{-\ht\treeboxone}{\usebox{\treeboxone}}%
\end{flushleft}%
\end{minipage}}%
\setlength{\treeoffsetone}{\parentoffset}%
\addtolength{\treeoffsetone}{0.5\wd\parentbox}%
\setlength{\treeshiftone}{0pt}%
\setlength{\treewidthone}{\treewidth}%
\sbox{\treeboxone}{\usebox{\treebox}}\poptree\poptree\poptree\poptree%
\else\typeout{QobiTeX warning--- Can't handle #1 branching}\fi\fi\fi\fi\fi}
\newcommand{\emptybranch}[1]{%
\setcounter{branchcount}{#1}%
\ifnum\value{branchcount}=1%
\setlength{\parentoffset}{\treeoffsetone}%
\setlength{\daughteroffset}{0in}%
\ifdim\parentoffset<0in%
\setlength{\daughteroffset}{-\parentoffset}%
\setlength{\parentoffset}{0in}\fi%
\setlength{\parentwidth}{\parentoffset}%
\setlength{\treeoffset}{\daughteroffset}%
\addtolength{\treeoffset}{\treeoffsetone}%
\setlength{\treewidth}{\wd\treeboxone}%
\addtolength{\treewidth}{\daughteroffset}%
\ifdim\treewidth<\parentwidth\setlength{\treewidth}{\parentwidth}\fi%
\sbox{\treebox}{\begin{minipage}{\treewidth}%
\begin{flushleft}%
{\setlength{\unitlength}{2ex}%
\hspace*{\treeoffset}\begin{picture}(0,1)%
\put(0,0){\line(0,1){1}}%
\end{picture}}\\
\vspace{-\baselineskip}
\hspace*{\daughteroffset}%
\raisebox{-\ht\treeboxone}{\usebox{\treeboxone}}%
\end{flushleft}%
\end{minipage}}%
\setlength{\treeoffsetone}{\parentoffset}%
\setlength{\treeshiftone}{0pt}%
\setlength{\treewidthone}{\treewidth}%
\sbox{\treeboxone}{\usebox{\treebox}}%
\else\ifnum\value{branchcount}=2%
\setlength{\branchwidthone}{\treewidthtwo}%
\addtolength{\branchwidthone}{\treeoffsetone}%
\addtolength{\branchwidthone}{-\treeshiftone}%
\addtolength{\branchwidthone}{-\treeoffsettwo}%
\setlength{\branchwidth}{\branchwidthone}%
\setlength{\daughteroffsetone}{\branchwidth}%
\addtolength{\daughteroffsetone}{-\branchwidthone}%
\addtolength{\daughteroffsetone}{-\treeshiftone}%
\setlength{\parentoffset}{0pt}
\addtolength{\parentoffset}{\treeoffsettwo}%
\addtolength{\parentoffset}{0.5\branchwidth}%
\setlength{\daughteroffset}{0in}%
\ifdim\parentoffset<0in%
\setlength{\daughteroffset}{-\parentoffset}%
\setlength{\parentoffset}{0in}\fi%
\setlength{\parentwidth}{\parentoffset}%
\setlength{\treeoffset}{\daughteroffset}%
\addtolength{\treeoffset}{\treeoffsettwo}%
\setlength{\treewidth}{\wd\treeboxone}%
\addtolength{\treewidth}{\daughteroffsetone}%
\addtolength{\treewidth}{\treewidthtwo}%
\addtolength{\treewidth}{\daughteroffset}%
\ifdim\treewidth<\parentwidth\setlength{\treewidth}{\parentwidth}\fi%
\sbox{\treebox}{\begin{minipage}{\treewidth}%
\begin{flushleft}%
{\setlength{\unitlength}{0.5\branchwidth}%
\hspace*{\treeoffset}\begin{picture}(2,0.5)%
\put(0,0){\line(2,1){1}}%
\put(2,0){\line(-2,1){1}}%
\end{picture}}\\
\vspace{-\baselineskip}
\hspace*{\daughteroffset}%
\makebox[\treewidthtwo][l]%
{\raisebox{-\ht\treeboxtwo}{\usebox{\treeboxtwo}}}%
\hspace*{\daughteroffsetone}%
\raisebox{-\ht\treeboxone}{\usebox{\treeboxone}}%
\end{flushleft}%
\end{minipage}}%
\setlength{\treeoffsetone}{\parentoffset}%
\setlength{\treeshiftone}{0pt}%
\setlength{\treewidthone}{\treewidth}%
\sbox{\treeboxone}{\usebox{\treebox}}\poptree%
\else\ifnum\value{branchcount}=3%
\setlength{\branchwidthone}{\treewidthtwo}%
\addtolength{\branchwidthone}{\treeoffsetone}%
\addtolength{\branchwidthone}{-\treeshiftone}%
\addtolength{\branchwidthone}{-\treeoffsettwo}%
\setlength{\branchwidthtwo}{\treewidththree}%
\addtolength{\branchwidthtwo}{\treeoffsettwo}%
\addtolength{\branchwidthtwo}{-\treeshifttwo}%
\addtolength{\branchwidthtwo}{-\treeoffsetthree}%
\setlength{\branchwidth}{\branchwidthone}%
\ifdim\branchwidthtwo>\branchwidth%
\setlength{\branchwidth}{\branchwidthtwo}\fi%
\setlength{\daughteroffsetone}{\branchwidth}%
\addtolength{\daughteroffsetone}{-\branchwidthone}%
\addtolength{\daughteroffsetone}{-\treeshiftone}%
\setlength{\daughteroffsettwo}{\branchwidth}%
\addtolength{\daughteroffsettwo}{-\branchwidthtwo}%
\addtolength{\daughteroffsettwo}{-\treeshifttwo}%
\setlength{\parentoffset}{-0.5\wd\parentbox}%
\addtolength{\parentoffset}{\treeoffsetthree}%
\addtolength{\parentoffset}{\branchwidth}%
\setlength{\daughteroffset}{0in}%
\ifdim\parentoffset<0in%
\setlength{\daughteroffset}{-\parentoffset}%
\setlength{\parentoffset}{0in}\fi%
\setlength{\parentwidth}{\parentoffset}%
\addtolength{\parentwidth}{\wd\parentbox}%
\setlength{\treeoffset}{\daughteroffset}%
\addtolength{\treeoffset}{\treeoffsetthree}%
\setlength{\treewidth}{\wd\treeboxone}%
\addtolength{\treewidth}{\daughteroffsetone}%
\addtolength{\treewidth}{\treewidthtwo}%
\addtolength{\treewidth}{\daughteroffsettwo}%
\addtolength{\treewidth}{\treewidththree}%
\addtolength{\treewidth}{\daughteroffset}%
\ifdim\treewidth<\parentwidth\setlength{\treewidth}{\parentwidth}\fi%
\sbox{\treebox}{\begin{minipage}{\treewidth}%
\begin{flushleft}%
\hspace*{\parentoffset}\usebox{\parentbox}\\
{\setlength{\unitlength}{0.5\branchwidth}%
\hspace*{\treeoffset}\begin{picture}(4,1)%
\put(0,0){\line(2,1){2}}%
\put(2,0){\line(0,1){1}}%
\put(4,0){\line(-2,1){2}}%
\end{picture}}\\
\vspace{-\baselineskip}
\hspace*{\daughteroffset}%
\makebox[\treewidththree][l]%
{\raisebox{-\ht\treeboxthree}{\usebox{\treeboxthree}}}%
\hspace*{\daughteroffsettwo}%
\makebox[\treewidthtwo][l]%
{\raisebox{-\ht\treeboxtwo}{\usebox{\treeboxtwo}}}%
\hspace*{\daughteroffsetone}%
\raisebox{-\ht\treeboxone}{\usebox{\treeboxone}}%
\end{flushleft}%
\end{minipage}}%
\setlength{\treeoffsetone}{\parentoffset}%
\addtolength{\treeoffsetone}{0.5\wd\parentbox}%
\setlength{\treeshiftone}{0pt}%
\setlength{\treewidthone}{\treewidth}%
\sbox{\treeboxone}{\usebox{\treebox}}\poptree\poptree%
\else\ifnum\value{branchcount}=4%
\setlength{\branchwidthone}{\treewidthtwo}%
\addtolength{\branchwidthone}{\treeoffsetone}%
\addtolength{\branchwidthone}{-\treeshiftone}%
\addtolength{\branchwidthone}{-\treeoffsettwo}%
\setlength{\branchwidthtwo}{\treewidththree}%
\addtolength{\branchwidthtwo}{\treeoffsettwo}%
\addtolength{\branchwidthtwo}{-\treeshifttwo}%
\addtolength{\branchwidthtwo}{-\treeoffsetthree}%
\setlength{\branchwidththree}{\treewidthfour}%
\addtolength{\branchwidththree}{\treeoffsetthree}%
\addtolength{\branchwidththree}{-\treeshiftthree}%
\addtolength{\branchwidththree}{-\treeoffsetfour}%
\setlength{\branchwidth}{\branchwidthone}%
\ifdim\branchwidthtwo>\branchwidth%
\setlength{\branchwidth}{\branchwidthtwo}\fi%
\ifdim\branchwidththree>\branchwidth%
\setlength{\branchwidth}{\branchwidththree}\fi%
\setlength{\daughteroffsetone}{\branchwidth}%
\addtolength{\daughteroffsetone}{-\branchwidthone}%
\addtolength{\daughteroffsetone}{-\treeshiftone}%
\setlength{\daughteroffsettwo}{\branchwidth}%
\addtolength{\daughteroffsettwo}{-\branchwidthtwo}%
\addtolength{\daughteroffsettwo}{-\treeshifttwo}%
\setlength{\daughteroffsetthree}{\branchwidth}%
\addtolength{\daughteroffsetthree}{-\branchwidththree}%
\addtolength{\daughteroffsetthree}{-\treeshiftthree}%
\setlength{\parentoffset}{0pt}
\addtolength{\parentoffset}{\treeoffsetfour}%
\addtolength{\parentoffset}{1.5\branchwidth}%
\setlength{\daughteroffset}{0in}%
\ifdim\parentoffset<0in%
\setlength{\daughteroffset}{-\parentoffset}%
\setlength{\parentoffset}{0in}\fi%
\setlength{\parentwidth}{\parentoffset}%
\setlength{\treeoffset}{\daughteroffset}%
\addtolength{\treeoffset}{\treeoffsetfour}%
\setlength{\treewidth}{\wd\treeboxone}%
\addtolength{\treewidth}{\daughteroffsetone}%
\addtolength{\treewidth}{\treewidthtwo}%
\addtolength{\treewidth}{\daughteroffsettwo}%
\addtolength{\treewidth}{\treewidththree}%
\addtolength{\treewidth}{\daughteroffsetthree}%
\addtolength{\treewidth}{\treewidthfour}%
\addtolength{\treewidth}{\daughteroffset}%
\ifdim\treewidth<\parentwidth\setlength{\treewidth}{\parentwidth}\fi%
\sbox{\treebox}{\begin{minipage}{\treewidth}%
\begin{flushleft}%
{\setlength{\unitlength}{0.5\branchwidth}%
\hspace*{\treeoffset}\begin{picture}(6,1)%
\put(0,0){\line(3,1){3}}%
\put(2,0){\line(1,1){1}}%
\put(4,0){\line(-1,1){1}}%
\put(6,0){\line(-3,1){3}}%
\end{picture}}\\
\vspace{-\baselineskip}
\hspace*{\daughteroffset}%
\makebox[\treewidthfour][l]%
{\raisebox{-\ht\treeboxfour}{\usebox{\treeboxfour}}}%
\hspace*{\daughteroffsetthree}%
\makebox[\treewidththree][l]%
{\raisebox{-\ht\treeboxthree}{\usebox{\treeboxthree}}}%
\hspace*{\daughteroffsettwo}%
\makebox[\treewidthtwo][l]%
{\raisebox{-\ht\treeboxtwo}{\usebox{\treeboxtwo}}}%
\hspace*{\daughteroffsetone}%
\raisebox{-\ht\treeboxone}{\usebox{\treeboxone}}%
\end{flushleft}%
\end{minipage}}%
\setlength{\treeoffsetone}{\parentoffset}%
\setlength{\treeshiftone}{0pt}%
\setlength{\treewidthone}{\treewidth}%
\sbox{\treeboxone}{\usebox{\treebox}}\poptree\poptree\poptree%
\else\ifnum\value{branchcount}=5
\setlength{\branchwidthone}{\treewidthtwo}%
\addtolength{\branchwidthone}{\treeoffsetone}%
\addtolength{\branchwidthone}{-\treeshiftone}%
\addtolength{\branchwidthone}{-\treeoffsettwo}%
\setlength{\branchwidthtwo}{\treewidththree}%
\addtolength{\branchwidthtwo}{\treeoffsettwo}%
\addtolength{\branchwidthtwo}{-\treeshifttwo}%
\addtolength{\branchwidthtwo}{-\treeoffsetthree}%
\setlength{\branchwidththree}{\treewidthfour}%
\addtolength{\branchwidththree}{\treeoffsetthree}%
\addtolength{\branchwidththree}{-\treeshiftthree}%
\addtolength{\branchwidththree}{-\treeoffsetfour}%
\setlength{\branchwidthfour}{\treewidthfive}%
\addtolength{\branchwidthfour}{\treeoffsetfour}%
\addtolength{\branchwidthfour}{-\treeshiftfour}%
\addtolength{\branchwidthfour}{-\treeoffsetfive}%
\setlength{\branchwidth}{\branchwidthone}%
\ifdim\branchwidthtwo>\branchwidth%
\setlength{\branchwidth}{\branchwidthtwo}\fi%
\ifdim\branchwidththree>\branchwidth%
\setlength{\branchwidth}{\branchwidththree}\fi%
\ifdim\branchwidthfour>\branchwidth%
\setlength{\branchwidth}{\branchwidthfour}\fi%
\setlength{\daughteroffsetone}{\branchwidth}%
\addtolength{\daughteroffsetone}{-\branchwidthone}%
\addtolength{\daughteroffsetone}{-\treeshiftone}%
\setlength{\daughteroffsettwo}{\branchwidth}%
\addtolength{\daughteroffsettwo}{-\branchwidthtwo}%
\addtolength{\daughteroffsettwo}{-\treeshifttwo}%
\setlength{\daughteroffsetthree}{\branchwidth}%
\addtolength{\daughteroffsetthree}{-\branchwidththree}%
\addtolength{\daughteroffsetthree}{-\treeshiftthree}%
\setlength{\daughteroffsetfour}{\branchwidth}%
\addtolength{\daughteroffsetfour}{-\branchwidthfour}%
\addtolength{\daughteroffsetfour}{-\treeshiftfour}%
\setlength{\parentoffset}{0pt}
\addtolength{\parentoffset}{\treeoffsetfive}%
\addtolength{\parentoffset}{2\branchwidth}%
\setlength{\daughteroffset}{0in}%
\ifdim\parentoffset<0in%
\setlength{\daughteroffset}{-\parentoffset}%
\setlength{\parentoffset}{0in}\fi%
\setlength{\parentwidth}{\parentoffset}%
\setlength{\treeoffset}{\daughteroffset}%
\addtolength{\treeoffset}{\treeoffsetfive}%
\setlength{\treewidth}{\wd\treeboxone}%
\addtolength{\treewidth}{\daughteroffsetone}%
\addtolength{\treewidth}{\treewidthtwo}%
\addtolength{\treewidth}{\daughteroffsettwo}%
\addtolength{\treewidth}{\treewidththree}%
\addtolength{\treewidth}{\daughteroffsetthree}%
\addtolength{\treewidth}{\treewidthfour}%
\addtolength{\treewidth}{\daughteroffsetfour}%
\addtolength{\treewidth}{\treewidthfive}%
\addtolength{\treewidth}{\daughteroffset}%
\ifdim\treewidth<\parentwidth\setlength{\treewidth}{\parentwidth}\fi%
\sbox{\treebox}{\begin{minipage}{\treewidth}%
\begin{flushleft}%
{\setlength{\unitlength}{0.5\branchwidth}%
\hspace*{\treeoffset}\begin{picture}(8,1)%
\put(0,0){\line(4,1){4}}%
\put(2,0){\line(2,1){2}}%
\put(4,0){\line(0,1){1}}%
\put(6,0){\line(-2,1){2}}%
\put(8,0){\line(-4,1){4}}%
\end{picture}}\\
\vspace{-\baselineskip}
\hspace*{\daughteroffset}%
\makebox[\treewidthfive][l]%
{\raisebox{-\ht\treeboxfour}{\usebox{\treeboxfive}}}%
\hspace*{\daughteroffsetfour}%
\makebox[\treewidthfour][l]%
{\raisebox{-\ht\treeboxfour}{\usebox{\treeboxfour}}}%
\hspace*{\daughteroffsetthree}%
\makebox[\treewidththree][l]%
{\raisebox{-\ht\treeboxthree}{\usebox{\treeboxthree}}}%
\hspace*{\daughteroffsettwo}%
\makebox[\treewidthtwo][l]%
{\raisebox{-\ht\treeboxtwo}{\usebox{\treeboxtwo}}}%
\hspace*{\daughteroffsetone}%
\raisebox{-\ht\treeboxone}{\usebox{\treeboxone}}%
\end{flushleft}%
\end{minipage}}%
\setlength{\treeoffsetone}{\parentoffset}%
\setlength{\treeshiftone}{0pt}%
\setlength{\treewidthone}{\treewidth}%
\sbox{\treeboxone}{\usebox{\treebox}}\poptree\poptree\poptree\poptree%
\else\typeout{QobiTeX warning--- Can't handle #1 branching}\fi\fi\fi\fi\fi}
\newcommand{\tree}{%
\usebox{\treeboxone}
\setlength{\treeoffsetone}{\treeoffsettwo}%
\sbox{\treeboxone}{\usebox{\treeboxtwo}}%
\poptree}

\def\diatop[#1|#2]{{\setbox1=\hbox{{#1{}}}\setbox2=\hbox{{#2{}}}%
                    \dimen0=\ifdim\wd1>\wd2\wd1\else\wd2\fi%
                    \dimen1=\ht2\advance\dimen1by-1ex%
                    \setbox1=\hbox to1\dimen0{\hss#1\hss}%
                    \rlap{\raise1\dimen1\box1}%
                    \hbox to1\dimen0{\hss#2\hss}}}%
 
\def\ipa{\ipatwelverm}
 
\def\inva{\edef\next{\the\font}\ipa\char'000\next}%
\def\scripta{\edef\next{\the\font}\ipa\char'001\next}%
\def\nialpha{\edef\next{\the\font}\ipa\char'002\next}%
\def\invscripta{\edef\next{\the\font}\ipa\char'003\next}%
\def\invv{\edef\next{\the\font}\ipa\char'004\next}%
 
\def\crossb{\edef\next{\the\font}\ipa\char'005\next}%
\def\barb{\edef\next{\the\font}\ipa\char'006\next}%
\def\slashb{\edef\next{\the\font}\ipa\char'007\next}%
\def\hookb{\edef\next{\the\font}\ipa\char'010\next}%
\def\nibeta{\edef\next{\the\font}\ipa\char'011\next}%
 
\def\slashc{\edef\next{\the\font}\ipa\char'012\next}%
\def\curlyc{\edef\next{\the\font}\ipa\char'013\next}%
\def\clickc{\edef\next{\the\font}\ipa\char'014\next}%
 
\def\crossd{\edef\next{\the\font}\ipa\char'015\next}%
\def\bard{\edef\next{\the\font}\ipa\char'016\next}%
\def\slashd{\edef\next{\the\font}\ipa\char'017\next}%
\def\hookd{\edef\next{\the\font}\ipa\char'020\next}%
\def\taild{\edef\next{\the\font}\ipa\char'021\next}%
\def\dz{\edef\next{\the\font}\ipa\char'022\next}%
\def\eth{\edef\next{\the\font}\ipa\char'023\next}%
\def\scd{\edef\next{\the\font}\ipa\char'024\next}%
 
\def\schwa{\edef\next{\the\font}\ipa\char'025\next}%
\def\er{\edef\next{\the\font}\ipa\char'026\next}%
\def\reve{\edef\next{\the\font}\ipa\char'027\next}%
\def\niepsilon{\edef\next{\the\font}\ipa\char'030\next}%
\def\revepsilon{\edef\next{\the\font}\ipa\char'031\next}%
\def\hookrevepsilon{\edef\next{\the\font}\ipa\char'032\next}%
\def\closedrevepsilon{\edef\next{\the\font}\ipa\char'033\next}%
 
\def\scriptg{\edef\next{\the\font}\ipa\char'034\next}%
\def\hookg{\edef\next{\the\font}\ipa\char'035\next}%
\def\scg{\edef\next{\the\font}\ipa\char'036\next}%
\def\nigamma{\edef\next{\the\font}\ipa\char'037\next}
\def\ipagamma{\edef\next{\the\font}\ipa\char'040\next}%
\def\babygamma{\edef\next{\the\font}\ipa\char'041\next}%
 
\def\hv{\edef\next{\the\font}\ipa\char'042\next}%
\def\crossh{\edef\next{\the\font}\ipa\char'043\next}%
\def\hookh{\edef\next{\the\font}\ipa\char'044\next}%
\def\hookheng{\edef\next{\the\font}\ipa\char'045\next}%
\def\invh{\edef\next{\the\font}\ipa\char'046\next}%
 
\def\bari{\edef\next{\the\font}\ipa\char'047\next}%
\def\dlbari{\edef\next{\the\font}\ipa\char'050\next}% ``dotless bar i''
\def\niiota{\edef\next{\the\font}\ipa\char'051\next}%
\def\sci{\edef\next{\the\font}\ipa\char'052\next}%
\def\barsci{\edef\next{\the\font}\ipa\char'053\next}% ``barred small cap i''
 
\def\invf{\edef\next{\the\font}\ipa\char'054\next}%
 
\def\tildel{\edef\next{\the\font}\ipa\char'055\next}%
\def\barl{\edef\next{\the\font}\ipa\char'056\next}%
\def\latfric{\edef\next{\the\font}\ipa\char'057\next}%
\def\taill{\edef\next{\the\font}\ipa\char'060\next}%
\def\lz{\edef\next{\the\font}\ipa\char'061\next}%
\def\nilambda{\edef\next{\the\font}\ipa\char'062\next}%
\def\crossnilambda{\edef\next{\the\font}\ipa\char'063\next}%
 
\def\labdentalnas{\edef\next{\the\font}\ipa\char'064\next}%
\def\invm{\edef\next{\the\font}\ipa\char'065\next}%
\def\legm{\edef\next{\the\font}\ipa\char'066\next}%
 
\def\nj{\edef\next{\the\font}\ipa\char'067\next}%
\def\eng{\edef\next{\the\font}\ipa\char'070\next}%
\def\tailn{\edef\next{\the\font}\ipa\char'071\next}%
\def\scn{\edef\next{\the\font}\ipa\char'072\next}%
 
\def\clickb{\edef\next{\the\font}\ipa\char'073\next}%
\def\baro{\edef\next{\the\font}\ipa\char'074\next}%
\def\openo{\edef\next{\the\font}\ipa\char'075\next}%
\def\niomega{\edef\next{\the\font}\ipa\char'076\next}%
\def\closedniomega{\edef\next{\the\font}\ipa\char'077\next}%
\def\oo{\edef\next{\the\font}\ipa\char'100\next}%
 
\def\barp{\edef\next{\the\font}\ipa\char'101\next}%
\def\thorn{\edef\next{\the\font}\ipa\char'102\next}%
\def\niphi{\edef\next{\the\font}\ipa\char'103\next}%
 
\def\flapr{\edef\next{\the\font}\ipa\char'104\next}%
\def\legr{\edef\next{\the\font}\ipa\char'105\next}%
\def\tailr{\edef\next{\the\font}\ipa\char'106\next}%
\def\invr{\edef\next{\the\font}\ipa\char'107\next}%
\def\tailinvr{\edef\next{\the\font}\ipa\char'110\next}%
\def\invlegr{\edef\next{\the\font}\ipa\char'111\next}%
\def\scr{\edef\next{\the\font}\ipa\char'112\next}%
\def\invscr{\edef\next{\the\font}\ipa\char'113\next}%
 
\def\tails{\edef\next{\the\font}\ipa\char'114\next}%
\def\esh{\edef\next{\the\font}\ipa\char'115\next}%
\def\curlyesh{\edef\next{\the\font}\ipa\char'116\next}%
\def\nisigma{\edef\next{\the\font}\ipa\char'117\next}%
 
\def\tailt{\edef\next{\the\font}\ipa\char'120\next}%
\def\tesh{\edef\next{\the\font}\ipa\char'121\next}%
\def\clickt{\edef\next{\the\font}\ipa\char'122\next}%
\def\nitheta{\edef\next{\the\font}\ipa\char'123\next}%
 
\def\baru{\edef\next{\the\font}\ipa\char'124\next}%
\def\slashu{\edef\next{\the\font}\ipa\char'125\next}%
\def\niupsilon{\edef\next{\the\font}\ipa\char'126\next}%
\def\scu{\edef\next{\the\font}\ipa\char'127\next}%
\def\barscu{\edef\next{\the\font}\ipa\char'130\next}%
 
\def\scriptv{\edef\next{\the\font}\ipa\char'131\next}%
 
\def\invw{\edef\next{\the\font}\ipa\char'132\next}%
 
\def\nichi{\edef\next{\the\font}\ipa\char'133\next}%
 
\def\invy{\edef\next{\the\font}\ipa\char'134\next}%
\def\scy{\edef\next{\the\font}\ipa\char'135\next}%
 
\def\curlyz{\edef\next{\the\font}\ipa\char'136\next}%
\def\tailz{\edef\next{\the\font}\ipa\char'137\next}%
\def\yogh{\edef\next{\the\font}\ipa\char'140\next}%
\def\curlyyogh{\edef\next{\the\font}\ipa\char'141\next}%
 
\def\glotstop{\edef\next{\the\font}\ipa\char'142\next}%
\def\revglotstop{\edef\next{\the\font}\ipa\char'143\next}%
\def\invglotstop{\edef\next{\the\font}\ipa\char'144\next}%
\def\ejective{\edef\next{\the\font}\ipa\char'145\next}%
\def\reveject{\edef\next{\the\font}\ipa\char'146\next}%
 
\def\upt{\edef\next{\the\font}\ipa\char'154\next}%   These are IPA pointers
\def\downt{\edef\next{\the\font}\ipa\char'155\next}%
\def\leftt{\edef\next{\the\font}\ipa\char'156\next}%
\def\rightt{\edef\next{\the\font}\ipa\char'157\next}%
 
\def\upp{\edef\next{\the\font}\ipa\char'164\next}
\def\downp{\edef\next{\the\font}\ipa\char'165\next}%
\def\leftp{\edef\next{\the\font}\ipa\char'166\next}%
\def\rightp{\edef\next{\the\font}\ipa\char'167\next}%
 
\def\stress{\edef\next{\the\font}\ipa\char'150\next}%     primary stress
\def\secstress{\edef\next{\the\font}\ipa\char'151\next}%  secondary stress
 
\def\syllabic{\edef\next{\the\font}\ipa\char'152\next}%   syllabic marker
 
\def\corner{\edef\next{\the\font}\ipa\char'153\next}%
 
\def\halflength{\edef\next{\the\font}\ipa\char'160\next}
\def\length{\edef\next{\the\font}\ipa\char'161\next}
 
\def\underdots{\edef\next{\the\font}\ipa\char'162\next}%
 
\def\ain{\edef\next{\the\font}\ipa\char'163\next}
 
\def\overring{\edef\next{\the\font}\ipa\char'170\next}%
\def\underring{\edef\next{\the\font}\ipa\char'171\next}%
 
\def\open{\edef\next{\the\font}\ipa\char'172\next}%
 
\def\midtilde{\edef\next{\the\font}\ipa\char'173\next}%
\def\undertilde{\edef\next{\the\font}\ipa\char'174\next}%
 
\def\underwedge{\edef\next{\the\font}\ipa\char'175\next}%
 
\def\polishhook{\edef\next{\the\font}\ipa\char'176\next}%

\newlength{\pausewidth}
\newlength{\longpausewidth}
\newcommand{\unib}{b}
\newcommand{\unid}{d}
\newcommand{\unig}{g}
\newcommand{\unip}{p}
\newcommand{\unit}{t}
\newcommand{\unik}{k}
\newcommand{\uniJ}{\v{j}}
\newcommand{\uniC}{\v{c}}
\newcommand{\unis}{s}
\newcommand{\uniS}{\v{s}}
\newcommand{\uniz}{z}
\newcommand{\uniZ}{\v{z}}
\newcommand{\unif}{f}
\newcommand{\uniT}{\nitheta}
\newcommand{\univ}{v}
\newcommand{\uniD}{\eth}
\newcommand{\unim}{m}
\newcommand{\unin}{n}
\newcommand{\uniG}{\eng}
\newcommand{\unil}{l}
\newcommand{\unir}{r}
\newcommand{\uniw}{w}
\newcommand{\uniy}{y}
\newcommand{\unih}{h}
\newcommand{\uniH}{\hookh}
\newcommand{\unii}{i}
\newcommand{\uniI}{\sci}
\newcommand{\uniE}{\niepsilon}
\newcommand{\unie}{e}
\newcommand{\uniA}{\ae}
\newcommand{\unia}{a}
\newcommand{\uniAH}{\invv} % ^
\newcommand{\uniO}{\openo}
\newcommand{\unio}{o}
\newcommand{\uniU}{\niupsilon}
\newcommand{\uniu}{u}
\newcommand{\uniAX}{\schwa} % *
\newcommand{\uniIX}{\bari} % &
\newcommand{\uniEPI}{-} % -

\font\ipaninerm=wsuipa9
\font\ipatenrm=wsuipa10

\begin{document}
\def\ipa{\ipatenrm}
\pagestyle{headings}

\begin{titlepage}
\begin{center}
{\large\bf Unsupervised Language Acquisition}\linebreak
\newline
by\linebreak
\newline
Carl G.\ de Marcken\linebreak
\newline
S.B.~Electrical Engineering and Computer Science (1990)\\
Massachussetts Institute of Technology\linebreak
%(1990)\linebreak
\newline
S.B.~Mathematics (1991)\\
Massachussetts Institute of Technology\linebreak
%(1991)\linebreak
\newline
S.M.~Electrical Engineering and Computer Science (1993)\\
Massachussetts Institute of Technology\linebreak
%(1993)\linebreak
\newline
\newline
Submitted to the Department of\\
Electrical Engineering and Computer Science\\
in partial fulfillment of the requirements for the degree of\\
Doctor of Philosophy\linebreak
\newline
at the\linebreak
\newline
Massachusetts Institute of Technology\linebreak
\newline
September, 1996\\
\vfill
\copyright 1996 Massachusetts Intstitute of Technology.  All Rights Reserved.\linebreak
\end{center}
\vfill
{\raggedright Signature of Author \hrulefill\mbox{}\\}
{\raggedleft Department of Electrical Engineering and Computer Science\\
September 6, 1996\\}
\bigskip\medskip
{\raggedright Certified by \hrulefill\mbox{}\\}
{\raggedleft Robert C.~Berwick\\
Professor of Computer Science and Engineering\\
Thesis Supervisor\\}
\bigskip\medskip
{\raggedright Accepted by \hrulefill\mbox{}\\}
{\raggedleft F.\ R.\ Morgenthaler\\
Chair, Department Committee on Graduate Students\\}
\end{titlepage}
\newpage
\setcounter{page}{2}
\thispagestyle{empty}
\mbox{ }
\newpage
\begin{center}
{\large\bf Unsupervised Language Acquisition}\\
\vspace{.15in}
by\\
\vspace{.15in}
Carl G.~de Marcken\\
\vspace{0.3in}
Submitted to the Department of Electrical Engineering and Computer Science\\
on September 6th, 1996, in partial fulfillment of the requirements for the degree
of\\
Doctor of Philosophy
\end{center}
\vfill
\begin{center}
{\large\bf Abstract}
\end{center}
\vfill

Children are exposed to speech and other environmental evidence, from which
they learn language.  How do they do this?  More specifically, how do
children map from complex, physical signals to grammars that enable them to
generate and interpret new utterances from their language?

This thesis presents a computational theory of unsupervised language
acquisition.  By {\em computational} we mean that the theory precisely
defines procedures for learning language, procedures that have been
implemented and tested in the form of computer programs.  By {\em
  unsupervised} we mean that the theory explains how language learning can
take place with no explicit help from a teacher, but only exposure to
ordinary spoken or written utterances.  The theory requires very little of
the learning environment.  For example, it predicts that much knowledge of
language can be acquired even in situations where the learner has no access
to the meaning of utterances.  In this way the theory is extremely
conservative, making few or no assumptions that are not obviously true of
the situation children learn in.

The theory is based heavily on concepts borrowed from machine learning and
statistical estimation.  In particular, learning takes place by fitting a
stochastic, generative model of language to the evidence.  Thus, the goal
of the learner is to acquire a grammar under which the evidence is
``typical'', in a statistical sense.  Much of the thesis is devoted to
explaining conditions that must hold for this learning strategy to arrive
at the desired form of grammar.  The thesis introduces a variety of
technical innovations, among them a common representation for evidence and
grammars that has many linguistically and statistically desirable
properties.  In this representation, both utterances and parameters in the
grammar are represented by composing parameters.  A second contribution is
a learning strategy that separates the ``content'' of linguistic parameters
from their representation.  Algorithms based on it suffer from few of the
search problems that have plagued other computational approaches to
language acquisition.

The theory has been tested on problems of learning lexicons (vocabularies)
and stochastic grammars from unsegmented text {\em and continuous speech}
signals, and mappings between sound and representations of meaning.  It
performs extremely well on various objective criteria, acquiring knowledge
that causes it to assign almost exactly the same linguistic structure to
utterances as humans do.  This work has application to data compression,
language modeling, speech recognition, machine translation, information
retrieval, and other tasks that rely on either structural or stochastic
descriptions of language.

\vfill
\begin{flushleft}
Thesis Supervisor: Robert C.~Berwick\\
Title: Professor of Computer Science and Engineering
\end{flushleft}
\newpage

\begin{center}
{\bf Acknowledgments}
\end{center}

The research presented here owes an enormous debt to Robert C.~Berwick, who
has supported and taught me for every one of the ten years I have been at
MIT.  Had I not too much respect for his work and ideas, the first
sentences of my thesis would have been borrowed from his.

Members of the MIT linguistics department have kept my interest in language
from flagging, by convincing and reconvincing me of its complexity and
beauty.  I thank Ken Hale, James Harris, Michael Kenstowicz, Alec Marantz,
David Pesetsky, Ken Wexler, and especially Morris Halle.

David Baggett, my favorite creative partner, first got me interested in
problems of phonological and lexical acquisition.  Had he remained at MIT
for the completion of this work, I have no doubt it would have benefited
greatly from his influence; I very much hope to work with him again.

Marina Meil\u{a} listened carefully to my ideas, read and fixed my
documents, taught me countless things I should have known, kept me up to
date on happenings in other fields, and provided emotional support without
which this thesis would have been impossible.

The thesis has benefited from discussions with many other people, including
Robert Thomas, Eric Ristad and Paul Viola.  Early conversations with Jeff
Siskind played a large role in my choice of thesis topic.  Gina Levow, Eric
Miller, Charles Isbell and Oded Maron all carefully read this document or
other presentations of the material and provided valuable feedback; any
qualities reflect their influence and all faults are my own.

And last there are those who cannot be thanked to the extent they deserve.
Jim Rees, Olaf Bleck, Greg Galperin, Gideon Stein.  Robin DeWitt.  Marina
Meil\u{a}.  Mom, Dad, Paya and Natasha.  Finally, David Lennart Marcelius--
Grandpa-- whom I dedicate this thesis to.

\vspace{1in}

The work has been supported through the Center for Biological and
Computational Learning at MIT, funded in part by NSF grant 9217041-ASC and
ARPA under the HPCC and AASERT programs.

\tableofcontents

\chapter{Introduction}\label{ch:introduction}

Children are exposed to speech and other environmental evidence, from which
they learn language.  How do they do this?  More specifically, how do
children map from complex, physical signals to grammars that enable them to
generate and interpret new utterances from their language?

This thesis presents a computational theory of unsupervised language
acquisition.  By {\em computational} we mean that the theory precisely
defines procedures for learning language, procedures that have been
implemented and tested in the form of computer programs.  By {\em
  unsupervised} we mean that the theory explains how language learning can
take place with no explicit help from a teacher, but only exposure to
ordinary spoken or written utterances.  The theory requires very little of
the learning environment.  For example, it predicts that much knowledge of
language can be acquired even in situations where the learner has no access
to the meaning of utterances.  In this way the theory is extremely
conservative, making few or no assumptions that are not obviously true of
the situation children learn in.

The theory is based heavily on concepts borrowed from machine learning and
statistical estimation.  In particular, learning takes place by fitting a
stochastic, generative model of language to the evidence.  Thus, the goal
of the learner is to acquire a grammar under which the evidence is
``typical'', in a statistical sense.  Much of the thesis is devoted to
explaining conditions that must hold for this learning strategy to arrive
at the desired form of grammar.  The thesis introduces a variety of
technical innovations, among them a common representation for evidence and
grammars that has many linguistically and statistically desirable
properties.  In this representation, both utterances and parameters in the
grammar are represented by composing parameters.  A second contribution is
a learning strategy that separates the ``content'' of linguistic parameters
from their representation.  Algorithms based on it suffer from few of the
search problems that have plagued other computational approaches to
language acquisition.

The theory has been tested on problems of learning lexicons (vocabularies)
and stochastic grammars from unsegmented text {\em and continuous speech}
signals, and mappings between sound and representations of meaning.  It
performs extremely well on various objective criteria, acquiring knowledge
that causes it to assign almost exactly the same linguistic structure to
utterances as humans do.  This work has application to data compression,
language modeling, speech recognition, machine translation, information
retrieval, and other tasks that rely on either structural or stochastic
descriptions of language.

\section{Summary}

Why is language learning so easy for children?  An instinctive answer to
this question is that parents trivialize the task, by speaking clearly,
pausing between words, pointing at the objects they are referring to, and so
on.  Indeed, many adults treat babies as idiots, no more intelligent than
dogs or foreigners, who are accorded similar treatment.

However, as chapter~\ref{ch:la} will argue, it is not clear that teaching
is a necessary (or even significant) part of language acquisition.  Many
societies raise children differently, speaking to them as adults.
Furthermore, there are important aspects of language that are not
highlighted in the evidence children receive.  To take a classic example,
every adult knows that nouns ending in \word{t} are pluralized with the
\word{s} sound (\word{cats}), while nouns ending in \word{g} get the
\word{z} sound (\word{dogz}).  But children must discover this fact-- no
mother employs special hand gestures to indicate the cause of the variation
in the plural marker, which she may not even be consciously aware of.
Similarly, in the sentence

\begin{center}
  What do you think they're going to do with the kangaroo?
\end{center}

\noindent \word{going} is pronounced without a pause between \word{go}
and \word{-ing}.  Yet children come to know that \word{going} is formed of
a root \word{go} that conveys meaning, and a suffix \word{-ing} that
conveys tense information.  In fact, most times such sentences are spoken,
they are spoken rapidly, with few or no pauses.  In casual speech, which
children seem to be quite capable of learning from, one word blends into
the next, with sequences like \word{what do you\ldots} jumbled together
into /\uniw\uniAH\unid\uniJ\uniAX/.\footnote{See
  Appendix~\ref{app:phonemes} for a description of the phonetic symbols
  used to transcribe sounds in this thesis.}

In this thesis language acquisition is treated as a problem of unsupervised
learning.  Rather than suppose the learner looks for explicit clues in the
evidence that give indications as to the underlying the structure of their
language, we assume the learner acquires this knowledge indirectly.  The
learner's active goal is to find the grammar that best {\em predicts} the
evidence the learner is exposed to.  More specifically, the learner
maintains a stochastic, generative model of language that assigns a
probability to every utterance $u$.  This model is defined by a grammar $G$
that attaches distributional information to its parameters.  Roughly
speaking, learning consists of finding the grammar that maximizes the joint
probability of all the utterances the learner has heard.  For example,
suppose the learner entertains two possible stochastic grammars, $G_{E}$
and $G_{F}$, that assign probabilities $p(u|G_{E})$ and $p(u|G_{F})$
respectively:

\begin{center}
\begin{tabular}{lcc}
\makebox[2in][l]{$u$} & \makebox[1in]{$p(u|G_{E})$} & \makebox[1in]{$p(u|G_{F})$}\\ \hline
{Hello.} & $\approx 10^{-2}$ & $\approx 10^{-5}$\\
{Bonjour.} & $\approx 10^{-5}$ & $\approx 10^{-2}$\\
{What's your name?} & $\approx 10^{-4}$ & $\approx 10^{-9}$\\
{Comment t'appelles tu?} & $\approx 10^{-9}$ & $\approx 10^{-4}$\\
\vdots & \vdots & \vdots
\end{tabular}
\end{center}

\noindent Then given evidence \word{Hello- what's your name?} the learner
would choose grammar $G_{E}$, because the probability of the evidence is
much higher under it ($10^{-2}\cdot 10^{-4}$) than under $G_{F}$
($10^{-5}\cdot 10^{-9}$).

For this to be a viable learning strategy, stochastic grammars must contain
the information necessary to generate and interpret utterances.  For
instance, in the above example the learner must be able to extract from
$G_{E}$ the words and syntax of English, or no useful learning has taken
place.  Furthermore, it must be true that the best stochastic model for a
language is an extension of its ``true'' grammar.  Unfortunately, this is
not always the case.  To understand why, realize that in fitting a
stochastic model to the evidence, the learner is in effect discovering
patterns in the evidence.  But patterns can arise from sources other than
language.  For example, a child learning English will often hear such
phrases as \word{eat your peas} and \word{clean your plate}, but not
\word{eat your plate} or \word{clean your peas}.  This fact is not
explained by the phrases' linguistic structure.  There is a significant
risk that the learner will nevertheless account for it using linguistic
mechanisms, perhaps by adding \word{eatyourpeas} and \word{cleanyourplate}
to the lexicon.  As words, the sound pattern is explained.

The existence of ``extralinguistic'' patterns has been the downfall of many
previous computational theories of language acquisition.
Chapter~\ref{ch:representation} introduces a representational framework for
language that is designed to work around this problem.  In this framework
words and other parameters in the grammar are represented in the same way
as sentences, by composing parameters from the grammar.  For example, just
as the sentence \word{I saw Mary} can be broken into \word{I}, \word{saw}
and \word{Mary}, so a word like \word{blueberry} can be broken down into
\word{blue} and \word{berry}.  This representation will turn out to have
many advantages for learning.  One of them is that if \word{eatyourpeas}
does makes its way into the grammar, it will be represented in terms of
\word{eat}, \word{your} and \word{peas}, just as it would be in the
``correct'' grammar (at the sentence level).  This mitigates the
consequences of such unavoidable mistakes.  The representation can also be
justified on purely linguistic grounds, offering a natural explanation for
why words like \word{blueberry} seem to inherit properties of their parts,
while still introducing new behaviors and meanings.

\begin{figure}[tbh]
\pageline
\begin{center}
\mbox{\ }
\psfig{figure=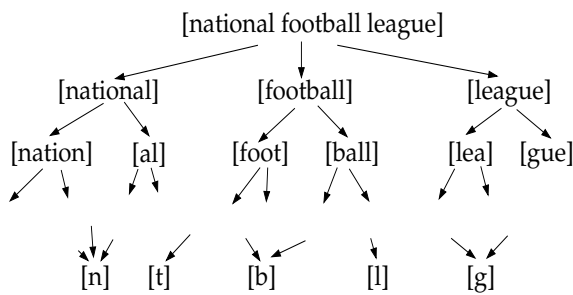,height=1.5in,width=3.0in}
\end{center}

\caption{\label{fig:nfl}The hierarchical representation of one word
  learned from a large body of text, \word{National Football League}, in
  terms of other ``words'' that were learned simultaneously.} \pageline
\end{figure}

Several instantiations of this framework are presented in the thesis, based
on simple models of language.  The first assumes that words and sentences
are character sequences, represented by concatenating words.  Thus, every
word and sentence is hierarchically decomposed.  Figure~\ref{fig:nfl}
presents an example of how one word learned from the Brown
corpus~\cite{Francis82} using this model is represented.  A second
instantiation of the framework extends this concatenative model with an
operator that adds meanings to words.  This model can be used to learn word
meanings from pairs of sentences and representations of meanings.  In
contrast to previous approaches to this problem, the model can account for
non-compositional behavior.  A third instantiation of the framework
explores phonetic and acoustic extensions; the resulting model is used to
learn words directly from continuous speech.  Finally, an instantiation of
the framework based on context-free grammars is explored.

In chapter~\ref{ch:algorithm} unsupervised learning algorithms are
presented for some of these models.  The algorithms start with a simple
stochastic grammar and iteratively refine it to increase the probability of
the training evidence.  Each iteration proceeds in two stages: first the
stochastic properties of the grammar are optimized while keeping the
underlying linguistic structure fixed, and then the linguistic structure is
altered in ways that are predicted to lead to a better stochastic model.
This general strategy has been used by others, who have found that their
algorithms get stuck in local optima-- grammars that are suboptimal but
which their algorithms can not improve upon.  The algorithms presented in
this thesis do not suffer from this problem to the same extent, because
they do not directly manipulate representations of grammars.  Instead, they
abstract to less-committal structures that are more closely tied to the
data than to the learner's model of the data.  To give an example, even
though the algorithms are based on the idea that sentences and words are
decomposed, sentences and words are stored as flat character sequences.
The best representation for a given word or sentence can be found by
parsing it.  Because of this, the representation for a word like
\word{watermelon} can go from \word{wa} $\circ$ \word{term} $\circ$
\word{el} $\circ$ \word{on} to \word{water} $\circ$ \word{melon} in a
single step that would stymie other algorithms.

Our algorithms are tested on problems of learning words and word meanings
from both unsegmented (spaceless) text and continuous speech.  The grammars
they produce are evaluated in terms of their linguistic and statistical
properties.  For example, after training on a large corpus of unsegmented
text, one algorithm produces hierarchical segmentations of the input such
as:

\[
\overline{\mbox{\tt $\,\overline{\mbox{\tt f$\,\overline{\mbox{\tt or}}\,$}}\,\overline{\mbox{\tt $\,\overline{\mbox{\tt t$\,\overline{\mbox{\tt he}}\,$}}\,\overline{\mbox{\tt $\,\overline{\mbox{\tt $\,\overline{\mbox{\tt p$\,\overline{\mbox{\tt ur}}\,$}}\,\overline{\mbox{\tt $\,\overline{\mbox{\tt $\,\overline{\mbox{\tt po}}\,$s}}\,$e}}\,$}}\,\overline{\mbox{\tt of}}\,$}}\,$}}\,$}}\,\overline{\mbox{\tt $\,\overline{\mbox{\tt $\,\overline{\mbox{\tt ma$\,\overline{\mbox{\tt in}}\,$}}\,\overline{\mbox{\tt ta$\,\overline{\mbox{\tt in}}\,$}}\,$}}\,\overline{\mbox{\tt $\,\overline{\mbox{\tt in}}\,$g}}\,$}}\,\overline{\mbox{\tt $\,\overline{\mbox{\tt $\,\overline{\mbox{\tt in}}\,\overline{\mbox{\tt t$\,\overline{\mbox{\tt er}}\,$}}\,$}}\,\overline{\mbox{\tt $\,\overline{\mbox{\tt n$\,\overline{\mbox{\tt a$\,\overline{\mbox{\tt t$\,\overline{\mbox{\tt i$\,\overline{\mbox{\tt on}}\,$}}\,$}}\,$}}\,$}}\,\overline{\mbox{\tt al}}\,$}}\,$}}\,\overline{\mbox{\tt $\,\overline{\mbox{\tt pe}}\,\overline{\mbox{\tt a$\,\overline{\mbox{\tt ce}}\,$}}\,$}}\,\overline{\mbox{\tt $\,\overline{\mbox{\tt an}}\,$d}}\,\overline{\mbox{\tt $\,\overline{\mbox{\tt p$\,\overline{\mbox{\tt ro}}\,$}}\,\overline{\mbox{\tt $\,\overline{\mbox{\tt mo}}\,$t}}\,\overline{\mbox{\tt $\,\overline{\mbox{\tt in}}\,$g}}\,$}}
\]
\[
\,\overline{\mbox{\tt t$\,\overline{\mbox{\tt he}}\,$}}\,\overline{\mbox{\tt $\,\overline{\mbox{\tt adv$\,\overline{\mbox{\tt a$\,\overline{\mbox{\tt n$\,\overline{\mbox{\tt ce}}\,$}}\,$}}\,$}}\,\overline{\mbox{\tt $\,\overline{\mbox{\tt $\,\overline{\mbox{\tt me}}\,$n}}\,$t}}\,$}}\,\overline{\mbox{\tt $\,\overline{\mbox{\tt of}}\,\overline{\mbox{\tt a$\,\overline{\mbox{\tt ll}}\,$}}\,$}}\,\overline{\mbox{\tt $\,\overline{\mbox{\tt pe}}\,\overline{\mbox{\tt op}}\,\overline{\mbox{\tt le}}\,$}}\,\overline{\mbox{\tt $\,\overline{\mbox{\tt $\,\overline{\mbox{\tt t$\,\overline{\mbox{\tt he}}\,$}}\,\overline{\mbox{\tt $\,\overline{\mbox{\tt $\,\overline{\mbox{\tt $\,\overline{\mbox{\tt un}}\,\overline{\mbox{\tt it}}\,$}}\,\overline{\mbox{\tt ed}}\,$}}\,\overline{\mbox{\tt $\,\overline{\mbox{\tt $\,\overline{\mbox{\tt st$\,\overline{\mbox{\tt at}}\,$}}\,$e}}\,$s}}\,$}}\,$}}\,\overline{\mbox{\tt $\,\overline{\mbox{\tt of}}\,\overline{\mbox{\tt a$\,\overline{\mbox{\tt me}}\,\overline{\mbox{\tt r$\,\overline{\mbox{\tt ic}}\,$}}\,$a}}\,$}}\,$}}\,\overline{\mbox{\tt $\,\overline{\mbox{\tt $\,\overline{\mbox{\tt jo}}\,\overline{\mbox{\tt in}}\,$}}\,\overline{\mbox{\tt ed}}\,$}}\,\overline{\mbox{\tt in}}
\]
\[
\overline{\mbox{\tt f$\,\overline{\mbox{\tt o$\,\overline{\mbox{\tt un}}\,$d}}\,$}}\,\overline{\mbox{\tt $\,\overline{\mbox{\tt in}}\,$g}}\,\overline{\mbox{\tt $\,\overline{\mbox{\tt t$\,\overline{\mbox{\tt he}}\,$}}\,\overline{\mbox{\tt $\,\overline{\mbox{\tt $\,\overline{\mbox{\tt $\,\overline{\mbox{\tt un}}\,\overline{\mbox{\tt it}}\,$}}\,\overline{\mbox{\tt ed}}\,$}}\,\overline{\mbox{\tt $\,\overline{\mbox{\tt n$\,\overline{\mbox{\tt a$\,\overline{\mbox{\tt t$\,\overline{\mbox{\tt i$\,\overline{\mbox{\tt on}}\,$}}\,$}}\,$}}\,$}}\,$s}}\,$}}\,$}}
\]

\noindent These segmentations are compared against word boundaries; the
results indicate that the algorithm produces structure that agrees
extremely well with humans' grammars.  On Chinese text, for example, 97\%
of word boundaries are matched and fewer than 1.3\% are violated.  On
statistical grounds the algorithms also fare very well: used as compression
algorithms they equal or better almost all other methods.
Chapter~\ref{ch:results} presents these and other results, including
dictionaries learned both from text and from speech.  In the most ambitious
test of any theory of language acquisition, we run on 68,000 utterances of
dictated Wall Street Journal articles-- complex, continuous speech produced
by many different speakers of both sexes.  Among the entries in the
resulting 9,600 word dictionary are

\begin{center}
\begin{tabular}{llll}
/\unip\unir\uniI\uniz\uniIX\unit\uniE\unin/ & president & /\unig\unio\uniu\unil\unid\unim\uniIX\unin\unis\uniA\unik\unis/ & Goldman-Sachs\\
/\unik\unim\unip\uniS\uniu\unit\unir/ & computer & /\unig\unia\univ\unir\unim\uniIX\unin/ & government\\
/\unim\uniIX\unin\uniIX\unis\unit\unir\unie\unii\uniS\uniIX\unin/ &
administration & /\unis\uniAH\unim\unip\uniD\uniIX\uniG/ & something\\
/\unib\unio\uniu\unis\unik\unig\unii/ & (Ivan) Boesky & /\unil\uniA\uniz\unid\uniJ\uniI\unir/ & last year\\
/\unih\unia\uniu\uniw\uniA\univ\unir/ & however & /\uniI\unin\uniI\unid\uniI\uniS\uniI\unin/ & in addition\\
\end{tabular}
\end{center}

Results like these demonstrate the power of our theory, both as an abstract
description of how children might learn, and as a foundation for the
machine acquisition of linguistic knowledge.

\section{Outline}

Chapter~\ref{ch:la} introduces the problem of language acquisition and
surveys the evidence available to the learner.  It argues that language
acquisition is best viewed as a problem of unsupervised learning, and
places constraints on theories of language acquisition, most importantly
that they be testable on data that is unequivocally available to children.
Finally, it argues that the phonological lexicon is the best starting point
for a complete theory of acquisition.

Chapter~\ref{ch:Bayes} introduces stochastic language models and the
statistical estimation technique of Bayesian inference.  It explains how
stochastic models can be used to differentiate between the many grammars
that are consistent with any given body of evidence, but cautions that
unless the class of language models satisfies certain conditions the
learning process may not produce the desired form of grammar.  Finally, it
discusses the problem of model selection and generalization from finite
evidence to grammars that explain new utterances.  The minimum description
length (MDL) principle is adopted as a substitute for structural risk
minimization.  Both of these strategies weigh the complexity of the set of
candidate grammars against the amount of evidence available.

Chapter~\ref{ch:representation} presents the compositional framework in
which both sentences and linguistic parameters are represented by
perturbing a composition of parameters.  Arguments for the framework are
given from the dual perspectives of learning and linguistics.  Four
instantiations of the framework are presented, that explore issues of
learning from speech and learning from simultaneous exposure to linguistic
and extralinguistic signals.

Chapter~\ref{ch:algorithm} describes two algorithms, one that learns
grammars from character sequences under the concatenative model, and
another that learns from character sequences paired with multiple
(ambiguous) representations of meaning.  A survey is made of related
algorithms and ideas from the fields of data compression, language
modeling, formal grammar induction, and orthographic segmentation.

Chapter~\ref{ch:results} presents the results of various applications of
the algorithms to large bodies of text and speech.  These tests explore
performance on tasks of segmentation, data compression, and lexical
induction.  Results are compared to other existing methods.

Chapter~\ref{ch:conclusion} summarizes the thesis and discusses possible
future work.

\chapter{The Problem of Language Acquisition}\label{ch:la}

At its most abstract, language acquisition is simply a mapping from some
input, consisting of speech and perhaps other evidence from the learning
environment, to ``knowledge of language''-- a grammar that can be used in
the generation and interpretation of new utterances.  An understanding of
language acquisition must therefore be founded on an understanding of the
nature of the input, the form and interpretation of grammars, and the
mapping itself.  These can each be understood at different levels.  For
example, Marr~\cite{Marr82} distinguishes between the broad goals of a
computation, the particular representations and algorithms employed, and
their hardware implementation.  Given our limited understanding of
language, computation and cognition, a complete theory of language
acquisition at all three levels is presently beyond reach.  This thesis
seeks formulate a computational theory that can be implemented using
specific algorithms and representations and tested on real input, by which
we mean evidence undeniably available to children.

This chapter serves as an introduction to problems and theories of language
acquisition.  It surveys the evidence available to learners and the
parameters\footnote{The word {\em parameter} here refers to any acquired
  piece of knowledge that contributes to language variation.  This
  definition extends the notion of a parameter as a characteristic constant
  (``the parameter that determines word ordering'') by also referring to
  such learned entities as words and rules.} that learners must acquire
from this evidence.  Switching attention, the chapter introduces several
conditions on theories of acquisition, in particular that theories be
testable and make as few unjustified assumptions as possible.  This leads
to a discussion of what assumptions can safely be made about the nature of
the input to the learning mechanism.  The chapter concludes by arguing two
important points: first, that language acquisition is best thought of as a
problem in unsupervised learning, where the goal is to identify structure
in the input that is not evident on its surface; and second, that the
logical starting point for a complete theory of language acquisition is a
theory of the acquisition of the phonological lexicon.

\section{An Introduction to Language Acquisition}

At its most abstract, language acquisition is the process of mapping from
environmental evidence-- spoken utterances and perhaps other clues-- to a
grammar that can be used to generate and interpret new utterances.
Therefore, language acquisition is best understood by understanding the
nature of the evidence, the form and interpretation of grammars, and the
mechanism that performs the mapping.  Here each of these are briefly
reviewed to provide a general background for further discussion.

\subsection{The Parameters}\label{ex-param}

Speakers express thoughts by causing rapid changes in air pressure.  The
production of this speech signal does not happen in one step but through a
complex derivational process~\cite{Levelt91} that involves many
intermediate representations, each generated in a manner that depends on
information the speaker has learned.  For example, in saying ``John caught
the weasels'' an English speaker relies on his knowledge that in English

\begin{itemize}
\item there is a proper name \word{John} and a noun \word{weasel} that
  refers to a kind of animal;
\item subjects are spoken before verbs, objects after verbs, and
  determiners before nouns;
\item tense is usually expressed through the main verb and plural nouns are
  marked with a suffix /\unis /;
\item proper names and ordinary noun phrases are not marked for case,
  contrasting with pronouns like \word{he} and \word{him};
\item \word{the} is unstressed and pronounced /\uniD\uniAX / but
  \word{catch} is stressed and (in a past tense sentence) pronounced
  /\unik\uniO\unit /;
\item the sound /\unil / can serve as the head of a syllable in
  \word{weasels} /\uniw\unii\uniz\unil\uniz / even though it is not a vowel;
\item the voicing (vocal cord vibration) in the /\unis /
  plural marker is determined by the voicing in the immediately preceding
  sound;
\item at the start of words stopped consonants like /\unik / are pronounced
  with a little puff of air;
\item declarative sentences are generally produced with a flat or decaying
  pitch.
\end{itemize}

\noindent These facts are peculiar to English and English speakers; they
have been learned.  Knowledge of language thus includes an acoustic
inventory; various motor skills; a lexicon that links phonological and
syntactic and semantic information; many phonological and morphological and
syntactic dictums; an understanding of conversational conventions; and
perhaps much more.  These parameters collectively constitute the grammar
that is the desired output of the language acquisition process, though of
course their exact form is open to debate.

From the standpoint of acquisition, grammars have several notable
properties.  One is that they contain a very large number of parameters,
many (like words) capable of seemingly infinite variety.  This implies that
the space of grammars can not be practically enumerated.  Parameters also
come in a great variety of forms; language seems to be built from different
modules that each require different types of information.  Despite this
fact, parameters in different modules are highly interdependent.  For
example, syntactic ordering rules have meaning only when combined with
part-of-speech tags found in the lexicon.  Furthermore, parameters interact
with the generation and interpretation mechanisms in such a way that there
are many parameter settings that could explain any piece of evidence.  For
example, \word{weasels} could be pronounced /\uniw\unii\uniz\unil\uniz /
because a root /\uniw\unii\uniz\unil/ combines with the plural marker and
the voicing rule, or it could simply be listed in the lexicon like
\word{caught}.  Finally, very few of the parameters relate directly to the
speech signal; almost all affect or link different hidden representations.

\subsection{The Evidence}

Children acquire their grammars principally from exposure to spoken
utterances, though it is widely conjectured that they also leverage
extralinguistic information derived from non-auditory senses like sight,
and expectations derived from their own internal state.  The difficulty of
language acquisition would seem to depend crucially on two things: first,
the amount of evidence available to the learner and second, the
transparency of the relationship between the input and the grammar that
produced it.  Even at the rate of ten utterances per minute for ten hours
each day, by the age of five a child can have heard no more than 11 million
utterances.  At this point most children have attained nearly all the
fluency and linguistic expertise of adults.  Though 11 million utterances
may seem like a lot, it is far less data than is commonly used to train
computer models of language~\cite{Brown92}, and allows for precious few
examples for each of the tens of thousands of words that must be learned.
However, the paucity of data is not nearly so troublesome for acquisition
as the opaque relation between the grammar and the input signal.

Much of the complexity of the relationship between the grammar of the
target language and the signal available to the learner is caused by
factors external to the language faculty.  Grammars are not the only source
of variation in the speech signal: language is a channel for the
transmission of information, and changing this information can have the
same effect on the speech signal as changing the grammar would.  Other
factors that confuse the relationship between the parameters and the signal
include background noise, starts and stops, coughs, other disfluencies,
ungrammatical structure and nonsense words.  An utterance may even reflect
incoherent thought or be from a different language.  Presumably, therefore,
a learner must be suspicious of all input and entertain the possibility
that it might not be useful evidence for the target language at all.

Even if speech signals could be taken at face value, they obscure the
parameters of the generating grammar quite effectively: without knowledge
of the generating grammar the derivational history of an utterance is
nearly invisible.  To take just one example, the phonological
representations that are the basis for speech production are rooted in
coarse articulatory gestures~\cite{Halle83b} like tongue movements that
have complex and sometimes subtle affects on the acoustic
signal~\cite{Rabiner93}. For this reason, it is extremely difficult to
determine the motor commands that produced a signal.  Even if they were
known, this would not uniquely determine the control sequence that caused
them, because in the process of speaking gestures are routinely (but not
necessarily predictably) omitted and otherwise corrupted in an attempt to
minimize muscular effort~\cite{Klatt92}.  Furthermore, unlike in the
English writing system, neither phonemes (primitive bundles of articulatory
gestures) nor words nor other units in speech are routinely separated by
delimiters.  The pause that is often supposed to exist between words is
usually a perceptual illusion apparent only to competent speakers: unknown
languages sound rapid and continuous.  Certainly word-internal boundaries
(such as between /\uniw\unii\uniz\unil / and /\uniz /) are almost never
highlighted, and there is little evidence that pause duration or other
information can be used to reliably segment higher structures like phrases.
The fact that the speech signal does not uniquely reflect phonological
representations and does not contain segmentation information means that
the naive learner can not determine the number or sounds of the words that
produced it, and therefore that no signal provides conclusive evidence for
any lexical parameters.  There are many additional ways that information
about the derivational process is lost.  For example, many phonological
processes destroy or hide information about underlying memorized
forms~\cite{Anderson88b}.  For these reasons and many more, the raw speech
signal offers few direct insights into the parameter settings of the
process that generated it.

Perhaps the best evidence that the speech signal provides relatively little
constraint on the derivational process (and hence the parameters that
control it) comes from the fact that {\em even if the generating grammar is
  known}, there are many possible interpretations for any given utterance.
Indeed, one of the principal components of any automatic speech recognition
device is a highly restrictive {\em language model} that attempts to filter
possible word sequences on the basis of language-specific usage
patterns~\cite{Rabiner93}.

This leaves open the possibility that parameter values can be easily
determined from extralinguistic input, such as the way a mother wiggles her
eyebrows, or (more plausibly) the manner in which she emphasizes different
parts of the speech signal.  This possibility is explored further in
section~\ref{input}; the conclusion there is that there is little evidence
such felicitous cues exist and even less that they are required for
learning.  Of course, it is clear that word meanings are not derived from
the speech signal alone, but it is doubtful that the evidence learners use
to acquire meaning also serves to determine low-level parameters.

\subsection{The Learning Process}

Very little is understood about the processes children employ to learn
language: researchers that have studied child language acquisition have
concentrated their efforts on characterizing children's knowledge at
various stages of life.  Roughly speaking, phonological distinctions,
syllable structure and other information concerning sound patterns are
learned early~\cite{Jusczyk93,Jusczyk94}, followed later by words and
syntax~\cite{Goodluck91,Pinker94}.  But such facts shed little insight into
the character of the process that maps evidence to grammar.  For this
reason, theorists have traditionally argued for or against hypothesized
learning mechanisms on the basis of how they accord with abstract
properties of the language learning problem (such as the seeming ease with
which children acquire language).  These properties are defined by the
nature of the input, the form of parameters and the mechanisms that
interpret them, and the purpose of language.

\subsection{Summary}

The preceding sections have shown that language acquisition is characterized
by the following important facts:

\begin{itemize}
\item there is relatively little evidence available to the learner, at
  least compared to the demands of existing computational models;
\item the learner chooses a grammar from among a high-dimensional parameter
  space, spanning many different types of parameters;
\item parameters are highly interdependent;
\item the relationship between parameters and observables is complicated
  and non-transparent;
\item the evidence available to the learner can be explained by many
  different parameter settings.
\end{itemize}

\noindent Thus, language acquisition has all the hallmarks of an extremely
difficult learning problem.  But these facts do not entirely specify the
task of researchers who seek to build theories of language acquisition;
this is the topic of the next section.

\section{Theories of Language Acquisition}\label{tla}

The purpose of a theory of language acquisition is to explain how a learner
can map from utterances (and perhaps other evidence available in the
learning environment) to a grammar that can be used in the generation and
interpretation of new utterances.  Theories can be evaluated on any of a
number of bases.  An engineer may be interested in theories that provide an
explicit algorithm for learning useful parameters from readily available
input.  A psychologist may be interested in theories that predict
acquisition in the same manner as children, perhaps even going so far as to
require a description at the level of neural anatomy.  Or an evolutionary
biologist might wish for an abstract characterization that makes plain what
classes of language are learnable by {\em any} mechanism.  Naturally, the
ideal situation would be to understand language acquisition at all levels
from neural implementation to computational theory~\cite{Marr82}.  As a
practical matter such an understanding is beyond current reach.

To understand the goals of this thesis, it is necessary to define
``knowledge of language'' more precisely.  In one sense, languages are sets
of sentences, or alternatively mappings between sound and meaning; this is
the traditional view of the structuralist and descriptivist schools (see,
for example, Bloomfield~\cite{Bloomfield33} and Lewis~\cite{Lewis75}).
Chomsky~\cite{Chomsky86} uses the term {\em E-language} (externalized
language) to refer to this notion.  Viewing languages this way, learning
language means to acquire knowledge sufficient to generate and interpret
new utterances in the same manner as the rest of the speech community.
This suggests that learning mechanisms should be judged by the
generalization performance of the grammars they produce; in fact, this
criteria has historically been the driving force behind theories of
language acquisition.  However, knowledge of language is also a property of
individuals.  Each speaker has internalized some particular knowledge in
some particular representation, and it is this knowledge that allows them
to generate and interpret languages.  Chomsky uses the term {\em
  I-language} (internalized language) to refer to ``the element of the mind
of the person who knows the language''.  To a scientist interested in
characterizing human cognitive processes, a theory of language learning
must also be judged on the basis of whether it produces the same internal
characterization of language that a child would attain in the same
circumstances.

This thesis seeks to formulate a theory of language acquisition that is
consistent with both views of language.  In other words, the theory (as
represented by a learning mechanism) will be evaluated both on the basis of
whether it produces grammars consistent with the E-language the learner is
exposed to {\em and} on the basis of whether it produces grammars that have
qualitatively similar internal representations to the grammars children
would produce in the same circumstances.

The remainder of this section argues several points relating to the
formulation of theories of language acquisition.  The first is that a
primary goal must be to produce theories that can be tested with only a
minimal number of additional assumptions.  The second is that, at the
present time, it is relatively unimportant that learning theories explain
the detailed manner in which children acquire language.  Finally, it is
argued that although the learning mechanism is at the heart of any theory
of acquisition, it must be justified in terms of general principles.  This
is essentially a statement that any theory at the level of algorithms and
representations must be related back to a more abstract description at the
level of computational theory.

\subsection{Testability and Theories of Acquisition}

In chapters~\ref{ch:representation} and \ref{ch:algorithm} a theory of
language acquisition is presented, formulated principally at the level of
representations and algorithms.  The justification for this level of
abstraction is that at this level theories are both sufficiently abstract
to shed insight into the general nature of the learning problem, and
sufficiently concrete to be testable.  There are at least six reasons to
concentrate effort on theories that can be evaluated with few additional
assumptions, and in particular, tested on real data.

\begin{itemize}
\item Any theory that can be tested on real data can be falsified or
  verified in a far more convincing way than a theory that is either
  phrased in vague terms, or that is removed from data by additional
  assumptions; it therefore has greater content.

\item Such theories, if verified, are existence proofs, demonstrating
  conclusively that certain parameters can be learned.  In this way they
  can form the foundations of further research that is predicated on language
  learnability, justifying certain assumptions.

\item As an existence proof, a tested theory also proves that it is not
  {\em necessary} to make assumptions beyond those that are in the theory.
  As discussed further in section~\ref{input}, many have assumed (without
  conclusive evidence) that the input children receive is quite rich; such
  input permits quite simple learning methods.  If it can be demonstrated
  that rich evidence is not necessary for learning, then theories that
  assume it are put under the additional onus of having to both demonstrate
  its existence {\em and} the fact that children rely on it.
  
\item In the course of applying algorithms and representations to real
  input, incorrect and implicit assumptions in abstract theories can be
  identified.  For example, without testing on real data it may not be
  apparent that a particular grammatical representation, while sufficient
  to model real language, cannot be correct because under it no plausible
  learning algorithm can identify a consistent grammar from unstructured
  evidence (see section~\ref{learning}).  In a similar vein, Ristad, Barton
  and Berwick~\cite{Barton87b,Ristad94} have argued that many theories can
  be dismissed on the basis of their computational complexity.  Such
  deficiencies usually become apparent immediately upon implementation.

\item In the course of applying algorithms and representations to real
  input, the most significant ``problems'' of language learning are
  identified.  This is not necessarily the case with more abstract theories
  of language.  For example, as discussed further in section~\ref{input},
  many abstract theories have concerned themselves with the issue of
  whether grammars can be uniquely identified on the basis of positive
  evidence.  But with the sort of grammatical theories that are necessary
  to explain real data, it quickly becomes clear that the answer is no.
  This suggests (see section~\ref{underspecified}) that the more important
  issue in language learning is how to select the correct grammar from
  among the set that are consistent with the input.

\item Since any learning theory that can be tested on real data necessarily
  includes an explicit, computationally feasible learning algorithm, it
  simultaneously serves as a solution to engineering problems involving
  the acquisition of human language.

\end{itemize}

\subsection{Conditions on Theories of Acquisition}\label{conditions}

In requiring that they be testable, various conditions have been placed on
theories of acquisition.  In particular, a theory must be {\em feasible}
(the learning mechanism embodied in it must make reasonable use of
computational resources and demand no more from the learning environment
than what is available), {\em complete} (the parameters, learning
mechanism, and form of the input must each be specified in sufficient detail
to be implemented and simulated) and {\em independent} (the theory must not
rely on the presence of other unattested or undemonstrated mechanisms to
preprocess evidence or otherwise aid the learning mechanism).

One condition not listed above is that a theory should predict learning in
the same {\em manner} as human beings~\cite{MacWhinney78}.  This is omitted
for several reasons.  First, in any scientific endeavor some
simplifications must be made and relaxing the manner condition is unlikely
to alter the fundamental character of the learning problem.  Secondly, it
is important to understand how language {\em can} be learned, irrespective
of mechanism.  For example, it is a goal of the engineering community to
create computer programs that mimic the end-to-end linguistic performance
of humans, though there is no desire for a neural implementation.  Even
within the realm of science, it is interesting to ask what the range of
possible learning mechanisms for language is.  Important questions include
``how much of language can be learned from sound alone?'' and ``to what
extent is the nature of language determined by the learning mechanism?''.
Finally, there is sufficiently little evidence for how children learn that
it is not clear a manner condition can be usefully and fairly applied.

These three conditions are quite restrictive; in particular, the
completeness and independence conditions leave little room for theories
that advance our understanding of acquisition without completely solving
the learning problem.  It could argued that by instilling these conditions,
scientific progress will be stifled, because they cannot be met at the
present time.  For example, researchers are almost totally ignorant of the
mechanisms that process extralinguistic information in the learning
environment and provide the child with the representations of meaning that
must eventually be associated with sound.  Plainly some artificial
substitute for these mechanisms must be used to test any current theory of
acquisition.  This is unavoidable, but it does not alter the fact that a
more desirable theory would dispense with the artificial input (and all the
assumptions associated with it) and work directly from attested evidence.
Regardless of whether the conditions can be met, they must be active goals.

\subsection{Assumptions and Modularity in Theories of Acquisition}\label{modularity}

No existing theory of language acquisition meets the above conditions.
Many assume grammatical and noiseless input.  Some assume the learner has
access to unlikely representations of sentence meaning
(section~\ref{input}) or similarly untestified segmentations of the speech
signal.  Most are based on linguistic theories that can account for only
small subsets of real utterances.  Almost all restrict the learning problem
to a small subset of linguistic parameters, assuming input neatly
preprocessed to eliminate all other aspects of acquisition (see below).
Some relax all computational constraints on the learning mechanism
(section~\ref{learning}).  And finally, many theories are so vague and
incomplete as to be entirely unimplementable.  Of course, some of these
violations are less detracting than others: a vague theory may be
contentless and a theory that assumes too much of the input may be
irrelevant, but a theory that adequately explains the acquisition of a
small part of language represents considerable progress, if it makes
plausible assumptions about the remainder of the acquisition process.  The
remainder of this section explores this issue in more detail.

Theories of language processing generally divide the language faculty into
various weakly interacting modules, such as acoustic processing, phonetics,
phonology, morphology, syntax and semantics.  The acquisition literature
reflects this split: most (reasonably well specified) theories of language
acquisition restrict their scope to the parameters of particular modules.
As a scientific practice this is not without risk, because the modules
themselves may be merely artifacts of current linguistic theory, and
because the boundaries between the theorized modules are unobservable and
hence uncertain.  There are two undesirable but common consequences of this:

\begin{itemize}
\item An acquisition theory for one part of language may make implausible
  demands of its evidence, such as requiring noiseless input, input in a
  linguistically implausible form, or input that cannot be computed without
  communication between modules.  Examples include theories of
  morphological acquisition that expect segmented, noiseless phoneme
  sequences as input and theories of syntactic acquisition that assume side
  semantic information is tree structured in a manner very similar to
  that of syntax.

\item An acquisition theory for one part of language may unreasonably
  assume that the parameters of other parts can be learned independently.
  Examples include theories of the acquisition of phonological rules that
  presume the underlying forms of words are already known (even though the
  underlying forms of words are difficult to derive without knowledge of
  phonological rules), and theories of the acquisition of syntax that
  assume word parts-of-speech are known (even though the principal source
  of information about word parts-of-speech is syntax).
\end{itemize}

\begin{figure}
\pageline

\begin{center}
\begin{tabular}{||lr|ccc||}\hline
\multicolumn{2}{||c|}{Paper} & Assumes & Input & Output \\ \hline\hline
Anderson 1977 & \cite{Anderson77} & I,NN,NH & MI,WM,SM & G,WS \\
Anderson 1981 & \cite{Anderson81} & I,NN,NH & MI,WM,SM & G,WS,MPHR \\
Berwick 1985 & \cite{Berwick85} & I,NN,OCWS & WI,WM,TR & G,WS \\
Brent 1993 & \cite{Brent93} & & WW & MPH,MPHR \\
Gibson \& Wexler 1994 & \cite{Gibson94} & NN & P,TR & G \\
Kazman 1994 &  \cite{Kazman94} & I,NH & WI,WM,WS & G,MPH,MPHR \\
Rayner {\it et al.} 1988 & \cite{Rayner88} & I,FG,NN,NH & WI & WS \\
Selfridge 1981 & \cite{Selfridge81} & I,NN,NH& WI,SM & WM \\
Siklossy 1972 & \cite{Siklossy72} & I,NN,NH & WI,SM & WM \\
Siskind 1992 & \cite{Siskind92} & I,NN,NH & WI,SM & G,WS,WM \\
Siskind 1994 & \cite{Siskind94} & I & WI,SM & WM \\ \hline
\end{tabular}

\vspace{.3in}
\begin{tabular}{|ll|}\hline
\multicolumn{2}{|c|}{Assumptions}\\ \hline
FG & \parbox[t]{4in}{Grammar fixed in program.}\\
NN & \parbox[t]{4in}{No noise or inconsistent input.}\\
NH & \parbox[t]{4in}{No homonymy: each identifier has a single interpretation.}\\
OCWS & \parbox[t]{4in}{Syntactic roles of open class words are known.} \\
I & \parbox[t]{4in}{Identity: words or morphemes are given unique identifiers.}\\
\hline
\hline
\multicolumn{2}{|c|}{Inputs}\\ \hline
WW & \parbox[t]{4in}{Sequence of separated written words.}\\
WI & \parbox[t]{4in}{Sequence of word identifiers.}\\
MI & \parbox[t]{4in}{Sequence of morpheme identifiers.}\\
P & \parbox[t]{4in}{Sequence of parts-of-speech.}\\
SM & \parbox[t]{4in}{Meaning of sentence as a whole.}\\
WM & \parbox[t]{4in}{Meaning of each word in sentence.}\\
WS & \parbox[t]{4in}{Syntactic role of each word in sentence.}\\
TR & \parbox[t]{4in}{Thematic roles (weaker form of sentence meanings).}\\
\hline
\hline
\multicolumn{2}{|c|}{Outputs}\\ \hline
G & \parbox[t]{4in}{Syntactic parameters/grammar.}\\
MPH & \parbox[t]{4in}{List of morphemes in lexicon.}\\
MPHR & \parbox[t]{4in}{Rules that constrain occurrence of morphemes.}\\
WM & \parbox[t]{4in}{Meaning of each word in lexicon.}\\
WS & \parbox[t]{4in}{Syntactic role of each word.}\\
\hline\end{tabular}

\end{center}

\caption{\label{fig:assumptions} Some notable papers on the machine
  acquisition of morphology, syntax, and the lexicon, cataloged by their
  assumptions and input-output behavior.}

\pageline
\end{figure}

Many theories fall into these traps: figure~\ref{fig:assumptions} catalogs
a selection of computational theories of language acquisition and their
input-output behavior.  Various assumptions are common: no ungrammatical
input, no input from languages other than the target languages, no
homonymy, etc.  These assumptions violate what we know about the real
environment children learn in.  Most theories also demand the extraordinary
from other parts of language: the existence of a remarkable preprocessor
that maps from acoustic signals to noiseless token sequences; access to a
similarly unerring module that extracts semantic structure from the
learning environment; a means of segmenting and uniquely identifying words
in the input; and so forth.  These requirements are far beyond the
capabilities of any known mechanisms.  Finally, all of these theories
assume that other modules can function without feedback and can be learned
independently.

It is of course not possible to construct a complete theory of language or
language acquisition in one step.  But the safest starting points are the
ones that require the fewest assumptions, and hence the ones nearest to
attested evidence.  This suggests that most effort should be devoted to
explaining how the most primitive parameters are learned; these might
include sound classes, constraints on syllable structure, and other
parameters close to the speech signal.  Of course, if it can be reasonably
argued (or demonstrated) that some parameters are not strictly necessary
for the acquisition of others, then their study can be reasonably deferred.
For example, it is possible that the phonological form of words can be
learned even without an understanding of syllable structure.

\subsection{Specification of The Learning Mechanism}\label{learning}

As has been mentioned, there are three principal components to any theory
of acquisition: the evidence, the parameters, and the learning mechanism.
The evidence is essentially fixed by what is available to children (though
what this evidence is is not entirely understood).  The parameters are
theory-internal, but are defined by the processes that interpret and
generate utterances, and these can be investigated independently of
acquisition.  Therefore theories of acquisition have relatively little
freedom to select the range and form of the parameters that must be
learned.  This would seem to imply that a theory of acquisition boils down
to a specification of a learning mechanism.  But if a theory emphasizes the
role of the learning mechanism, then it is under an increased obligation to
justify its function in terms of general principles.  For this reason, it
is unsatisfying to assume a baroque mechanism.

To understand the importance of the learning mechanism, it is worth
introducing a simple one (discussed in more length in the following
section).  Imagine an algorithm that enumerates grammars in some
predetermined order and stops at the first one that is consistent with the
evidence, under some simple definition of consistency.  Given the number of
possible grammars and the possibility of noise in the input, it is clear
that this algorithm is merely a theoretical tool; it cannot possibly be
computationally feasible or reliable.  These issues cannot be lightly
dismissed on the grounds that the algorithm is merely being described at
the level of a computational theory and abstracts from various details
necessary to handle real-life situations.  Efficiency, convergence,
robustness and other properties of learning mechanisms all indirectly bear
on other parts of the learning framework.  For example, there is
significant evidence that the known induction algorithms for certain
classes of grammars (such as stochastic context-free
grammars~\cite{Carroll92,deMarcken95b,Pereira92}) are systematically
incapable of learning linguistically relevant languages; this reflects back
on the appropriateness of the grammar class as a model of human language.
Hence, a complete theory of language acquisition, even at the abstract
level of computational theory, must be explicit about the details of the
learning mechanism.

Unfortunately, there are good reasons not to overly burden the learning
mechanism.  Complex learning algorithms are notoriously difficult to
analyze and make categorical statements about.  In most cases, the only
means of evaluating them is to simulate their execution.  Thinking in terms
of general principles provides greater insight into the language learning
process as a whole.  It is for similar reasons that optimization
researchers think in terms of an objective function, even though their
algorithms may only consider its derivative when searching.  An example
serves to illustrate the problematic nature of complex learning algorithms.
Dresher and Kaye~\cite{Dresher90}, arguing that brute-force enumeration
strategies are unsuitable models of human language acquisition, propose a
{\em cue-based} learning algorithm for the parameters of a metrical stress
system.  In cue-based strategies, the learner is aware of the relationship
between various sentences and parameter values.  Thus, in Dresher and
Kaye's model evidence of a certain stress pattern might trigger the
resetting of a parameter from its default value to a marked one.  They
describe cues appropriate for their simple parameter system and argue that
the cues are sufficient for learning.  Unfortunately, the cues are not so
simple as to be easily derivable from the parameter system, and thus must
be a hardwired part of the learning algorithm, selected presumably by
evolution.  Little can be said about the nature of the cues without
reference to the details of the parameter system; for any change in the
model of stress the feasibility of a cue-based strategy must be
re-justified.  In contrast, Gibson and Wexler's~\cite{Gibson94} simpler
``TLA'' parameter-setting algorithm is easily analyzed~\cite{Niyogi94},
though its success is similarly dependent on the structure of the parameter
system.

\section{The Nature of the Input}\label{input}

In section~\ref{modularity} it was argued that theories of language
acquisition should be built up from the evidence that is available to the
learner.  This forces us to examine in more detail the nature of the input.
There are two important questions.  The first is whether the learner has
access to feedback and evidence for what utterances are {\em not} in the
target language.  The second is the extent to which extralinguistic input
serves to directly transmit parameter values.  These are both discussed
here in the context of one particular framework for theories of
acquisition.

Chomsky writes~\cite{Chomsky61,Chomsky65} that any theory of language must
provide

\begin{itemize}
\item (i) an enumeration of the class $s_1, s_2,\ldots$ of possible
  sentences;
\item (ii) an enumeration of the class $SD_1, SD_2, \ldots$ of possible
  structural descriptions;
\item (iii) an enumeration of the class $G_1, G_2, \ldots$ of possible
  generative grammars;
\item (iv) specification of a function $f$ such that $SD_{f(i,j)}$ is the
  structural description assigned to sentence $s_i$ by grammar $G_j$;
\item (v) specification of a function $m$ such that $m(i)$ is an integer
  associated with the grammar $G_i$.
\end{itemize}

\noindent In this abstraction, a {\em language} is a set of {\em
  sentences}.\footnote{Here, the word language is used in the E-language
  sense (see section~\ref{tla}).  In more recent work Chomsky has treated
  learning as a problem of learning an I-language.}  Presumably these
sentences represent some slight abstraction of the acoustic stream, though
Chomsky is not specific about this.  A grammar is a set of parameters for a
process that generates sentences; thus, a grammar $G$ defines a language
$L(G)$, the set of all sentences that can be generated under the parameter
setting $G$.  By {\em structural description} Chomsky is collectively
referring to information that reflects the derivation of a sentence under a
grammar, such as sentence meaning and syntactic structure.  This ``side
information'' might be extractable by the learner from the learning
environment and used to disambiguate between grammars, by means of the
function $f$.  The function $m$ is a preference function over grammars,
reflecting some arbitrary criterion such as simplicity.

Chomsky imagines the following learning strategy: a teacher with target
grammar $G$ presents a set of sentences drawn from $L(G)$ to a child, along
with their structural descriptions under $G$; the child enumerates grammars
in order of their image under $m$, and selects the first grammar consistent
with the input.  Thus, the child's grammar is a complex function of the
input and the class of grammars available to the child.  Having learned a
grammar, the child can use it to determine whether a sentence is in her
language, and if it is, assign it a structural description.

In this framework Chomsky is implicitly assuming that learning takes place
from {\em positive examples}-- sample sentences from the target language.
This is consistent with Brown and Hanlon's~\cite{Brown70} assessment (see
also Marcus~\cite{Marcus93}) that children receive no {\em negative
  evidence}, a term that refers to both feedback from the teacher to the
learner and {\em negative examples}-- sentences labeled as outside of the
target language.  But this assumption introduces an apparent paradox, since
it can be shown in Chomsky's framework that under reasonable definitions of
learnability, most classes of formal languages that are similar to human
languages are not learnable from positive examples alone.  Restricting
attention to the input, one way out of this paradox is to assume the
learner has access to side information, such as ``meaning'', culled from
the extralinguistic environment or derived independently from the speech
stream.  This is consistent with what is known, but from a scientific
standpoint it is important to explore the possibility that such side
information plays a limited role in the learning process.

\subsection{Positive and Negative Examples and Restricted Language Classes}
\label{positive}

Gold~\cite{Gold67} presents a framework for the study of the induction of
formal languages that is very similar to Chomsky's, but allows for negative
examples.  There it is assumed that examples (labeled {\em positive} or
{\em negative}) are presented to the learner in a felicitous sequence, such
that all possible examples are eventually presented.  After each example
the learner names a language.  If there is a learning strategy that
guarantees that for any target language, the learner will eventually name
the target language and never again change its hypothesis, then the class
of languages the learner is choosing among is {\em identifiable in the
  limit}.  It is possible to place strong bounds on what classes of
languages are identifiable in the limit from positive examples
alone~\cite{Angluin80}, even assuming a preference ordering on languages
like Chomsky's $m$ function.  Gold proved that many linguistically relevant
classes of languages, such as the regular and context-free languages, are
identifiable from both positive and negative examples but not from positive
examples alone.

It is not surprising that powerful classes of languages are not
identifiable from positive examples alone.  Any learning algorithm that
guesses a language that is a superset of the target will never receive
correcting evidence; this is especially relevant when the possibility of
noise (input outside of the target language) is taken into account.  But
more fundamentally, for powerful classes of languages there are simply too
many languages consistent with any set of data.  Nevertheless, many
restricted classes of languages can be identified from positive data alone.
This is the case, for instance, if every language contains a sentence that
is unique to that language.  Some have proposed that the class of grammars
that children consider is highly restricted, with particular properties
that render it identifiable (see Berwick~\cite{Berwick85} and Wexler and
Culicover~\cite{Wexler80} for discussion).  This possibility has generally
been raised in the context of syntax.  Regardless of whether it holds,
other parts of language, such as the lexicon, are not so limited.  For this
reason, it is difficult to construct linguistically plausible classes of
grammars that are unambiguous with respect to natural input.  As an
example, most sentences are logically decomposable into words, but there
also exist idiomatic phrases that must be memorized.  Given the two
possibilities, it seems that a child could account for any sentence
as either following from parts or being a lengthy idiom.  To rule out
the second possibility while still permitting rote-memorized passages is
difficult, and leads to baroque and unwieldy theories of language.  Any
natural class of grammars must allow for both possibilities, and hence
arbitrary ambiguity.

The fact that most powerful classes of formal languages are not
identifiable in the limit from positive examples still leaves a variety of
possible outs for human language acquisition.  One is that the child has
access to a generous source of negative examples.  Many have contested
Brown and Hanlon (see Sokolov and Snow~\cite{Sokolov94} for review), and
suggested that in fact implicit and explicit negative evidence does appear
in the input children receive.  Unfortunately, evidence for significant
amounts of feedback is tenuous (it is not clear how much is present, or of
what sort) and there is little evidence that children rely on it; some
cultures do not even direct speech at pre-linguistic infants
\cite{Lieven94}.  For this reason, although it is {\em possible} that
children make use of negative evidence, it appears more promising to look
for alternative explanations of learnability.

\subsection{Side Information}\label{meaning}

Chomsky allows that the learner may have access to structural descriptions
as well as sentences.  More generally, it is possible that side information
extracted from beyond the speech stream or derived independently from the
speech stream could be used to disambiguate between grammars, if the side
information reflects properties of the derivation of input sentences under
the target grammar.  For example, Gleitman~\cite{Gleitman90} suggests that
syntactic parse trees can be reconstructed from prosodic information alone.
Perhaps more plausibly, the actions taking place around a child may suggest
various possible ``meanings'' for the sentences the child is hearing.  This
in turn could provide the child with information about the words in the
sentences it is hearing, as well as the manner in which the words are
composed.

Providing the learner with linguistically structured information like
syntactic trees or semantic formulae can trivialize the learning process,
by making the grammar explicit in the input.  Some recent papers argue that
there are powerful classes of languages identifiable from positive data
alone~\cite{Kanazawa94,Sakakibara92,Shinohara90}; these learnability proofs
assume access to structural descriptions in the input.  Sakakibara
\cite{Sakakibara92}, for example, has shown that a significant subset of
context-free languages (those generated by {\em reversible} context-free
grammars) are identifiable from positive data, if the example sentences
come structured into unlabeled derivation trees.  Similarly, various
algorithms have been constructed that use artificial semantic
representations to aid the acquisition of syntax~\cite{Siskind92,Siskind94}
and the phonological lexicon~\cite{deMarcken94b}.  Indeed, much work on
syntactic acquisition has assumed that the thematic roles of noun phrases
are known to the learner~\cite{Gibson94}.  Finally, it has been shown that
children do not learn much, if anything, from sound patterns in isolation
\cite{Sachs72,Snow76}; some environmental clues are probably necessary for
learning.

Despite the fact that it provides an easy way around troublesome learning
problems, there are a number of arguments against {\em relying} on side
information to explain language learnability:

\begin{itemize}

\item There is only shaky evidence as to what side information is available
  to children, and no conclusive evidence that children make use of it in
  learning (other than the uncontested fact that meaning is not learned from
  sound alone).

\item It may be that significant learning needs to have taken place before
  side information becomes useful.  For example, it seems unlikely that
  children pair sounds to extralinguistic events before they are capable of
  at least rudimentary segmentation of the sound stream.

\item It is not clear how much extralinguistic information is {\em
    necessary} for learning language.  Thus, there is a substantial risk
  that we will incorrectly attribute all that we do not understand to magic
  in extralinguistic processing mechanisms.
  
\item The use of side information as an aid to language learning falls out
  naturally in some learning frameworks, and need not receive a special
  role in the learning model.  See section~\ref{rep:meaning} for further
  discussion.

\item There are many engineering tasks that demand learning about language
  from speech or text alone, such as the automated construction of
  automatic speech recognition systems.

\end{itemize}

To summarize, it is possible and even likely that children use other
information for learning than just the teacher's speech signal.  Even in
the speech stream, it is quite possible that occasional clues like pause
duration, accent and stress are used by the child in addition to the
sentence-like properties of the signal.  However, given that we do not know
the extent that children rely on such information, it is important to make
as few assumptions as possible and to determine lower bounds on the amount
of side information that is necessary for learning language.

\section{Conclusions}

This chapter has surveyed the problem of language acquisition, describing
the evidence available to the learner and the obligations of the learning
mechanism.  In doing so, it has promoted certain conditions on theories of
acquisition, in particular testability.  Two statements that have been made
need reemphasis, as they motivate the focus of the remainder of this
document.  The first, from section~\ref{modularity}, is that a theory of
acquisition should be built up from the evidence available to the learner,
because this guards against unjustified (and quite possibly incorrect)
assumptions.  The second, from section~\ref{input}, is that the only
evidence that is {\em known} to be available to the learning mechanism, at
least during early stages of acquisition, is the speech signal.  As
discussed below, these two facts determine the most natural starting point
for a theory of acquisition (the phonological lexicon) and the fundamental
challenge to acquisition (the {\em unsupervised} nature of the problem).

\subsection{The Phonological Lexicon}

The acquisition of the phonological lexicon is a natural starting point for
a complete theory of acquisition.  This is the problem of mapping from
continuous speech to a discrete lexicon of phonological representations,
perhaps for English including words like /\uniD\uniAX / (\word{the}) and
/\unik\uniO\unit / (\word{caught}) and morphemes like /\uniI\uniG /
(\word{-ing}).  A theory of this process must predict the acquisition of
parameters that enable a new speech signal to be segmented into a sequence
of these representations.  There are several justifications for the
primacy of this task:

\begin{itemize}
\item The lexicon is close to the speech signal, so it seems likely that
  theories of lexical acquisition would rely on fewer assumptions about the
  nature of the input, and data is readily available for testing purposes.

\item The phonological lexicon is a natural foundation for other
  acquisition processes.  All the work summarized in
  figure~\ref{fig:assumptions} assumes the existence of a mechanism that
  can map from an acoustic signal to a sequence of morpheme identifiers, in
  particular identifiers that can be used as attachment points for
  syntactic and semantic information.

\item Given that the acquisition of syntax and semantics is likely to be
  dependent on at least a rudimentary understanding of lexical parameters,
  it seems probable that at least the early stages of phonological
  acquisition occur without reference to extralinguistic
  information,\footnote{Some have argued that the acquisition of the
    phonological lexicon {\em is} dependent on knowledge of stress and
    intonational patterns~\cite{Cutler94,Jusczyk93,Jusczyk94}.} and
  consequently fewer potentially incorrect assumptions have to be made
  about extralinguistic processing mechanisms.
  
\item Even if assumptions must be made about the nature of acoustic
  processing, the use of (unsegmented) written text as a substitute input
  does not alter many of the fundamental aspects of the learning problem.

\item Although humans' phonological lexicons are not directly observable,
  the plausibility of a learned lexicon can be judged on the basis of its
  predictions about pause (or space) placement and whether there is a
  natural correspondence between parameters and what are considered roots
  and affixes in standard dictionaries.  Thus, theories of lexical
  acquisition can be objectively evaluated.

\item The lexicon accounts for a large portion of the total variability in
  language.  Therefore any viable theory of lexical acquisition is a
  significant contribution to a complete theory of language acquisition.

\item Very few theories have been proposed that attempt to explain the
  acquisition of the lexicon from speech-like input; it is a fundamental
  topic that remains mostly unexplored.
\end{itemize}

These facts motivate the emphasis of chapters~\ref{ch:representation} and
\ref{ch:algorithm}, which formulate representations and algorithms for the
induction of the phonological lexicon.

\subsection{Underdetermined Parameters and Unsupervised Learning}\label{underspecified}

The fact that language (in the E-language sense) is a mapping between sound
and meaning would seem to imply that the learning problem is fundamentally
one of choosing the grammar that best reproduces the mapping of the target
language.  In such a case the actual parameter values that are hypothesized
by the learning mechanism are of little concern; only their collective
performance matters.  Unfortunately, the principal challenge to theories of
acquisition is that the choice of parameter values is extremely important,
but underdetermined by the evidence available to the learner.

There are two reasons why the choice of parameter values is a fundamental
issue.  First, different speakers of the same language generalize
consistently, which is explained only if they have similar parameter
settings; this similarity is not predicted from the evidence available to
the learner, since this evidence varies and any finite sample is consistent
with many grammars.  Second, many underdetermined layers of representation
separate sound and meaning, so at least the early stages of learning must
be performed on the basis of the speech signal alone, which has been argued
to contain few explicit clues about the source grammar.  These stages must
therefore produce parameter values that are consistent with the mapping
even though they have no access to it.

In arguing that the choice of parameters values is important, and that
language is learned from signals that provide few explicit clues about the
source grammar, we are concluding that language acquisition involves
unsupervised learning.  The term {\em unsupervised learning} is generally
applied to problems where the goal is to identify structure that is not
evident on the surface of the input.  In the case of language acquisition this
structure can be thought of as the parameters.  Scientists interested in
formulating a theory of child language acquisition are faced with a
doubly-difficult task.  Not only must they propose an unsupervised learning
mechanism that can acquire a grammar that accounts for the evidence and
generalizes to new sound-meaning pairs, but this mechanism must also
acquire the {\em same} I-language that a child would attain in the same
circumstances.  The nature of this I-language can be partially deduced by
experiments performed on adult speakers' generation and interpretation
mechanisms-- this has been the primary goal of modern linguistics.

The next chapter presents a particular framework for unsupervised learning,
and explains various conditions that must be met for learning mechanisms
based on the framework to acquire grammars that accord with human
performance.

\chapter{Stochastic Grammars, Model Selection and Language Acquisition}~\label{ch:Bayes}
\pagestyle{headings}
\markboth{CHAPTER 3.\ \ STOCHASTIC GRAMMARS, MODEL SELECTION AND LANGUAGE ACQ.}
{CHAPTER 3.\ \ STOCHASTIC GRAMMARS, MODEL SELECTION AND LANGUAGE ACQ.}

In the previous chapter it was shown that during language acquisition a
single grammar must be selected from a set of many that are consistent with
the input signal; the lack of any explicit evidence favoring one over
another is one of the fundamental reasons language acquisition is a
difficult problem.  Here it is shown that if grammars are given stochastic
interpretations, those grammars under which the input is typical can be
favored over those under which it is unusual.  This evaluation metric
favors linguistically plausible grammars, and can be justified by the
statistical estimation technique of {\em Bayesian inference}.  Although
Bayesian inference has a number of advantages over competing learning
frameworks, there are various subtleties involved in its application that
largely determine whether it will produce the correct target grammar.  The
most important of these are the manner in which stochastic interpretations
are tied to linguistic reality, and the manner in which generalization
takes place from a small amount of evidence to a grammar that explains unseen
data.  Discussions of these two topics form the bulk of this chapter.

In the Bayesian inference framework, the language learning problem can be
expressed as follows: through some process hidden to the learner a target
grammar $G$ is chosen from a class $\cal{G}$.  Various utterances $U =
u^1,u^2,\ldots,u^n$ are generated in a manner that depends on the target
grammar, and this evidence is presented to the learner, who must select a
single hypothesis grammar from among the possibilities, presumably the
one that was most likely to have generated the evidence.  If the learner
has access to two fundamental pieces of information, the {\em prior}
probability distribution $p(G)$ of the grammar $G$ being selected, and the
{\em conditional} probability distribution $p(U|G)$ of the evidence $U$
being generated given that the grammar $G$ was selected, then there is a
principled way for the learner to choose a hypothesis.  Bayes' formula, a
rewriting of the definition of conditional probability, is a mathematically
sound expression of the {\em posterior} probability of a grammar $G$ given
evidence $U$:

\begin{equation}\label{eq:Bayes}
  p(G|U) = \frac{p(U|G)p(G)}{p(U)}.
\end{equation}

\noindent The value $p(G|U)$ can be interpreted as the proper degree of
belief in a grammar $G$ after observing evidence $U$, given an initial
belief $p(G)$.  If at the conclusion of the presentation of evidence the
learner hypothesizes the grammar in which she has the highest belief, then
the hypothesis grammar $G$ is determined by

\begin{equation}\label{eq:bayes}
  G = \begin{array}[t]{c}\mbox{argmax}\\{\scriptstyle
    G'\in\cal{G}}\end{array} p(U|G')p(G').
\end{equation}

\noindent Equation~\ref{eq:bayes} includes most of the important components
of a formal theory of language acquisition.  The hypothesis class $\cal{G}$
is the class of all grammars the learner is capable of representing.  The
sequence $U$ is the data available to the learner.  The maximization over
$\cal{G}$ can be thought of as a search the learning mechanism performs for
the best grammar in $\cal{G}$ given the input $U$.  $p(G)$ is the learner's
default preference for certain grammars over others.  Finally, $p(U|G)$
captures the relation between grammars and evidence.  In a complete theory
of language acquisition, each of these components must be explicitly
defined.  For expository convenience we will generally assume that
utterances are produced relatively independently of one another, so that
the conditional probability $p(U|G)$ can be expressed in a factored form
$p(U|G) = \prod_{u\in U} p(u|G)$.

\section{Stochastic Language Models}\label{slms}

With respect to language acquisition, the principal advantage of the
Bayesian framework over those of Chomsky (section~\ref{input}) and Gold
(section~\ref{positive}) is that it evaluates grammars with respect to a
graded judgment of the {\em typicality} of the evidence.  A simple example
illustrates this.  Suppose a learner choosing over the class of finite
context-free grammars is given input $aba$, $abba$, $abbbba$, $abbbbba$.
Consider two grammars, both consistent with this evidence: $S\Rightarrow
aBa, B\Rightarrow Bb|b$ and $S\Rightarrow a|b|SS$.  Which is the prefered
one?  The intuitive answer is the first, because it explains better why the
observed evidence conforms to the pattern $ab^{+}a$.  This fact can be
captured naturally in the Bayesian framework, if grammars are given a
probabilistic interpretation.  In particular, compare the following two
{\em stochastic} context-free grammars (SCFGs~\cite{Baker79,Jelinek90}),
where the choice of nonterminal expansion is governed by probabilities:

\hspace{1in}
\begin{tabular}{clc}
\multicolumn{3}{c}{Grammar 1}\\ \hline
$S$ & $\Rightarrow aBa$ & (1) \\
$B$ & $\Rightarrow Bb$ & $(\frac{1}{2})$\\
& $\Rightarrow b$ & $(\frac{1}{2})$
\end{tabular}
\hfill
\begin{tabular}{clc}
\multicolumn{3}{c}{Grammar 2}\\ \hline
$S$ & $\Rightarrow SS$ & $(\frac{1}{2})$ \\
& $\Rightarrow a$ & $(\frac{1}{4})$\\
& $\Rightarrow b$ & $(\frac{1}{4})$
\end{tabular}
\hspace{1in}

\noindent
The probability of the sentence $aba$ under Grammar~1 is $\frac{1}{2}$.
Under Grammar~2 there are two possible derivations of the sentence, each
with probability $\frac{1}{256}$, for a combined probability of
$\frac{1}{128}$: {\em aba} is substantially more likely under Grammar~1.
The particular evidence $aba$, $abba$, $abbbba$, $abbbbba$ is of course
unlikely under both grammars, but it is much more probable under
the first one: $p(U|G_{1}) \gg p(U|G_{2})$.  So long as the prior
probabilities of the two grammars are comparable, equation~\ref{eq:Bayes}
gives us $p(G_{1}|U) \gg p(G_{2}|U)$, exactly in line with the intuition
that the first grammar is to be prefered.  In learning frameworks that do
not allow for such graded judgments of ``grammaticality'', heuristics (such
as the Subset Principle~\cite{Angluin80,Berwick85}) must be introduced to
favor Grammar~1 over Grammar~2.

Generative grammars with probabilistic interpretations (in other words,
grammars that implicitly or explicitly define $p(U|G)$) are commonly called
{\em stochastic language models}.  The discriminatory power of stochastic
language models comes at a steep price.  Unless probabilities are computed
arbitrarily, grammars must include extra parameters (such as the expansion
probabilities in the above example) that define the exact probability of
each utterance; the estimation of these extra parameters presumably
complicates the learning problem.  More fundamentally, stochastic language
models burden the grammar with the task of specifying the probability of
utterances, which is decidedly counterintuitive given that the source of
utterances lies outside of language altogether: the sentence \word{please
  remove this egret from my esophagus} is undoubtedly rare in English, but
not because of linguistic parameters; the frequency that it occurs is
principally determined by the circumstances of life.  This issue is one of
the reasons why many researchers have denied the appropriateness of
stochastic language models.  But the fact that the grammar is not the
principal cause of frequency variation does not mean that stochastic
extensions to traditional grammars cannot be valuable aids to learning.
In particular, because a stochastic grammar's ability to assign high
probability to evidence can be tied to the quality of the (non-stochastic)
fit of the grammar to that evidence, statistical measures such as
equation~\ref{eq:Bayes} can discriminate between multiple consistent
grammars without relying on extralinguistic evidence like utterance
meanings.  This is important in the early stages of learning when such
information may not be available to the learner (or the learner may not
know enough to make use of the information).

\subsection{Typicality and Linguistic Plausibility}

Under equation~\ref{eq:Bayes}, a stochastic English grammar that faithfully
approximates the distribution of English sentences should be a better model
for English input than a French grammar under which the input is highly
atypical.  In this way statistical properties of the grammar serve as an
alternative to extralinguistic evidence (that would be in conflict with the
French grammar for different reasons).  For equation~\ref{eq:Bayes} to be a
successful evaluation metric, however, statistical properties of language
models must mirror psychological reality: were a French stochastic grammar
to predict English-like output with high probability (maybe by predicting
frequent, pernicious misspellings) then the wrong grammar could be favored.
Thus, the important question is: given evidence $U$ produced from a
(non-stochastic, teacher's) grammar $G$, does the stochastic grammar that
maximizes the likelihood of $U$ have the same core (non-stochastic)
structure as $G$?\footnote{Note that as more and more extralinguistic
  evidence that constrains derivations becomes available to the learner the
  answer tends towards yes, because regardless of its stochastic nature a
  grammar with the wrong underlying structure will be inconsistent with the
  input.} The answer, discussed at length in de
Marcken~\cite{deMarcken95b}, depends crucially on the way that the
stochastic properties of language models are tied to linguistic structure.

A natural way to estimate stochastic parameters for a language model is to
find the parameters that maximize the likelihood of the observed evidence;
this puts each grammar in its best possible light with respect to
equation~\ref{eq:Bayes}.  Empirical
tests~\cite{Carroll92,deMarcken95b,Pereira92} using various naive classes
of stochastic grammars indicate that the stochastic grammars that maximize
the probability of linguistic evidence do not in general have
``linguistically plausible'' structure.  For example, although Grammar~3 is
a closer approximation of how sentences are generated in English, both of
the stochastic context-free grammars below perfectly account for the
distribution of evidence on the left:

\begin{center}
\begin{tabular}{c|c|c}
The Evidence & Grammar 3 & Grammar 4\\ \hline
\begin{tabular}{lc}
\word{Pron Verb} & $(\frac{1}{2})$ \\
\word{Pron Verb Noun} & $(\frac{1}{4})$\\
\word{Pron Verb Det Noun} & $(\frac{1}{4})$
\end{tabular}
&
\begin{tabular}{clc}
\word{S} & \word{$\Rightarrow$ Pron VP} & $(1)$ \\
\word{VP} & \word{$\Rightarrow$ Verb} & $(\frac{1}{2})$\\
& \word{$\Rightarrow$ Verb NP} & $(\frac{1}{2})$\\
\word{NP} & \word{$\Rightarrow$ Noun} & $(\frac{1}{2})$\\
& \word{$\Rightarrow$ Det Noun} & $(\frac{1}{2})$
\end{tabular}
&
\begin{tabular}{clc}
\word{S} & \word{$\Rightarrow$ Pron Verb} & $(\frac{1}{2})$ \\
& \word{$\Rightarrow$ Pron NP} & $(\frac{1}{2})$\\
\word{NP} & \word{$\Rightarrow$ VP Noun} & $(1)$\\
\word{VP} & \word{$\Rightarrow$ Verb} & $(\frac{1}{2})$\\
& \word{$\Rightarrow$ Verb Det} & $(\frac{1}{2})$
\end{tabular}
\end{tabular}
\end{center}

\noindent
These simple stochastic grammars, however, do not make significant use of
the mechanisms of language in their definition of the conditional
probability $p(U|G)$; for example, they do not take advantage of the
agreement relations that commonly exist between pairs of elements in a
common phrase.  In a more linguistically sophisticated class of stochastic
grammars, the agreement relation that exists between determiners and nouns
in English might be incorporated into Grammar~3.  This extra constraint
would enable a better statistical fit between the stochastic grammar and
English evidence.  For example, if the grammar contains the following
co-occurrence information on determiner-noun agreement

\begin{center}
\begin{tabular}{ccc}
\multicolumn{3}{c}{\word{NP $\Rightarrow$ Det Noun}} \\ \\
determiner type & noun type & probability\\ \hline
definite & singular & $(.47)$\\
indefinite & singular & $(.20)$\\
definite & plural & $(.32)$\\
indefinite & plural & $(.01)$
\end{tabular}
\end{center}

\noindent then it will assign higher probability to English evidence than
one that naively wastes probability on the
indefinite-determiner-plural-noun possibility.  Since under Grammar~4
determiners and nouns are not in the proper structural relation to be
constrained by agreement, the extra stochastic machinery would not aid that
grammar.  Of course, the Grammar~4 could use this sort of agreement model
to account for any statistical dependency between the verb and the
determiner, but given the way English is produced, there is no reason to
believe that a strong dependency exists there.  This is one example of how,
as stochastic models are tied to linguistic mechanisms, they increasingly
favor linguistically plausible grammars.

One could argue in this example that the stochastic agreement model is
merely playing the same role that a traditional, non-stochastic mechanism
would.  However this is a misinterpretation.  It is true that a mechanism
that merely ruled out the possibility of indefinite/plural pairs would
model English almost as effectively as the stochastic agreement model
(though noise and the occasional ungrammatical sentence might pose a
problem).  But the real issue is whether agreement would be learned at all
without the stochastic interpretation.  Since English evidence is
``grammatical'' whether or not an English grammar incorporates the
agreement restriction, there is no obvious incentive to acquire this
information (determiner-noun agreement is not a {\em necessary} component
of a grammar).  In contrast, in the Bayesian inference framework there {\em
  is} an incentive to understand agreement, because it enables the learner
to better predict the input $U$.  In fact, the statistical nature of the
learning problem gives the learner an incentive to acquire as much
knowledge of the target language as possible, since a stochastic grammar
that incorporates such knowledge is more likely to assign a high
probability to $U$.\footnote{This can be argued more formally by assuming
  that the utterances the language learner receives are produced
  independently, each in a manner that depends not only on the source
  grammar but also on other hidden information such as the teacher's
  thoughts.  Thus, as far as the learner is concerned, $U$ is produced
  piecemeal by a stochastic process with approximate distribution
  $p_T(U)=\prod_{u\in U} p_T(u)$.  (This is not to imply that the teacher
  necessarily uses a stochastic grammar; here the uncertainty in $p_T(u)$
  is principally due to the learner's ignorance of the input to the
  language mechanism.) If the learner's stochastic language model is also
  factored over individual utterances ($p_L(U)=\prod_{u\in U} p_L(u)$), then it
  can easily be shown that as the number of sample utterances grows,
  $p_L(U)$ is maximized when the learner's grammar is chosen to minimize
  the Kullback-Leibler distance $D(p_T\parallel p_L)$ between the
  distributions $p_T$ and $p_L$, where the Kullback-Leibler distance is
  defined by

\[ D(p_T \parallel p_L) = \sum_u - p_T(u) \log \frac{p_L(u)}{p_T(u)}.\]

\noindent It is possible~\cite{Berger96,DellaPietra95} (and indeed
effective) to construct stochastic language models by defining $p_L$ to be
the least-committal (maximum-entropy) distribution consistent with known
properties of the target language distribution $p_T$.  Using this class of
models, as more properties of the target language are incorporated into
$p_L$, the Kullback-Leibler distance between $p_T$ and $p_L$ decreases.  In
this sense, the grammar with the greatest chance of being selected by
equation~\ref{eq:bayes}, ignoring for now the prior term, is the one that
incorporates the most knowledge of the target language.}

All of these arguments rely on stochastic language models being defined in
such a way that their statistical modeling power is greatest when the
linguistic structure of the learner's grammar is naturally aligned with the
linguistic structure of the evidence.  In the above example, for instance,
the reason the linguistically plausible grammar is favored is because it
brings the stochastic agreement model to bear on a regularity (the
determiner-noun co-occurrence pattern), whereas under the linguistically
implausible grammar this mechanism is wasted.  Fortunately, language is not
entirely uniform, so stochastic models tailored for certain phenomena (say,
explaining morphological agreement) are unlikely to function well when
applied to other phenomena (explaining phonetic assimilation).  Thus, the
more finely tuned stochastic models are to their expected role, the more
likely Bayesian inference is to converge to desired grammars.  Of course,
if a regularity exists in the data but no statistical mechanism is built
into the class of language models to account for it, then there is a great
risk that some other (inappropriate) mechanism will be coopted to explain
it, confusing the estimation of whatever linguistic parameters that
mechanism was meant to be used for.  This is a very important practical
matter: language models that offer only a single mechanism to explain
statistical regularities (such as SCFGs) will necessarily end up using that
mechanism to account for all regularities.  The greatest risk is that
regularities that are not due to language but to the surrounding
environment that influences language will end up being modeled by
linguistic parameters; this is the subject of the next section.

\subsection{Linguistic and Extralinguistic Sources of Regularity}\label{extralinguistic}

In Bayesian inference, a stochastic grammar fares well if it assigns high
probability to evidence produced by the target grammar.  This is
accomplished by specifying a distribution that reproduces the {\em
  regularities} of the target language-- properties that are generally true
of signals produced by the target grammar but not of all possible signals.
Regularity in the input arises from two sources.  One is language; examples
of linguistic sources of regularity include words, agreement, syllable
structure, syntax, and in general any mechanism or parameter that reduces
the space of possible utterances in a language or favors some over others.
These are the regularities that the learner is interested in modeling,
since in doing so the learner will hopefully acquire the correct linguistic
parameters of the target language.  Unfortunately, there is another source
of regularity in the evidence available to the learner, and that is the
``control signal'' to language-- the outside world and all of the rest of
the teacher's brain.  This both complicates and simplifies the problem of
language acquisition.

Patterns in the input that are caused by mechanisms external to language,
but which appear similar to those imposed by language, can obviously
distract and mislead the learner.  For instance, all learners will hear
certain phrases repeated often-- examples include conversational cliches
like \word{beg your pardon}, prayers, legal idioms, and popular quotes-
whose frequency will not fall out of their linguistic basis.  One
possibility the learner must entertain is that each is merely a single
(long) word.  As words, the statistical regularity of the sounds within
these phrases is explained, and thus there is a motivation in the
stochastic framework for placing all passages which occur with unusual
frequency in the lexicon, regardless of whether they are linguistically
interesting.  These problems can be partially alleviated by introducing
extra parameters into language models that serve only to capture
extralinguistic regularity; this is a principal motivation for the class of
language models introduced in chapter~\ref{ch:representation}.

More problematic are cases where extralinguistic regularities cross
linguistic boundaries.  Consider the potential consequences of evidence
that can be bisected into a set of sentences involving John and Mary, and
another set involving Alice and Bob.  In the first case there might be many
sentences of the form \word{John verb Mary} and in the second of the form
\word{Alice verb Bob}.  To a learner with no access to sentence meanings,
there might appear to be an agreement phenomena between the first and last
positions in the sentence (that could have been imposed by the language
faculty).  Since languages do not generally exhibit agreement between
subject and object positions, the learner might be led to suppose a
different structure than subject-verb-object (perhaps treating \word{Bob}
and \word{Mary} as main verbs rather than direct objects).  Fortunately,
given carefully constructed classes of stochastic grammars and sufficient
evidence such pernicious examples are rare.  Furthermore, as
extralinguistic evidence becomes available it can be used to separate
regularities imposed by the language faculty from external regularities.

The John-Mary-Alice-Bob example above is unusual: because ideas are
generally mapped to language in a compositional fashion, regularities due
to extralinguistic causes often (indirectly) provide evidence about
linguistic structure.  Take for example the phrases \word{walked the mangy
  dog}, \word{bought a new car} and \word{ate a red apple}.  Each is more
likely to occur than arbitrary verb-determiner-adjective-noun sequences,
because each reflects natural associations of actions and modifiers with
objects.  The fact that all of these associations take the same form
(adjectives attached to the left of nouns and noun phrases attached to the
right of verbs) suggests that common syntactic mechanisms are being used to
capture semantic relations.  Thus, even nonlinguistic regularities are good
indicators of underlying linguistic structure.  This fact is one of the
primary reasons that unsupervised learning schemes can be successful at
elucidating linguistic structure.

Extralinguistic patterns have been the downfall of many computational
theories of language acquisition, that have modeled them at the expense of
linguistic ones (see for example Olivier~\cite{Olivier68} and Cartwright
and Brent~\cite{Cartwright94}).  In chapter~\ref{ch:representation} a
representation for language is presented that does not prevent
extralinguistic patterns from making their way into the grammar, but does
ensure that they do not preclude desired parameters.

\section{Generalization, Model Selection and the Prior}\label{generalization}

It was argued informally that the grammar $S\Rightarrow aBa, B\Rightarrow
Bb|b$ is a better hypothesis than $S\Rightarrow a|b|SS$ for the input
$aba$, $abba$, $abbbba$, $abbbbba$, because under it the input is more
typical.  On this measure the grammar $S\Rightarrow
aba|abba|abbbba|abbbbba$ is better yet.  Nevertheless, our intuition is
that this grammar is an undesirable choice, because it merely encodes the
observations and is unlikely to generalize to other sentences from the
target language.  In language acquisition, where only a very small sample
of the target language is available to the learner, generalization from
available evidence to a grammar that also explains other data is a key
issue.  This is a problem of {\em model selection}: which of many models
consistent with the data is best?  In Bayesian inference, this question is
answered by equation~\ref{eq:bayes}, which depends on the prior probability
distribution $p(G)$.  Thus, the prior can be used to manipulate
generalization performance.  However, Wolpert and
others~\cite{Schaffer94,Wolpert95} have shown that unless assumptions are
made about the learning problem, no generalization strategy (and hence no
prior) performs better than any other.  In this section various properties
of grammars and the language acquisition problem are used to motivate a
prior that favors simple grammars over complex ones, where simplicity is
defined syntactically.

By evolutionary necessity different speakers, exposed to different small
samples of a single target language, must each with high probability
converge to a language very close to the target language.  With suitable
formalization it can be shown that for this to be possible, the class of
hypothesis languages must be heavily constrained; for example, in the PAC
learning framework~\cite{Valiant84} it can be shown that the VC-dimension
of the hypothesis class is bounded by the number of samples available to
the learner~\cite{Ehrenfeucht89}, up to a factor that depends on the
allowable error rate.\footnote{The VC-dimension of a set of functions is,
  roughly speaking, a measure of the effective coverage of the
  set~\cite{Vapnik71}.  For a set of indicator functions $\cal{F}$ it is
  defined to be the size of the largest set of elements that can be labeled
  in all possible ways by functions in $\cal{F}$.  This definition can be
  extended to measure the VC-dimension of functions with arbitrary ranges,
  such as probability distributions like $p(U|G)$.}  This means that the
complexity\footnote{Here the word \word{complexity} is used with no special
  technical connotations.} of the class of grammars that can be entertained
by the learner is inherently constrained by the amount of data available
for parameter estimation.  Perhaps surprisingly, given this result, there
does not seem to be an upper bound on the number or complexity of
individual languages-- new words can always be added to an existing
language, for example.  One escape from this apparent paradox is for the
learner to adjust the hypothesis class of grammars to reflect the amount of
evidence available for estimation.

\subsection{Structural Risk Minimization}

In the Bayesian inference framework, where the language learner attempts to
optimize a stochastic language model $p(U|G)$, generalization performance
can be measured by the divergence of this conditional distribution from the
``true'' teacher's distribution over evidence, $p_T(U)$; this divergence is
computed as an expected value over all utterances, not just the sample the
learner is exposed to.  Conceptually, generalization error arises from two
sources.  The first is the choice of the hypothesis class and the fidelity
of its members to the true distribution $p_T(U)$.  If the hypothesis class
is too restrictive even the best possible grammar in it may be a poor
approximation to the true distribution.  The second is the possibility that
the learner will choose incorrectly from among the members of the
hypothesis class; the higher the ratio of the VC-dimension of the
hypothesis class to the amount of evidence, the more likely the learner is
to select a grammar that generalizes more poorly than is
necessary~\cite{Vapnik82} (given sufficient evidence for a given
VC-dimension, any function consistent with the evidence will generalize
well~\cite{Kearns94}).  

Vapnik~\cite{Vapnik82} advocates the {\em structural risk minimization}
framework in which the learner selects a hypothesis class (from among a
{\em structural hierarchy} of classes) with VC-dimension that minimizes the
sum of these two contributions to the generalization error.  In the case of
language, given a small amount of evidence the learner might restrict
attention to a small class of grammars, none of which are likely to
approximate the true function well, and as more evidence becomes available
expand the search to include a greater number of grammars, some of which
will be better approximators.  Niyogi~\cite{Niyogi95} explores this idea in
more mathematical detail, also with respect to language acquisition; see
also literature on the bias-variance tradeoff~\cite{Breiman84,Geman92}.

At face value structural risk minimization seems to be irrelevant to the
language acquisition problem.  After all, the learner does not get to
choose what the class of human grammars is; that is defined externally to
learning altogether.  This contrasts with the function approximation tasks
that motivated Vapnik, where parameters play a secondary role to the
quality of the approximation.  In language acquisition as we have defined
it, the conditional probability distribution $p(U|G)$ is merely an
algorithmic tool.  Approximating it is useful only insofar as the members
of the hypothesis class serve to identify human grammars, and this
precludes artificially simplifying stochastic grammars to conform to a
structural hierarchy.  Fortunately, the nature of human language is such
that stochastic language models can be defined over {\em partial} parameter
sets, in such a way that a structural hierarchy of stochastic grammar
classes of increasing complexity can be defined, each identifying a greater
portion of the target grammar.  For example, one might imagine structuring
grammars by the size of the lexicon.  Asked to choose among lexicons with
only one word the learner might opt for the lexicon containing the word
\word{the}.  Given access to more data, the learner might select between
lexicons containing ten words each.  Although there is obviously some risk
that the constraint of modeling with an artificially small parameter set
will lead the learner astray (perhaps, forced to choose the single ``word''
that best improves the model $p(U|G)$, selecting \word{howareyoutoday} over
\word{the}), the expectation is that as the amount of evidence is
increased, and with it the modeling power of the grammars, core parameters
will remain constant and additional parameters will be devoted to
explaining ever less important phenomena.

\subsection{The Minimum Description Length Principle}\label{mdl}

To implement structural risk minimization on top of a class of grammars two
items must be defined: a structural hierarchy over the grammars and a
function that determines the appropriate class in the hierarchy for a given
amount of evidence.  Unfortunately, this function is dependent on the
VC-dimension of each class, as well as the expected fit of each class of
grammars to the target language.  Both of these quantities are extremely
difficult if not impossible to compute in practice.  For this reason,
heuristic approximations must be used in place of structural risk
minimization.  One effective heuristic is Rissanen's {\em minimum
  description length} (MDL)
principle~\cite{Rissanen78,Rissanen89,Rissanen91}, in which description
length is used as a substitute for informational complexity measures like
the VC-dimension.  The minimum description length principle, as applied to
stochastic grammars, says that the best grammar $G$ minimizes the
combined description length of the grammar and the evidence.  More
formally,

\begin{equation}\label{eq:mdl}
  G = \begin{array}[t]{c}\mbox{argmin}\\{\scriptstyle
    G'\in\cal{G}}\end{array} |G'| + |U|_{G'}
\end{equation}

\noindent where $|G'|$ is the length of the shortest encoding of $G'$ and
$|U|_{G'}$ is the length of the shortest encoding of $U$ given knowledge of
the grammar $G'$.  Using near-optimal coding schemes, Shannon's source
coding theorem~\cite{Shannon48} implies that $|U|_{G'}$ can be made to
closely approach $-\log p(U|G')$, and therefore equation~\ref{eq:mdl}
can be rewritten

\begin{equation}\label{eq:mdl2}
  G = \begin{array}[t]{c}\mbox{argmin}\\{\scriptstyle
    G'\in\cal{G}}\end{array} |G'| - \log p(U|G'),
\end{equation}

\noindent a more intuitive formulation from the standpoint of stochastic
grammars.  The duality between description lengths and probabilities is
convenient.  It means, among other things, that any coding scheme for
utterances can be interpreted as a stochastic grammar, and vice versa (see
section~\ref{coding} for further discussion).  It also means that if the
prior probability $p(G)$ is defined by $p(G) = 2^{-|G|}$ then
equations~\ref{eq:bayes} and \ref{eq:mdl2} coincide.  Thus, MDL can be
interpreted as a Bayesian prior that is biased against grammars with high
syntactic complexity.  Rather than try to argue for MDL from first
principles,\footnote{See~\cite{Kolmogorov65,Li91,Li93,Rissanen78,Rissanen89,Rissanen91}
  for attempted justifications of MDL and the closely related Kolmogorov
  complexity.  Other relevant arguments for simplicity as measured by
  description length
  include~\cite{Berwick85,Blumer87,Chomsky51,Halle61,Solomonoff60}.} we
note that it is merely a heuristic, but point out three important ways in
which it mimics the philosophy of the better-justified structural risk
minimization:

\begin{itemize}
\item In very many cases the VC-dimension of a parameterized class of
  functions is linear or near-linear in the number of free parameters in
  the class~\cite{Baum89,Kearns94,Vapnik82}.  Given an efficient coding
  scheme, the length of a description of a set of (independent) parameters
  is linear in the number of parameters.  Hence, in a structural hierarchy
  where classes consist of functions with the same number of free
  parameters, the description length of a grammar should be linearly
  related to the VC-dimension of the class it is in.  By penalizing
  grammars with high description length $|G|$, MDL therefore weighs against
  classes that have too high VC-dimension for good generalization
  performance.

\item With sufficient evidence, for a class of a given VC-dimension good
  generalization performance can be achieved by selecting the function that
  models the evidence best~\cite{Vapnik71}; for stochastic grammars, this
  is the one that maximizes $p(U|G)$.  Hence, the $-\log p(U|G)$ term
  biases toward grammars that are likely to generalize well.
  
\item Assuming a nearly stationary class of stochastic grammars, to a first
  approximation the probability distribution $p(U|G)$ can be factored over
  individual utterances: $p(U|G) = \prod_{u\in U} p(u|G)$, which tends
  towards $\hat{p}^n$ where $\hat{p}$ is the (geometric) mean probability
  per utterance and $n$ is the number of utterances.  Thus, the term $-\log
  p(U|G) = -n\log \hat{p}$ grows linearly with the amount of evidence
  available to the learner.  As it grows, so does the incentive to increase
  $\hat{p}$ (by moving to a grammar from a broader class with better
  approximation properties).  In this way the choice of the VC-dimension of
  the hypothesis class is made to depend on the amount of evidence
  available to the learner.
\end{itemize}

\noindent Although MDL has had successful applications in language
inference, it depends on a syntactic definition of complexity and therefore
its effectiveness is tied to the encoding scheme used for stochastic
grammars.  Despite its motivations, it does not trade VC-dimension against
evidence in the theoretically optimal way, and in no way guarantees that
generalization performance is maximized: although results vary by
application~\cite{Murphy95}, as is to be expected, practical experience
indicates (see~\cite{Murphy94,Ristad95,Webb96} and
section~\ref{results:compression}) that MDL as commonly used tends to
underestimate the number of parameters necessary for optimum
generalization.  From a Bayesian perspective this is not surprising: the
$2^{-|G|}$ prior very heavily biases towards grammars that are improbably
simple from the linguistic perspective.  Despite the fact that MDL is only
a heuristic approximation to more desirable model-selection schemes such as
structural risk minimization, it will be used in the learning schemes
presented in the remainder of this thesis, because description lengths can
be conveniently computed and manipulated.

\section{Example}

At this point it is worth looking at a very simple example of how the
minimum description length principle (as embodied in equation~\ref{eq:mdl})
can be used for language acquisition.  The example is chosen to illustrate
ideas that will be relevant in the following chapters.  Let us suppose the
learner receives evidence in the form of a sequence of characters, such as
{\tt iateicecream}.  The grammars the learner entertains each consist of a
set of words, where each word is a sequence of characters.  Thus, one
possible grammar is \{ {\tt i}, {\tt ate}, {\tt ice}, {\tt cream} \}.

In the Bayesian inference framework, two distributions must be defined.
The first is a prior distribution over possible grammars, $p(G)$, and the
second is a conditional distribution over possible character sequences
$p(u|G)$.  The MDL principle is more simply expressed in terms of
description length than probabilities, so for the moment let us concentrate
on coding schemes rather than distributions.  Suppose that every word in a
grammar is assigned a prefix-free codeword.  Then the evidence $u$ is
encoded by writing down a sequence of codewords.  For example, given the
grammar

\begin{center}
\begin{tabular}{l|cccccccc}
Word & {\tt c} & {\tt a} & {\tt i} & {\tt e} & {\tt r} & {\tt m} & {\tt t} & {\tt ice}\\
Codeword & 00 & 010 & 011 & 100 & 101 & 110 & 1110 & 1111
\end{tabular}
\end{center}

\noindent then the evidence {\tt iateicecream} can be encoded in 30 bits
as {\tt i}$\cdot${\tt a}$\cdot${\tt t}$\cdot${\tt e}$\cdot${\tt ice}$\cdot${\tt c}$\cdot${\tt r}$\cdot${\tt e}$\cdot${\tt a}$\cdot${\tt m}:

\[ 011\cdot 010\cdot 1110\cdot 100\cdot 1111\cdot 00\cdot 101\cdot 100\cdot
010\cdot 110. \]

A coding scheme for grammars must also be specified.  Suppose that all
grammars include the 26 letters of the alphabet, so they don't need to be
explicitly encoded into grammars.  The words in a grammar that are more
than one character long are encoded by writing out the codewords of their
component characters.  The word {\tt ice} in the above grammar, for
example, is encoded $011\cdot 00\cdot 100$ ({\tt i}$\cdot${\tt
  c}$\cdot${\tt e}).  There are many details being glossed over here, such
as how codewords are assigned to words; for the time being it is more
important to focus on fundamental issues.

Given this model of language, let us compare three grammars for the evidence
{\tt themanonthemoon}.

\begin{center}
\begin{tabular}{cl}
(A) & 
\begin{tabular}{l|ccccccc}
Word & {\tt o} & {\tt n} & {\tt t} & {\tt h} & {\tt e} & {\tt m} & {\tt a}\\
Codeword & 00 & 01 & 100 & 101 & 110 & 1110 & 1111
\end{tabular}\\
\\
(B) &
\begin{tabular}{l|cccccccc}
Word & {\tt o} & {\tt n} & {\tt the} & {\tt m} & {\tt t} & {\tt h} & {\tt e} & {\tt a}\\
Codeword & 00 & 01 & 100 & 101 & 1100 & 1101 & 1110 & 1111
\end{tabular}\\
\\
(C) & \begin{tabular}{l|cccccccc}
Word & {\tt o} & {\tt n} & {\tt t} & {\tt h} & {\tt e} & {\tt m} & {\tt a} & {\tt themanonthemoon}\\
Codeword & 00 & 01 & 100 & 101 & 1100 & 1101 & 1110 & 1111
\end{tabular}
\end{tabular}
\end{center}

Each of these grammars defines a total description length for {\tt
  themanonthemoon}.  For Grammar~A, which has no words other than single
characters, this is simply the length of the best encoding of the evidence.
Grammars~B and C must add to this the cost of representing extra words in
the grammar.

\begin{center}
\begin{tabular}{cll}
(A) & Evidence & 100$\cdot$101$\cdot$110$\cdot$1110$\cdot$1111$\cdot$01$\cdot$00$\cdot$01$\cdot$100$\cdot$101$\cdot$110$\cdot$1111$\cdot$00$\cdot$00$\cdot$01 \\
& & ( {\tt t}$\cdot${\tt h}$\cdot${\tt e}$\cdot${\tt m}$\cdot${\tt a}$\cdot${\tt n}$\cdot${\tt o}$\cdot${\tt n}$\cdot${\tt t}$\cdot${\tt h}$\cdot${\tt e}$\cdot${\tt m}$\cdot${\tt o}$\cdot${\tt o}$\cdot${\tt n} )\\
& Length & 42 bits.\\
\\
(B) & Evidence & 100$\cdot$101$\cdot$1110$\cdot$01$\cdot$00$\cdot$01$\cdot$100$\cdot$1101$\cdot$00$\cdot$00$\cdot$01 ( {\tt the}$\cdot${\tt m}$\cdot${\tt a}$\cdot${\tt n}$\cdot${\tt o}$\cdot${\tt n}$\cdot${\tt the}$\cdot${\tt m}$\cdot${\tt o}$\cdot${\tt o}$\cdot${\tt n} )\\
& Grammar & 1100$\cdot$1101$\cdot$1110 ( {\tt t}$\cdot${\tt h}$\cdot${\tt e} )\\
& Length & 40 bits.\\ \\
(C) & Evidence & 1111 ( {\tt themanonthemoon} )\\
& Grammar & 100$\cdot$101$\cdot$1100$\cdot$1101$\cdot$1110$\cdot$01$\cdot$00$\cdot$01$\cdot$100$\cdot$101$\cdot$1100$\cdot$1101$\cdot$00$\cdot$00$\cdot$01 \\
& & ( {\tt t}$\cdot${\tt h}$\cdot${\tt e}$\cdot${\tt m}$\cdot${\tt a}$\cdot${\tt n}$\cdot${\tt o}$\cdot${\tt n}$\cdot${\tt t}$\cdot${\tt h}$\cdot${\tt e}$\cdot${\tt m}$\cdot${\tt o}$\cdot${\tt o}$\cdot${\tt n} )\\
& Length & 48 bits.\\
\end{tabular}
\end{center}

The minimum description length principle says that the best grammar is the
one that results in the shortest description length for the evidence {\em
  and} the grammar.  That is Grammar~B, at 40 bits.  Grammar~C has a very
short description of the evidence, but at the expense of an extremely long
and overly specific grammar.  Grammar~A has too general a grammar and fails
to capture an important pattern in the evidence.  Grammar~B, which moves
the word {\tt the} into the lexicon and thus saves bits every time it is
used (the codeword for {\tt the} is considerably shorter than the combined
length of the codewords for {\tt t}, {\tt h} and {\tt e}), strikes a happy
medium.  Thus, in this case the MDL principle favors the grammar with the
most linguistically appealing structure.

Notice that the coding scheme for utterances is equivalent to a stochastic
language model $p(u|G)$.  In particular, to stochastically generate an
utterance $u$ under a grammar $G$, first generate a random sequence of bits
by flipping a coin, and then use $G$ to decode that sequence into an
utterance $u$.  This is why it doesn't matter whether we think in terms of
stochastic language models or in terms of probability distributions.

\section{The Search Procedure}

In section~\ref{learning} it was argued that the learning mechanism must be
given a principled foundation.  In the Bayesian inference framework the
function of the learning mechanism is to find the grammar with the maximum
posterior probability; at a conceptual level, therefore, it is entirely
defined by the class of grammars, the prior probability distribution, and
the conditional probability distribution.  In practice, however, the class
of grammars will be large, if not infinite, precluding maximization via
enumeration and necessitating heuristic searches that take advantage of
the qualities of specific grammar classes.

\section{Related Work}\label{uns:related}

Bayesian inference and MDL each have rich histories, and have been
routinely applied to problems of language acquisition.  Some of the
earliest work on the inductive inference of language was performed by
Solomonoff~\cite{Solomonoff59,Solomonoff60}, who would later play a major
role in defining the theory that motivates MDL~\cite{Solomonoff64}.  In his
language work the importance of penalizing complexity is already
emphasized.  As far back as 1955 Chomsky wrote in {\em The Logical
  Structure of Linguistic Theory}~\cite{Chomsky55}

\begin{quote}
  In applying this theory to actual linguistic material, we must construct
  a grammar of the proper form\ldots\ Among all grammars meeting this
  condition, we select the simplest.  The measure of simplicity must be
  defined in such a way that we will be able to evaluate directly the
  simplicity of any proposed grammar\ldots\ It is tempting, then, to
  consider the possibility of devising a notational system which converts
  considerations of simplicity into considerations of length.
\end{quote}

Stochastic methods have also been applied from very early on.  One of the
first demonstrations of Markov models~\cite{Markov13} was an elucidation of
the dependencies between adjacent characters in the text of Pushkin's {\em
  Eugene Onegin}.  Olivier~\cite{Olivier68} uses stochastic models in an
early computational study of language acquisition.  However, very few in
the natural language community have looked carefully at the necessary
relation between stochastic models and the problems they are applied to; as
a consequence most experiments in the unsupervised learning of language
have tended to result in parameter values that fare well on statistical
criteria, but not on linguistic ones.

\section{Conclusions}

This chapter has surveyed the issues surrounding the application of
Bayesian inference to the problem of unsupervised language acquisition.
This framework for statistical estimation evaluates grammars largely on the
basis of whether they explain the typicality of the evidence, and hence can
discriminate between grammars even in absence of binary grammaticality
judgments and without reference to information from beyond the speech
signal, such as sentence meanings, that may not always be available to the
learner.  Various subtleties have been discussed at length, in particular
the need for certain relations to hold between the structure of stochastic
language models and the linguistic parameters that are the desired output
of the learning process.  The difficult problem of ensuring good
generalization from a small amount of evidence was used to promote a
bias in the learning algorithm towards simple grammars.

The main purpose of this chapter has been to provide an objective function
(namely, the posterior probability given a prior that is defined in terms
of description length) by which a learning algorithm can evaluate a
grammar.  Neither the form of grammars nor the learning algorithm has been
specified; these are the topics of the next two chapters.  The choices
there will determine whether the MDL-based inference procedure is
successful.  In particular, they will determine whether the entire learning
process converges to linguistic parameters that agree with what is known
about human language and human performance.

It is important to note that stochastic grammars and the description-length
prior are serving here as tools to aid the learning algorithm.  This
chapter has {\em not} argued that language is best viewed as a random
process, or even that analogs of stochastic parameters are present in the
grammars used by adults for generation and interpretation.  However, the
discussion is equally relevant to human language acquisition as it is to
engineering applications in which it is necessary to estimate stochastic
language models for use in disambiguation and compression.

\pagestyle{headings}

\chapter{A Representation for Lexical Parameters}\label{ch:representation}

This chapter presents the principal innovation of this thesis, a framework
for the representation of linguistic knowledge.  In it, parameters like
words are represented in the lexicon as a perturbation of the composition
of other lexical parameters.\footnote{In this thesis the word {\em lexicon}
  refers to the store of memorized, irregular knowledge about language.  As
  a matter of convenience the word {\em word} will often be used to refer
  to any lexical parameter, though a more proper term would be {\em
    listeme} (defined by Di Sciullo and Williams~\cite{DiSciullo87} as an
  item that must be memorized).  Listemes include morphemes, many syntactic
  words, idioms, and perhaps syllables.  Here even syntactic rules are
  treated as part of the lexicon, if there is reason to believe that they
  are memorized.  Under these definitions the lexicon does not include
  objects that can be derived using completely regular processes, even if
  they are words in the traditional sense; see Spencer~\cite{Spencer91} for
  further discussion.} This recursive decomposition of knowledge in the
lexicon is similar in spirit to the hierarchical phrase structures commonly
associated with sentence processing, distinguished by the fact that at
every level in the hierarchy perturbations introduce changes to default
compositional behaviors.  As a theory at the computational level, the
framework abstracts from details of linguistic theory while highlighting
issues of memory organization that are central to language acquisition.
When used in conjunction with the inference framework presented in
chapter~\ref{ch:Bayes}, it neatly circumvents many of the potential
pitfalls of unsupervised learning raised there, such as the propensity for
the learner to model extralinguistic patterns in the signal.  In this way
it is a theory of language acquisition as well as a theory of lexical
organization.  The success of the theory is demonstrated through learning
algorithms and results presented in chapters~\ref{ch:algorithm} and
\ref{ch:results}.

The chapter begins with an introduction to the representational framework,
culminating in a simple example in which parameters are character sequences
built by concatenating other character sequences.  This example is used as
background to present various motivations for the framework, principally
from the standpoint of unsupervised learning but also with respect to the
nature of language.  The issue of coding is then explored in more depth.
Finally, four instantiations of the framework are defined in greater
detail.

\section{The Representational Framework}

A central tenet of modern linguistic theory is that language makes
``infinite use of finite means''~\cite{Chomsky65,Humboldt1836}, or in
plainer terms, that language combines a finite set of lexical parameters to
produce an infinite variety of sentences.  This chapter argues that these
lexical parameters, the primitive units of sentence processing, {\em are
  themselves built by composing parts}, inside the lexicon.  Thus, each
lexical parameter is constructed very much like a sentence, with idioms
built from words, words from morphemes, and so on. What distinguishes the
lexicon from the sentence processing mechanism is that the composition
occurs off-line, and more importantly, that parts combine to produce a
whole that is greater (or at least different) than the sum of the parts.
This idea is captured here by a framework for lexical representation in
which each parameter $w$ in the lexicon is represented as the perturbation
of a composition of other parameters $w_1\ldots w_n$,

\[ w = (w_1 \circ \cdots \circ w_n) + \mbox{\sc perturbations}. \]

\noindent Here the composition operator $\circ$ is taken to represent the
same process that combines words and other elements from the lexicon during
on-line processing.  The intuition behind this representation is that $w$
inherits the linguistic properties of its components $w_1\ldots w_n$.  At
the same time the perturbations introduce changes that give $w$ a unique
identity: a word that acts exactly as the composition of its parts could be
removed from the lexicon and reconstructed on-line during normal sentence
processing.  Conceptually, this framework is quite similar to the class
hierarchy of a modern programming language, where classes can modify
default behaviors that are inherited from superclasses.  The more of its
properties a parameter inherits from its components, the fewer need to be
specified via perturbations.

\begin{figure}[tbh]
\pageline
\begin{center}
\begin{tabular}{l|l}
  Parameter & Possible Representation \\ \hline

  \word{cat} & (\word{c} $\circ$ \word{a} $\circ$ \word{t}) + {\sc
    is-a-noun} + {\sc meaning} + {\sc freq}\\ 

  \word{motor} & (\word{mo} $\circ$ \word{tor}) + {\sc is-a-noun} + {\sc
    meaning} + {\sc freq}\\

  \word{blueberry} & (\word{$\langle$Noun $\Rightarrow$ Adj Noun$\rangle$}
  $\circ$ \word{blue} $\circ$ \word{berry}) + {\sc meaning} + {\sc freq}\\ 

  \word{wanna VP} & (\word{$\langle$VP $\Rightarrow$ Verb to VP$\rangle$}
  $\circ$ \word{want}) + {\sc sound-change} + {\sc freq}\\ 

  \word{Verb Prep NP} & (\word{$\langle$VP $\Rightarrow$ Verb PP$\rangle$}
   $\circ$ \word{$\langle$PP $\Rightarrow$ Prep
    NP$\rangle$}) + {\sc freq}\\ 

  \word{take off NP} & (\word{$\langle$VP $\Rightarrow$ Verb Prep NP$\rangle$}
  $\circ$ \word{take} $\circ$ \word{off}) + {\sc meaning} + {\sc freq}\\ 

  \word{kick the bucket} & (\word{$\langle$VP $\Rightarrow$ Verb
    NP$\rangle$} $\circ$ \word{kick} $\circ$ \word{$\langle$NP
    $\Rightarrow$ Det Noun$\rangle$} $\circ$ \word{the} $\circ$
  \word{bucket}) + {\sc meaning} + {\sc freq}

\end{tabular}
\end{center}
\caption{\label{fig:perturbations} Some informal examples of how different
  lexical parameters can be represented by perturbing a composition of
  other parameters, ranging from phonemes and syllables to words and
  syntactic rules.  Here perturbations are represented with capital
  letters, with {\sc meaning} denoting a change in meaning and
  {\sc freq} a change in frequency.} \pageline
\end{figure}

Figure~\ref{fig:perturbations} presents several (very informal) examples
that should help convey the intended use of this abstract framework.  In
each case parameters are constructed by composing several parameters and
perturbing the result.  Perturbations include sound changes (\word{want to}
becomes \word{wanna}), changes to syntactic properties (\word{cat} and
\word{motor} are nouns), changes to meaning (a \word{blueberry} is more
than just a blue berry and \word{kick the bucket} has nothing to do with
kicking or buckets), and changes to frequency.  Frequency information is
used to give a stochastic interpretation to the lexicon during unsupervised
learning of the sort described in chapter~\ref{ch:Bayes}.  Its use and
importance will be discussed in greater detail later.  The parameters that
are composed in these examples range from phonemes and syllables to words
and syntactic rules.  The definition of the composition operator dictates
how parameters combine.  Ideally, the composition operator encodes most of
a detailed theory of language, explaining how phonemes and syllables come
together in words like \word{cat} and \word{motor}, how syntactic rules
combine, and even how semantic interpretations are constructed by composing
words under standard syntactic relations (as with \word{blueberry}).  Note
that in most of these examples relatively little information needs to be
added via perturbations.  For example, although \word{blueberry} does mean
something different than \word{blue berry}, much of the meaning and all of
the syntactic and phonological properties of the word are inherited at the
mere price of references to one syntactic rule and two other words.
Without such a means of sharing structure, each parameter would include an
enormous amount of redundant information.  For example, the irregular
passive form \word{taken} would need to be memorized twice, once for
\word{take} and once for \word{take off}.  As is, the framework can neatly
explain how \word{take off} can have a meaning that is quite independent
of \word{take} and \word{off}, but nevertheless share properties with its
components.

Many objects not traditionally considered ``word-like'' are included in
these examples, such as syntactic rules and syllables.  This is because the
representational framework is relatively independent of details of
linguistic theory, and conveys its advantages at any level of the
linguistic hierarchy.  The fact that a single symbol $\circ$ is used to
represent the composition operator in each of the examples in
figure~\ref{fig:perturbations} is not meant to imply that in realistic
instantiations the same combinatory process would be applied universally;
presumably, for example, the mechanism that combines phonemes into
syllables should function differently than the one that composes syntactic
rules.  Because it is the abstract framework that is studied here, rather
than the details of linguistic theory, all instantiations of the framework
that will be discussed use only a single composition operator each, general
enough to approximate processes ranging from morphology to
syntax.\footnote{With a single composition operator, the framework offers
  no internal means of distinguishing between ``words'' and other
  parameters-- all are treated alike, and any test of ``word-dom'' must be
  applied externally.  This agrees with the fact that it is extremely
  difficult to find language-independent definitions that agree with our
  intuition of what a word is~\cite{Spencer91}.  In contrast, if multiple
  composition operators are used, then parameters can be classified
  according to the compositional process that they are built with.}
Furthermore, no parameters beyond the lexicon are studied, and therefore in
the remainder of this thesis the lexicon effectively acts as a grammar.
(The two words will be used largely interchangeably below).  In fact, since
parameters are represented in almost the same way as utterances, {\em the
  lexicon is the grammar both for utterances and for itself}.

As with any kind of grammar, lexicons can be given stochastic
interpretations for the purposes of Bayesian inference.  As a simple
example, one which will be discussed at much greater length below, each
word in the lexicon could be associated with a probability that determines
the relative frequency of that word.  In such a case, words serve both as
points at which perturbations attach new information and also as a means to
refine a stochastic model.  The word \word{motor}, for example, might allow
a grammar to explain why the components \word{mo} and \word{tor} occur
together so much more often than would be expected given their independent
probabilities.  The fact that parameters can be motivated from the
standpoint of Bayesian inference as well as on the basis of where
perturbations need to occur is what allows the framework to be used for
unsupervised learning.  The fact that the lexicon serves as a stochastic
language model both for the input and itself means that description lengths
for utterances and parameters are computed in the same way.  This
simplifies the statement of the MDL principle, allowing
equation~\ref{eq:mdl} to be rewritten as

\begin{eqnarray}
  G & = & \mbox{\begin{array}[t]{c}\mbox{argmin}\\{\scriptstyle
    G'\in\cal{G}}\end{array}} |G'| + |U|_{G'} \nonumber \\
& \approx & \mbox{\begin{array}[t]{c}\mbox{argmin}\\{\scriptstyle
    G'\in\cal{G}}\end{array}} \sum_{w\in G'} |w|_{G'} + \sum_{u\in
    U} |u|_{G'}.\label{eq:smdl}
\end{eqnarray}

\noindent where $|x|_{G'}$ is the description length of $x$ under the
grammar (lexicon) $G'$.  A concrete example of how the representational
framework can be instantiated and interpreted with respect to
equation~\ref{eq:smdl} is given below.  It will be used as the basis for
further discussion of the framework.

\subsection{Concatenative Example}\label{concat}

Let us look at a linguistically naive instantiation of the above framework,
that ignores all details of phonology, syntax and semantics.  Each word in
the lexicon is simply a sequence of characters, linked to a codeword that
serves as a pointer.  For example, one word might be

\begin{center}
\begin{tabular}{cc}
Character & Codeword \\ \hline
\word{badminton} & 0011
\end{tabular}.
\end{center}

The composition operator is concatenation: each word is represented as the
concatenation of the character sequences of other words, plus its codeword
(the only perturbation).  This process bottoms out in words that are single
characters.  In this way, \word{badminton} can be represented as \word{bad}
$\circ$ \word{min} $\circ$ \word{ton} + 0011.  For realistically sized
examples, clever coding schemes can nearly eliminate the cost of coding the
perturbation (0011) and the cost of terminating the encoding of the
composition.  Assuming a prefix-free code, the representational cost of
each word then reduces to the cost of writing down in sequence the
codewords of its components.  For example, if \word{bad} is coded as 10,
\word{min} as 011, and \word{ton} as 010, then \word{badminton} can be
encoded in 8 bits as $10011010$.

\begin{figure}[tbh]
\pageline
\begin{center}
\begin{tabular}{llcccr}
code & $w$ & representation & encoding & count & $|w|$\\ \hline
000 & the & t $\circ$ h $\circ$ e & 01001101011 & 2 & 11\\
001 & at & a $\circ$ t & 1100010 & 2 & 7\\
010 & t &&& 2 & \\
0110 & h &&& 2 & \\
0111 & cat  & c $\circ$ at & 1101001 & 1 & 7\\
1000 & hat  & h $\circ$ at & 0110001 & 1 & 7\\
1001 & thecat & the $\circ$ cat & 0000111 & 1 & 7\\
1010 & thehat & the $\circ$ hat & 0001000 & 1 &  7\\
1011 & e &&& 1 & \\
1100 & a &&& 1 & \\
1101 & c &&& 1 & \\
1110 & i &&& 1 & \\
1111 & n &&& 1 & \\
\multicolumn{2}{l}{$u =$ thecatinthehat} & thecat $\circ$ i $\circ$ n $\circ$ thehat &
1001111011111010 &  & 16\\  \hline
 & & & \multicolumn{3}{r}{$|u| + \sum |w| = 62$}
\end{tabular}
\end{center}
\vspace{-.15in}
\caption{\label{fig:model}A 62-bit, suboptimal description of {\em
    thecatinthehat}.  The complete description length of the input is
  computed by adding the length of the representation of the input to the
  lengths of the representations of the parameters; this ignores several
  minor coding costs.  Terminals have no representation.} \pageline
\end{figure}

Figure~\ref{fig:model} presents a lexicon for the character sequence
\word{thecatinthehat} (though not a good one).  Representations and their
encodings are provided for the input and each (nonterminal) parameter in
the lexicon.  The count of how many times parameters are referenced in the
complete description of both \word{thecatinthehat} and the lexicon
determines the length of the codeword for each parameter (here a Huffman
code~\cite{Huffman52} was used).  The description length of each parameter
is the sum of the lengths of its components' codewords (since the cost of
perturbations and terminators is negligible).

The lexicon in figure~\ref{fig:model} does not minimize the description
length of the input; this small amount of evidence is not sufficient to
justify words like \word{cat} and \word{hat}.  This example is meant only
to demonstrate how the abstract representational framework can be turned
into a concrete coding scheme, that {\em could} be used to search for a
lexicon with minimum description length.  Despite its naivete, this simple
concatenative model is quite powerful.  Chapter~\ref{ch:algorithm} presents
a search algorithm for the model that attempts to find the lexicon that
minimizes the total description length of some input.  Tests on large texts
(presented in chapter~\ref{ch:results}) indicate that this algorithm learns
a lexicon that agrees closely with human judgments.  For example, when
tested on an unsegmented (spaceless) version of the Brown
corpus~\cite{Francis82}, one of the parameters learned is
\word{nationalfootballleague}.  The representation of this phrase is
\word{national} $\circ$ \word{football} $\circ$ \word{league}.  A larger
portion of the recursive decomposition of the phrase in the lexicon is
presented in figure~\ref{fig:nfl}.  The reason that the optimal lexicon
agrees closely with our intuitions, despite the fact that the learning
mechanism has no access to syntactic and semantic information, was given in
chapter~\ref{ch:Bayes}: given appropriate representations, the learner is
best able to model the statistical properties of the input by reproducing,
at least in part, the process that generated it.

Before presenting various motivations for the representational framework, it
is worth looking a little closer at the statistical properties of this
concatenative model.  Assuming codewords are chosen to minimize the total
description length, codeword lengths $l(w)$ will be related to word
frequencies $p(w)$ according to the standard relation $l(w) = - \log p(w)$,
where frequencies are defined over the representations of both the input
and the lexicon.  Thinking in terms of probabilities rather than codewords,
it is clear that this coding system defines a stochastic language model
under which both the input and the parameters are generated by
concatenating parameters chosen by an independent and identically
distributed ({\em i.i.d.}) process.  Thus, the probability of the character
sequence $u$ under this language is

\[ p(u) = \sum_{n} p(n) \sum_{w_1 \ldots w_n\ s.t.\ u =
  w_1\circ\cdots\circ w_n} p(w_1)\cdots p(w_n), \]

\noindent where $p(n)$ is effectively defined by the manner in which
compositional encodings are terminated.  This stochastic language model has
been called a {\em multigram}~\cite{Deligne95} and used for a variety of
language modeling applications.  Multigrams account for statistical
dependencies by assigning probabilities $p(w)$ to lengthy character
sequences: they are essentially variable-length block codes.  For example,
the fact that $p(\mbox{\word{the}}) \gg p(\mbox{\word{t}})
p(\mbox{\word{h}}) p(\mbox{\word{e}})$ is captured in
figure~\ref{fig:model} by the word \word{the}, which is assigned a codeword
much shorter than the combined length of the codewords for \word{t},
\word{h} and \word{e}.  Since multigrams do make independence assumptions
at parameter boundaries, they have difficulty reproducing complex
distributions.  Their modeling power can be increased by increasing the
length of parameters (thereby reducing the number of independence
assumptions), but this increases the number of parameters exponentially,
and also makes it difficult to assign linguistic interpretations to the
parameters.  One of the fundamental advantages the hierarchical framework
conveys upon the multigram model is that, since each parameter is itself
decomposed, statistical modeling power need not be at the expense of
linguistic structure.  For instance, in figure~\ref{fig:nfl} the parameter
\word{nationalfootballleague} captures a statistical dependence that spans
22 characters, while its internal representation provides information about
linguistic structure at finer scales.

\section{Motivations}

The preceding discussion gives some hints as to the advantages the
composition and perturbation framework offers with respect to language
acquisition, and in particular language acquisition in the Bayesian
framework presented in chapter~\ref{ch:Bayes}.  The framework can in fact
be motivated from many standpoints, among them that it leads to simple
incremental learning algorithms, explains how the learner can avoid being
confused by extralinguistic patterns, and accords with what is known
about language and language change.

\subsection{Learning}\label{motiv:learnability}

In order to understand how the representational framework aids learning, it
is first necessary to understand how the representation interacts with the
minimum description length evaluation function (equation~\ref{eq:smdl}).
To simplify discussion, two assumptions will be made: first, that the
composition operator is associative ($a \circ (b \circ c) = (a \circ b)
\circ c$), and second that the perturbation operator commutes with the
composition operator ($(a \circ b) + \mbox{\sc p} = (a + \mbox{\sc p})
\circ b = a \circ (b + \mbox{\sc p})$).  These assumptions hold for
concatenation and the meaning perturbation operator presented below in
section~\ref{rep:meaning}.  More complex operators will usually violate
these assumptions to varying extents, but most intuitions remain the same.
Given the assumptions, any representation $w_1 \circ \cdots \circ w_n +
\mbox{\sc p}_1 + \cdots + \mbox{\sc p}_m$ is equivalent to the same
representation with $w_i \circ \cdots \circ w_{j-1} + \mbox{\sc p}_1 +
\cdots + \mbox{\sc p}_{k-1}$ removed and replaced with a parameter $W$, so
long as $W$ is equivalent to the removed portions of the representation:

\begin{eqnarray*}
  \lefteqn{w_1 \circ \cdots \circ w_n + \mbox{\sc p}_1 + \cdots + \mbox{\sc
      p}_m =} \\
  & & w_1 \circ \cdots \circ w_{i-1} \circ w_i \circ \cdots \circ w_{j-1}
  \circ w_j \circ \cdots \circ w_n + \mbox{\sc p}_1 + \cdots \mbox{\sc
    p}_{k-1} + \mbox{\sc p}_k + \cdots + \mbox{\sc p}_m = \\ 
  & & w_1 \circ \cdots \circ w_{i-1} \circ (w_i \circ \cdots \circ w_{j-1}
  + \mbox{\sc p}_1 + \cdots + \mbox{\sc p}_{k-1}) \circ w_j \circ \cdots
  \circ w_n + \mbox{\sc p}_k + \cdots + \mbox{\sc p}_m = \\ 
  & & w_1 \circ \cdots \circ w_{i-1} \circ W \circ w_j \circ \cdots \circ
  w_n + \mbox{\sc p}_k + \cdots + \mbox{\sc p}_m. \label{r2}\\ \\ 
  W & = & w_i \circ \cdots \circ w_{j-1} + \mbox{\sc p}_1 + \cdots +
  \mbox{\sc p}_{k-1}.
\end{eqnarray*}

\noindent This means that it does not matter whether information is written
explicitly into a representation or referenced indirectly via another
parameter, at least as far as linguistic interpretation is concerned.  One
consequence of this is that {\em the internal representation of a parameter
  does not affect its use}.  In fact, if perturbations could occur at the
utterance level during on-line processing, then the simplest grammar,
consisting only of primitive terminals, could account for as much input as
any other grammar.\footnote{Although the representational framework
  specifically disallows perturbations at the utterance level, it will turn
  out to be useful to allow for such perturbations during learning; this
  will ensure that all input can be analyzed, even in the earliest stages
  of incremental learning when only the most rudimentary of grammars is
  available.} In general then, under this framework grammars cannot be
favored on the basis of whether or not they account for the input.
Instead, in accordance with the inference framework of
chapter~\ref{ch:Bayes}, grammars are judged on the basis of description
length, a measure that trades the typicality of the input against the
complexity of the grammar.

\subsubsection{The Statistical Interpretation of a Parameter}

The important question becomes: when does a parameter reduce the total
description length?  To answer this, it helps to imagine each parameter as
having two parts.  The first is linguistic in nature (and the desired
output of the learning mechanism) and the second is statistical.  The
linguistic portion of a parameter can be thought of as a predicate (a test)
that is either true of part of an utterance or is not.  For example, the
word \word{the} is true of the first three letters in \word{the cat} but
not of the first three letters in \word{a dog}.  Similarly, a phonological
rule like ``voice the plural marker \word{-s} after voiced consonants''
could be expressed by the predicate ``voiced plural marker or unvoiced
preceding consonant''.  The second half of each parameter, the statistical
portion, is information that, very roughly speaking, determines the
proportion of time the predicate is true of utterances generated by the
stochastic grammar.  In the concatenative model presented in
section~\ref{concat} this information took the form of a codeword, or
equivalently, a probability.  More sophisticated models might represent
statistical properties differently, perhaps in a manner better suited for
combining multiple pieces of information (see, for example, the
maximum-entropy language modeling scheme described by Della Pietra {\em et
  al.}~\cite{Berger96,DellaPietra95}).  The more parameters are in a
lexicon, the more control points the stochastic model has, and the better
it will be able to model the target language.  Hence, {\em any} lexical
parameter should reduce (or at least not increase) the description length
of the input and existing parameters.  However, as discussed in
section~\ref{generalization}, to ensure good generalization performance it
is necessary to penalize parameters by their own description length.
Hence, to be included in the lexicon, a parameter must not only reduce the
description length of the input and the remainder of the lexicon, but
reduce it by more than the length of its own representation.

In order to answer the question of when a parameter reduces the total
description length, it is therefore first necessary to ask when a parameter
leads to {\em significant savings} in the representations of the input and
other parameters.  Imagine a parameter $w = w_1 \circ \ldots \circ w_n +
\mbox{\sc p}_1 + \cdots + \mbox{\sc p}_{m}$.  Leaving the issue of the
perturbations aside, this parameter is a means of defining the statistical
properties of a linguistic predicate whose behavior is already governed by
the parameters $w_1 \ldots w_n$.  Therefore, $w$ improves the stochastic
language model only in so much as the predicate it represents has different
statistical behavior than expected given the behavior of its parts.  For
example, in figure~\ref{fig:nfl} \word{nationalfootballleague} improved the
lexicon because \word{national}, \word{football} and \word{league} occur
together far more often than the multigram language model predicts given
their individual probabilities.  But in contrast, a grammar that includes a
parameter \word{NP $\Rightarrow$ Det Noun} that predicts that determiners
and nouns co-occur frequently, and a parameter \word{the} that is a
determiner and a parameter \word{dog} that is a noun, would not be expected
to gain significant statistical advantage from a parameter \word{the dog}.
The parameter would yield substantial reduction only if it was extremely
important to model the statistical behavior of the word \word{dog} in fine
detail, which would only be the case if \word{dog} were very frequent.
This reflects an important point.  Parameters that are only infrequently
true must introduce substantial savings to be worth including in a lexicon;
parameters with widespread usage are beneficial even if they introduce only
incremental improvements to the statistical model.

\subsubsection{The Compositional Prior}

The second half of the answer to the question of when a parameter reduces
the total description length relates to its description length, since
parameters are penalized by their description length in an attempt to bias
against over-fitting to the training data.\footnote{It is an interesting
  question whether the description length prior can be interpreted as a
  confidence test.  One way to view the learning problem is that for each
  possible parameter $w = w_1 \circ \ldots \circ w_n + \mbox{\sc p}_1 +
  \cdots + \mbox{\sc p}_{m}$ the learner is faced with the problem of
  determining whether the finite evidence indicates at some confidence
  level that the true probability of $w$ is greater than the probability
  defined by the parameters $w_1\ldots w_n$; if so, $w$ is justified.  In
  certain situations this condition can be formalized.  Imagine in the
  multigram model the problem of deciding whether to add a parameter
  $x_{12} = x_1 \circ x_2$, and suppose it will be added if at the 95\%
  confidence level $p(x_{12}) > p(x_1)p(x_2)$.  Given the number of words
  $N$, probabilities $p(x_1)$ and $p(x_2)$, and counts $c(x_1)$ and
  $c(x_{12})$, define $\hat{p}(x_{12}) = c(x_{12})/N$ and $\hat{p}(x_2|x_1)
  = c(x_{12})/c(x_1)$.  Then (see an introductory statistics textbook such
  as Keeping~\cite{Keeping62}) various assumptions and approximations lead
  to the condition to add $x_{12}$ if

  \[ \hat{p}(x_{12}) - p(x_1)p(x_2) > \frac{1.96
    \sqrt{p(x_{12})(1-\hat{p}(x_{2}|x_{1}))}}{\sqrt{N}}. \]

  \noindent This condition says that $x_{12}$ is justified if its empirical
  probability exceeds the model's prediction by more than a certain
  threshold that depends inversely on the amount of evidence.  If the
  representational prior could be justified as a confidence test, the
  numerator $1.96 \sqrt{p(x_{12})(1-\hat{p}(x_{2}|x_{1}))}$ would be a
  monotonic function of the description length of $x_{12}$, $-\log
  p(x_1)p(x_2)$, but it does not seem to be.  Thus, the representational
  prior is better viewed as combining a statistical test with a bias toward
  parameters with certain linguistic properties.} The length of a
representation is independent of the number of times a parameter is used,
so given enough evidence the benefits of any parameter will outweigh
its costs.  The description length of a parameter is mostly a function of
the length of its linguistic representation, since the cost of the
statistical information associated with each parameter tends to be
relatively small under efficient coding schemes.  This has several
implications:

\begin{itemize}
\item Since perturbations increase the representation cost, parameters are
  favored if they behave as expected given their parts; it requires more
  evidence to justify a parameter that introduces new linguistic behavior
  than one that does not.
\item Parameters are favored if they look like other parameters.  This
  follows from the fact that the learner is under an incentive to explain
  patterns in parameters as well as in the input.  If a parameter has a
  long description length, it indicates that the parameter doesn't fit into
  any discernible pattern found within other parameters.
\item Parameters are favored if they share information common to other
  parameters.  This can be viewed as a means to ensure that there is
  sufficient evidence to estimate the information in a parameter.  Less
  evidence is required to justify a parameter built from common parts
  because most of its properties are inherited from well-established
  parameters.
\end{itemize}

With this background in place, it is possible to look at several ways in
which the representational framework aids learning, and in particular
unsupervised learning.  These include the manner in which it allows for
incremental learning, explains how linguistic structure emerges from within
extralinguistic patterns, and separates on-line and off-line processing
issues.

\subsubsection{Incremental Learning}\label{incremental}

Section~\ref{mdl} argued that for language to be learnable from small
amounts of data, it must be the case that the learner chooses from among a
restricted set of grammars, the complexity of the set determined in part by
the amount of evidence available to the learner.  In such a situation
learning is incremental, with the size of the grammar increasing as more
evidence becomes available.  Incremental learning falls out naturally from
the compositional representation, since parameters do not introduce new
behavior so much as they group behaviors that are already present.

\begin{figure}[tbh]
\pageline
\begin{center}
\begin{tabular}{l|l}
Representation of the Input & Lexicon\\ \hline
(\word{t} $\circ$ \word{h} $\circ$ \word{e} $\circ$ \word{m} $\circ$
\word{o} $\circ$ \word{o} $\circ$ \word{n}) + {\sc moon}
& \word{t}, \word{h}, \word{e}, \word{m}, \word{o}, \word{n}\\
\\
(\word{the} $\circ$ \word{moon})
& \word{t}, \word{h}, \word{e}, \word{m}, \word{o}, \word{n}\\
& \word{the} = (\word{t} $\circ$ \word{h} $\circ$ \word{e}),\\
& \word{moon} = (\word{m} $\circ$ \word{o} $\circ$ \word{o} $\circ$ \word{n}) + {\sc moon}
\end{tabular}
\end{center}

\caption{\label{fig:incremental}Two lexicons for the
  input \word{the moon} with ``meaning'' {\sc moon}, where meanings are
  captured via perturbations.  The top lexicon, the simplest possible,
  consists only of terminals and must capture the meaning as an
  utterance-level perturbation.  The more mature lexicon on the bottom has
  grouped various terminals and moved the perturbation into the lexicon.}
\pageline
\end{figure}

In fact, this is not quite accurate given the framework as stated above.
This is because perturbations do not occur during on-line sentence
processing.  Hence, the simplest lexicons, containing only terminals, can
not be composed to explain an utterance that requires perturbations.  The
reason that perturbations are not permitted at the utterance level is that
it is not clear what interpretation would be given to them: the sorts of
perturbations in the lexicon that allow a phrase like \word{kick the
  bucket} to mean ``to die'' do not occur on an utterance by utterance
basis.  When someone says a sentence it does not mean different things at
different times randomly in ways that cannot be explained by the grammar
and the situation.  However, if the {\em learning mechanism pretends that
  perturbations occur at the utterance level}, treating them as a sort of
unpredictable noise, and {\em represents each utterance in exactly the same
  way that parameters are represented}, then (under the associative and
distributive assumptions) any utterance can be explained by any grammar.
Under such a scheme at the earliest stages of learning the grammar is the
simplest possible-- a lexicon that contains only terminals.  Each utterance
is analyzed as an essentially random sequence of terminals that undergo
random perturbations.  This randomness leads to long description lengths,
and as evidence is presented the learner is motivated to group terminals
and move perturbations into the lexicon to reduce the description length.
If learning were ever complete, the only use for perturbations at
the utterance level would be to explain random sound variations and other
noise-like behavior that is beyond the ability of any grammar to account
for.  Figure~\ref{fig:incremental} contains an example showing how
perturbations can be moved into the lexicon during learning.

\subsubsection{Extralinguistic Patterns}

In section~\ref{extralinguistic} it was argued that one of the fundamental
difficulties of unsupervised language acquisition is that learning systems
model patterns in the input signal regardless of whether their root cause
is linguistic in nature.  As a simple example, in the Brown corpus of
English text~\cite{Francis82}, the phrase \word{kicking the bucket} is used
five times.  That is surprisingly high, given the relative infrequency of
the words \word{kicking} and \word{bucket}.  A learner might (correctly)
take this frequency as a sign that \word{kicking the bucket} has a special
linguistic role, and include it in the lexicon; given an encoding scheme as
in the example in section~\ref{concat}, its inclusion will reduce the
description length of the input.  But the phrase \word{scratching her
  nose} also occurs in the Brown corpus five times.  This phrase has no
special linguistic role, and its unusual frequency follows from causes
external to language.\footnote{Indeed, from this bizarre but appropriate
  passage in the Brown corpus:

 \begin{quote}
   He could not make out, but he knew that again she was scratching her
   nose. Mollie the Mutton was scratching her nose.  The words ran crazily
   in his head: Mollie the Mutton is scratching her nose in the rain.  Then
   the words fell into a pattern: "Mollie the Mutton is scratching her
   nose, Scratching her nose in the rain. Mollie the Mutton is scratching
   her nose in the rain".  The pattern would not stop.
 \end{quote}
 } A language learner, faced with discriminating between these two phrases
on the basis of purely statistical information, has a
nearly impossible task.  Without access to meaning, both are extremely
similar-- relatively infrequent infinitive action verbs followed by
determiners and equally infrequent nouns referring to physical objects.

This implies that the language learner, at least in the early stages of
learning, can not identify all the linguistic parameters without also
identifying many false positives.  Here the compositional representation
offers a particularly pretty solution, by reducing if not eliminating the
undesirable consequences of including ``extralinguistic parameters'' in the
lexicon.  It was argued in section~\ref{extralinguistic} that most
extralinguistic patterns are built from linguistic units.  This is
certainly true of \word{scratching her nose}.  Given a reasonable grammar
(lexicon), almost certainly the optimal (shortest-length, most probable)
representation will decompose such parameters into linguistically
meaningful units.  In the lexicon, therefore, we would expect to find
\word{kicking the bucket} represented as something like \word{kicking}
$\circ$ \word{the} $\circ$ \word{bucket}, and \word{scratching her nose}
represented as something like \word{scratching} $\circ$ \word{her} $\circ$
\word{nose}.  In both of these cases the interpretation implied by the
representational framework is that the parameters inherit their properties
from their parts.  To understand the advantage this conveys, it is
important to recall the role of {\em unsupervised} learning, or in this
case, learning in absence of clues about word meanings.  It is to provide a
base linguistic structure for further learning.  In this case, if at a
latter stage of learning the learner is presented with the input
\word{Methuselah'll be kicking the bucket soon} and hints that it means
something like \word{Methuselah'll be dying soon}, then \word{kicking the
  bucket} provides the perfect point for the death meaning to be attached,
via perturbations.  In the case of \word{scratching her nose} the learner
will never have cause to introduce additional perturbations (beyond the
statistical information that caused the phrase to be included in the
lexicon in the first place), because the phrase behaves exactly as the
composition of its parts would imply.  The phrase will by default inherit
the correct interpretation, and act as if it were not in the lexicon at
all.

It is worth returning to a point made at the end of section~\ref{concat}.
Because the compositional framework eliminates most of the undesirable
consequences of having extralinguistic parameters like \word{scratching her
  nose} in the lexicon, the learner is essentially free to include them in
the lexicon.  In fact, because parameters have compact representations in
terms of other parameters, from the minimum description length standpoint
parameters are extremely cheap.  This allows the lexicon to model detailed
statistical properties of the input even if its underlying model of
language is poor, by multiplying the number of parameters in the lexicon.
This makes the framework an excellent choice for language modeling and
compression applications, and, as discussed in chapter~\ref{ch:Bayes},
helps ensure that the learner does not devote linguistic mechanisms to the
explanation of extralinguistic patterns.

\subsubsection{The Lexicon as Linguistic Evidence}

An important point made in section~\ref{extralinguistic} is that because
the majority of extralinguistic patterns are built upon linguistic
structure, they serve as evidence for linguistic parameters.  The
representational framework captures this intuition by forcing each
parameter to be represented in the same way as utterances from the input.
Common word sequences like

\begin{center}
\begin{tabular}{l}
\word{kicking the bucket}\\
\word{scratching her nose}\\
\word{walk the dog}\\
\word{waxed the car}\\
\word{caught a cold}\\
\ldots
\end{tabular}
\end{center}

\noindent are all likely to make their way into the lexicon, because with
suitable interpretation in a stochastic language model, they can be made to
reduce the description length of English input.  In the lexicon each must
be represented, and these representations contribute to the total
description length of the input.  A naive coding scheme that encodes each
component word independently (as with \word{kicking the bucket} =
\word{kicking} $\circ$ \word{the} $\circ$ \word{bucket} in the
concatenative model) fails to capture an important pattern, namely that
each of these parameters is a sequence of a verb followed by a noun phrase.
Because of this, under such a scheme the description length of these
parameters is longer than is necessary.  A better model (see
section~\ref{rep:cfg} below) that can represent \word{kicking the bucket}
as something like \word{$\langle$VP $\Rightarrow$ verb NP$\rangle$} $\circ$
\word{kicking} $\circ$ \word{$\langle$NP $\Rightarrow$ det noun$\rangle$}
$\circ$ \word{the} $\circ$ \word{bucket} can reduce the description length
of these parameters by taking into account the conditional dependency
between the three parts of speech, captured here by the rules
\word{$\langle$VP $\Rightarrow$ verb NP$\rangle$} and \word{$\langle$NP
  $\Rightarrow$ det noun$\rangle$}.  In this way, the existence of these
word patterns justifies the inference of these rules.  

It might be argued that there will be plenty of \word{verb det noun}
sequences in normal input to justify the creation of these syntactic rules,
independently of the need to represent the lexicon.  This is not
necessarily the case, for two reasons.  First, {\em given enough evidence}
learning mechanisms of the sort discussed here will incorporate {\em every}
particular instance of this general pattern into the lexicon, in an effort
to model the statistical properties of the input as closely as
possible.\footnote{Many other statistical induction schemes used for
  language inference have suffered from this problem; for examples see the
  results of Olivier~\cite{Olivier68} and Cartwright and
  Brent~\cite{Cartwright94}.  Their schemes, like the one presented in
  section~\ref{concat}, increase the number and length of the parameters
  learned as the size of the input increases, in an effort to model the
  statistics of the input as closely as possible (with a block code, in
  effect).  But since their parameters are not represented meaningfully in
  the lexicon, the ever lengthening parameters become ever more devoid of
  linguistic relevance.} More importantly, there are many linguistically
important parameters that manifest themselves only within other parameters.
Common examples include morphemes like {\em sub-} and {\em -ed}, and
syllables that are not also words.  As an extreme example, consider a case
in which the learner's evidence $U$ is a sequence of utterances $U'$
repeated twice: $U = U'U'$.  The learner, by placing $U'$ in the lexicon
(in the same way that the learner might memorize a long pattern like a song
or a prayer), can quickly halve the representation cost of the input.
After such a move, the only way for linguistically interesting learning to
take place from $U'$ is if it occurs in the lexicon.

\subsubsection{The Relation Between On-Line and Off-Line Processing}

It is important to understand the implications of representing parameters
in a certain way.  So far only two have been discussed.  The first is that
representations are the basis for codes, which define description lengths
and hence the prior for Bayesian inference.  Thus, the representation of
parameters in part determines the fitness of grammars.  The other
implication that has been mentioned, peculiar to the compositional
framework, is that in absence of conflicting evidence properties of
parameters are inherited; this has not been formalized in any sense.  The
compositional framework has two other important implications for the
learning mechanism that are worth mentioning briefly; both will be
explored in greater depth in the next chapter.

As discussed above, one way to allow for incremental learning is for the
learning mechanism to represent utterances in {\em exactly} the same way as
parameters, as perturbations of compositions.  This means that the same
processing mechanisms can be used for both the lexicon and on-line
processing.  While this may seem like a small point, it leads to very
simple learning methods that treat the parameter acquisition process as
that of memorizing common actions taken by the processing mechanism.
In fact, the computation of the expected change in description length from
adding or deleting a parameter is often quite simple, because the process
of adding or deleting can be thought of as merely moving parts of
representations back and forth between the lexicon and the processing
mechanism.

\begin{figure}[t]
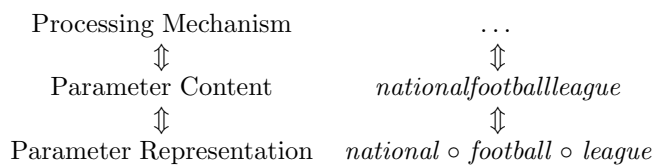

\pageline
\begin{center}
\begin{tabular}{cc}
Processing Mechanism & \ldots\\
$\Updownarrow$ & $\Updownarrow$ \\
Parameter Content & \word{nationalfootballleague}\\
$\Updownarrow$ & $\Updownarrow$ \\
Parameter Representation & \word{national} $\circ$ \word{football} $\circ$ \word{league}
\end{tabular}
\end{center}
\caption{\label{fig:proc} The representation of a parameter is conceptually
  separated from its content; the processing mechanism depends only on the
  content.  Therefore the representation is free to restructure so long as
  its content remains the same.}
\pageline
\end{figure}

A much more important implication of the framework is that, as mentioned at
the start of this section, the internal representation of a parameter does
not affect its use.  There are several ways to interpret this.  One is that
the representation of a parameter is strictly separated from the processing
mechanism.  One might imagine (as in figure~\ref{fig:proc}) that each
parameter has a special buffer that holds its content (the end result of
composing and perturbing) in whatever form is best suited for use by the
processing mechanism; this is convenient for computer implementation but
the extra storage requirements belie the compression properties of the
compositional representation.  In some ways a more attractive understanding
of the separation between the representation and use of parameters is that
the processing mechanism can directly interpret the representation of
parameters, but that the lexicon is free to write the content of a
parameter to a buffer and then reparse it, potentially restructuring the
parameter.  This separation between representation and processing mechanism
is supremely desirable.  To a large extent it separates the recognition of
patterns from the representation of patterns.  One way to build parameters
is to move common representational substructure into the lexicon.  But
alternatively, patterns in the input can be stored as parameters without
regard to representation.  For example, a child hearing a word like
\word{hypothermia} could memorize the sound pattern, storing it as
something like \word{h} $\circ$ \word{y} $\circ\cdots\circ$ \word{a}.
Later when the child has learned the roots \word{hypo} and \word{therm} the
parameter can be reanalyzed.  This reanalysis can take place without worry
that a change in representation will affect the representations of other
parameters and utterances.\footnote{This is not absolutely true.  To take
  full advantage of the inheritance framework, the learning mechanism must
  allow information to propagate up from components to parameters.
  Changing the representation changes the information that propagates up.
  This is not an issue if the parent parameter requires the information,
  since then the representation is constrained.  But if the parameter has
  the equivalent of ``don't cares'', then some extra complexities are
  introduced.} In contrast, many other representational frameworks that
merge the representation of patterns with on-line processing, such as
neural networks, have difficulty changing internal representations, because
in intermediate stages their performance is degraded (the pervasive
local-minima problem; see for example de Marcken~\cite{deMarcken95b}).

\subsection{Language}\label{motiv:linguistics}

All of the above arguments in favor of the representational framework
center around learning issues and are independent of ``linguistics''.  It
remains important that the compositional prior favor linguistically
plausible parameters.  Here several linguistic arguments in favor of the
representational framework are presented.

\subsubsection{Language as a Hierarchy}

The hierarchical nature of the representational framework mirrors the
hierarchical nature of language.  An utterance exists at many levels of
representation (see the example at the start of section~\ref{ex-param}):
linguistic constraints are defined on sequences of phonetic features; on
sequences of syllables; on trees of morphemes and words; etc.  Although
there is evidence that these different levels of representation are at
least partially orthogonal,\footnote{For example, in languages like Spanish
  syllables can cross word boundaries~\cite{Harris93}.  See the literature
  on {\em bracketing paradoxes}~\cite{Kenstowicz94,Spencer91}.}  to a first
approximation they are structured as trees, with constraints at one level
exerting themselves within boundaries imposed at other levels.  For
example, in English syllabification generally occurs within word
boundaries.  For this reason, words in English can usually be decomposed
into a sequence of syllables.  There is nothing implausible in English
about a word like \word{lublick}, but \word{ludbnick} would not be expected,
because \word{bnick} violates syllabicity constraints.  By forcing
parameters to be represented in terms of parts, the framework captures this
intuition.  A word like \word{ludbnick} will have a comparatively long
description because it can not be represented in terms of knowledge about
syllables.  As a consequence, more evidence will be required to justify the
inclusion of the word in the lexicon.  In the same way, \word{kicking the
  bucket} is prefered over \word{icking the bucke} as an English parameter,
another fact that follows from the compositional representation.

\subsubsection{Irregular Forms and Compilation}

One of the great mysteries of language is how the processing mechanism
rapidly reconstructs word sequences from speech.  During language
production the underlying forms of words are transformed by various
corrupting and distorting morphological, phonological and phonetic
processes.  Standard computational models of recognition attempt to invert
these processes~\cite{Anderson88b,Koskenniemi83,Klatt92} during
recognition; since the forward processes are many-to-one, inversion seems
to require (expensive) search.  And yet speech recognition is quick and
easy for people.  The representational framework offers a partial
explanation.  Many of the corrupting processes are non-deterministic, but
not entirely random.  This is especially true for phonological and phonetic
alterations that occur during fast speech.  For example, \word{want to} is
often (though not always) pronounced \word{wanna}, and \word{grandpa} is
often pronounced \word{grampa}.  In both of these cases the sound changes
are naturally accounted for by certain phonological assimilation and
deletion mechanisms that can be treated as perturbations.  These
perturbations are not entirely random, and therefore in terms of
statistical language modeling it behooves the learner to move them into the
lexicon, building parameters like

\begin{center}
\begin{tabular}{l}
\word{wanna VP} = (\word{$\langle$VP $\Rightarrow$ Verb to VP$\rangle$}
  $\circ$ \word{want}) + {\sc sound-change} + {\sc freq}\\ 
\word{grampa} = (\word{grandpa}) + {\sc sound-change} +
{\sc freq}
\end{tabular}
\end{center}

\noindent where in the first case the sound change is captured by an
assimilation of nasality from the /\unin/ to the /\unit/ and a reduction of
the vowel in \word{to}, and in the second case also an assimilation of the
nasality of the /\unin/ to the /\unid/ (see de Marcken~\cite{deMarcken95d}
for a more detailed definition of a phonological perturbation scheme that
can account for such phenomena).  These parameters inherit their syntax and
meaning from underlying (uncorrupted) words, and yet contain statistical
information that indicates that the sound changes are to be expected.
Given the proper implementation of the framework, this information can
either be used to direct and constrain search during word recognition, or
render it unnecessary (because changes have been compiled into underlying
forms).

In fact, the framework's ability to compile out common patterns of usage
extends well beyond phonology.  Another example from
figure~\ref{fig:perturbations} is 

\begin{center}
  \word{take off NP} = (\word{$\langle$VP $\Rightarrow$ Verb Prep NP$\rangle$}
  $\circ$ \word{take} $\circ$ \word{off}) + {\sc meaning} + {\sc freq}.
\end{center}

\noindent Here a verb-particle pair is explained in terms of standard
syntactic rules.  This representation has many advantages: it explains why
the case of the noun phrase is determined by the particle (this can be
tested in languages other than English), explains why particles are chosen
from among the class of prepositions, etc.  At the same time, the fact that
this parameter compiles out a sequence of syntactic compositions into the
surface pattern \word{take off NP} explains why \word{take off} is
recognized so easily as a single linguistic entity.

The idea that common changes are compiled out in the lexicon receives
internal support in the representational framework.  Many frequent words
incorporate unusual sound changes:\footnote{It is no surprise that
  perturbations are concentrated on frequent words.  In this learning
  framework it takes substantial evidence to justify perturbations,
  evidence available for frequent words but not for rare ones that are
  learned from small numbers of examples.} at the top of the list in
English are suppletive alternations such as
\word{be-am-is-was-were-are-being} and \word{gone-went}.  Slightly less
common examples include \word{want to-wanna}, \word{going to-gonna} and
irregular alternations like \word{think-thought} and \word{catch-caught}.
If these sound changes are not compiled into the lexicon but handled
on-line by the processing mechanism, then they should be predicted with
frequency proportional to the frequency of the words in which they occur.
But Baayen and Sproat~\cite{Baayen96} have determined that the best
indicator of the frequency of such phenomena in new or unknown words is the
frequency of the phenomenon in the lexicon, unweighted by word frequency.
This is exactly as would be expected if the changes are compiled into the
lexicon.

\subsubsection{Diachronic Arguments}

A final argument for the composition and perturbation framework arises from
the historical evolution of language.  Most irregular forms and idioms are
not completely devoid of internal structure.  Even the \word {to be}
paradigm-- \word{be}, \word{am}, \word{is}, \word{are}, \word{was},
\word{were}, \word{being}-- has some regularities, that reflect the
historical derivation of the paradigm.  Over time irregularities are
introduced into commonly used parameters, and are slowly weeded out of
rarer ones (enabling them to be learned from smaller amounts of data).  In
this way \word{Wednesday} has acquired a meaning and pronunciation distinct
from its Scandinavian root \word{Wodnesdaeg}, while the spelling of
\word{night} tends towards the more intuitive \word{nite}.  Similarly,
\word{kicking the bucket} has attained a meaning that no longer has any
obvious relation to its original usage.  If perturbations are viewed as a
means of capturing changes that occur over time, then the representational
framework can be seen as a means for ontogeny to recapitulate phylogeny:
the learner acquires words by representing them in a manner that reflects
their historical derivation.  This is of course not because the learner has
access to true history of the target language.  Instead, the manner in
which language evolves leads to shared patterns among parameters, that the
compositional framework can use to shorten descriptions.  In this way,
parameters are favored if they can be explained in terms of expected
historical processes.

\section{Coding}\label{coding}

As has been mentioned many times, codes and probabilities are fundamentally
and simply related by Shannon's source coding theorem~\cite{Shannon48},
which says that a code can be designed for a distribution such that the
expected description length under the code is almost exactly the entropy of
the distribution; as a practical matter this implies that a code can be
designed such that the length of a description of $u$ almost exactly equals
$-\log p(u)$, and that on average no code can do better.  Hence, thinking
of minimizing code lengths is almost always equivalent to thinking of
maximizing probabilities, and vice versa.  Nevertheless, it is often the
case that one view is more intuitive than another in a given situation.  For
example, the probability of an utterance $u$ under the multigram model was
expressed as

\[ p(u) = \sum_{n} p(n) \sum_{w_1 \ldots w_n\ s.t.\ u =
  w_1\circ\cdots\circ w_n} p(w_1)\cdots p(w_n). \]

\noindent Here probabilities are summed over multiple possible derivations
of the utterance $u$, or thinking in terms of codes, multiple
representations of $u$.  Thus, the fact that there are multiple
representations for an utterance should mean that it can be coded in fewer
bits than if there were only one; yet it is not obvious how to design a
coding scheme that fulfills this requirement.  In fact it is
possible,\footnote{One way to see this is to imagine the choice between two
  equal-length representations as a ``free bit'' of information that is
  conveyed to the decoder.  This bit can be applied to other parts of the
  encoding.} although usually impractical.  For most language modeling
applications this is not an important issue: a small number of derivations
tend to be much more probable than others, and the difference in
probability between the sum over all derivations (the {\em complete}
probability) and the single best derivation (the {\em maximum-likelihood},
or {\em Viterbi}, probability) is usually insignificant (even a factor of
two is only a single bit, a small amount relative to the total cost of
encoding a parameter).  This example illustrates one reason why it is often
more convenient to think in terms of probabilities than codes.  Another
reason concerns roundoff in codeword length.  In the example in
figure~\ref{fig:model} integral-length codewords are used.  But in the
ideal situations codewords are chosen according to the the equation $l(w) =
-\log p(w)$, which does not in general imply integral length codewords.  In
fact codes can be designed that circumvent this problem (arithmetic
codes~\cite{Pasco76,Rissanen76} are a practical solution).  But again,
since the principal purpose of codes here is to compute description lengths
for use in the already heuristic MDL criterion, it is much more convenient
to simply ignore details of code construction and use $-\log p(w)$ directly
in description length computations.  Of course, for compression
applications it {\em is} necessary to design concrete and practical coding
schemes, but the inefficiencies introduced are usually small relative to
the ``fundamental'' cost $-\log p(w)$.

Just as it can be more convenient to think in terms of probabilities than
codes, the converse is also true.  For example, in the above equation there
is a probability $p(n)$ that determines how many parameters are output in
the generation of an utterance.  Rather than worry about the estimation and
representation of this distribution, it is more convenient to realize that
in practice most parameters are built from a small number of others (less
than four).  Thus, two bits is probably an upper bound on the mean cost of
encoding the length of a parameter; as this is small relative to the cost
of specifying the components, it can be safely ignored.  In fact this will
lead to much more efficient learning algorithms, and in those applications
(such as text compression) where it is important to completely specify the
code, most any simple code can be used to encode the length of each
parameter.

\section{Examples}\label{rep-examples}

The representational framework abstracts from details of coding and the
composition and perturbation operators.  Thus far only one instantiation of
the framework has been discussed in any detail, the concatenative model of
section~\ref{concat}.  Below, it is expanded upon and three variations are
presented.  The first extends the composition operator by grouping
parameters into classes.  These classes act as the nonterminals of
traditional context-free grammars, and the composition operator is
nonterminal expansion.  This model can encode linguistically important
statistical dependencies that can not be captured succinctly in the
concatenative model.  The second instantiation varies along a different
dimension: it introduces a perturbation operator that can be used to learn
artificial representations of meaning, and serves as an example of how the
learning framework can be used to solve the ``complete'' language learning
problem.  The final variation, discussed only briefly, is a perturbation
operator that encodes significant phonological knowledge, and that can be
used with other extensions to learn directly from raw speech signals.

\subsection{Composition by Concatenation}\label{rep:concat}

Section~\ref{concat} introduced a multigram model in which the composition
operator $\circ$ is concatenation, terminals are characters, and a
stochastic interpretation is defined by associating with each parameter a
probability.  The complete probability of an utterance (or a parameter) $u$
is therefore

\[ p(u) = \sum_{n} p(n) \sum_{w_1 \ldots w_n\ s.t.\ u =
  w_1\circ\cdots\circ w_n} p(w_1)\cdots p(w_n). \]

\subsubsection{Relation to Other Finite-State Models}

The multigram model is finite-state, in that it has only a finite memory of
previous events.  This memory extends at most the length of the longest
parameter.  As a finite-state model, it has obvious similarities to other
finite-state models such as Markov models (MMs) and hidden Markov models
(HMMs)~\cite{Bell90,Rabiner89}.  Both Markov models and hidden Markov
models define a stochastic model on top of a finite state machine, where
the state $q_i$ of the system at time $i$ is drawn from a finite set $Q$.
At each time step a symbol (a character) $o_i$ is generated in accordance
with a distribution that depends only on the state $q_i$.  This state is a
stochastic function of the state at the previous time step, $q_{i-1}$.  In
hidden Markov models the stochastic transition matrix $a_{jk} = p(q_i =
k|q_{i-1} = j)$ is arbitrary.  In Markov models the state $q_i$ is defined
by recent outputs, $q_i = o_{i-m}\ldots o_{i-1}$, where $m$ is the order of
the Markov model (more general {\em context
  models}~\cite{Bell90,Rissanen83} select the context $o_{i-m}\ldots
o_{i-1}$ from among a set of variable-length suffixes).  In the multigram
model the generating parameter $w$ acts as a hidden state, though the fact
that the output function is deterministic gives the model the feel of an
ordinary Markov model.

The fact that parameters are generated independently in the multigram model
means that there are distributions for which no multigram model performs as
well as a more general MM or HMM.  For example, a simple first order Markov
model that can not be simulated by any multigram is one in which characters
are divided into consonants and vowels, and generated in a manner that
ensures that consonants and vowels alternate.  However, by expanding the
parameter set to include ever longer strings, a multigram can be made to
approach the entropy of any MM or HMM arbitrarily closely.  As a
consequence, in practical language modeling applications multigrams can be
competitive with more general finite-state models.

Multigrams have several advantages for learning over other types of
finite-state models.  Most importantly, they are easy to assign linguistic
interpretations to, because parameters can be associated with words.  In
Markov models and hidden Markov models each state is given equal status;
potential linguistic boundaries can only be defined by ad-hoc functions
applied to transition probabilities.  Furthermore, for language modeling
applications multigrams often have much smaller representations than
equivalent MMs and HMMs.  For example, Ristad and Thomas~\cite{Ristad95}
use the MDL criterion to learn a context model.  The equivalent information
found in the single multigram parameter \word{Mississippi} is in their
context model captured by many parameters:

\begin{center}
\begin{tabular}{l}
$p(\mbox{\word{M}}|\cdot) = \ldots $\\
$p(\mbox{\word{i}}|\mbox{\word{M}}) = \ldots $\\
$p(\mbox{\word{s}}|\mbox{\word{Mi}}) = \ldots $\\
$p(\mbox{\word{s}}|\mbox{\word{Mis}}) = \ldots $\\
$p(\mbox{\word{i}}|\mbox{\word{Miss}}) = \ldots $\\
$p(\mbox{\word{s}}|\mbox{\word{Missi}}) = \ldots $\\
$p(\mbox{\word{s}}|\mbox{\word{Missis}}) = \ldots $\\
$p(\mbox{\word{i}}|\mbox{\word{Mississ}}) = \ldots $\\
$p(\mbox{\word{p}}|\mbox{\word{Mississi}}) = \ldots $\\
$p(\mbox{\word{p}}|\mbox{\word{Mississip}}) = \ldots $\\
$p(\mbox{\word{i}}|\mbox{\word{Mississipp}}) = \ldots $\\
\end{tabular}
\end{center}

\noindent The cost of representing these (redundant) context strings is
high.  This unnecessarily multiplies the cost of parameters, adversely
affecting the performance of the context model.

\subsubsection{Coding}

Although section~\ref{coding} has argued that details of coding schemes are
not an important issue with respect to learning, a coding scheme for the
concatenative model is presented in enough detail here to illustrate how an
efficient coding scheme can be created for the purposes of text
compression.  In chapter~\ref{ch:results} this scheme is in fact used to
prove that the concatenative model makes for an extremely good
compression algorithm.  

In the concatenative representation, the input and each parameter are
described by the composition of a sequence of parameters.\footnote{In fact,
  there are many different representations for the input and each
  parameter.  For coding purposes, only a single representation is
  considered, the most probable representation.} The coding scheme
described here references each parameter by a codeword.  Codewords are
determined by a Huffman code~\cite{Huffman52} that is constructed in
accordance with parameter frequencies.  In practice, Huffman codes very
closely approach the theoretical optimum efficiency that would result from
non-integral length codewords; this is true both because the number of
parameters is usually large and because parameter frequencies follow a
smooth inverse-frequency distribution~\cite{Zipf49}.  The number of
parameters in each parameter representation is also coded via a Huffman
code.  The two Huffman codes must themselves be specified; fortunately
Huffman codes can be specified quite efficiently so long as the objects
they reference are ordered by frequency.  In particular, since the length
of codewords is monotonically decreasing function of frequency, codes can
be assigned in increasing lexicographic order to parameters of decreasing
frequency.  Then, a Huffman code is completely defined by specifying the
number of codewords of each length.  For codes over large numbers of
objects this is a very compact representation.  Finally, each terminal
codeword must be identified and associated with its denotation, namely its
character.

\begin{figure}[tbh]
\pageline
\begin{small}
\begin{tabbing}
{\em Specification of Huffman code for parameter codewords:}\\
\hspace{.2in} \= Write the Elias-coded length $l$ of the longest parameter codeword.\\
\> For each length $n$ in the range $1\ldots l$,\\
\> \hspace{.2in} \= Write an $n$-bit integer specifying the number of
codewords of length $n$.\\
{\em Specification of Huffman code for representation length codewords:}\\
\> Write the Elias-coded length $l$ of the longest representation length codeword.\\
\> For each length $n$ in the range $1\ldots l$,\\
\> \> Write an $n$-bit integer specifying the number of codewords of length $n$.\\
\> \> \hspace{.2in} \= For each representation length with codeword of length $n$,\\
\> \> \> \hspace{.2in} \= Write an Elias-coded integer specifying the
representation length.\\
{\em Association of terminals with parameter codewords:}\\
\> Write 128 bits that specify whether terminals 1\ldots128 are used in the
code.\\
\> In predetermined order, for each terminal used in code,\\
\> \>  Write the codeword for that terminal.\\
{\em Representations of nonterminal parameters:}\\
\> In order of decreasing parameter frequency, for each nonterminal parameter,\\
\> \> Write the codeword for its representation length. \> \hspace{4.5in} \= (A)\\
\> \> For each component parameter in its representation,\\
\> \> \> Write the codeword for that parameter. \> (B)\\
{\em Representation of input:}\\
\> Write the Elias-coded number of parameters in the representation of the input.\\
\> For each parameter in the representation,\\
\> \> Write the codeword for that parameter. \> \> (C)
\end{tabbing}
\end{small}
\caption{\label{fig:code}A complete and compact coding scheme for any
  reasonably sized input and lexicon.  In practice lines (B) and (C)
  account for the overwhelming majority of the total description length.} \pageline
\end{figure}

Figure~\ref{fig:code} summarizes the coding scheme, which assumes an
inventory of 128 terminals.  In practice lines (B) and (C) account for the
overwhelming majority of the total description length, dwarfing the only
other factor that grows super-logarithmically with the size of the lexicon,
line (A).  This motivates and justifies the following derivation:

\begin{eqnarray}
-\log p(u) & = & - \log \sum_{n} p(n) \sum_{w_1 \ldots w_n\ s.t.\ u =
  w_1\circ\cdots\circ w_n} p(w_1)\cdots p(w_n) \label{eq:real} \\
& \le & \sum_{n} \left( - \log p(n) + - \log \sum_{w_1 \ldots w_n\ s.t.\ u =
  w_1\circ\cdots\circ w_n} p(w_1)\cdots p(w_n) \right) \nonumber \\
& \approx & - \log \sum_{w_1 \ldots w_n\ s.t.\ u =
  w_1\circ\cdots\circ w_n} p(w_1)\cdots p(w_n).\label{eq:upper}
\end{eqnarray}

\noindent The first step uses Jensen's inequality to substitute a close
upper bound for the true description length.  The approximation in the
second step, which assumes that $-\log p(n)$ is insignificant in comparison
with the cost of writing down parameter codewords, dramatically simplifies
the computation of this upper bound, for reasons that will become clear in
the next chapter.  The learning algorithms presented there minimize
\ref{eq:upper} rather than the true description length of
equation~\ref{eq:real}.

\subsubsection{Deficiencies}

As chapter~\ref{ch:results} will show, the simple concatenative model has
surprising statistical modeling power, and is remarkably effective at
learning linguistic parameters.  Nevertheless, it does suffer from a number
of fundamental deficiencies.  Abstracting from its obvious linguistic
shortcomings, two issues stand out above all others.  First, it can not
describe statistical dependencies except by referring to individual
parameters.  This prevents it from describing relationships that are true
of broad classes of objects.  For example, it can not express the simple
syntactic rule \word{NP $\Rightarrow$ Det Noun}.  As a consequence, in
practice it builds up multitudes of highly redundant parameters

\begin{center}
\begin{tabular}{llll}
\word{the car} & \word{a car} & \word{some people} & \word{a dog}\\
\word{the fliers} & \word{any one} & \word{an apple} & \word{some apples}\\
\word{many people} & \word{no one} & \word{few apples} & \word{the dogs}\\
\word{\ldots}
\end{tabular}
\end{center}

\noindent that still can't explain why new nouns and determiners should
fall into the same pattern.  Similarly simple phonological constraints
between phonological classes can not be described, and there is no way to
describe the fact that a phrase like \word{red apples} behaves very much
like \word{apples}.  This problem will be partially addressed by the next
instantiation of the representational framework, in which the grammar can
describe and make reference to classes of objects.

The second fundamental deficiency, one that remains true of the other
instantiations discussed in this thesis, is that the chained stochastic
model\footnote{A chained model like a Markov model, hidden Markov model, or
  context-free grammar is one in which the probability of a derivation is
  computed by successively multiplying conditional probabilities that
  reflect subcomponents of the derivation.} provides no way to combine
orthogonal knowledge sources to improve coding efficiency, without
introducing redundant parameters.  Consider a model in which one parameter
captures the fact that the determiner \word{an} usually precedes vowels,
and another parameter captures the fact that it usually modifies singular
nouns.  The parameters can not be combined to explain the double
improbability of a sequence like \word{an cows} without creating a third
parameter (that can inherit from only one of the first two) that explicitly
refers to the class of singular nouns that start with vowels.  Language
models that can combine multiple knowledge sources are described by Della
Pietra {\em et al.}~\cite{Berger96,DellaPietra95}, but result in complex and
computationally burdensome learning algorithms.

\subsection{Composition by Substitution}\label{rep:cfg}

\newcommand{\cfgword}[1]{{$\langle$\word{#1}$\rangle$}}

One of the most significant deficiencies with the concatenative model is
that it can not capture relations that hold of broad classes of objects
without multiplying the number of parameters in the grammar.  This section
briefly explains how this can be partially remedied by basing the
composition operator on stochastic context-free grammars
(SCFGs~\cite{Baker79,Jelinek90}).  Although we have performed successful
experiments with the type of model described here, a number of fundamental
deficiencies remain, and it is presented only as an illustration.

The parameters of the concatenative model all fall under a single class, in
the sense that their probabilities are all defined with respect to one
another.  Suppose that the number of classes is increased, and parameters
include information specifying the class they are in.  Possible classes
include nouns, verbs, consonants, vowels, days of the week, etc.  Write $w
= X\Rightarrow\lambda$ to mean a parameter $w$ with class $X$ and pattern
$\lambda$.  Furthermore, suppose that each parameter now captures a pattern
over {\em both terminals and classes}.  For example, a (partial) grammar
might look something like

\begin{center}
\begin{tabular}{lclc}
Parameter & Prob. & Parameter & Prob.\\ \hline
\word{NP $\Rightarrow$ Det Noun} & $(1)$ &\word{Noun $\Rightarrow$ caterpillar}&$(\frac{1}{4})$\\
\word{VP $\Rightarrow$ Verb to VP} & $(\frac{1}{2})$ & \word{Noun $\Rightarrow$ hat}&$(\frac{3}{4})$\\
\word{VP $\Rightarrow$ take off NP}&$(\frac{1}{2})$ & \word{Det $\Rightarrow$ the}&$(\frac{1}{3})$\\
\word{Verb $\Rightarrow$ want}&$(\frac{1}{3})$ & \word{Det $\Rightarrow$ my}&$(\frac{1}{3})$\\
\word{Verb $\Rightarrow$ ask}&$(\frac{2}{3})$ & \word{Det $\Rightarrow$ a}&$(\frac{1}{3})$
\end{tabular}
\end{center}

\noindent where classes are written with an upper-case letter.  Notice that
the probabilities of all parameters with a common class sum to one.  The
probability of an utterance $u$ under a grammar $G$ is defined by a
rewriting process that starts from a single distinguished class $R$.  In
particular, $R$ is used to initialize a sequence $\Psi$: $\Psi = R$.
Generation proceeds as follows:

\vspace{-.2in}
\begin{enumerate}
\item If $\Psi$ consists only of terminals, let $u = \Psi$.  Stop.
\item Otherwise, let $\Psi = \alpha X\beta$ where $\alpha$ is the longest
  class-free prefix of $\Psi$ and $X$ is a class.  Choose with probability
  $p_G(w)$ any parameter $w = X\Rightarrow\lambda$ and let $\Psi =
  \alpha\lambda\beta$.  Go to step 1.
\end{enumerate}
\vspace{-.2in}

This process generates a sequence of terminals $u$ through successive
substitution.  Each step can be thought of as the application of a
(non-associative) composition operator.  For example, if the distinguished
class is VP then the utterance \word{want to take off my hat} can be
represented as the composition of six of the above parameters:

\begin{center}
\begin{tabular}{l|ll}
Representation & Derivation So Far ($\Psi$) & Probability\\ \hline
\cfgword{VP $\Rightarrow$ Verb to VP} $\circ$ & \word{Verb to VP} & $\frac{1}{2}$\\
\cfgword{Verb $\Rightarrow$ want} $\circ$ & \word{want to VP} & $\frac{1}{2}\frac{1}{3} = \frac{1}{6}$\\
\cfgword{VP $\Rightarrow$ take off NP} $\circ$ & \word{want to take off NP} & $\frac{1}{2}\frac{1}{3}\frac{1}{2} = \frac{1}{12}$\\
\cfgword{NP $\Rightarrow$ Det Noun} $\circ$ & \word{want to take off Det Noun}&$\frac{1}{2}\frac{1}{3}\frac{1}{2}1 = \frac{1}{12}$\\
\cfgword{Det $\Rightarrow$ my} $\circ$ & \word{want to take off my Noun}&$\frac{1}{2}\frac{1}{3}\frac{1}{2}1\frac{1}{3} = \frac{1}{36}$\\
\cfgword{Noun $\Rightarrow$ hat} & \word{want to take off my hat}&$\frac{1}{2}\frac{1}{3}\frac{1}{2}1\frac{1}{3}\frac{3}{4} = \frac{1}{48}$\\
\end{tabular}
\end{center}

As defined so far, this model of language is simply an SCFG.  To give it
the power of the compositional framework it is necessary to find a way to
represent parameters in terms of other parameters.  Several examples were
given at the start of this chapter, for example

\vspace{-.1in}
\begin{center}
\cfgword{VP $\Rightarrow$ Verb Prep NP} = \cfgword{VP $\Rightarrow$ Verb PP} $\circ$ \cfgword{PP $\Rightarrow$ Prep NP}.
\end{center}
\vspace{-.1in}

Given the generation process defined above this is not actually a valid
representation, since it has not exhaustively expanded the classes.  To
handle such cases, introduce for every class $X$ a special ``stop''
parameter $X\Rightarrow\diamond$.  Expanding a class with the stop
parameter marks the class to remain in the final sequence.  Thus,

\vspace{-.2in}
\begin{eqnarray*}
\lefteqn{\mbox{\cfgword{VP $\Rightarrow$ Verb Prep NP}} =}\\
& & \mbox{\cfgword{VP $\Rightarrow$ Verb PP}} \circ
\mbox{\cfgword{Verb $\Rightarrow \diamond$}} \circ\\
&& \mbox{\cfgword{PP $\Rightarrow$ Prep NP}} \circ
\mbox{\cfgword{Prep $\Rightarrow \diamond$}} \circ
\mbox{\cfgword{NP $\Rightarrow \diamond$}}.
\end{eqnarray*}
\vspace{-.1in}

There are many more details that need to be filled in.  For example, how is
the class of a parameter specified, and is it somehow determined by the
classes of the parameters used in the representation?  How are parameters
represented that do not have obvious tree structure, such as \word{Noun
  $\Rightarrow$ caterpillar}?  Rather than provide answers to these
questions here (there are many possible answers), we step back and look at
what extensions of this sort offer, and what their shortcomings are.

The principal advantage of this substitution model is that it allows
patterns over broad classes of objects to be captured.  This makes for more
succinct grammars with better generalization properties.  For example, in
tests we have performed with this type of model patterns that are learned
include number sequences like {\tt \$D,DDD.DD} and {\tt (D0D) DDD-DDDD},
where {\tt D} is a class that has been learned and includes the digits
0,\ldots, 9.\footnote{The telephone-number pattern {\tt (D0D) DDD-DDDD} is
  specialized to include a 0 in the area code, since all U.S.\ area codes
  contain either a 0 or a 1 in second position.}  Such patterns
substantially improve grammars' ability to predict the behavior of digits.
To achieve the equivalent in the concatenative model would required a
parameter for every possible telephone number!

Implementing the substitution model on top of the compositional framework
means that dependencies between successive class expansions can be modeled
without sacrificing linguistic structure.  For example, \word{want to take
  off my hat} may occur in some document with surprising frequency.  A
learner using a SCFG in an ordinary way could account for this fact by
adding a long, flat rule \word{VP $\Rightarrow$ want to take off my hat};
in doing so all of the linguistic structure within the phrase will be lost.
In the compositional framework the phrase will be represented in the
grammar in terms of other parameters, implicitly defining a tree structure
over the words.  There are many similarities between this type of model and
tree-grammars~\cite{Joshi75}.

However, the substitution model as defined above is not pursued further in
this thesis, because it has significant linguistic and statistical
shortcomings, and is not a sufficient improvement over the concatenative
model to warrent extensive investigation.  In particular, it has the
fundament flaw that it assigns every linguistic object to a single class.
But in fact every linguistic object falls into many ``classes''.  For
example, a phrase like \word{red apples} is a noun phrase, and also a
plural noun phrase, and a phrase about apples, and a phrase about red
apples, and so on.  Another parameter should be able to refer to any subset
of these properties when defining a pattern.

\subsection{Learning from Multiple Input Streams}\label{rep:meaning}

This section extends the concatenative model with a perturbation operator
that endows parameters with artificial representations of
meaning.\footnote{The extensions apply identically to the substitution
  model, but the concatenative model makes for a simpler exposition.} This
extended model can be applied to the complete language acquisition task of
learning to map between sound and meaning.

Recall that in the MDL framework learning is equivalent to signal
compression.  Up until now the only signal that has been considered is $U$,
the sequence of utterances.  In the real language acquisition problem the
audio signal is paired with other input.  Let us assume for simplicity's
sake that this other input can be distilled in a sequence of utterance
``meanings'' $V$.  The language learner's goal is to learn the relation
between $U$ and $V$.  One way to capture this goal without leaving the MDL
framework is for the learner to compress the pair $U$ and $V$
simultaneously.  Ignoring for the moment the cost of parameters, this is
accomplished by minimizing the entropy $H(U,V)$ of the learner's joint
model of $U$ and $V$.\footnote{In reality, the minimization is of the
  cross-entropy between the learner's model and the true distribution, but
  it is convenient to drop the distracting {\em cross-} terminology.} This
entropy can be rewritten

\begin{equation}
 H(U,V) = H(U) + H(V) - I(U,V).\label{eq:joint}
\end{equation}

\noindent where $H(U)$ and $H(V)$ are the marginal entropies of the two
signals and $I(U,V)$ is the mutual information between them.  The learning
framework as discussed so far devotes its efforts to minimizing $H(U)$,
which (as is apparent from equation~\ref{eq:joint}) is one part of
minimizing $H(U,V)$.  Similar strategies could be applied to $H(V)$,
compressing the two signals independently.  But if there is mutual
information $I(U,V)$ between the two signals, as would be expected in the
language acquisition problem, the learner can do better yet by compressing
both signals simultaneously.  Here, this will be accomplished by attaching
both meaning and sound information to parameters.  A single sequence of
parameters then suffices to represent both $U$ and $V$, as in
figure~\ref{fig:incremental}) Thus (allowing for perturbations at the
utterance level),

\[ (u, v) = w_1 \circ \cdots \circ w_n + \mbox{\sc p}_1 + \cdots +
\mbox{\sc p}_m. \]

\newcommand{\sem}[1]{{\tt #1}}
\newcommand{\sems}[1]{\{\sem{#1}\}}
\newcommand{\sword}[2]{{\em #1} \sems{#2}}

The goal here is to explore the induction of word meanings in as abstract a
manner as possible.  This motivates a simple and obviously toy
representation for meanings: the meaning of an utterance is merely a set of
arbitrary symbols (call them {\em sememes} for convenience).  For example,
a possible meaning for the sentence \word{john walked} is \sems{john walk}.
Here the sememes \sem{john} and \sem{walk} have no inherent denotation-
they are gensyms.  In examples and tests, utterance meanings will be
constructed in such a way that sememes can be associated in an intuitive
fashion with meaning-bearing linguistic units.  Sememe sets are unordered,
and therefore the most natural extension of the concatenative composition
operator is one in which the meaning of the composition of two parameters
is the union of the meanings of each parameter.  Writing a parameter with
character sequence $x$ and sememe set $s$ as $(x,s)$, composition is
therefore defined $(x,s) \circ (y,t) = (xy,s\cup t)$.  Perturbations add or
delete sememes from the default meaning of a composition.  Terminals are
defined to have empty sememe sets.  Figure~\ref{fig:meaning} presents
various examples of the use of this composition and perturbation
scheme.

The naivete of the meanings-as-sets representation does not imply that it
is without value.  It captures the fundamental aspect of semantic
acquisition, the apportionment of primitives in utterance meanings to
smaller linguistic units.  It is generally compositional (as with most
theories of semantic representation) yet acknowledges the possibility that
the meaning of a structure might not follow from its parts, which many more
complicated theories do not.  Siskind~\cite{Siskind94} argues that once
semantic symbols have been apportioned, it is a relatively trivial matter
to learn the relational structures found in more complicated semantic
representations based on tree-like functional composition.

\begin{figure}[tbh]
\pageline
\begin{center}
\begin{tabular}{ll}
Parameter & Representation \\ \hline
\sword{cat}{cat} & \sword{c}{} $\circ$ \sword{a}{} $\circ$ \sword{t}{} + \sem{cat}\\
\sword{cats}{cat} & \sword{cat}{cat} $\circ$ \sword{s}{}\\
\sword{blueberry}{blue berry soft} & \sword{blue}{blue} $\circ$
\sword{berry}{berry} + \sem{soft}\\
\sword{strawberry}{red berry sweet} & \sword{straw}{straw} $\circ$ \sword{berry}{berry}
 + \sem{red} + \sem{sweet} - \sem{straw}\\
\sword{cranberry}{red berry tart} & \sword{c}{} $\circ$ \sword{r}{}
$\circ$ \sword{a}{} $\circ$ \sword{n}{} $\circ$ \sword{berry}{berry}
 + \sem{red} + \sem{tart}\\
\sword{bank}{} & \sword{b}{} $\circ$ \sword{a}{} $\circ$ \sword{n}{} $\circ$
\sword{k}{} \\
\sword{bank}{tilt} & \sword{bank}{} + \sem{tilt}\\
\sword{bank}{river-edge} & \sword{bank}{} + \sem{river-edge}\\
\sword{bank}{financial-institution} & \sword{bank}{} + \sem{financial-institution}
\end{tabular}
\end{center}
\caption{\label{fig:meaning}Some examples of the use of the concatenative
  model extended with the meaning perturbation operator.  Notice how the
  inheritance mechanism lets many words inherit meaning from a common root
  (as from \word{cat} and \word{berry}), and also how the ability to
  perturb meanings at any level of the lexical hierarchy can explain how a
  \word{cranberry} can be a specific kind of berry even though there is no
  such thing as a \word{cran}.}
\pageline
\end{figure}

\subsubsection{Ambiguity}

One significant simplification that has been made here is that the learner
can reliably extract the unique ``meaning'' of every utterance from the
extralinguistic environment.  More realistically, the extralinguistic
environment will often provide few or no clues about the meaning of an
utterance, and in other cases the learner will be more sure but still not
certain.  In fact, it is not difficult to extend the learning framework to
accommodate these possibilities.  Suppose that for each utterance the
learner receives as input $u$ and, instead of a meaning $v$, some
extralinguistic information $z$ (the combination of a visual signal and the
internal state of the learner, perhaps).  Assume that from the contextual
information $z$, the learner can compute a function that expresses a prior
expectation over possible meanings $v$.  One way to interpret such a
function is as a conditional probability $p(z|v)$.\footnote{Of course, this
  is not meant to imply that language learners actually estimate
  probabilities of extralinguistic evidence given utterance meanings (it is
  difficult to imagine how they could).  Again, the statistical
  interpretation is merely a convenience that leads to learning algorithms
  with known properties.  $p(z|v)$ is simply a term that weights different
  meanings by their contextual likelihood.  Only its relative magnitude is
  important.} Then the joint probability of $u$ and $z$ under the learner's
language model is $p(u,z) = \sum_{v} p(z|v)p(u,v)$, and compressing $u$ and
$z$ simultaneously amounts to weighting meanings according to a prior
expectation of their naturalness in a given extralinguistic context.  The
posterior probability of a meaning $v$ can be computed as

\[ p(v|u,z) =  \frac{p(z|v)p(u,v)}{p(u,z)}. \]

\noindent Notice that this is a function of both how linguistically natural
the relation is between $u$ and $v$ (the $p(u,v)$ term) and the learner's
prior expectations (the $p(z|v)$ term).  Thus, prior expectations can be
overwhelmed by linguistic evidence, yet can still contribute to learning in
cases where linguistic evidence is underconstraining.

\subsubsection{Coding}

There is no need to define a careful coding scheme for parameter meanings,
as in this case description length matters only in as much as it serves as
a fitness function.  There are only two significant changes from the simple
concatenative framework.  The first is the perturbations that occur both at
the parameter and the utterance level.  In each case a list of sememes is
appended to the list of component parameters (whether a sememe is added or
deleted from the sememe set follows automatically from the prior content of
the set).  If a special ``stop'' sememe is used to terminate the list, then
the description length of a list of sememes is a small constant plus the
cost of representing each sememe.  Assuming a fixed-length code for
sememes, each sememe contributes a fixed cost $\log |S|$ where $S$ is the
complete set of sememes.  Thus the description length of a representation
$w_1 \circ \cdots \circ w_n + \mbox{\sc p}_1 + \cdots + \mbox{\sc p}_m$ is
approximately

\[ m\log |S| + \sum_{i = 1}^{n} -\log p(w_i). \]

\noindent This is essentially equivalent to defining $p(u,v)$ by

\begin{equation}
  p(u,v) = \sum_{v'} 2^{-|S|\cdot |v\otimes v'|} \sum_{w_1\ldots w_n s.t.
    (u,v') = w_1\circ\cdots\circ w_n} \prod_{i=1}^{n} p(w_i) \label{eq:meaninglength}
\end{equation} 

\noindent where $v\otimes v'$ is the set of sememes that occur in one but
not both of $v$ and $v'$.  In conjunction with the conditional probability
term $p(z|v)$, equation~\ref{eq:meaninglength} defines the joint description
length of $u$ and $z$,

\[ |u,z| = -\log \sum_{v} p(z|v)p(u,v). \]

It will turn out to be very convenient when building learning algorithms to
move the cost of representing perturbations into the $p(z|v)$ term; this
eliminates much of the need to think explicitly about perturbations during
utterance processing.  Define

\begin{equation}
  p'(z|v') = \sum_{v} p(z|v) 2^{-|S|\cdot |v\otimes v'|}. \label{eq:pprime}
\end{equation}

\noindent Then

\begin{equation}
  |u,z| = -\log \sum_{v'} p'(z|v') \sum_{w_1\ldots w_n s.t.
    (u,v') = w_1\circ\cdots\circ w_n} \prod_{i=1}^{n}
  p(w_i). \label{eq:meaninglength2}
\end{equation}

\noindent Thus, by slightly altering $p(z|v)$ to produce $p'(z|v')$ the
computation of the joint description length has been simplified, and made
into a form that more closely reflects the calculation in the concatenative
model without the meaning perturbation operator.

\subsection{Phonology and Speech}\label{rep:speech}

For the most part this thesis has been vague about the nature of the signal
available to the learner.  Given that children acquire language from raw
speech, one might ask the question whether the terminals of the
compositional representation must be air-pressure measurements.  The answer
is no.  We have been implicitly assuming that language is produced and
interpreted in stages.  At some point at the border between ``linguistic''
processing and the physical act of speech production the compositional
framework ceases to play a role. The mechanisms beyond that point behave
very differently than those that motivate the framework.

Suppose that language production is modeled as a three-stage process.  The
first stage encompasses most of the mechanisms commonly associated with
higher-level linguistic processing and terminates in a sequence of
phonemes.  A phoneme is a primitive object used to represent sound in the
lexicon~\cite{Halle83b}.  Each one defines a set of desired positions for
various vocal articulators.  For example, the /\unim/ phoneme specifies
that the lips should be closed, that the velum should be lowered so that
air flows through the nose, that the vocal cords should be vibrating, and
so on.  During the actual act of speaking articulators do not always attain
the positions specified by the phoneme sequence.  For example, when
pronouncing \word{want you} the tongue may anticipate the /\uniy / sound
during the production of the immediately preceding /\unit /.  As a
consequence, /\unit / may be pronounced /\uniC /, turning \word{want you}
into \word{wanchya} (a common phenomenon in fast speech).  Thus, the second
stage of our model encompasses the phonetic processes that transform
commands into muscular behavior.  The final stage of the model accounts for
the remainder of the language production process, from muscular motion all
the way to the pressure variations that register on the learner's ear.

\begin{figure}
\pageline
\begin{center}
\mbox{\ }
\psfig{figure=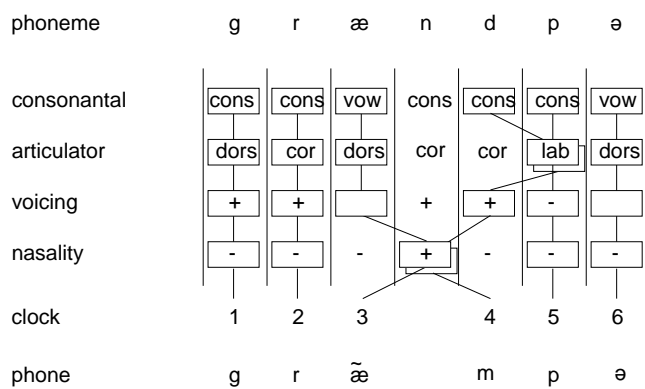,height=2in}
\end{center}

\caption{\label{fig:phonetic}A depiction of how
  /\unig\unir\uniA\unin\unid\unip\uniAX/ might surface as
  /\unig\unir\~{\uniA}\unim\unip\uniAX/.  Each phoneme is a bundle of
  articulatory features (4 as depicted here, more in the real model).  Each
  feature is copied from the input to the output, but this process is
  (stochastically) affected by clock skew and copying errors.  For example,
  the nasalization of the /\uniA/ and /\unid/ and the place-of-articulation
  assimilation of the /d/ are explained by clock skew.  No surface phoneme
  is output for the underlying phoneme /n/.} \pageline
\end{figure}

We have constructed a stochastic model of language production with this
structure.  The first stage is the concatenative model as described in this
chapter, with phonemes as terminals.  The second stage is actually an
extension to the first stage, a phonological perturbation operator that can
capture sound changes that are expected given the physical nature of the
production process.  To understand how this operator functions, realize
that each word in the lexicon is a sequence of phonemes.  The composition
operator, as before, concatenates words in the lexicon to produce longer
sequences of phonemes.  The phonological perturbation operator
stochastically transforms these phoneme sequences by inserting, deleting,
and mutating phonemes.  For example, the word (\word{grandpa})
/\unig\unir~{\uniA}\unim\unip\uniAX/ might be represented as \word{grand}
/\unig\unir\uniA\unin\unid/ $\circ$ \word{pa} /\unip\uniAX/ + {\sc
  sound-change}.  The description length of a sound change from a sequence
$\Phi$ to a sequence $\Theta$ is determined by a stochastic model
$p(\Theta|\Phi)$.  $p(\Theta|\Phi)$ is constructed to reflect a simple
theory of phonetics.  This is described in more detail in de
Marcken~\cite{deMarcken95d}.  Figure~\ref{fig:phonetic} gives a flavor for
how this model works.

The final output of the first two stages of our model is still a sequence
of phonemes.  The third and final stage of the stochastic production
process maps from phoneme sequence to acoustic signals.  We model this
using linguistically uninteresting techniques that are standard to the
automatic speech recognition community.  They are detailed in
section~\ref{results:speech}, where an experiment is described in which
words are learned directly from speech.  For various computational reasons
the experiment described there does not utilize the phonological
perturbation operator described above.

\section{Related Work}\label{rep:related}

There are really two fundamental ideas in the representational framework
described in this chapter.  The first idea is that the same composition
operators traditionally used during sentence processing can also be used to
construct parameters in the grammar and lexicon.  The second idea is that
parameters in the grammar and lexicon receive their identity through
perturbations that alter their behavior.  Curiously, neither of these ideas
has received concentrated attention before.  In the linguistics community
it is commonly assumed that there is internal structure in the lexicon, but
the relationship between this structure and on-line processing is not
generally made explicit.  Similarly, while the lexicon is commonly viewed
as the source of behavior that does not otherwise fall out of the grammar,
we are not aware of any work that attempts to make explicit how such
behavior is specified and related to the normal functioning of the grammar.

As mentioned in section~\ref{uns:related}, computational studies of
language acquisition have routinely made use of the ideas of Bayesian
inference, stochastic language models and MDL.  Despite this fact, little
emphasis has been put on the importance of efficient coding for grammars.
Very often the complexity of grammars is measured using coding schemes that
treat the grammar as a sequence of symbols to be written out on a piece of
paper for viewing.  Exceptions include the work of
Ellison~\cite{Ellison92,Ellison94} (where linguistically interesting
representations for grammars are explored) and Stolcke~\cite{Stolcke94}
(where statistically principled means are used to estimate description
length).

The data compression community has put more emphasis on efficient coding of
parameters, and has produced several representations for parameters that
are similar to ours; these are described in more detail in
section~\ref{textcomp}.  However, in the data compression community little
emphasis has been put on the interpretation of parameters, and as a
consequence no consideration is given to complex or
linguistically-meaningful composition operators.

\chapter{Learning Algorithms}\label{ch:algorithm}

Chapter~\ref{ch:representation} defines a representation for utterances and
grammars (lexicons).  Various arguments are made in chapters~\ref{ch:Bayes}
and \ref{ch:representation} that the learner can acquire language by
choosing the grammar that minimizes the combined description length of the
grammar and the evidence available to the learner.  This leaves open the
question of the learning mechanism: how in practice does the learner find a
grammar that results in a short description length?  As argued in
section~\ref{learning}, this is a fundamental question, since the space of
grammars is far too large for simple enumeration strategies to be
practical, and algorithmic issues of efficiency, convergence and robustness
all reflect back on the appropriateness of the abstract learning framework.
This chapter presents concrete, efficient learning algorithms for two of
the instantiations of the compositional framework presented in the previous
chapter, the concatenative model and the concatenative model extended with
the meaning perturbation operator.  In doing so it demonstrates the
feasibility of the learning framework that has been built up over the
preceding chapters.

The algorithms that are presented here are not meant to reflect
psychological reality.  They are phrased in terms of classical computation,
not biological mechanisms, and have some properties that are in conflict
with what we know about human performance.  For example, the algorithms
make multiple passes over the entire body of evidence available to the
learner.  The principal purpose in presenting the algorithms is, in line
with the goals set out in chapter~\ref{ch:la}, to demonstrate that language
can be learned from real data using the representations and techniques that
have been discussed so far.  Although additional constraints must indeed be
imposed on theories that attempt to explain human cognitive processing,
these constraints are not enforced here, and as a natural consequence, the
algorithms no doubt deviate in many important ways from the mechanisms
people use for language acquisition.  Experience has shown that once
abstract issues are understood, it is often a relatively simple matter to
restate and otherwise transform algorithms to conform with what is known
about human processing mechanisms.

The chapter begins by presenting a general architecture for learning
algorithms that evaluate compositional representations with respect to the
minimum description length principle.  Algorithmic details are dependent on
the choice of the composition and perturbation operators, and various other
decisions.  Specific algorithms are presented for two instantiations of the
compositional framework, the concatenative model and the concatenative
model extended with the meaning perturbation operator.  These
implementations are among the simplest possible, but only small changes are
necessary to handle significantly more complex operators (such as in the
substitution model).

\section{General Architecture}

Under the inference scheme presented in chapter~\ref{ch:Bayes}, learning as
expressed at the computational level is the search for the grammar that
minimizes the combined description length of the input and the grammar.
The MDL principle imposes a trivial upper bound on the length of a
plausible grammar, namely the length of the input, but even so grammars can
be extremely large (millions of bits long in the examples in
chapter~\ref{ch:results}).  The space of possible grammars is therefore
enormous, and precludes learning by brute-force enumeration.  At the same
time, parameters can not be evaluated independently, but only with respect
to a complete grammar.  This and the desire for incremental learning
strategies (see section~\ref{mdl}) motivates the use of heuristic
algorithms that attempt to minimize the description length by iteratively
updating and improving a grammar (in this case a lexicon), by adding,
deleting and otherwise manipulating parameters.  The algorithms that will
be presented all follow this general strategy:

\begin{center}
\begin{tabular}{l}
Start with the simplest lexicon.\\
Iterate until convergence:\\
\hspace{.2in} Refine the parameters of the lexicon to reduce the description length.
\end{tabular}
\end{center}

\noindent Since the change in description length that a new parameter (or a
change to an existing parameter) causes is determined mostly by any
improvement in statistical modeling performance that it brings, an
important part of the learning process is the continual collection of
information that describes the performance of the current lexicon in
predicting the evidence.  Such information can be used by the learner to
{\em estimate} the effect on the description length of some change to the
lexicon.  Of course, it is impossible for the learner to know exactly what
effect a change will have on the description length.  This is because
changes have complex repercussions-- they alter parameter usage patterns,
which in turn motivates further changes in the lexicon, ad infinitum.  Thus
a better statement of the last line of the learning strategy would be
``refine the parameters of the lexicon in any way that is {\em predicted}
to reduce the description length''.

As discussed in section~\ref{motiv:learnability}, each parameter can be
thought of as pairing a linguistic predicate with some information that
determines the stochastic properties of the language model.  Assuming that
the description length of this stochastic information is relatively
independent of its content ({\em i.e.}, assuming a uniform prior on the
stochastic information), it follows that for any fixed set of linguistic
parameters the description length is minimized by the stochastic language
model that best models the evidence and the parameters.  Thus, the learning
process can be separated into a stage where stochastic properties are
optimized assuming a fixed linguistic structure in the lexicon, and a stage
where the linguistic structure of the lexicon is altered assuming
relatively fixed stochastic properties.  This general procedure of
alternating between {\em structural} and {\em parametric} (in the
traditional sense) updates to Bayesian models is a popular strategy for
structural induction problems; see for example the stochastic grammar
induction schemes of Stolcke~\cite{Stolcke94} and Chen~\cite{Chen95}, and
the extended literature on structural induction of neural networks and
Bayes' nets.

In the representational framework of chapter~\ref{ch:representation},
parameters from the lexicon are represented in exactly the same way as
utterances from the input.  Furthermore, under the minimum description
length principle (as expressed in equation~\ref{eq:smdl}, echoed below),

\[ G = \mbox{\begin{array}[t]{c}\mbox{argmin}\\{\scriptstyle
    G'\in\cal{G}}\end{array}} \sum_{w\in G'} |w|_{G'} + \sum_{u\in
  U} |u|_{G'}, \]

\noindent the representation cost of parameters is weighted equally with
the representation cost of utterances.  At the algorithmic level, this
implies that the learner should treat any parameter in the lexicon as just
another utterance in the input.  The combined set of utterances and
parameters is denoted in subsequent discussion by $U' = U + G$, with
elements still called utterances for want of a better term.

\begin{figure}
\pageline
\begin{tabbing}
Let $G$ be the simplest lexicon.\\
Iterate until convergence:\\
\hspace{.2in} \= Let $U' = U + G$.\\
\> \hspace{.2in} \= Optimize stochastic properties of $G$ over $U'$.\\
\> \> Collect statistics describing performance of $G$ over $U'$.\\
\> \> Refine linguistic properties of $G$ to improve expected performance over $U'$.
\end{tabbing}
\caption{\label{fig:alg} The general architecture of the learning algorithms
  that will be considered in this chapter.}
\pageline
\end{figure}

Summarizing the preceding paragraphs, figure~\ref{fig:alg} presents the
architecture of the learning algorithms described in this chapter.  As
mentioned, there are two major subroutines in each learning algorithm: a
routine that performs stochastic optimization and a routine that performs
structural optimization.  Each of these is surveyed immediately below, and
then expanded upon in the presentations of the specific algorithms for the
two instantiations of the representational framework.

\subsection{Stochastic Optimization}

The problem of stochastic optimization is to find the stochastic settings
that minimize the description length of the parameters and evidence $U' = U
+ G$, assuming the linguistic structure of the lexicon remains fixed.  This
is equivalent to maximizing the probability of $U'$.  In the specific case
of the concatenative model of section~\ref{concat} it is the problem of
finding codelengths that minimize the total description length.  In that
example, if the counts of parameters are known it is a simple matter to
derive the optimal parameter probabilities (by normalizing), and from that
the codelengths.  Unfortunately, parameter counts are determined by
utterance and parameter representations, which are hidden (underdetermined
by the evidence).  There are many representations of any character
sequence, each with a description length determined by parameter
probabilities.  Hence, the optimization procedure seems cyclic: the optimal
stochastic model is a function of representations, which are in turn a
function of the stochastic model.

The expectation-maximization (EM) procedure of Dempster {\em et
  al.}~\cite{Baum70,Csiszar84,Dempster77} is a standard method for solving
optimization problems involving hidden representations.  It alternates an
E-step (expectation-step) in which the posterior probability of
representations are computed under the current stochastic model with an
M-step (maximization-step) in which the stochastic model is adjusted to
maximize the expected log-likelihood of the representations, where the
expectation is under the posterior probabilities defined by the E-step.
This is equivalent to minimizing the expected description length of the
representations.  Each iteration of the EM algorithm is guaranteed to
monotonically decrease the complete description length, asymptotically
approaching a local optimum.  Expressed somewhat more formally, the E-step
consists of determining for every $u\in U'$ the posterior probability of
the representation $r$ (a sequence of compositions and perturbations)

\[ p_{G}(r|u) = \frac{p_{G}(u,r)}{p_{G}(u)} =
\frac{p_{G}(u,r)}{\sum_{r'} p_{G}(u,r')}. \]

\noindent Since a representation completely determines an utterance,
$p_G(u,r) = p_G(r)$ if $r$ is a valid representation for $u$, and 0
otherwise.  The M-step then produces an improved grammar $G^{\ast}$ defined
by

\begin{eqnarray}
  G^{\ast} & = & \mbox{\begin{array}[t]{c}\mbox{argmin}\\{\scriptstyle
      G'\in\cal{G}}\end{array}}  \mbox{E}\left[ \sum_{u\in U'} -\log p_{G'}(u,r)
\right]\nonumber \\
& = & \mbox{\begin{array}[t]{c}\mbox{argmin}\\{\scriptstyle
      G'\in\cal{G}}\end{array}} \sum_{u\in U'} \sum_{r} -p_{G}(r|u) \log
  p_{G'}(u,r).\label{eq:mstep}
\end{eqnarray}

Here $G^{\ast}$ has the same linguistic structure as $G$, but different
stochastic properties.  For the types of language models considered here,
the E-step is simple but not trivial, whereas the M-step is simply a
normalization.  For other types of models both steps can be complex, and it
is often not possible to perform the optimization involved in the M-step
exactly.  In such cases it may still be possible to choose $G$ so as to
decrease the right hand side of equation~\ref{eq:mstep}.  In such cases the
EM algorithm is still guaranteed to monotonically reduce the complete
description length, though often the procedure converges substantially more
slowly.

The EM procedure is only guaranteed to approach a local optimum, not a
global one.  The effectiveness of the procedure at finding a global optimum
is a function of the complexities of the search space as well as the
starting point for the algorithm; in many cases the procedure is woefully
incapable of finding a global optimum, and this can have significant
effects on learning strategies based on the EM algorithm alone (see
\cite{deMarcken95b} for discussion).  In the context of the algorithms
discussed here, the other optimization step, which modifies the linguistic
structure of the lexicon, often provides a means of escaping from local
optima.  This is an advantage of algorithms that manipulate the structure
of the grammar over algorithms that start with complete structures (for
example, all possible grammatical rules) and attempt to learn solely by
manipulating stochastic properties~\cite{Briscoe92,Pereira92}.

Only rarely will the EM-algorithm converge to the exact (local) optimum in
a finite number of iterations.  However, for the implementations considered
below, after two or three iterations improvements to the description length
tend to be so small as to be irrelevant.  This is because the learning
algorithms (see figure~\ref{fig:alg}) start the stochastic optimization
procedure from a lexicon that has undergone only incremental changes since
the previous optimization step.

\subsection{Structural Refinement}

During each iteration of the learning algorithm the linguistic structure of
the lexicon is refined in an effort to reduce the description length
of the evidence and lexicon.  This is an incremental learning strategy: the
learning procedure starts with a minimal lexicon, just sufficient to
explain any utterance, and expands this lexicon over time through local
changes.  Local changes are those whose effects are confined to small
portions of the lexicon (such as single parameters), so that it is
reasonable to assume that the usage properties of the remainder of the
lexicon stay relatively fixed under the change.  By constraining changes to
be local, it is possible to design procedures that can efficiently and
reasonably accurately estimate the effect of a change on the description
length.  Then the following strategy can be used to refine the linguistic
structure of the lexicon:

\begin{center}
\begin{tabular}{l}
Hypothesize a set of changes to the lexicon.\\
For each change, estimate the effect on the total description length.\\
Implement each change that is estimated to reduce the description length.
\end{tabular}
\end{center}

\noindent By considering large numbers of changes to the lexicon in
parallel, the number of iterations of the learning algorithm necessary for
convergence is made small.  Of course, it can be very difficult to estimate
the effects of large numbers of changes implemented simultaneously.  The
effect of each change can be calculated under an assumption of
independence, but as this assumption is often incorrect, it results in many
undesirable changes (such as two parameters being added when one will do).
However, the undesirable consequences of the independence assumption can be
mostly eliminated by considering changes that ``undo'' previous
modifications to the lexicon, such as by deleting parameters that were
previously created, or creating parameters that were previously deleted.
In this way, many changes are made during each iteration, and those that
are not justified are compensated for during the next iteration.

There are two parts to this refinement procedure.  The first is the
generation of a set of candidate changes to the lexicon, and the second is
an evaluation of each change.  The types of changes that can be considered
are constrained by the types of changes whose effects can be efficiently and
accurately estimated.  The effect of a change to the lexicon, such as the
creation of a new parameter, is very dependent on the performance of the
existing language model.  For this reason, the evaluation procedure must
have access to information that describes the performance of the current
lexicon, and that gives indications as to what sorts of changes should be
favored.

In the algorithms described below, changes are motivated by the
interpretation of parameters given at the start of
section~\ref{motiv:learnability}, where each parameter is viewed as a way
of expressing a statistical dependence among its component perturbations
and compositions.  Two types of changes are considered: adding a parameter,
and deleting a parameter.  A parameter with representation $w_1 \circ
\cdots \circ w_n + \mbox{\sc p}_1 + \cdots + \mbox{\sc p}_m$ is added if
the benefit of representing the chain of compositions and perturbations by
a single reference is expected to exceed the description length of the
parameter.  To estimate this, it is necessary to have some idea of how many
times these components occur together in the representations of the
evidence and the lexicon, and compare this with the number expected given
an independence assumption.  Since it would require too much storage space
to record this information for every {\em possible} representation, new
parameters are considered only if they appear as a subpart of the most
probable representation of some utterance or parameter.  In fact, the
algorithms described below only consider parameters that can be built by
composing two other parameters or by perturbing a single parameter:
parameters are built up by the pairwise combination of existing objects
that are composed in existing representations.  A parameter is deleted if
it appears that the cost of substituting its representation for it is less
than the cost of its description length.  There are many other types of
changes that could be considered.\footnote{The types of changes to the
  lexicon that should be considered depend heavily on the nature of the
  composition and perturbation operators.} For example, new parameter
candidates could be hypothesized by looking for long repeated sequences in
the evidence (as opposed to considering only candidates that are the
pairwise combination of existing parameters).  This would enable the
learning algorithm to create some parameters that the algorithms below will
not, because they have too limited a view.  On the other hand, such a
strategy would complicate the collection of usage statistics necessary to
evaluate such candidates.

In many cases it can be important to consider various second order effects.
For example, the creation of one parameter may justify the deletion of
another.  This deletion will reduce the description length, and should be
taken into account when computing the benefit of the first parameter.  Of
course, there are limits to the effects that can be considered.  A guiding
principle used here is that no effect is considered if it would require
reanalysis of the evidence.

\section{Concatenative Model}\label{concatmodel}

The concatenative model of section~\ref{rep:concat} allows for a
particularly simple and efficient learning algorithm, presented in this
section.  Stochastic optimization via EM is accomplished by the Baum-Welch
procedure~\cite{Baum72}, and fairly simple estimation procedures are used
to predict the effects of adding and deleting parameters.

To simplify and shorten the exposition, it will be assumed that the
evidence $U$ available to the learner is a sequence of characters drawn
from some alphabet (letters, phonemes, etc.), possibly presegmented
into utterances $u\in U$.  In some applications it is necessary to allow
for less certain input; in many of these cases the input is logically
viewed as a stochastic lattice over characters, where transition
probabilities reflect the source of uncertainty.  This would be true of the
extended model presented in section~\ref{rep:speech} where a phoneme
sequence serves as an intermediate representation that generates the speech
signal.  In that case the transition probabilities would reflect the
probability of the speech signal given the phonemes.  Using a stochastic
lattice as an input rather than a sequence slightly complicates the
stochastic and structural optimization procedures, but not in any
fundamental way.\footnote{If the lattice is extremely dense, performance
  may be reduced substantially.} For this reason, the learning algorithm
for the concatenative model is presented in a form that only handles the
simple case of a single, ``noiseless'' character sequence.  It is not
difficult to extend this basic algorithm to handle more interesting cases.

Two methods for refining the linguistic structure of the lexicon will be
considered.  First, new parameters can be created.  Although this is by no
means a necessity, the algorithm will consider as new parameters only
parameters that can be formed by composing two or more existing parameters.
Second, parameters can be deleted from the lexicon.  All parameters except
the terminal characters are considered for deletion during every iteration.
The learning algorithm alternates between creating and deleting parameters.
Figure~\ref{fig:calg} summarizes the learning algorithm, which will be
explained in more detail below.

\begin{figure}
\pageline
\begin{tabbing}
Let $G$ be the set of terminals with uniform probabilities.\\
Iterate until convergence:\\
\hspace{.2in} \= Let $U' = U + G$.\\
\> \hspace{.2in} \= {\em Optimize stochastic properties of $G$ over $U'$. }\\
\> \> \hspace{.2in} \= Perform optimization via 2 steps of the forward-backward algorithm.\\
\> \> \> During second step record parameter co-occurrence counts and Viterbi representations.\\
\> \> {\em Refine linguistic properties of $G$ to improve expected performance over $U'$.}\\
\> \> \> Add new parameters to $G$ that are the composition of existing ones.\\
\> Set $U' = U + G$.\\
\> \> {\em Optimize stochastic properties of $G$ over $U'$. }\\
\> \> \> Perform optimization via 3 steps of the forward-backward algorithm.\\
\> \> {\em Refine linguistic properties of $G$ to improve expected performance over $U'$.}\\
\> \> \> Delete parameters from $G$.
\end{tabbing}
\caption{\label{fig:calg} The learning algorithm for the concatenative
  model.}
\pageline
\end{figure}

Section~\ref{motiv:learnability} mentions that the representation of each
parameters does not affect its use.  The algorithm presented here takes
advantage of this fact, internally storing each parameter as a sequence
$u_1\ldots u_l$ of characters, just like an utterance.  Of course, there
are many possible representations for each parameter, and these determine
its description length.  But only rarely does the algorithm need to have
access to parameter representations.  It is therefore not necessary to
explicitly maintain representations for each parameter throughout the
learning process.  In circumstances where representations become important
(in particular, during structural refinement of the lexicon)
representations for a parameter are extracted by parsing its character
sequence.  Among the advantages this conveys is that there is no need to
update representations as parameters are added and deleted from the
lexicon.  Other advantages will be discussed in
section~\ref{relatedlearning}.

\subsection{Optimization of Stochastic Parameters}\label{stochopt}

As discussed in section~\ref{rep:concat}, multigrams are a special form of
hidden Markov models.  The EM procedure for HMMs is known as the Baum-Welch
algorithm~\cite{Baum72}, and is rederived below in a simpler form more
appropriate for multigrams.  As with all EM procedures, the Baum-Welch
procedure alternates E-steps and M-steps.  This procedure rapidly
converges.  Although it is generally possible to test for convergence by
setting thresholds on changes to the total description length, in this case
it is as effective to simply execute a fixed number of iterations (as
described in figure~\ref{fig:calg}).

\subsubsection{The Maximization Step}

Recall that a representation $r$ is a sequence of parameters $w_1\ldots
w_n$, and (following from equation~\ref{eq:upper}) that the joint
probability of a sequence and a representation $p(u,r)$ can be expressed
$p(u,r) = \prod_i p(w_i)$ if $u = w_1\circ\cdots\circ w_n$ ($p(u,r) = 0$
otherwise).  Following equation~\ref{eq:mstep}, define for the multigram
model the expected description length $L$:

\[ L = \sum_{u\in U'} \sum_{w_1\ldots w_n} -p_{G}(w_1\ldots w_n|u) \log
\prod_{j=1}^{n} p_{G}(w_j). \]

\noindent Then the optimal lexicon is one in which $\partial L/\partial
p(w) = 0$ for each parameter $w$.  As this is a constrained optimization
problem (the total probability of all parameters must sum to one), a
Lagrange multiplier term is introduced ($L' = L + \lambda \sum_w p_G(w)$),
giving

\begin{eqnarray*}
  \frac{\partial L'}{\partial p_{G^{\ast}}(w)} & = & \sum_{u\in U'}
  \sum_{w_1\ldots w_n} -p_{G}(w_1\ldots w_n|u) \sum_{j=1}^{n}
  \frac{\partial p_{G^{\ast}}(w_j)/\partial
    p_{G^{\ast}}(w)}{p_{G^{\ast}}(w_j)} + \lambda\\ & = & \sum_{u\in U'}
  \sum_{w_1\ldots w_n}
  -p_{G}(w_1\ldots w_n|u)\frac{c(w\in w_1\ldots w_n)}{p_{G^{\ast}}(w)} +
  \lambda\\ & = & 0,
\end{eqnarray*}

\noindent where $c(w \in w_1\ldots w_n)$ is the number of times the
parameter $w$ appears in the representation $w_1\ldots w_n$.  Thus,

\[ p_{G^{\ast}}(w) =  \frac{\sum_{u\in U'} \sum_{w_1\ldots w_n} p_{G}(w_1\ldots
  w_n|u)c(w\in w_1\ldots w_n)}{\lambda}. \]

\noindent The Lagrange multiplier $\lambda$ merely acts as a normalization
constant.  Therefore the optimal probability for each parameter $w$ is
given by

\begin{equation}
 p_{G^{\ast}}(w) = \frac{c_{G}(w)}{\sum_{w'} c_{G}(w')}, \label{eq:norm}
\end{equation}

\noindent where $c_{G}(w)$ is the expected number of times that the
parameter $w$ is used in the complete description of $U'$ under the lexicon
$G$:

\begin{equation}
 c_{G}(w) = \sum_{u\in U'} \sum_{w_1\ldots w_n} p_{G}(w_1\ldots
w_n|u) c(w\in w_1\ldots w_n).\label{eq:counts}
\end{equation}

\noindent As might be expected, the maximization step optimizes
probabilities by normalizing the expected counts of parameters under the
lexicon $G$.

\subsubsection{The Expectation Step}

\newcommand{\awb}{{a\stackrel{w}{\rightarrow}b}}
\newcommand{\awwb}{{a\stackrel{w_1,w_2}{\rightarrow}b}}

The E-step for the multigram model consists of computing the posterior
counts $c_{G}(w)$ used in equation~\ref{eq:norm}.  This would appear from
equation~\ref{eq:counts} to involve a sum over all possible representations
for each utterance.  Since the number of representations can be exponential
in the length of an utterance, this might appear intractable.  However, a
dynamic programming\footnote{Here {\em dynamic programming} is used in the
  algorithmic sense, though the backward portion of the forward-backward
  algorithm is also a dynamic programming algorithm in the optimization
  sense~\cite{Bellman57}!} technique known as the forward-backward
algorithm~\cite{Baum72} enables this computation to be performed
efficiently.  Here the forward-backward algorithm is presented in a
somewhat simplified form appropriate for the multigram model.  The
algorithm consists of two steps for each utterance.  First, {\em forward}
and {\em backward} probabilities are computed for each location (character
index) in the utterance.  Then these probabilities are used to compute for
each parameter $w$ and each starting location $a$ and each ending location
$b$ the posterior probability $p_{G}(\awb|u)$ of $w$ generating
$u_{a}\ldots u_{b}$ in a derivation of $u = u_1\ldots u_l$.  Thus, in the
course of computing posterior counts the forward-backward algorithm
essentially parses each utterance into representations.

For an utterance (a character sequence) $u = u_1\ldots u_l$ let the {\em
  forward probability} $\alpha_i(u)$ be the probability of the stochastic
model generating any complete parameter sequence $w_1\ldots w_o w_p\ldots
w_n$ such that $u_1\ldots u_i = w_1\circ\cdots\circ w_o$.  Then
$\alpha_0(u) \equiv 1$ and

\[
\alpha_i(u) = \sum_{j=0}^{i} \alpha_j(u) \sum_{w=u_{j+1}\ldots
  u_{i}\in G} p_G(w).
\]

\noindent Further let the {\em backward probability} $\beta_i(u)$ be the
probability of the stochastic model generating any complete parameter
sequence $w_1\ldots w_o w_p \ldots w_n$ such that $u = w_1\circ\cdots\circ
w_n$, {\em given that} $u_{1}\ldots u_{i} = w_1\circ\cdots\circ w_o$.  Then
$\beta_l(u) \equiv 1$ and \footnote{The fact that $\beta_l(u) \equiv 1$ follows
  from the simplification made in equation~\ref{eq:upper}, that ignores in
  the stochastic model the number of parameters in a representation.  The
  use of a special terminating parameter would eliminate the need for
  this approximation.}

\[
\beta_i(u) = \sum_{j=i}^{l} \beta_j(u) \sum_{w= u_{i+1}\ldots u_{j}\in G}
p_G(w).
\]

\noindent Notice that $p_G(u) \equiv \alpha_0(u) \equiv \beta_l(u)$.  It
follows from the independence of parameter generation in the multigram
model that the conditional probability $p_G(\awb|u)$ of a parameter $w$
spanning a region $u_{a+1}\ldots u_{b}$ during the generation of an
utterance $u_1\ldots u_l$ is given by

\begin{equation}
  p_G(\awb|u) = \frac{\alpha_{a}(u)
    p_G(w)\beta_{b}(u)}{p_G(u)}\label{eq:span}
\end{equation}

\noindent if $w = u_{a+1}\ldots u_b$. $p_G(\awb|u) \equiv 0$ otherwise.

Conveniently, the distributive properties of the expectation operator imply
that the calculation of the expected count of a parameter $w$ over an
utterance $u$ can be reexpressed as a sum over all subsequences of the
utterance of the expected probability of that parameter spanning that
subsequence.  Hence, equation~\ref{eq:counts} can be rewritten as

\begin{equation}
 c_{G}(w) = \sum_{u_1\ldots u_l\in U'}\sum_{a=0}^{l} \sum_{b=a}^{l}
 p_{G}(\awb|u_1\ldots u_l).\label{eq:counts2}
\end{equation}

For each utterance the calculation of the forward and backward
probabilities is linear in the length of the utterance and linear in the
length of the longest parameter.  The calculation of expected parameter
counts has the same complexity.  Given the expected counts, the
maximization step is a simple matter of normalization.  Hence the
computational complexity of each step of the EM-algorithm is essentially
linear in the total length of the evidence and the parameters (as measured
by the number of characters) and linear in the length of the longest
parameter.  By representing the lexicon as a character tree, this cost can
be further reduced.

\subsubsection{Maintaining a Logically Consistent Lexicon}

Although there are advantages to representing parameters as character
sequences during the execution of the learning algorithm, it does introduce
a significant complication.  Because each parameter is stored by its
``content'' rather than its representation, there is no guarantee that the
representation of the lexicon is internally consistent, such that the
internal hierarchy (as in figure~\ref{fig:nfl}) is a directed acyclic
graph.  For example, with the forward and backward probabilities defined as
they are each parameter will be represented by a single component-- itself!
While a remarkably efficient representation, this obviously defeats the
purpose of the compositional framework.  One way to ensure that the lexicon
remains internally consistent is to impose a complete ordering on
parameters that depends only on parameter content.  Then so long as each
parameter $w$ is represented in terms of other parameters $w_1\ldots w_n$
such that $\forall i, w_i < w$, the lexicon is consistent.  This constraint
can be imposed by slightly altering the definitions of the forward and
backward probabilities:

\begin{eqnarray*}
  \alpha_i(u) & = & \sum_{j=0}^{i} \alpha_j(u) \sum_{w=u_{j+1}\ldots
    u_{i}\in G, w<u} p_G(w).\\ \beta_i(u) & = & \sum_{j=i}^{l} \beta_j(u)
  \sum_{w=u_{i+1}\ldots u_{j}\in G, w<u} p_G(w).
\end{eqnarray*}

For the concatenative model it suffices to define the ordering over
parameters in terms of parameter length: $u_1\ldots u_l < v_1\ldots v_m$ if
$l < m$.  This prevents a parameter from being used as its own
representation.  In models where perturbations play a bigger role, such as
the meaning model presented further below, the ordering constraint must be
more complex if it is to prevent various cyclic representations.

\subsubsection{Recording Parameter Cooccurrence Statistics}

In order to refine the linguistic structure of the lexicon by adding and
deleting parameters, it is necessary for the learning algorithm to first
record statistics about the usage patterns of parameters.  The methods for
refining the lexicon described in the following section require that two
kinds of information be recorded: the optimal (most probable)
representation of the evidence and each parameter, and expected counts of
how often two parameters are composed.  This information can be extracted
as part of the forward-backward algorithm.

The most probable (Viterbi) representation $R(u)$ for an utterance $u =
u_1\ldots u_l$ can be computed by the following procedure that mirrors the
calculation of the forward probabilities:

\begin{center}
\begin{tabular}{l}
  Set $R_0(u) = \emptyset$, $\alpha^v_0(u) = 1$.\\ For $i = 1$ to $l$,\\ 
  \hspace{.2in} Set $R_i(u) = \emptyset$, $\alpha^v_i(u) = 0$.\\ 
  \hspace{.2in} For $j = 0$ to $i$,\\
  \hspace{.4in} For $w\in G, w = u_{j+1}\ldots u_{i}$,\\
  \hspace{.6in} Let $\alpha^v = \alpha^v_j(u) p_G(w)$.\\
  \hspace{.6in} If $\alpha^v > \alpha^v_i(u)$ then\\ 
  \hspace{.8in} Set $R_i(u) = \langle R_j(u), w\rangle$, $\alpha^v_i(u) =
  \alpha^v$.\\
  Then $R(u) = R_l(u)$.
\end{tabular}
\end{center}

The counts $c_G(w_1,w_2)$ of the expected number of times two parameters
$w_1$ and $w_2$ are composed under the grammar $G$ can be computed in a
similar fashion to the counts $c_G(w)$.  Following
equation~\ref{eq:counts2},

\begin{equation}
 c_{G}(w_1,w_2) = \sum_{u_1\ldots u_l\in U'}\sum_{a=0}^{l} \sum_{b=a}^{l}
 p_{G}(\awwb|u_1\ldots u_l) \label{eq:wwcount}
\end{equation}

\noindent where the probability of the composition $w_1\circ w_2$ spanning
the subsequence $u_{a+1}\ldots u_{b}$ during the generation of $u$ is given
by (following equation~\ref{eq:span} and assuming that $w_1 w_2 =
u_{a+1}\ldots u_{b})$

\begin{equation}
  p_G(\awwb|u) = \frac{\alpha_{a}(u) p_G(w_1)p_G(w_2)
    \beta_{b}(u)}{p_G(u)}. \label{eq:wwspan}
\end{equation}

\subsection{Refinement of Model Structure}

As figure~\ref{fig:calg} makes explicit, the linguistic structure of the
lexicon is refined in two separate stages, first by creating new parameters
and then by deleting existing parameters.  In each case a set of candidate
changes is considered, each one evaluated under the assumption that it is
the only change.  The evaluation process consists of estimating the
approximate counts $c_{G^{\ast}}(w)$ of each parameter after the change.
By comparing with the counts $c_{G}(w)$ before the change, an estimate of
the approximate change $\Delta$ in description length can be made:

\begin{equation}
  \Delta \approx \sum_{w\in G^{\ast}} -c_{G^{\ast}}(w)\log p_{G^{\ast}}(w) - \sum_{w\in G}
  -c_G(w)\log p_G(w), \label{eq:delta}
\end{equation}

\noindent where $p_{G^{\ast}}(w) = c_{G^{\ast}}(w) / \sum
c_{G^{\ast}}(w)$.\footnote{Equation~\ref{eq:delta} is only an approximation,
  but one that is generally quite accurate.  The complete description
  length (before changes) is, echoing equation~\ref{eq:upper},

\[ |U'| = \sum_{u\in U'} - \log \sum_{w_1 \ldots w_n\ s.t.\ u =
  w_1\circ\cdots\circ w_n} p(w_1)\cdots p(w_n). \]

\noindent In contrast, equation~\ref{eq:delta} is derived by moving the
logarithm inside the summation,

\begin{eqnarray*}
 |U'| & \approx & \sum_{u\in U'} \sum_{w_1 \ldots w_n\ s.t.\ u =
   w_1\circ\cdots\circ w_n} - \log p(w_1)\cdots p(w_n).\\
 & = & \sum_{w\in G} -c_{G}(w)\log p_{G}(W).
\end{eqnarray*}

\noindent This approximation is valid because the Viterbi representation
tends to contribute the vast majority of the probability of an utterance.
For example, if there are two representations for an utterance, one of
length 10 bits, and another of length 15 bits, then the correct description
length is $-\log (2^{-10}+2^{-15}) = 9.95$ bits, whereas the approximation
gives $10\cdot \frac{2^{-10}}{2^{-10}+2^{-15}} + 15\cdot
\frac{2^{-10}}{2^{-10}+2^{-15}} = 10.15$ bits, a difference of only 2\%.
Even in the case where there are two representations of equal length, the
difference amounts to only one bit.  Furthermore, when used in
equation~\ref{eq:delta} even these small approximation errors tend to
cancel out.} So long as only a small number of parameter counts change (the
premise of local updates to the lexicon), equation~\ref{eq:delta} can be
evaluated efficiently.  This is because it can be rewritten in an even more
convenient form:

\[
  \Delta \approx \left(C_{G}-\sum_{w\in G-H} c_{G}(w) \right) \log
  \frac{C_{G}+\Delta C}{C_{G}} - \sum_{w\in G^{\ast} - H} c_{G^{\ast}}(w)\log
  p_{G^{\ast}}(w) + \sum_{w\in G-H} c_G(w)\log p_G(w),
\]

\noindent where $H$ is the set of all parameters $w$ whose counts do not
change from $G$ to $G^{\ast}$, $C_{G} = \sum_{G} c_{G}(w)$ is the total
count under $G$, and $\Delta C = \sum_{G^{\ast}-H} c_{G^{\ast}}(w) -
\sum_{G-H} c_{G}(w)$ is the total change in parameter counts.  In this way,
so long as the total count $C_{G}$ is known, the calculation of $\Delta$
does not involve terms for every parameter, but only those that are added
or deleted or that change counts.

If $\Delta < 0$, then the change from $G$ to $G^{\ast}$ is estimated to reduce
the combined description length of the evidence and the lexicon.  For
convenience, changes are hypothesized and evaluated in parallel.  Then all
changes for which $\Delta < 0$ are implemented simultaneously.

\subsubsection{Adding Parameters}

The set of new parameter candidates is constructed from pairs of parameters
that are composed in the representation of the evidence and lexicon.  For
example, if under the grammar $G$ a representation for \word{thecat} is
\word{t} $\circ$ \word{h} $\circ$ \word{e} $\circ$ \word{cat} then the
parameters \word{th}, \word{he}, and \word{ecat} are all candidates.  Since
for any utterance there may be many representations, most of which are
fantastically unlikely (they have much longer description lengths than the
best representation), only parameter pairs that occur in the best (Viterbi)
representation are considered; this substantially reduces the total set
under consideration.  Viterbi representations are computed during the
stochastic optimization process (see figure~\ref{fig:calg}) by the method
described at the end of section~\ref{stochopt}, and used to construct the
set of candidate parameters.  In the concatenative model no parameter will
be added to the model if it only occurs once (this is not true of more
complicated instantiations like the substitution model); therefore the
set of candidates can be pruned by eliminating all parameters that occur
fewer than two times in Viterbi representations.

For each candidate parameter $W$ with Viterbi representation $w_1 \circ
w_2$, the expected count $c_G(w_1,w_2)$ of the composition $w_1 \circ w_2$
is computed as described in section~\ref{stochopt}.  This produces counts
that differ only slightly from those that would result from simply adding
the number of times the pair occurred in Viterbi representations.  The
count $c_G(w_1,w_2)$ is used to estimate the changes in parameter counts
that would result from adding $W$ to the lexicon.

To estimate the expected changes in parameter counts from adding $W$ to the
lexicon, various assumptions must be made.  The fundamental assumption will
be that representations change only in so much as $W$ replaces, in whole or
in part, its representation.  For example, if the parameter \word{th} is
added to the lexicon, then the representation \word{th} $\circ$ \word{e}
$\circ$ \word{cat} will compete with \word{t} $\circ$ \word{h} $\circ$
\word{e} $\circ$ \word{cat}, presumably substantially reducing the counts
of \word{t} and \word{h}.  Other parameter counts will remain the same.
Imagine that the count $c_{G^{\ast}}(W)$ of $W$ under the updated lexicon
is known.  Further define the count of a parameter $w$ in the
representation of an utterance $u = u_1\ldots u_l$ by

\[ c_{G}(w\in u) = \sum_{a=0}^{l} \sum_{b=a}^{l} p_{G}(\awb|u_1\ldots
u_l). \]

\noindent Then each occurrence of $W$ will reduce the count of the members
of its representation by their count in its representation; on the other
hand, $W$ must be represented, and this will increase the counts of the
members of its representation by $c_{G^{\ast}}(w\in W)$.  Thus,

\begin{equation}
  c_{G^{\ast}}(w) \approx c_{G}(w) + c_{G^{\ast}}(w\in W) -
  c_{G^{\ast}}(W)c_{G}(w\in W).\label{eq:adddelta}
\end{equation}

To compute with equation~\ref{eq:delta} the expected change in description
length then, estimates are needed of $c_{G^{\ast}}(W)$ and
$c_{G^{\ast}}(w\in W)$.  It is possible to get accurate estimates through
various iterative methods and these can very slightly improve performance,
but in practice it is more than adequate to use quite simple
approximations: for concreteness, let $c_{G^{\ast}}(W) \approx c_G(w_1,
w_2)$ and $c_{G^{\ast}}(w\in W) \approx c_{G}(w\in W)$.  Note that the
computation of equation~\ref{eq:adddelta} therefore requires a pass of the
forward-backward algorithm over $W$ to estimate $c_{G}(w\in W)$.

It is sometimes necessary to consider the secondary effects of adding a
parameter.  A particularly common case is when adding a parameter
eliminates the motivation for one or more of the parameters in its
representation.  For example, if \word{banana} = \word{bana} $\circ$
\word{na} is added to the lexicon, then \word{banana} will largely or
entirely replace \word{bana} and it is quite likely that the only use of
\word{bana} will be in the representation of \word{banana}.  In such a case
\word{bana} would be deleted from the lexicon in the next stage of the
learning algorithm for a net reduction in description length.  However, the
increased length at the intermediate stage where both \word{banana} and
\word{bana} exist might prevent \word{banana} from being added in the first
place.  The next section describes an estimation procedure that determines
the expected savings from deleting a parameter.  This procedure is used to
calculate the expected secondary changes in description length $\Delta_1$
and $\Delta_2$ from deleting the words $w_1$ and $w_2$ after $W$ has been
added.  The revised condition is to add $W$ if

\[ \Delta + \min(\Delta_1, 0) + \min(\Delta_2, 0) < 0. \]

\noindent This lookahead does mean that it is common after the parameter
creation stage for there to be an increase in description length.  After
the deletion stage there is almost always a net reduction.

\subsubsection{Deleting Parameters}

In each iteration, all parameters except the terminals are candidates to be
deleted from the lexicon.  Parameters are generally deleted because other
parameters have rendered them superfluous.  To estimate the changes in
parameter counts that result from deleting a parameter $W$, the assumption
is made that each occurrence of $W$ is replaced by its representation.  Of
course, the parameters in the representation of $W$ under $G$ also have
their count reduced in $G^{*}$ because $W$ no longer needs to be
represented.  Then $c_{G^{\ast}}(W) = 0$ and (compare with
equation~\ref{eq:adddelta})

\[ c_{G^{\ast}}(w) \approx c_{G}(w) - c_{G}(w\in W) + c_{G}(W)c_{G}(w\in W). \]

In some cases the independence assumption considerably slows the
convergence of the learning algorithm.  In particular, it can be the case
that exactly one of a set of parameters is necessary (any one), but that
the entire set is deleted because the algorithm computes changes in
description length under the assumption that only one parameter is deleted
at a time.  This problem can be mostly eliminated by deleting parameters
sequentially and checking whether the Viterbi representation for a
parameter has changed before deleting it.  If it has changed, it is an
indication various assumptions made in the description length calculations
have been violated, and the parameter should be retained: it can always be
deleted during the next iteration of the learning algorithm.

\subsection{Convergence}

The algorithm given above, as outlined in figure~\ref{fig:calg}, does not
necessarily converge.  Parameters are added and deleted if it is {\em
  estimated} that this will reduce the description length of the evidence
and lexicon.  Although these estimates are remarkably accurate, in some
cases when parameters are only marginally justified they may be added and
deleted in an endless cycle because of mismatches in the errors of the
creation and deletion estimation procedures.  This phenomena has almost no
effect on either linguistic structure or description length, and generally
occurs only after the vast majority of the lexicon has been fixed.

Various tests can be imposed to stop the algorithm.  For example, the
algorithm could stop after any iteration that increases the net description
length, or when the number of parameters added or deleted drops below some
threshold.  On the data sets that the algorithm has been tested on, the
algorithm has always ceased any significant learning after 15 iterations,
and it is as convenient to simply run the algorithm for 15 iterations
regardless.

\subsection{Computational Complexity}

Let $i$ be the number of iterations of the learning algorithm, $l$ be the
length of the evidence (in characters), $g$ be the length of the largest
lexicon attained during training (in characters), $p$ be the length of the
longest parameter in the lexicon during training (in characters), and $c$
be the size of the largest set of candidate changes to the lexicon.  Then
the time complexity of the stochastic optimization steps in each iteration
of the learning algorithm is ${\cal O}((l+g)p)$.

The process of adding and deleting parameters involves two steps, the
recording of statistics and the estimation of $\Delta$'s.  For each
candidate change the estimation of $\Delta$ involves one pass of the
forward-backward algorithm with cost ${\cal O}(p^2)$ and then some simple
algebra that can be performed in essentially constant time.  Thus, the time
complexity of that portion of the algorithm is ${\cal O}(cp^2)$.

The statistics that must be recorded are the counts $c_G(w_1,w_2)$ for
every parameter pair $w_1$, $w_2$ that are composed in the Viterbi
representation of some utterance.  The calculation of the Viterbi
representations adds only a constant factor to the existing cost of the
forward-backward algorithm.  The calculation of the numbers $p_G(\awwb|u)$
is linear in the total length of the evidence and parameters and quadratic
in the length of the longest parameter, ${\cal O}((l+g)p^2)$.  The total
time complexity of the structural optimization step is therefore ${\cal
  O}(cp^2 + (l+g)p^2) = {\cal O}(p^2(c + l))$.  But in the implementations
that have been experimented with, the real cost of computing the
$c_G(w_1,w_2)$ statistics is the space complexity of their storage, ${\cal
  O}(c)$.  The number of parameter pairs that co-occur in Viterbi
representations can number in the millions for a large corpus.  It is the
expense of storing these pairs (before the pruning of all pairs that only
occur together once) that dominates the cost of the algorithm.  Experiments
have been performed in which triples $w_1 \circ w_2 \circ w_3$ are
considered as candidates for new parameters, and it is imperative under
such schemes to prune triples for which $c_G(w_1,w_2) < 2$ or $c_G(w_2,w_3)
< 2$, or the number of triples quickly exceeds reasonable storage
requirements on even moderately sized data sets.

The total time complexity of the algorithm is ${\cal O}(ip^2(c+l))$,
essentially linear in the length of the evidence.  This is as efficient as
could reasonably be expected and turns out to be quite practical (total
execution times on million-character inputs tend to be in the tens of
minutes on standard 1995 workstations).  It is the $c$ term that is the
limiting factor, and $c$ grows with $l$.  For natural-language evidence of
the type described in chapter~\ref{ch:results}, the algorithm can be run on
input tens of millions of characters long (on standard 1995 workstations)
without memory storage requirements becoming prohibitive.  For longer
input, more complex strategies may be necessary to reduce the effects of
the $c$ term.

\section{Extensions for Meaning}

The addition of the meaning perturbation operator described in
section~\ref{rep:meaning} does not alter the learning algorithm in any
fundamental way, though it does complicate some parts of it.  The
``parsing'' process in the E-step of the stochastic optimization subroutine
must be extended to simultaneously analyze the character sequence and the
meaning of an utterance.  The parameter creation and deletion procedures
must be slightly altered to take into account perturbations.  But roughly
speaking, the same architecture suffices for both models.  Only those
aspects of the meaning induction algorithm that differ from the previous
algorithm are discussed here; all else is assumed to be identical.

\begin{figure}
\pageline
\begin{center}
\begin{tabular}{ccc}
$u$ & $v$ & $p(z|v)$ \\ \hline
\word{john walked} & \sems{john walk} & .5\\
& \sems{john walk slow} & .2\\
& \sems{mary see john} & .1\\
& \sems{john see mary} & .1\\
\end{tabular}
\end{center}
\caption{\label{fig:meaninginput} A sample $u,z$ pair.  A sequence of such
  pairs is the input to the learning algorithm.}
\pageline
\end{figure}

The input to the learning algorithm under the extended model is a sequence
of pairs $u,z$, where $u = u_1\ldots u_l$ is a character sequence and $z$
is some summary of the extralinguistic environment.  It is assumed that
from $z$ the learner can compute the function $p(z|v)$ over $v$, where $v$
is a possible ``meaning'' for the utterance $u$, a set of sememes.  In the
tests that will be made of the learning algorithm, $p(z|v)$ is provided
explicitly for each $v$ that assigns $z$ a positive probability.  Thus, for
each utterance the input to the learning algorithm looks like the example
given in figure~\ref{fig:meaninginput}, where the interpretation is that
\word{john walked} must mean one of four things, with the meaning
\sems{john walk} slightly favored on the basis of extralinguistic
information alone.

Each parameter in the lexicon is stored as a character sequence and a set
of sememes.  The meaning of a parameter $w$ will be written $m(w)$.
Parameters do not have ambiguous meanings, unlike utterances.  To use the
same mechanisms for dealing with both, $p(z|v)$ is simply defined for
parameters to be 1 if $v = m(w)$, and 0 otherwise.  The algorithm starts
with a lexicon that consists only of the terminals, each assigned the empty
meaning.

\subsection{Optimization of Stochastic Parameters}

Stochastic optimization is again accomplished via the EM algorithm and the
maximization step (equation~\ref{eq:norm}) remains as stated, but the
computation of parameter counts $c_G(w)$ is complicated by the fact that
each utterance is now a two-tiered object.  In particular, the calculation
of the posterior probability $p_G(\awb|u)$ of an utterance spanning the
region $u_{a}\ldots u_{t-1}$, as expressed in equation~\ref{eq:span} must be
revised to take into account the influence of utterance meanings.

It is necessary to make the forward and backward probabilities a function
of meanings.  Let $\alpha_i(u,q)$ be the probability of the stochastic
model generating any complete representation $w_1\ldots w_o w_p\ldots w_n$ such that
$(u_1\ldots u_i,q) = w_1\circ\cdots\circ w_o$.  In other words,
$\alpha_i(u,q)$ is the probability that after some number of parameters
have been composed the stochastic model will have generated the prefix
$u_1\ldots u_i$ and the sememe set $q$.  Then $\alpha_0(u,\emptyset) = 1$
and

\[
\alpha_i(u,q) = \sum_{j=0}^{i} \sum_{q'\subset q} \alpha_j(u,q')
\sum_{w=u_{j+1}\ldots u_{i}\in G, w<u} p_G(w) \delta(q'\cup m(w),q).
\]

Further let the backward probability $\beta_i(u,z|q)$ be the probability of
the utterance-extralinguistic pair $u,z$ given that the stochastic model
generated the partial parameter sequence $w_1\ldots w_o$ such that
$(u_{1}\ldots u_{i},q) = w_1\circ\cdots\circ w_o$.  Then, following
equation~\ref{eq:meaninglength2}, $\beta_l(u,z|q) = p'(z|q)$, and

\[
\beta_i(u,z|q) = \sum_{j=i}^{l} \sum_{w=u_{i+1}\ldots u_{j}\in G, w<u}
p_G(w) \beta_j(u,z|q\cup m(w)).
\]

\noindent Notice that $p_G(u,z) = \beta_0(u,z|\emptyset)$.  The revised form
of equation~\ref{eq:span} is

\begin{equation}
  p_G(\awb|u,z) = \frac{\sum_{q} \alpha_{a}(u,q) p_G(w) \beta_{b}(u,z|q\cup
    m(w))}{p_G(u,z)}.\label{eq:qspan}
\end{equation}

\noindent (Equation~\ref{eq:wwcount} can be similarly transformed.)  The
final calculation of parameter counts remains as in
equation~\ref{eq:counts2}: $c_{G}(w) = \sum_{(u,z) \in U'}\sum_{a} \sum_{b}
p_{G}(\awb|u,z)$.  It will turn out to be useful to be able to compute for
each parameter and utterance the posterior probability $p_G(s|u,z)$ that
the representation includes a perturbation that adds or deletes the sememe
$s$.  This is analogous to the probability $p_G(\awb|u,z)$, and can be
computed (following equation~\ref{eq:pprime}) by

\begin{equation}
p_G(s|u,z) = \frac{\sum_{v'} \alpha_{l}(u,v')2^{-|S|}p(z|v'\otimes
  \{s\})}{p_G(u,z)}.\label{eq:scount}
\end{equation}

\noindent The total expected count of a perturbation $c_G(s)$ is then
$\sum_{(u,z)\in U'} p_G(s|u,z)$.  Calculations involved in the parameter
building process also require the expected count $c_G(w,s)$ of how many
times the parameter $w$ is used in a representation that also involves a
perturbation that adds or deletes the sememe $s$.  This is analogous to the
count $c_G(w_1,w_2)$ and can be computed by $c_{G}(w,s) = \sum_{(u,z)\in
  U'}\sum_{a} \sum_{b} p_{G}(\awb,s|u,z)$ where $p_G(\awb,s|u,z)$ is
computed by

\begin{equation}
  p_G(\awb,s|u,z) = \frac{\sum_{v'}2^{-|S|}p(z|v'\otimes\{ s \})
    \sum_{q}\alpha_{a}(u,q) p_G(w) \beta_{b}(u,v'|q\cup
    m(w))}{p_G(u)}.\label{eq:swcount}
\end{equation}

\subsubsection{A Factorial Representation of Probabilities}

Unfortunately, although these changes to the forward-backward algorithm are
conceptually simple, they turn it from a polynomial-time algorithm into one
that is exponential in the number of sememes.  This is because various
summations are made over the entire space of sememe sets.  Intuitively,
what has happened is that amount of information necessary to summarize the
state of the generation process has been expanded.  In the concatenative
model, given knowledge of $u$ all that is necessary to describe the state
of the generation process is an utterance location.  As a consequence, the
calculation of the forward and backward probabilities involves a sum over
utterance locations, and the number of forward and backward probabilities
that must be stored is equal to the length of the utterance.  In the
meaning model, given knowledge of $u$ and $z$ the state of the generative
process is summarized by both the utterance location {\em and} the sememes
that have been generated.  Thus, the calculations of forward and backward
probabilities involves a double sum, and probabilities must be stored for
every combination of location and possible sememe set.

There are several escapes from what seems to be a computational overload.
First realize that for any finite lexicon, only some of the forward
probabilities $\alpha_i(u,q)$ will be non-zero.  If only these are stored,
and backward probabilities for which forward probabilities are zero are
ignored, then the algorithm as it stands may be practical; this depends
heavily on the ambiguity of the lexicon.  It is also possible to use a
beam-search strategy, storing for each location only those forward
probabilities that are within some factor of the highest forward
probability for that location.  This risks introducing errors, but is
likely to be a viable strategy.  Another possibility, discussed at further
length here, is to store forward and backward probabilities using a
factorial representation.  This introduces various approximation errors,
but can substantially reduce computation in cases where the size of the
lexicon precludes using the other strategies.

The idea is to assume that the probability of a sememe being in the sememe
set of an utterance is independent of other sememes.  In other words, if
$p_G(u,v)$ is the probability of the language model $G$ generating an utterance
$u$ with meaning $v$, then

\[ p_G(u,v) = p_G(u) \prod_{s \in v} p_G(s|u) \prod_{s \in S-v} \overline{p_G(s|u)}, \]

\noindent where $s$ is a sememe drawn from the total set of sememes $S$.
This is of course not true in general.  For example, the probability of
\word{kicking the bucket} meaning \sems{kick bucket} is not the product of
two independent probabilities: either the phrase means ``to die'', in which
case neither sememe is in the meaning, or it means ``kicking the bucket'',
in which case both are.  However, the approximation can be surprisingly
effective, and has the advantage that the number of probabilities that need
to be computed stored is a linear function of $|S|$.

Let $\alpha_i(u,q) = \alpha_i(u) A_i(q|u)$ where $\alpha_i(u)$ is as
defined in section~\ref{stochopt} and $A_i(q|u)$ is the probability that
the representation $w_1\ldots w_o$ has collective meaning $q$ given that it
generates $u_1\ldots u_i$.  Let $\beta_i(u,v|q) = \beta_i(u) B_i(v|u,q)$
where $\beta_i(u)$ is as defined in section~\ref{stochopt} and $B_i(v|u,q)$
is the probability that $w_1\ldots w_n$ has collective meaning $v$ given
that $w_1\circ\cdots\circ w_o = (q,u_1\ldots u_i)$.  $B_i(u,z|q)$ is
defined in terms of $B_i(u,v|q)$ by

\[ B_i(u,z|q) = \sum_{v'} p'(z|v') B_i(u,v'|q). \]

Write $\langle q\rangle^s$ to mean 1 if $s\in q$ and 0 if $s \not\in q$.
Then the factorial approximation proceeds by assuming that

\begin{center}
  $A_i(q|u) \approx \prod_{s\in S} A^s_i(\langle q\rangle^s|u)$
  \hspace{.3in} and \hspace{.3in} $B_i(v|u,q) \approx \prod_{s\in S}
  B^s_i(\langle v\rangle^s|\langle q\rangle^s,u)$,
\end{center}

where $A^s_i(\langle q\rangle^s|u)$ and $B^s_i(\langle v\rangle^s|\langle
q\rangle^s,u)$ are marginal probabilities that can be computed by

\begin{eqnarray*}
  A^s_i(1|u) & = & \frac{1}{\alpha_i(u)}\sum_{j=0}^{i} \alpha_j(u)
  \sum_{w=u_{j+1}\ldots u_{i}\in G, w<u} p_G(w) (\langle m(w)\rangle^s +
  \overline{\langle m(w)\rangle^s} A^s_j(1|u))\\ 
  B^s_i(1|0,u) & = & \frac{1}{\beta_i(u)}\sum_{j=i}^{l} \beta_j(u)
  \sum_{w=u_{i+1}\ldots u_{j}\in G, w<u} p_G(w) (\langle m(w)\rangle^s +
  \overline{\langle m(w)\rangle^s} B^s_j(1|0,u))
\end{eqnarray*}

\noindent where $A^s_0(1|u) = 0$, $A^s_i(0|u) = \overline{A^s_i(1|u)}$,
$B^s_l(1|0,u) = 0$, $B^s_i(1|1,u) = 1$ and $B^s_i(0|x,u) =
\overline{B^s_i(1|x,u)}$.  The calculation of these marginal forward and
backward probabilities does not involve summations over all possible
meanings, and is hence linear in the size of the sememe set.  This still
leaves a summation over all possible meanings in the calculation of
parameter counts, in equations~\ref{eq:qspan}, \ref{eq:scount} and
\ref{eq:swcount}.  Fortunately, under the factorial assumption these
summations are equivalent to a more efficient product.  In the case of
equation~\ref{eq:qspan},

\begin{eqnarray*}
\lefteqn{p_G(\awb|u,z)}\\
& = & \frac{\sum_{q} \alpha_{a}(u,q)
  p_G(w) \beta_{b}(u,z|q\cup m(w))}{p_G(u,z)}\\
& = & \frac{\sum_{q} \alpha_{a}(u) \prod_s A_{a}^s(\langle
  q\rangle^s|u) p_G(w)
  \sum_{v'} p'(z|v') \beta_{b}(u) \prod_s B_{b}^s(\langle
  v'\rangle^s|u,\langle q\cup m(w)\rangle^s)}{p_G(u,z)}\\
& = & \frac{\alpha_{a}(u) p_G(w)
  \beta_{b}(u) \sum_{q} \prod_s A_{a}^s(\langle q\rangle^s|u) \sum_{v'}
  p'(z|v') \prod_s B_{b}^s(\langle v'\rangle^s|u,\langle q\cup
  m(w)\rangle^s)}{p_G(u,z)}\\
& = & \frac{\sum_{v'} p'(z|v') \alpha_{a}(u) p_G(w)
  \beta_{b}(u) \sum_{q} \prod_s A_{a}^s(\langle q\rangle^s|u)
  B_{b}^s(\langle v'\rangle^s|u,\langle q\cup m(w)\rangle^s)}{p_G(u,z)}\\
& = & \frac{\sum_{v'} p'(z|v') \alpha_{a}(u) p_G(w)
  \beta_{b}(u) \prod_s (A_{a}^s(0|u) B_{b}^s(\langle
  v'\rangle^s|u,\langle m(w)\rangle^s) + A_{a}^s(1|u)\langle
  v'\rangle^s)}{\alpha_l(u) \sum_{v'} p'(z|v') \prod_s A_l^s(\langle v'\rangle^s|u)}.
\end{eqnarray*}

This last form is much more efficient to compute, but still involves a sum
over the giant space of all utterance meanings $v'$.  There are many
different approximations that can be used to eliminate or simplify this
sum.  A surprisingly effective one, adopted here, is to first partition the
set of meanings into $n$ disjoint subsets $V_1\ldots V_n$ (where $n$ is
small).  Then assume that $p'(z|v' \in V_k) = \prod_s f_k^s(\langle
v'\rangle^s)$ where $f_k^s(x) = \sum_{v'\in V_k, \langle v'\rangle^s = x}
p'(z|v')$.  This allows $p_G(\awb|u,z)$ to be computed efficiently by

\begin{eqnarray}
  \lefteqn{p_G(\awb|u,z) =} \nonumber \\ 
  && \frac{\alpha_{a}(u) p_G(w) \beta_{b}(u)
    \sum_k \prod_s (f_k^s(0) A_{a}^s(0|u) B_{b}^s(0|u,\langle m(w)\rangle^s)
    + f_k^s(1) A_{a}^s(1|u))}{\alpha_l(u) \sum_k \prod_s (f_k^s(0) A_l^s(0|u) +
    f_k^s(1) A_l^s(1|u)}. \label{eq:qqspan}
\end{eqnarray}

Assuming input as in figure~\ref{fig:meaninginput} (or in a variety of
other natural forms), equation~\ref{eq:pprime} can be rewritten in a manner
that allows $f_k^s(1)$ and $f_k^s(0)$ to be computed efficiently.  

So long as the partition of utterance meanings into $V_1\ldots V_n$ is done
in such a way as to maximize the effectiveness of the factorial
representation, equation~\ref{eq:qqspan} results in an efficient and fairly
accurate method of approximating parameter counts.
Equations~\ref{eq:wwspan}, \ref{eq:scount} and \ref{eq:swcount} can be
similarly transformed to efficiently approximate $p_G(w_1,w_2)$,
$p_G(s|u,z)$ and $p_G(\awb,s|u,z)$.  For example,

\[
p_G(s|u,z) = \frac{2^{-|S|}\sum_{k}
  (A_l^s(1|u)f_k^s(0)+A_l^s(0|u)f_k^s(1)) \prod_{s'} (A_l^{s'}(1|u) f_k^{s'}(0)+
  A_l^{s'}(0|u)f_k^{s'}(1))}{\sum_k \prod_{s'} (f_k^{s'}(0) A_l^{s'}(0|u) +
  f_k^{s'}(1) A_l^{s'}(1|u)}.
\]

\subsubsection{Maintaining a Logically Consistent Lexicon}

It was possible in the base concatenative model to ensure consistency in
the lexicon by imposing on each component $w_i$ of a parameter $w$ the
requirement that $w_i<w$, where $<$ is defined in terms of the length of
the character sequences.  The meaning perturbation operator complicates
things.  The existing constraint is undesirable, because it prevents
representations as in figure~\ref{fig:meaning} where different forms
inherit from a common base:

\begin{center}
\begin{tabular}{lll}
\sword{bank}{} &=& \sword{b}{} $\circ$ \sword{a}{} $\circ$ \sword{n}{} $\circ$
\sword{k}{} \\
\sword{bank}{tilt} &=& \sword{bank}{} + \sem{tilt}\\
\sword{bank}{river-edge} &=& \sword{bank}{} + \sem{river-edge}\\
\sword{bank}{financial-institution} &=& \sword{bank}{} +
\sem{financial-institution}
\end{tabular}
\end{center}

A solution that seems plausible at first glance is to redefine the $<$
operator to be true of parameters of equal length if they have different
meanings.  But this still allows for cyclic representations like

\begin{center}
\begin{tabular}{lll}
\sword{bank}{tilt} &=& \sword{bank}{river-edge} + \sem{tilt} - \sem{river-edge}\\
\sword{bank}{river-edge} &=& \sword{bank}{tilt} + \sem{river-edge} - \sem{tilt}
\end{tabular}
\end{center}

\noindent where the learning algorithm never actually represents the
characters of \word{bank}.  Although this is an interesting problem that
becomes even more complicated when phonological perturbations are allowed,
a fairly uninteresting solution is adopted here.  The $<$ predicate orders
parameters first by length of character sequence, and then by number of
sememes.  Thus, \sword{bank}{tilt} can not be represented in terms of
\sword{bank}{river-edge} because the component parameter has an equal
number of characters and an equal or greater number of sememes.

\subsection{Refinement of Model Structure}

The procedures for adding and deleting parameters are not altered much when
the concatenative model is extended with the meaning perturbation operator.
The procedure for creating new parameters from the composition of two
existing ones is retained in {\em exactly} the same form.  The calculation
of the change in description length from deleting a parameter is only very
slightly altered by the fact that the parameter may include meaning
perturbations.  One additional type of change to the lexicon is considered,
the creation of a new parameter by combining an existing parameter with a
meaning perturbation.

It is necessary to extend equation~\ref{eq:delta} to take into account
changes in the number of perturbations in the complete representation.

\[
\Delta \approx \sum_{w\in G^{\ast}} -c_{G^{\ast}}(w)\log p_{G^{\ast}}(w) - \sum_{w\in G}
-c_G(w)\log p_G(w) + (\log |S|) \sum_{s\in S} (c_{G^{\ast}}(s) - c_{G}(s)).
\]

\subsubsection{Adding Parameters}

In addition to the method of building a new parameter from two existing
ones, a new type of change to the lexicon is considered: a new parameter
can be created by adding or removing a sememe from an existing parameter's
sememe set (leaving the original parameter intact).  The set of new
parameter candidates is constructed from parameter-perturbation pairs that
cooccur in the representation of the evidence and lexicon.  For example, if
under the grammar $G$ a representation for \sword{thecat}{cat} is
\sword{the}{} $\circ$ \sword{cat}{} + \sem{cat} then the parameters
\sword{the}{cat} and \sword{cat}{cat} are both candidates.  Again, only
pairs that occur in the Viterbi representation are considered, and again
the set of candidates can be pruned by eliminating all pairs that occur
fewer than two times in Viterbi representations.

For each candidate parameter $W$ with Viterbi representation $w \pm
s$, the expected count $c_G(w, s)$ is computed.  Then estimates of new
counts are made under the same assumptions used for the two-parameter case,
resulting in

\begin{eqnarray*}
c_{G^{\ast}}(w) & \approx & c_{G}(w) + c_{G^{\ast}}(w\in W) -
c_{G^{\ast}}(W)c_{G}(w\in W).\\
c_{G^{\ast}}(s) & \approx & c_{G}(s) + p_{G^{\ast}}(s|u,z) -
c_{G^{\ast}}(W)p_{G}(s|u,z).
\end{eqnarray*}

The computation of $\Delta$ thus requires estimates of $c_{G^{\ast}}(W)$,
$c_{G^{\ast}}(w\in W)$ and $p_{G^{\ast}}(s|W)$.  Here, we simply let
$c_{G^{\ast}}(W) \approx c_G(w,s)$, $c_{G^{\ast}}(w\in W) \approx
c_{G}(w\in W)$ and $p_{G^{\ast}}(w|W) \approx p_{G}(w|W)$.  Thus, the
parameter $W$ is parsed to find its representation under the existing
grammar, and this representation is assumed to be the one it will have
after the change also.  The effect of a subsequent deletion of the
parameter $w$ is added in to the computation of $\Delta$.

This estimation procedure is not very faithful to the compositional
framework, because it does not take into account the inheritance properties
as well as it might.  Consider the case where three parameters exist,
\word{cat}, \word{a cat} and and \word{the cat}, with the last two
parameters represented in terms of the first.  If only \word{a cat} and
\word{the cat} occur at the top level, then they may be considered for the
addition of the \sem{cat} sememe, but not \word{cat}.  \word{cat} will only
acquire it later, in an effort to reduce the description length of \word{a
  cat} and \word{the cat}.  Although the algorithm may eventually converge
to the ``right'' grammar, it does so in an unnecessarily circuitous
fashion.

\subsubsection{Deleting Parameters}

Consider the question of how much the total description length changes when
a parameter $W$ with representation $w_1 \circ \ldots \circ w_n + s_1
+\ldots + s_k - s_{k+1} - \ldots - s_m$ is deleted.  The assumption made
previously was that when a parameter is deleted, its representation takes
its place; this assumption is generally valid because a parameter's
representation is the shortest description of its content (at least before
the deletion in turn causes various other changes to the lexicon), and
hence the best substitute for the parameter.  This remains true when the
meaning perturbation operator is introduced.  Therefore, the only change to
the deletion procedure is a formula for estimating the changes in
perturbation counts that mirrors the original formula for estimating
changes in parameter counts.

\begin{eqnarray*}
  c_{G^{\ast}}(w) & \approx & c_{G}(w) - c_{G}(w\in W) + c_{G}(W)c_{G}(w\in
  W).\\
  c_{G^{\ast}}(s) & \approx & c_{G}(s) - p_{G}(s|W) + c_{G}(W)p_{G}(s|W).
\end{eqnarray*}

\section{Related Work}\label{relatedlearning}

The learning algorithms that have been presented in this chapter are
similar in many respects to algorithms presented by others who have
explored grammar induction and related fields.  These similarities arise
because of the domain (language), the specific task (the acquisition of a
lexicon), the nature of the underlying stochastic models (finite-state
machines), and the particular learning methods employed (alternating
stochastic and structural refinement).  Several bodies of research that
seem particularly relevant are discussed below, and compared to the
approach taken here.

\subsection{Grammatical Inference and Language Acquisition}

There have been many attempts to build computer programs that learn the
underlying structure of sequences; a common name for this line of research
is {\em grammatical inference}.\footnote{Historically (in line with Gold's
  view of (E-)language~\cite{Gold67}) the term grammatical inference has
  referred to the learning of a classification procedure from positive and
  negative examples that can predict whether a sentence is or is not in a
  language; see Biermann and Feldman~\cite{Biermann72} for a review.  More
  recently, researchers interested in building mechanisms that acquire the
  specific generative grammar believed to underly some input have also
  adopted the term to refer to their work; it is this (I-language) sense of
  grammatical inference that is used here.} Much of this effort has been
directed at human language, though DNA sequences, music scores, computer
traces and cryptographic codes are other common subjects of interest.
Grammatical inference is distinguished from language modeling, text
compression and many other tasks that may benefit from a predictive model
of the data in that the grammar is the objective, rather than merely a
tool.  Thus, researchers in grammatical inference often directly evaluate
grammars (or grammatical derivations) rather than the languages generated
by a grammar or a grammar's predictive ability.

This line of research has lead to many approaches that are similar to ours.
For example, Olivier~\cite{Olivier68},
Wolff~\cite{Wolff77,Wolff80,Wolff82}, Brent {\em et al.}~\cite{Brent95},
and Cartwright and Brent~\cite{Brent96,Cartwright94} all present algorithms
for the induction of word-like linguistic units from character and phoneme
sequences; these algorithms all rely on dictionary-based representations
similar to our multigram model (though usually no stochastic interpretation
is assigned).  With the exception of Olivier, all of this work has relied
on metrics similar to MDL to evaluate dictionaries.  Nevertheless this work
has not achieved impressive results, in the sense that the resulting
dictionaries and segmentations of the input have not agreed particularly
well with linguistic intuitions; this in part motivated this thesis.  The
reasons behind the failures harken back to the discussions of
chapter~\ref{ch:Bayes}: extralinguistic patterns are learned at the expense
of linguistic ones and words are made long in an effort to improve
stochastic models.

Much recent work has focused on the induction of context-free grammars or
variations
thereof~\cite{Baker79,Briscoe92,Carroll92,Carroll95,Chen95,Chen96,Cook76,Lari90,Pereira92,Stolcke94}.
The hierarchical nature of these grammars would seem on the surface to be
quite similar to our hierarchical, concatenative representation.  However,
algorithms designed for the induction of context-free grammars have not
performed well in practice.  Pereira and Schabes~\cite{Pereira92} attempt
to learn an English grammar by applying the inside-outside
algorithm~\cite{Baker79} (the EM-algorithm for stochastic context-free
grammars) to a grammar that contains all possible binary rules over a fixed
set of nonterminals and terminal parts-of-speech.  Although the end
grammars model the input moderately well from a predictive viewpoint, the
derivations assigned to sentences do not agree with human judgments.
Follow up work by Carroll and Charniak~\cite{Carroll95,Carroll92} achieves
similar results.  Stolcke~\cite{Stolcke94} and Chen~\cite{Chen95,Chen96},
by emphasizing structural induction to a greater extent, achieve better
results on artificial languages but again are unable to learn
natural-language grammars that reflect human judgments from real data.
Some of the reasons for these failures are given in de
Marcken~\cite{deMarcken95b}, and motivate the compositional representation
we use.  They can be divided into two categories.  First, search in the
space of context-free grammars is fraught with local optima.  This is
discussed at greater length below.  Second, grammars that contain many long
rules are favored over linguistically plausible grammars containing smaller
number of simpler rules, because such grammars involve fewer expansions,
and therefore fewer independence assumptions.  Again, because these
researchers have not adopted a compositional representation for their
grammars, they can not have the best of both worlds.

At the algorithmic level, there are three major differences between our
approach and the range of algorithms explored by the above researchers.
First and most fundamentally, our algorithm lumps the parameters in with
the input, giving them internal representations and forcing the grammar to
explain regularities within parameters.  Second, all of these algorithms
search by directly manipulating a single representation of the grammar and
the input.  In contrast, our algorithm does commit to one representation,
but stores parameters in terms of their surface content, leaving the
reconstruction of a representation as a parsing problem.  Finally, while
many of the above algorithms are motivated by MDL, they do not in general
invoke it explicitly.  Instead, ad hoc estimates of description length are
often used, usually based on symbol counts rather than adaptive generative
models.  The last two issues and their implications are taken up further
below.

\subsection{Induction of Finite-State Automata}

Stochastic finite-state automata, exemplified by Markov models and hidden
Markov models, are traditional modeling tools for sequences.  The
literature on the induction of finite-state automata has traditionally been
divided.  On the one hand there has been a great deal of study put into the
induction of non-stochastic finite-state automata from examples; see
Pitt~\cite{Pitt89} for a survey.  Because this problem taken at face value
is trivial (merely encode the positive examples directly into the model),
various optimization criteria have been imposed; for example,
Angluin~\cite{Angluin78} and Gold~\cite{Gold78} show that identification of
the minimum-size automaton consistent with a finite set of examples is
NP-complete.  This literature has not generally considered linguistic
applications (though see Berwick and Pilato~\cite{Berwick87b}, who use
Angluin's~\cite{Angluin82} notion of $k$-reversibility to acquire automata
for the English auxiliary system).  Since the classes of automata that are
generally used allow for arbitrary states and arbitrary transitions, it is
often difficult to imagine how these automata could be given linguistic
interpretations.

The other half of the literature on finite-state induction comes from the
stochastic modeling community, which has generally assumed fixed
finite-state backbones (often fully-connected) and concentrated on the
estimation of transition probabilities.  The classical solution to the
problem of estimating the transition probabilities of a hidden Markov model
is the Baum-Welch algorithm~\cite{Baum72}.  Again there is no obvious way
to assign a linguistic interpretation to either the resulting transition
probabilities or to the sequence of state transitions that occurs during
the generation of a sequence.

\subsection{Language Modeling}

The language engineering community has studied the problem of creating
stochastic models of word and characters sequences in depth, usually with
an eye to using such models as the prior probability in speech and
handwriting recognition applications.\footnote{Speech or handwriting
  production is modeled as a two stage process: an underlying sequence $x$
  (a word or character sequence) is generated and then an observable signal
  $y$ (speech or handwriting) is generated from $x$.  The recognition
  problem is to find the most likely underlying sequence $x$ given the
  observable $y$.  Then

\[ x = \mbox{\begin{array}[t]{c}\mbox{argmax}\\{\scriptstyle x'}\end{array}} p(x'|y) =
\mbox{\begin{array}[t]{c}\mbox{argmax}\\{\scriptstyle x'}\end{array}}
p(y|x')p(x'). \]

\noindent Thus, an important part of a recognition system is a prior
probability over word or character sequences, $p(x')$.  This same
noisy-channel methodology has been applied to problems of language
translation~\cite{Brown90}.} Markov models have generally been the tool of
choice, because there are no hidden aspects to the derivation of a
sequence, and therefore the stochastic optimization process is trivial (see
Kupiec~\cite{Kupiec92b} for a notable exception).  The most impressive
stochastic language model reported to date, with an entropy rate of 1.75
bits per character over the Brown corpus, was achieved by the IBM Language
Modeling Group~\cite{Brown92} using a Markov model over words with a
two-word context (a {\em trigram}); as with most work in language modeling,
their algorithms had access to a predefined lexicon.  Almost all successful
language models have relied on techniques like Markov and hidden Markov
models that do not assign linguistic interpretations to the generated
sequences.  Nevertheless, there have been some experiments in language
modeling that used underlying structures with natural linguistic
interpretations, such as the long-range trigram model of Della Pietra {\em
  et al.}~\cite{DellaPietra94}, based loosely on the link grammars of
Sleator and Temperley~\cite{Sleator91}.  The only cases that have met with
significant success (on language modeling grounds) have not demonstrated
that they actually produce derivations that agree with linguistic
intuitions.  The most pointed example of this is the multigram model,
discussed in the context of language modeling by Deligne and
Bimbot~\cite{Deligne95}.  Although this is the same model that is used here
(and that has been studied by others; see below), the model is used by them
in a different way (without the compositional representation) and does not
produce linguistically plausible segmentations of the input.  In fact,
Deligne and Bimbot do not seriously address the induction problem, starting
with all possible words and merely adjusting word probabilities.  As a
consequence, implausible words remain in the lexicon, though they may be
assigned low probabilities.

Thus, little of the considerable language modeling literature bears
directly on the language acquisition problem.  It is quite possible, and in
fact common practice, to model the stochastic properties of language
without using techniques that reflect linguistic reality.  It is an
interesting question, answered in chapter~\ref{ch:results}, whether the
linguistically motivated algorithms presented here perform better than
traditional language modeling techniques on the stochastic modeling task.

\subsection{Text Compression}\label{textcomp}

The data compression community has also studied finite-state models in
depth.  Text compression techniques are in general more relevant to
language acquisition than language modeling techniques, because little
prior knowledge tends to be encoded into compression algorithms, and thus
they usually incorporate structural induction mechanisms.  Bell {\em et
  al.}~\cite{Bell90} provide an excellent introduction to the problems and
methods of compression, and in particular, text compression.  Popular text
compression schemes can be divided into four classes: those based on
adaptive frequency techniques like Huffman codes; those based on context
models~\cite{Rissanen83} (such as the PPM
algorithm~\cite{Cleary84,Moffat90,Teahan96}, probably the most effective
widely-used method for text compression); those based on hidden Markov
models~\cite{Bell90} (these are less common); and those based on dictionary
methods.  Only the dictionary methods, exemplified by the LZ78~\cite{Ziv78}
and LZW~\cite{Welch84} algorithms, have underlying models that can easily
be assigned linguistic interpretations.

Dictionary-based text compression techniques are variable-length block
coding schemes, very similar to the multigram model.  They compress text by
building a dictionary of words, each word a character string.  Words are
referenced via codewords.  The difference between dictionary-based
compression techniques and our methods stems from the manner in which
dictionaries are constructed.  In our algorithm, the dictionary is
iteratively refined.  Compression algorithms are generally designed for
speed, and make a single pass over the input incrementally building the
dictionary (often this improves compression by allowing the algorithm to
adapt to nonstationary input).  As a consequence, deterministic (and
usually greedy) strategies are used to build the dictionary.  For example,
given some remaining input $u$, the LZW coder proceeds by writing the
codeword of the longest prefix $w$ of $u$ that is in the dictionary, and
then a fixed code for the following character, $c$.  Both the encoder and
the decoder then add the new word $wc$ to their dictionary.  Thus, for
every codeword that is written a new word is also created.  This
compression technique has been proven to asymptotically approach the
entropy of any Markov source~\cite{Ziv78}.

Through the derivational history of words, algorithms like LZW implicitly
define a hierarchical structure in the lexicon (in the case of LZW, a
left-branching tree).  The LZMW algorithm~\cite{Miller84}, which is like
LZW except that the dictionary is built by concatenating two words rather
than one word and a character, constructs a hierarchy that is very similar
in spirit to our compositional representation.  However, because these
algorithms do not iteratively restructure the dictionary and rely on greedy
on-line parsing strategies, the lexical hierarchies they generate do not
agree very well with linguistic intuitions.  In fact, in one of the
earliest empirical works in natural language grammar induction,
Olivier~\cite{Olivier68} built an algorithm very similar to LZMW, and its
failings were a principal motivation for this thesis.

\subsubsection{Nevill-Manning's {\tt Sequitur}}

Recently Nevill-Manning~\cite{NevillManning96} has described {\tt
  Sequitur}, a text compression algorithm with remarkable similarities to
our concatenative algorithm, also motivated in part by arguments related to
language acquisition.  {\tt Sequitur} constructs a deterministic
context-free grammar that generates the input.  The grammar obeys the
following constraints: no symbol sequence in the grammar is repeated ($S
\Rightarrow abcdbc$ violates this constraint, whereas $S \Rightarrow aBdB$,
$B\Rightarrow bc$ does not), and every rule in the grammar is used at least
twice.  Figure~\ref{fig:sequitur} presents a trace of {\tt Sequitur}'s
execution on the input $abcdbcabcd$, taken from
Nevill-Manning~\cite{NevillManning96}.  As should be clear, the end result
is a grammar that is similar to the representations our algorithm would
produce.

\begin{figure}
\pageline

\begin{center}
\begin{tabular}{lcll}
Input So Far & & Resulting Grammar & Violated Constraints\\ \hline
$a$ & & $ S \Rightarrow a$ \\
$ab$ & & $ S \Rightarrow ab$ \\
$abc$ & & $ S \Rightarrow abc$ \\
$abcd$ & & $ S \Rightarrow abcd$ \\
$abcdb$ & & $ S \Rightarrow abcdb$ \\
$abcdbc$ & & $ S \Rightarrow abcdbc$ & $bc$ occurs twice \\
& $\dagger$ & $ S \Rightarrow aAdA, A \Rightarrow bc$ \\
$abcdbca$ & & $ S \Rightarrow aAdAa, A \Rightarrow bc$ \\
$abcdbcab$ & & $ S \Rightarrow aAdAab, A \Rightarrow bc$ \\
$abcdbcabc$ & & $ S \Rightarrow aAdAabc, A \Rightarrow bc$ & $bc$ occurs twice\\
& $\dagger$ & $ S \Rightarrow aAdAaA, A \Rightarrow bc$ & $aA$ occurs twice\\
& $\dagger$ & $ S \Rightarrow BdAB, A \Rightarrow bc, B \Rightarrow aA$\\
$abcdbcabcd$ & & $ S \Rightarrow BdABd, A \Rightarrow bc, B \Rightarrow
aA$ & $Bd$ occurs twice\\
& $\dagger$ & $ S \Rightarrow CAC, A \Rightarrow bc, B \Rightarrow
aA, C \Rightarrow Bd$ & $B$ used only once\\
& $\ddagger$ & $ S \Rightarrow CAC, A \Rightarrow bc, C \Rightarrow aAd$\\
\end{tabular}
\end{center}

\caption{\label{fig:sequitur}A trace of {\tt Sequitur}'s execution on the
  input $abcdbcabcd$.  Lines marked $\dagger$ depict rule creation
  operations very similar to our create-parameter-from-two-parameters
  operation, and lines marked $\ddagger$ depict rule-deletion operations
  very similar to our parameter deletion operation.}

\pageline
\end{figure}

There are several key differences between {\tt Sequitur} and our algorithm.
First, {\tt Sequitur} is in one sense incremental-- it proceeds in a single
pass over the input from left to right, adding characters to the top-level
rule.  It avoids many of the drawbacks of the greedy schemes of LZW and
LZMW by restructuring the grammar whenever it violates one of the two
constraints, by adding and deleting rules and changing rule
representations.\footnote{In this sense, it is not incremental: it must
  store the entire input in the current grammar so that it can make changes
  arbitrarily far back.} Unlike our algorithm, these updates do not involve
completely reparsing the input and grammar, but only local modifications.
Second, although {\tt Sequitur} is motivated with description-length
arguments, there is no evaluation function for the grammar-- the grammar is
only restructured to ensure that every repeated sequence is represented by
a rule and that every rule is used at least twice (this is why {\tt
  Sequitur} does not need to reparse).  Although this makes for an
efficient algorithm, it means that there are many possible valid grammars.
Nevill-Manning acknowledges this, and also that grammars often do not
conform to linguistic intuitions.  He proposes that these problems be
solved by using domain-specific heuristics to decide how to modify the
grammar.  He does consider our solution, the reparsing of the grammar under
a global evaluation function, but rejects it for several reasons:

\begin{enumerate}
\item It is not clear how different grammars can be compared.
\item Local changes to the grammar propagate, forcing other changes.
\item Changes are difficult to undo.
\end{enumerate}

Note that our algorithm solves all of these problems.  First, since the
notion of description length is taken seriously and stochastic grammars are
used, representations can be compared according to the MDL principle.
Second, since the use of parameters (rules) is independent of their
representation, parameters can be restructured without worry that this will
force other changes.  And finally, since parameters are represented in the
algorithm by their content rather than their representation, there is never
a worry that changes to representations can not be undone.

Comparisons of the compression performance of the two algorithms is given
in chapter~\ref{ch:results}.  Nevill-Manning discusses the problem of text
segmentation and presents some hierarchies (similar to
figure~\ref{fig:nfl}) for sample sentences, but does not present
segmentation results in a form suitable for comparison.

\subsection{Orthographic Segmentation}

Languages such as Chinese do not separate words in their orthography, just
as in English writing no explicit divisions are made between sub-word units
like syllables.  Since Chinese words are of variable length, most sentences
are ambiguous with respect to word boundaries, even given knowledge of a
dictionary.  As a consequence, even the most rudimentary language
processing tasks require a complex segmentation process (see for review Wu
and Tseng~\cite{Wu93}).  Most researchers attacking the segmentation
problem have assumed access to a dictionary.  The standard approach is to
build a stochastic finite-state model of sentences based on words (perhaps
a multigram) and then find the maximum-likelihood segmentation of a
sentence using the forward-backward algorithm.  The greatest challenge to
this problem comes from unknown words and proper names that are not in
dictionaries~\cite{Sproat94,Wang92}.  Thus, an important problem in
processing text in languages like Chinese is the discovery of words in an
environment where word boundaries are uncertain.  The only difference
between this problem and ours is that we start with no prior knowledge of
the lexicon.

However, most of the techniques commonly used to discover new words for
segmentation tasks are either application specific (Sproat {\em et
  al.}~\cite{Sproat94} and Chang {\em et al.}~\cite{Chang92} discuss methods
for learning Chinese names that are based on their idiosyncratic
properties, and Chang {\em et al.}~\cite{Chang95} judge new Chinese words by
their similarity to existing words) or very similar to the more general
lexical induction schemes of Olivier~\cite{Olivier68}, Cartwright and
Brent~\cite{Cartwright94}, etc.  Thus, if applied to the task of learning
words from scratch, most of these algorithms would either be inappropriate
or suffer from many of the same problems as the algorithms already
discussed in the section on grammatical inference.  One potential exception
to this is Luo and Roukos~\cite{Luo96}, who learn words in Chinese starting
from scratch and use a cross-validation technique to keep from building
too-large words.

\subsection{Search Procedures}

The search procedures used for grammatical inference and language modeling
generally fall into one of two classes.  Members of the first class, found
here and in the work of Olivier~\cite{Olivier68}, Cook {\em et
  al.}~\cite{Cook76} Wolff~\cite{Wolff80,Wolff82},
Ellison~\cite{Ellison92,Ellison94}, Nevill-Manning~\cite{NevillManning96},
Cartwright and Brent~\cite{Cartwright94}, Chen~\cite{Chen95,Chen96},
Stolcke~\cite{Stolcke94} and others, iteratively update the underlying
structure of the grammar.  (Some, like our algorithms, start with the most
general grammar possible while others, like Stolcke's, start with the most
specific grammar possible.)  Members of the second class, exemplified by
the work of Pereira and Schabes~\cite{Pereira92}, Deligne and
Bimbot~\cite{Deligne95}, Briscoe and Waegner~\cite{Briscoe92} and Lari and
Young~\cite{Lari90}, pick an extremely general structural backbone for a
stochastic model, and proceed by optimizing its stochastic properties,
usually through the EM procedure.  For example, Pereira and Schabes train a
giant stochastic context-free grammar containing all possible rules of a
certain form.  The language-specific properties of their grammar emerge
through the rule probabilities.  It is difficult to evaluate the linguistic
properties of grammars produced by the second class directly, but they can
be judged on the basis of the (maximum-likelihood) derivations they assign
to utterances.

In general, the second class of learning algorithms has fared more poorly
than the first.  The reason, as discussed by Pereira and
Schabes~\cite{Pereira92} and de Marcken~\cite{deMarcken95b}, is that the
hill-climbing inside-outside algorithm is incapable of making the complex
moves in grammar-space necessary to escape local optima.  As a consequence,
these learning algorithms quickly get stuck near their starting point, with
little learning having taken place.  The first class of learning algorithms
has a potential escape from this problem.  These algorithms (including
ours) incorporate mechanisms for altering the linguistic structure (and
stochastic properties) of the grammar that can be designed to perform
essentially arbitrary moves, including those that would stump the EM
algorithm.  Of course some of these algorithms (like ours) {\em also} use
the EM algorithm to optimize stochastic properties along the way.

Almost all algorithms of the first type define a set of candidate changes,
and an evaluation function.  Some, like Ellison's~\cite{Ellison92}, use a
simulated-annealing approach where a change may be accepted even if it
results in a poorer score from the evaluation function.
Stolcke~\cite{Stolcke94} uses incremental count-change techniques very
similar to ours to estimate changes in description length.  Others define a
simpler evaluation function and do not need to utilize formulas like
equation~\ref{eq:delta}.  However, all of the algorithms mentioned still
suffer from local minima problems, though admittedly to a lesser extent
than the purely stochastic methods of the second type.  This is because
these algorithms maintain a single grammar, stored and manipulated in terms
of its representation.  As pointed out in de Marcken~\cite{deMarcken95b},
moves that are relatively simple to express at a conceptual level may
involve quite substantial changes to the representation of a grammar.  For
example, imagine a context-free grammar that generates the structure on the
left:

\newcommand{\AP}{\mbox{\em AP}}
\newcommand{\SAP}{\mbox{\small\em AP}}
\newcommand{\BP}{\mbox{\em BP}}
\newcommand{\SBP}{\mbox{\small\em BP}}
\newcommand{\CP}{\mbox{\em CP}}
\newcommand{\SCP}{\mbox{\small\em CP}}

\begin{center}
\begin{small}
\begin{tabular}{ccc}
\hspace{-.4in}
\leaf{$A$}
\leaf{$B$}
\branch{1}{\SBP}
\branch{2}{\SAP}
\leaf{$C$}
\branch{2}{\SCP}
\branch{1}{$S$}
\tree & \hspace{1.5in} &
\hspace{-.4in}
\leaf{$A$}
\branch{1}{\SAP}
\leaf{$B$}
\branch{2}{\SBP}
\leaf{$C$}
\branch{2}{\SCP}
\branch{1}{$S$}
\tree \\
\end{tabular}
\end{small}
\end{center}

To capture the idea that $A$ adjoins to $B$, rather than the other way
around (the sort of change one might well imagine a learning algorithm
wanting to make), the learning algorithm must change the grammar to produce
the structure on the right.  This involves changes to three nonterminals!

\newcommand{\rrule}[2]{\(#1 \Rightarrow #2\)}
\begin{center}
\begin{small}
\begin{tabular}{lcl}
\rrule{\SAP}{A\ \SBP} & $\longrightarrow$ & \rrule{\SAP}{A} \\
\rrule{\SBP}{B} & $\longrightarrow$ & \rrule{\SBP}{\SAP\ B} \\
\rrule{\SCP}{\SAP\ C} & $\longrightarrow$ & \rrule{\SCP}{\SBP\ C} \\
\end{tabular}
\end{small}
\end{center}

These sort of big changes are in general too complex to code into the
hypothesis-generating mechanisms, and as a consequence the stochastic
context-free grammar induction algorithms based on structural updates fare
only slightly better than those based on stochastic changes alone.

The learning algorithms presented in this thesis are fundamentally
different than those just mentioned.  The grammar is not stored in terms of
a single representation.  Instead, the parameters (words) of the grammar
are stored in terms of their content.  This is a character sequence in the
concatenative model and a character sequence and a set of sememes in the
meaning model.  In an instantiation of the framework based on context-free
grammars, the above trees would be stored $S \Rightarrow A B C$.  As a
consequence, the algorithm implicitly stores many different possible
representations, and can reconstruct them at any time by parsing the input
and parameters.  Thus, there is no idea of incrementally changing the
representation of a word or a rule.  Every iteration of the learning
algorithm recreates representations from scratch.  For this reason, very
substantial changes can occur (or be undone) in one step.  For example, it
is quite common for the following sort of change to happen (here to the
most likely representation of the word \word{watermelon})

\begin{center}
\word{wa} $\circ$ \word{term} $\circ$  \word{el}
$\circ$ \word{on}\\
$\downarrow$ \\
\word{water} $\circ$ \word{melon}
\end{center}

\noindent Such a change in representation might be triggered by an increase
in probability of the word \word{water}, and would not involve multiple
steps as it would in Nevill-Manning's {\tt Sequitur} or other algorithms
based on the standard technique of directly manipulating representations.

\subsection{The Use of MDL}

Much of the related work that has been presented relies on evaluation
functions that are based on notions of description length.  However, our
methodology is unique in that the evaluation function used (based on
equation~\ref{eq:delta}) is a very close approximation of the description
length actually achieved by versions of our algorithms that generate a
complete description (for text compression).  In contrast, Brent {\em et
  al.}~\cite{Brent95}, Cartwright and Brent~\cite{Cartwright94},
Chen~\cite{Chen95,Chen96}, Ristad and Thomas~\cite{Ristad95} and others
that invoke the MDL principle all compute ad hoc estimates of description
length (often based on symbol counts) that do not closely reflect the best
possible encodings of their grammars (Stolcke~\cite{Stolcke94} is more
careful).  Although it is not clear exactly how much this affects
performance, it is worth noting that by assuming naive, nonadaptive
encoding schemes for parameters, these researchers are unnecessarily
penalizing parameters.  Ristad and Thomas, for example, demonstrate that by
accepting parameters that their evaluation function estimates to increase
total description length, generalization performance is improved.

\subsection{Learning Meanings}

There have been many efforts to build computer programs that learn word
``meanings'' from paired sequences of text and semantic representations.
This work includes studies of language acquisition (see
Selfridge~\cite{Selfridge81}, Siklossy~\cite{Siklossy72} and
Siskind~\cite{Siskind92,Siskind93b,Siskind94}); parameter estimation
schemes for machine translation, where sentences in a second language
substitute for semantic input (see Brown {\em et al.}~\cite{Brown93} and
Berger {\em et al.}~\cite{Berger96}); and parameter estimation schemes for
systems that classify utterances (see Tishby and Gorin~\cite{Tishby94}).

The learning algorithm presented in this chapter for the concatenative
model extended with the meaning perturbation operator advances previous
work in many ways.  First, unlike all of the other work cited, it does not
assume presegmented input.  This is a very substantial difference.  Most
other work has relied on knowing exactly what words are in each sentence;
many do not cope well with homonomy.  Our algorithm functions despite the
possibility of massive ambiguity in both the utterance meaning and in the
segmentation of the text stream.  Second, to our knowledge ours is the only
algorithm that learns a representation that shares structure.  Other
algorithms, treating words of the input as arbitrary symbols, must learn
the meanings of \word{walk} and \word{walked} independently.  In contrast,
our algorithm allows \word{walked} to be represented in terms of
\word{walk}, and to share its sememes.  Third, to our knowledge ours is the
only algorithm that allows meanings to be mapped to lexical units that are
not presented in the input.  For example, \word{walk} can receive meaning
even if it never appears in the input (if \word{walks} and \word{walking}
do).  Furthermore, \word{kicking the bucket} can be assigned a meaning even
though it is a 3-word sequence.  Finally, our algorithm is the only one
that offers an alternative to purely compositional behavior.  All other
methods, like ours, assume that when two words are combined, their meanings
compose in some natural way.  This allows them to explain the meaning of
unremarkable phrases like \word{red ball}, but not idiomatic ones like
\word{random variable}.  To handle \word{random variable}, it must be
marked in the input as a single word, and then its meaning will be learned
independently of \word{random} and \word{variable}.  In contrast, our
algorithm can explain how the word can inherit meaning from its components
while still introducing idiosyncratic properties.

\chapter{Results}\label{ch:results}

This chapter presents the results of various tests of the two learning
algorithms presented in chapter~\ref{ch:algorithm}.  The tests explore both
the linguistic and the statistical properties of the lexicons produced by
the algorithms.  Given the compositional framework underlying the
algorithms, it is hoped that they will produce lexicons that conform to our
linguistic intuitions and at the same time accurately reproduce the
statistical properties of the input.

Several different types of tests are presented.  First, the basic
concatenative algorithm is applied to the Brown and Calgary text corpora.
Both are standard benchmarking suites for language modeling and
compression, and the statistical performance of our algorithm is evaluated
and compared to well-known compression algorithms and language modeling
techniques.  Then, to test the linguistic properties of the same algorithm,
it is applied to the Brown corpus again (this time with punctuation and
segmentation information removed) and also to a large corpus of
(unsegmented) Chinese.  The resulting hierarchical segmentations are
compared to the ``true'' segmentations of the input.  The algorithm is also
applied to phoneme sequences derived automatically from continuous speech.
This demonstrates the algorithm's ability to learn words from input that is
in many ways {\em more} complex than that children are exposed to.
Finally, the extended algorithm is applied to unsegmented text paired with
artificial representations of sentence meanings.  Performance is measured
by the algorithm's ability to reconstruct meanings of new sentences given
access only to their text.

\section{Compression and Language Modeling}

Although the primary focus of this thesis is language acquisition, it is
important to explore the purely statistical performance of learning
algorithms independently of the linguistic representations they produce.
Such tests provide the simplest introduction to the algorithms.
Furthermore, language modeling and compression are important applications
in their own right.

\subsection{Input}

The concatenative algorithm of chapter~\ref{ch:algorithm} was run on two
bodies of text, the Brown corpus~\cite{Francis82} and the Calgary
corpus~\cite{Bell90}.  The Brown corpus is a diverse million-word
(approximately 40,000 sentence) corpus of English text, divided into 15
sections by document type and further into 500 documents of about 2000
words each.  The text ranges from romance novels to political commentary to
music reviews, and dates from 1961.  The Calgary corpus is a standard
collection of documents used to test compression schemes; the text portions
consists of a fiction and nonfiction book, a bibliography, USENET articles,
a console transcript and some computer programs.

\subsection{Method}

\begin{figure}[tbp]
\pageline
\begin{small}
\vspace{-.05in}\verb*|   the fulton county grand jury said friday an investigation of atlanta's recent primary electi|\\
\verb*|on produced "no evidence" that any irregularities took place.|\\
\vspace{-.05in}\verb*|    the jury further said in term-end presentments that the city executive committee, which had|\\
\vspace{-.05in}\verb*| over-all charge of the election, "deserves the praise and thanks of the city of atlanta" for t|\\
\verb*|he manner in which the election was conducted.|\\
\vspace{-.05in}\verb*|    the september-october term jury had been charged by fulton superior court judge durwood pye|\\
\vspace{-.05in}\verb*| to investigate reports of possible "irregularities" in the hard-fought primary which was won b|\\
\verb*|y mayor-nominate ivan allen jr&.|\\
\vspace{-.05in}\verb*|    "only a relative handful of such reports was received", the jury said, "considering the wid|\\
\verb*|espread interest in the election, the number of voters and the size of this city".|\\
\vspace{-.05in}\verb*|    the jury said it did find that many of georgia's registration and election laws "are outmod|\\
\vspace{-.05in}\verb*|ed or inadequate and often ambiguous".|
\end{small}
\caption{\label{fig:brown}The first five sentences of the Brown corpus as
  used for statistical tests.}
\pageline
\end{figure}

The text of the Brown corpus was broken into sentences\footnote{Where a
  sentence is a character sequence ended by a period, exclamation point or
  question mark.  Word-internal punctuation (as the period in ``Mr.'')  is
  denoted with ampersands in the Brown corpus).} and converted to lower
case; the resulting alphabet is 64 characters.  A small sample of the
corpus as seen by the algorithm is given in figure~\ref{fig:brown}.  The
Calgary corpus was broken into units at 1024 character intervals, but not
otherwise altered.  The only consequence of the pre-segmentation of the
input into smaller units is that words can not cross these boundaries.
The segmentations are introduced for implementational convenience, so that
the forward-backward algorithm does not need analyze the entire input in
one step.  The Brown corpus was converted to lower case so that the
learning algorithm does not introduce additional parameters to model
capitalized words at the start of sentences; Brown {\em et
  al.}~\cite{Brown92} demonstrate that case distinctions contribute at most
0.04 bits per character to the entropy rate of the Brown corpus.

For compression tests, the learning algorithm is run for 15 iterations,
each iteration (as per figure~\ref{fig:calg}) a two-step process where
first new words are added to the lexicon, and then existing words are
deleted.  The version of the algorithm tested here is a slight variation of
that presented in chapter~\ref{ch:algorithm}: it builds new parameters $w$
by considering both two parameter sequences ($w = w_1 \circ w_2$) {\em and}
three parameter sequences ($w = w_1 \circ w_2 \circ w_3$).  Because the
number of new parameters that can be added to the lexicon in a single
iteration is sometimes computationally burdensome, the algorithm is
arbitrarily limited to adding no more than 20,000 words in each iteration.

The coding scheme of figure~\ref{fig:code} is used to compute the final
description length of the input: a special pass of the stochastic
optimization routine is made over the input and parameters in which only
the most likely (Viterbi) representations are considered.  This produces
the counts and representations needed for the coding scheme.

\subsection{Brown Corpus Compression Results}\label{results:compression}

\begin{figure}[tbp]
  \pageline
\begin{center}
\mbox{\ }\psfig{figure=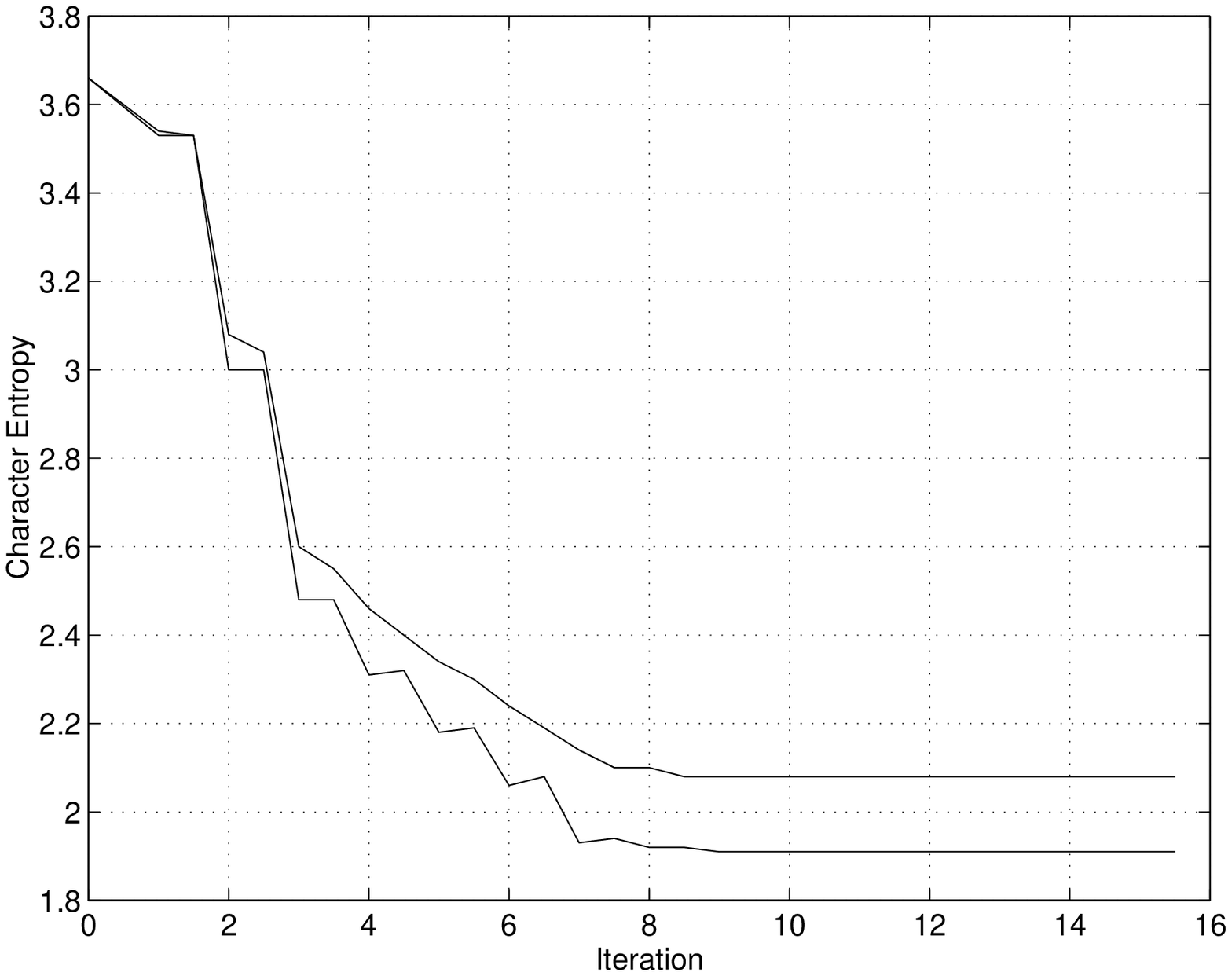,height=2in,width=4.0in}
\mbox{\ }\psfig{figure=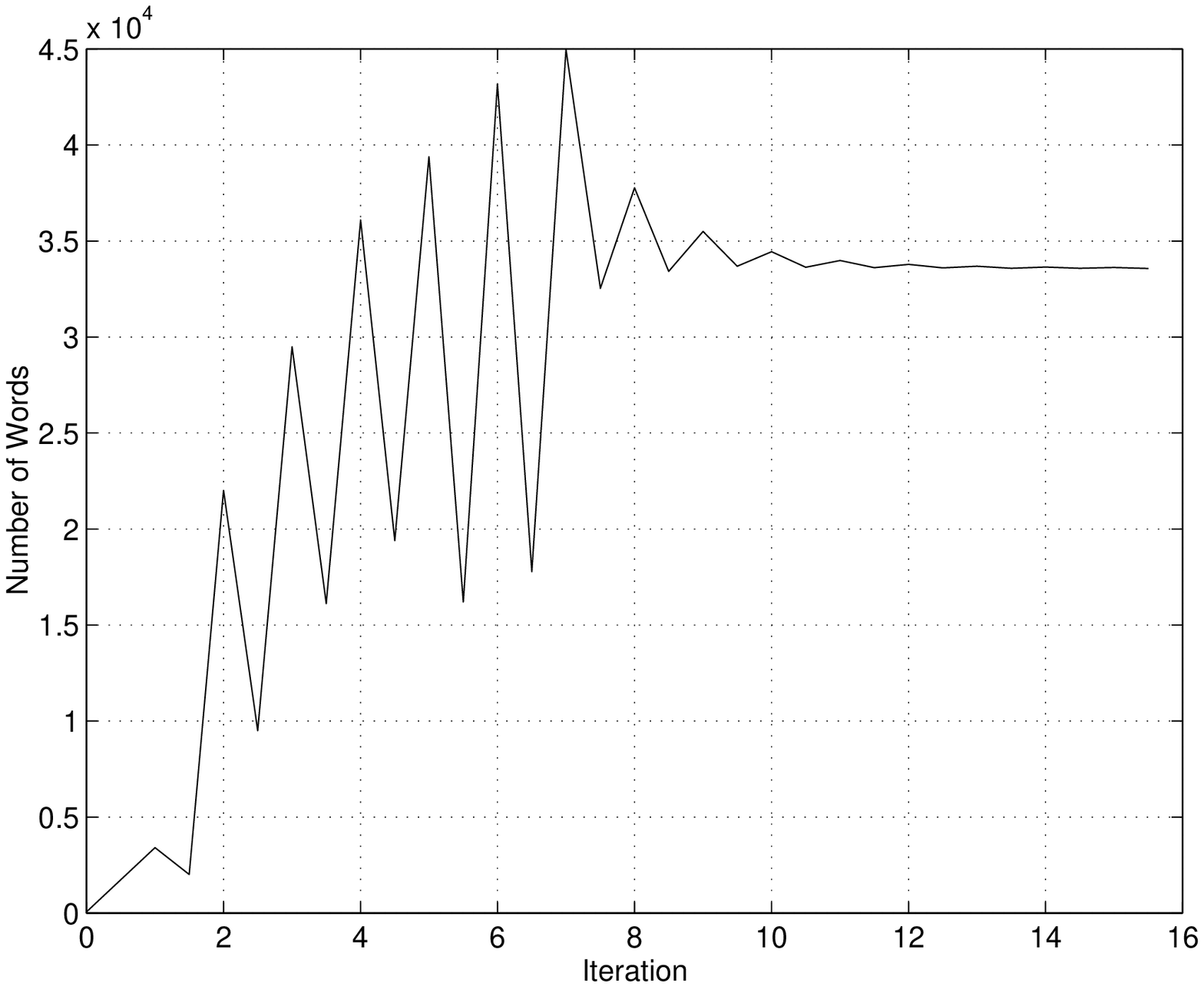,height=2in,width=4.0in}
\end{center}
\vspace{-.25in}
\caption{\label{fig:learning}Learning curves for the algorithm on the Brown
  corpus.  The top graph plots model performance.  The upper line is the
  compression rate-- the complete description length divided by length of
  the input.  The lower line discounts the cost of parameters: it is the
  cross-entropy rate of the model with the input.  The bottom graph plots
  the number of words in the lexicon.} \pageline
\end{figure}

When run on the Brown corpus, our algorithm compresses the input from
48,032,256 bits (each character stored as an 8-bit byte) to 12,530,415
bits, a ratio of 3.83:1 and a compression rate of 2.09
bits/char.\footnote{During the learning process, when probabilities are
  computed over all representations, and details of practical coding
  schemes are not considered, the estimated rate is 2.08 bits/char.
  Actually writing out all bits necessary to reproduce the input using the
  coding scheme in figure~\ref{fig:code} gives the 2.09 figure.} Compare to
3.40 bits/char (2.35:1) for the LZ78-based~\cite{Welch84,Ziv78} UNIX {\tt
  compress} program and 3.02 bits/char (2.65:1) for the
LZ77-based~\cite{Ziv77} UNIX {\tt gzip} program.  Figure~\ref{fig:learning}
presents learning curves for the algorithm across the 15 iterations.  Each
iteration receives two data points, the first depicting performance after
new words have been added to the lexicon, and the second point depicting
performance after existing parameters have been deleted (the 0 point is
performance with the 64 terminals alone).  As would be hoped, the complete
description length monotonically decreases.  However, as is visible from
the lower line on the top graph, the description length of the input does
{\em not} monotonically decrease: when words are deleted the description
length of the input increases, though this is more than compensated for by
the savings in the lexicon.  The number of words in the lexicon increases
non-monotonically from 64 to 33,569.  During some iterations (iteration 5,
for instance) the number of words decreases, though the model improves.
Near the end of the learning process changes are still taking place but
they have almost no effect on modeling performance.

\begin{figure}[p]
  \pageline \vspace{-.15in}
\begin{small}
\begin{center}
\begin{tabular}{rrrrll}
Rank & $-\log p_G(w)$ & $|w|_G$ & $c_G(w)$ & w & $\mbox{rep}(w)$\\ \hline 
    0&  4.644&  & 42101.60& \verb|.| & {\em terminal}\\
    1&  4.890&  & 35507.32& \verb|,| & {\em terminal}\\
    2&  5.622& 21.293& 21381.42& \verb|[ the]| & \verb|[ [the]]| \\
    3&  5.656& 17.665& 20873.68& \verb|[ and]| & \verb|[[ an]d]| \\
    4&  5.793&  & 18992.36& \verb|s| & {\em terminal}\\
    5&  6.433& 22.885& 12186.06& \verb|[ of]| & \verb|[ [of]]| \\
    6&  6.798& 18.196& 9461.31& \verb|[ a]| & \verb|[ a]| \\
    7&  6.898& 18.566& 8826.39& \verb|[ in]| & \verb|[ [in]]| \\
    8&  6.971& 21.311& 8389.09& \verb|[ to]| & \verb|[ [to]]| \\
  100& 10.333& 23.135&  816.11& \verb|[ two]| & \verb|[ [two]]| \\
  101& 10.342& 16.093&  811.01& \verb|[ it was]| & \verb|[[ it][ was]]| \\
  102& 10.347& 21.721&  808.46& \verb|[ time]| & \verb|[ [time]]| \\
  103& 10.348& 18.786&  807.69& \verb|["?]| & \verb|["?]| \\
  104& 10.415& 22.439&  771.02& \verb|[ like]| & \verb|[ [like]]| \\
  105& 10.416& 23.505&  770.67& \verb|[ (]| & \verb|[ (]| \\
  106& 10.466& 22.218&  744.37& \verb|[ our]| & \verb|[ [our]]| \\
  107& 10.469& 23.052&  742.74& \verb|[ my]| & \verb|[ [my]]| \\
  108& 10.473& 16.954&  740.73& \verb|[ there]| & \verb|[[ the][re]]| \\
  500& 12.466& 16.283&  186.06& \verb|[    but]| & \verb|[[   ][ but]]| \\
  501& 12.467& 21.486&  185.91& \verb|[ized]| & \verb|[[ize]d]| \\
  502& 12.469& 18.645&  185.68& \verb|[ling]| & \verb|[l[ing]]| \\
  503& 12.469& 17.212&  185.67& \verb|[ like a]| & \verb|[[ like][ a]]| \\
  504& 12.470& 30.686&  185.52& \verb|[ period]| & \verb|[[ peri][od]]| \\
  505& 12.474& 25.611&  185.00& \verb|[ second]| & \verb|[ [second]]| \\
  506& 12.477& 22.997&  184.60& \verb|[ town]| & \verb|[ [town]]| \\
  507& 12.481& 19.682&  184.21& \verb|[ine]| & \verb|[[in]e]| \\
  508& 12.482& 22.068&  184.02& \verb|[ best]| & \verb|[[ be][st]]| \\
15000& 16.684& 47.086&   10.00& \verb|[ pakistan]| & \verb|[[ pa]k[ist][an]]| \\
15001& 16.684& 40.181&   10.00& \verb|[ creativity]| & \verb|[ [creat][ivity]]| \\
15002& 16.684& 45.745&   10.00& \verb|[ misleading]| & \verb|[[ mis][lea]d[ing]]| \\
15003& 16.684& 39.732&   10.00& \verb|[ criterion]| & \verb|[[ cri][ter][ion]]| \\
15004& 16.684& 39.017&   10.00& \verb|[ barbed wire]| & \verb|[[ barb][ed][ wire]]| \\
15005& 16.684& 40.711&   10.00& \verb|[ drexel]| & \verb|[[ dr][ex][el]]| \\
15006& 16.684& 38.713&   10.00& \verb|[ shrewd]| & \verb|[[ shr][ew]d]| \\
15007& 16.684& 40.047&   10.00& \verb|[ nonetheless]| & \verb|[[ none][the][less]]| \\
15008& 16.684& 40.885&   10.00& \verb|[ configuration]| & \verb|[[ con][figur][ation]]| \\
27167& 18.006& 33.412&    4.00& \multicolumn{2}{l}{\verb|[[ massachusetts][ institute of technology]]|} \\
33500& 19.006& 44.044&    2.00& \verb|[, dionys]| & \verb|[,[ di][on]ys]| \\
33501& 19.006& 44.245&    2.00& \multicolumn{2}{l}{\verb|[[ reflected][ from the][ ionosphere]]|} \\
33502& 19.006& 40.688&    2.00& \multicolumn{2}{l}{\verb|[[ the belgians][, and][ appealed to]]|} \\
33503& 19.006& 43.168&    2.00& \verb|[ ionosphere]| & \verb|[[ ion]o[sphere]]| \\
33504& 19.006& 52.399&    2.00& \verb|[ and bogus material.]| & \verb|[[ and][ bo][gus][ material].]| \\
33505& 19.006& 41.010&    2.00& \verb|[ of sant']| & \verb|[[ of][ san]t']| \\
33506& 19.006& 42.336&    2.00& \verb|[ paprika]| & \verb|[[ pa][pri][ka]]| \\
33507& 19.006& 57.078&    2.00& \multicolumn{2}{l}{\verb|[[ north][ atlantic][ treaty][ organization]]|} \\
33508& 19.006& 110.659&    2.00& \multicolumn{2}{l}{\verb|[[ to the][ person or persons][ found][ by the][com|}\\
&&&                            & \multicolumn{2}{l}{\verb|ptroller general of the united states][ to be][ ent|}\\
&&&& \multicolumn{2}{l}{\verb|titled][ thereto]]|}
\end{tabular}
\end{center}
\end{small}
\vspace{-.1in}
\caption{\label{fig:dict}Some words from the lexicon with
  their representations, ranked by probability.}
\pageline
\end{figure}

Figure~\ref{fig:dict} presents some selections from the final lexicon.
Words are ranked by their probability, and listed along with the length of
their codeword $-\log p_G(w)$, the length of their description $|w|_{G}$,
their count $c_G(w)$, and their Viterbi representations.  Notice that
lengths and counts are non-integral; this is because these are as computed
over all possible representations during the execution of the learning
algorithm, not as produced by the compression coding scheme that uses only
Viterbi representations.  The information in figure~\ref{fig:dict} makes
plain why the lexicon compresses the input.  The 15,000th parameter
(\verb*| pakistan|), for example, has a representation that is about 47
bits long.  In contrast, the length of its codeword is about 17 bits.  Thus,
each of the 10 occurrences of the word saves about 30 bits-- 300 bits in
all.\footnote{This is not exactly true.  Were the parameter not in the
  lexicon, each of its components would have higher counts, and thus
  slightly shorter codewords.} Of course, 47 bits are spent representing
the word in the lexicon, but the net savings is still around 250
bits.  More common words like \verb*| the| can save hundreds of thousands of
bits.  Notice that the algorithm seems to have adopted a uniform policy of
placing spaces at the start of words.

\subsection{Brown Corpus Language Modeling Results}

To test language modeling performance, where only the generalization rate
over new input matters, a slightly different methodology is required.  Each
of the 500 documents in the Brown corpus was split, with the first 90\%
used for training and the last 10\% reserved for testing.  The algorithm
was run on all of the training text and created a lexicon of 30,347 words.
This lexicon was then used to calculate the probability of all of the
held-out test data.  The cross-entropy rate on the test text is 2.04
bits/char (compare with 1.92 bits/char for the training text).  Running
this experiment again with slightly different conditions for creating words
produces a lexicon of 42,668 words that has slightly poorer compression
performance on the training text (2.19 bits/char vs.\ 2.12 bits/char) but a
cross-entropy rate of 1.97 bit/char on the test text.

\subsection{Calgary Corpus Compression Results}

\begin{figure}[tbhp]
\pageline
\begin{center}
\begin{tabular}{lrccccc}
Source & size (bytes) & {\tt compress} & {\tt gzip} & {\tt Sequitur} & PPM & our scheme \\ \hline
{\tt bib}&111,261     &3.35    &2.51   &2.48   &2.12   &2.33\\
{\tt book1}&768,771     &3.46   &3.25   &2.82   &2.52   &2.56\\
{\tt book2}&610,856&     3.28&   2.70&   2.46&   2.28&   2.27\\
{\tt news}&377,109&      3.86&   3.06&   2.85&   2.77&   2.78\\
{\tt paper1}&53,161&     3.77&   2.79&   2.89&   2.48&   2.73\\
{\tt paper2}&82,199&     3.52&   2.89&   2.87&   2.46&   2.63\\
{\tt progc}&39,611&      3.87&   2.68&   2.83&   2.49&   2.75\\
{\tt progl}&71,646&      3.03&   1.80&   1.95&   1.87&   1.95\\
{\tt progp}&49,379&      3.11&   1.81&   1.87&   1.82&   1.87\\
{\tt trans}&93,695&      3.27&   1.61&   1.69&   1.75&   1.73\\ \hline
\multicolumn{2}{l}{mean rate (unweighted by size)} & 3.45& 2.51&   2.47&   2.26&   2.36
\end{tabular}
\end{center}
\caption{\label{fig:calgary}Compression rates over the Calgary corpus,
  compared with four other methods: the UNIX {\tt compress} and {\tt gzip}
  programs, Nevill-Manning's {\tt Sequitur}, and a PPM-based program.}
\pageline
\end{figure}

Run separately on each of the 10 files of the Calgary corpus, the algorithm
produces compression rates that beat other dictionary-based compression
algorithms, and are competitive with the context models produced by the PPM
algorithm, especially on longer files.  Figure~\ref{fig:calgary} presents
results over the corpus, compared with the LZ78-based~\cite{Welch84,Ziv78}
{\tt compress} program, the LZ77-based~\cite{Ziv77} {\tt gzip} program,
{\tt Sequitur}~\cite{NevillManning96} and a PPM-based
program~\cite{Cleary84,Moffat90}.  The performance figures for other
programs are taken from Nevill-Manning~\cite{NevillManning96}.

\subsection{Discussion}

The algorithm compresses the Brown corpus to 2.09 bits/char.  This is the
best result we have seen reported on the Brown corpus, and is substantially
better than standard compression algorithms like {\tt gzip} achieve.  Of
course, the algorithm is substantially slower than one-pass compression
algorithms.  On the Calgary corpus of shorter texts, the algorithm beats
other dictionary algorithms, including Nevill-Manning's {\tt Sequitur},
indicating that there are substantial savings to be had by using stochastic
grammars and optimizing the internal structure of the lexicon.  For short
texts, context models such as PPM outperform our algorithm, taking
advantage of the fact that they do not introduce independence assumptions
at word boundaries.  On the other hand, one of the interesting advantages
our algorithm has over Markov-model based compression schemes like PPM is
that it represents the input in terms of linguistic structure (this will be
shown in the next section).  As a consequence, it is possible to perform
``linguistic'' operations like search, text-indexing and summarization
directly on compressed documents.

The algorithm achieves a cross-entropy rate of 1.97 bits/char on a portion
of the Brown corpus not used for training (though a portion fairly similar
to the training data).  This happens to be the same rate achieved by Ristad
and Thomas~\cite{Ristad95} using a context model on the same data.  The
best rate over the entire Brown corpus, achieved by Brown {\em et
  al.}~\cite{Brown92} with a trigram Markov model over words, is 1.75
bits/char.  This upper bound on the ``true'' entropy of English (or at
least of the Brown corpus) is significantly closer to the rates of 1.3 and
1.25 bits/char achieved by human subjects as tested by
Shannon~\cite{Shannon51} and Cover and King~\cite{Cover50}.\footnote{Those
  rates were over much smaller samples of text, and used a smaller
  alphabet, but it is widely believed that human subjects would best the
  1.75 bits/char figure on the Brown corpus.}  However, that result came
after training on almost 600 million words of text, starting with
substantial knowledge of language.  The resulting model would have dwarfed
the Brown corpus in size, and hence is difficult to compare with a {\em
  compression} algorithm.  Without performing the test, it is not easy to
guess what entropy rate our algorithm would achieve after training on such
a large amount of data, though it is not likely to best Brown {\em et
  al.}'s 1.75 bit figure: although the lexicons the algorithm produces
model ``lexical'' phenomena fairly well, the independence assumptions made
at parameter boundaries prevent the algorithm from modeling many
regularities that have syntactic and semantic roots.

\section{Segmentation}

The algorithm's statistical performance is pleasing, but the principal goal
of this thesis is not statistical, but linguistic.  The most important
question is how well the lexicons produced by the algorithm agree with
linguistic reality.  There are two ways this might be investigated:
directly, by looking at the lexicons, or indirectly, through the
derivations the algorithm produces when it analyzes text.  Here the second
possibility is chosen.  One challenge is to find a gold standard to compare
against.  For want of a better substitute, the hierarchical structures
produced by the algorithm are judged against segmentations of text as
defined by spaces in the case of English input, and the output of another
computer program (that has access to a lexicon) in the case of Chinese
output.

\subsection{Input}

\begin{figure}[tbp]
\pageline
\begin{small}
\vspace{-.05in}\mbox{\tt thefultoncountygrandjurysaidfridayaninvestigationofatlantasrecentprimaryelectionproducednoevide}\\
\mbox{\tt ncethatanyirregularitiestookplace}\\
\vspace{-.05in}\mbox{\tt thejuryfurthersaidintermendpresentmentsthatthecityexecutivecommitteewhichhadoverallchargeofthee}\\
\mbox{\tt lectiondeservesthepraiseandthanksofthecityofatlantaforthemannerinwhichtheelectionwasconducted}\\
\vspace{-.05in}\mbox{[{\tt the}][{\tt fulton}][{\tt county}][{\tt grand}][{\tt jury}][{\tt said}][{\tt friday}][{\tt an}][{\tt investigation}][{\tt of}][{\tt atlantas}][{\tt recent}][{\tt primary}][{\tt election}]}\\
\mbox{[{\tt produced}][{\tt no}][{\tt evidence}][{\tt that}][{\tt any}][{\tt irregularities}][{\tt took}][{\tt place}]}\\
\vspace{-.05in}\mbox{[{\tt the}][{\tt jury}][{\tt further}][{\tt said}][{\tt in}][{\tt termend}][{\tt presentments}][{\tt that}][{\tt the}][{\tt city}][{\tt executive}][{\tt committee}][{\tt which}][{\tt had}][{\tt ove}}\\
\vspace{-.05in}\mbox{{\tt rall}][{\tt charge}][{\tt of}][{\tt the}][{\tt election}][{\tt deserves}][{\tt the}][{\tt praise}][{\tt and}][{\tt thanks}][{\tt of}][{\tt the}][{\tt city}][{\tt of}][{\tt atlanta}][{\tt for}][{\tt the}][{\tt mann}}\\
\mbox{{\tt er}][{\tt in}][{\tt which}][{\tt the}][{\tt election}][{\tt was}][{\tt conducted}]}
\end{small}

\vspace{.15in}
\begin{center}
\mbox{\,}\psfig{figure=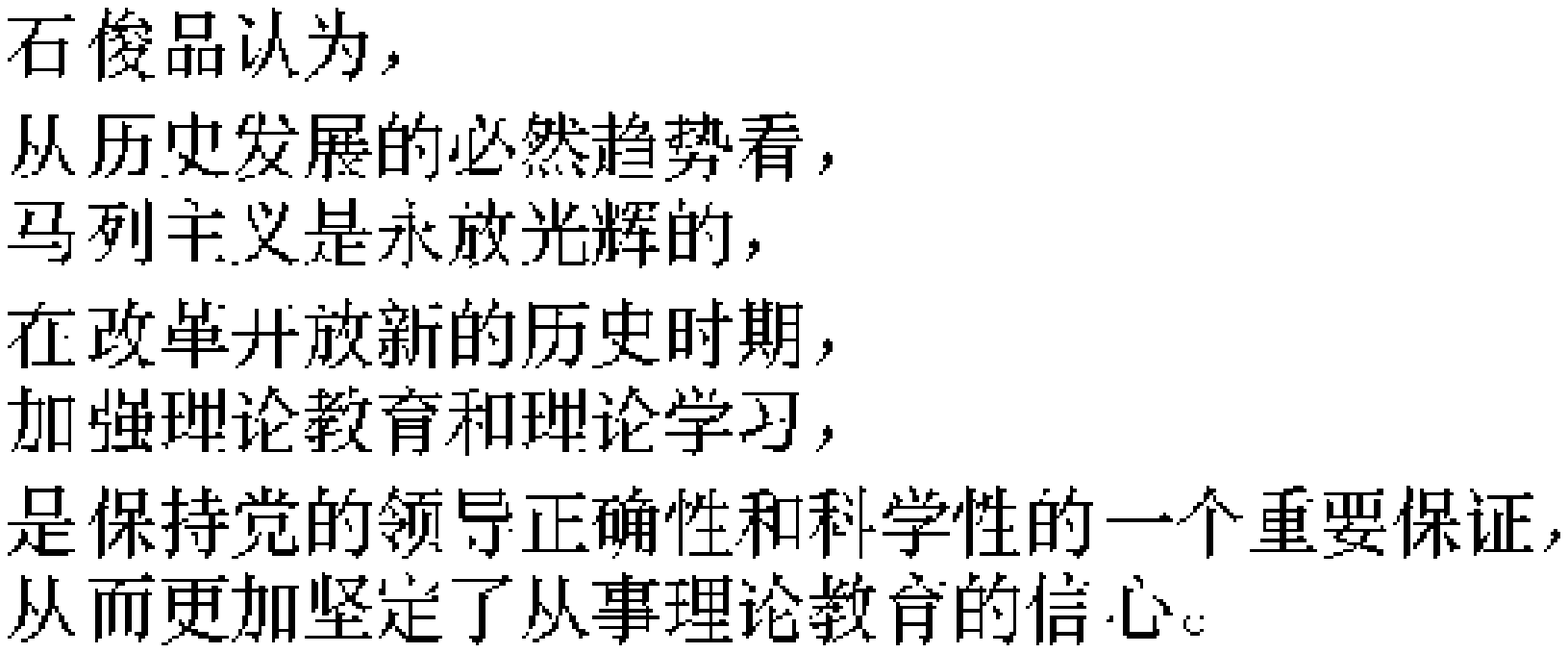,height=1in,width=2.2in}
\hfill
\mbox{\,}\psfig{figure=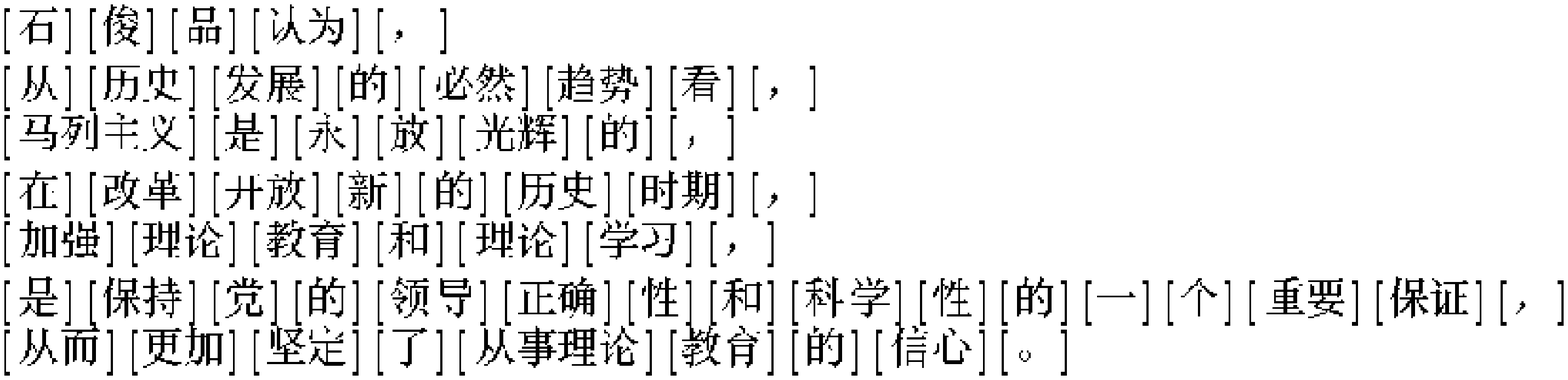,height=1in,width=3.8in}
\end{center}
\vspace{-.2in}
\caption{\label{fig:brown2}At top, the first two sentences of the Brown corpus as
  modified for segmentation tests, first as the algorithm sees them, and
  then with the bracketings that define true word boundaries.  Below,
  seven of the ``sentences'' (phrases) from the Xinhua corpus of Chinese
  news articles.  At left is the input the algorithm gets (each character
  is actually presented as a two-byte code) and at right is the true
  segmentation as defined by a segmentation program that had access to a
  human-made dictionary.}  \pageline
\end{figure}

Two different corpora are used for segmentation tests.  The Brown corpus is
used again, segmented into sentences as before and shifted to lower case,
but with spaces and punctuation removed (only alphanumeric characters are
retained).  The original locations of spaces are stored for segmentation
tests: spaces (along with sentence boundaries) are used to bracket the
sentence into words.  Figure~\ref{fig:brown2} presents the first two
sentences of the corpus as the algorithm sees them.  The second corpus is 4
million characters of Chinese text, a collection of news articles from
China's official Xinhua news agency dating from 1990 and
1991.\footnote{Selected and processed by Guo Jin {\em et al.} at the
  National University of Singapore, as made available by the Linguistics
  Data Consortium.}  The text is divided into phrases at punctuation marks
and has an alphabet of 4725 characters.  These characters are coded as
two-byte sequences.  The algorithm is provided the raw byte stream (a
256-character alphabet) and is not in any way specially modified for the
two-byte format.  In practice, the algorithm groups bytes into characters
before it builds bigger units.  For testing purposes, the characters have
been segmented into words (Chinese words generally range in length from one
to three characters) using a segmentation algorithm that has access to a
50,000 word dictionary but has no mechanisms for dealing with unknown words
and names.  As a consequence, the segmentation that is tested against is
good, but not ideal.  Figure~\ref{fig:brown2} presents seven sample
``sentences'' (phrases) from the corpus, along with their true
segmentations.

\subsection{Method}\label{segex}

The algorithm is applied to each corpus, producing a lexicon.  This lexicon
is used to produce representations of the input.  For example, the
following is the Viterbi representation of a typical sentence from the
Brown corpus, followed by the true segmentation as defined by where spaces
occur:

\begin{center}
\begin{tabular}{l}
\mbox{[{\tt forthepurposeof}][{\tt maintaining}][{\tt international}][{\tt peace}][{\tt and}][{\tt promoting}][{\tt the}][{\tt advancement}]}\\
\mbox{[{\tt ofall}][{\tt people}][{\tt theunitedstatesofamerica}][{\tt joined}][{\tt in}][{\tt found}][{\tt ing}][{\tt theunitednations}]}\\
\\
\mbox{[{\tt for}][{\tt the}][{\tt purpose}][{\tt of}][{\tt maintaining}][{\tt international}][{\tt peace}][{\tt and}][{\tt promoting}][{\tt the}][{\tt advancement}]}\\
\mbox{[{\tt of}][{\tt all}][{\tt people}][{\tt the}][{\tt united}][{\tt
    states}][{\tt of}][{\tt america}][{\tt joined}][{\tt in}][{\tt
    founding}][{\tt the}][{\tt united}][{\tt nations}]}
\end{tabular}
\end{center}

Since the lexicon is represented as a hierarchy, each of the words in the
algorithm's representation also has a Viterbi representation.  Expanding
this hierarchy down to terminals gives:

\begin{center}
\begin{tabular}{l}
\mbox{[[{\tt f}[{\tt or}]][[{\tt t}[{\tt he}]][[[{\tt p}[{\tt ur}]][[[{\tt po}]{\tt s}]{\tt e}]][{\tt of}]]]][[[{\tt ma}[{\tt in}]][{\tt ta}[{\tt in}]]][[{\tt in}]{\tt g}]][[[{\tt in}][{\tt t}[{\tt er}]]][[{\tt n}[{\tt a}[{\tt t}[{\tt i}[{\tt on}]]]]][{\tt al}]]]}\\
\mbox{[[{\tt pe}][{\tt a}[{\tt ce}]]][[{\tt an}]{\tt d}][[{\tt p}[{\tt ro}]][[{\tt mo}]{\tt t}][[{\tt in}]{\tt g}]][{\tt t}[{\tt he}]][[{\tt adv}[{\tt a}[{\tt n}[{\tt ce}]]]][[[{\tt me}]{\tt n}]{\tt t}]][[{\tt of}][{\tt a}[{\tt ll}]]][[{\tt pe}][{\tt op}][{\tt le}]]}\\
\mbox{[[[{\tt t}[{\tt he}]][[[[{\tt un}][{\tt it}]][{\tt ed}]][[[{\tt st}[{\tt at}]]{\tt e}]{\tt s}]]][[{\tt of}][{\tt a}[{\tt me}][{\tt r}[{\tt ic}]]{\tt a}]]][[[{\tt jo}][{\tt in}]][{\tt ed}]][{\tt in}][{\tt f}[{\tt o}[{\tt un}]{\tt d}]][[{\tt in}]{\tt g}]}\\
\mbox{[[{\tt t}[{\tt he}]][[[[{\tt un}][{\tt it}]][{\tt ed}]][[{\tt n}[{\tt a}[{\tt t}[{\tt i}[{\tt on}]]]]]{\tt s}]]]}
\end{tabular}
\end{center}

(The brackets around terminals are not printed.)  An easier format to read
is given below; horizontal bars are used in place of bracket pairs.  Notice
the linguistically natural structure assigned to the sentence.

\[
\overline{\mbox{\tt $\,\overline{\mbox{\tt f$\,\overline{\mbox{\tt or}}\,$}}\,\overline{\mbox{\tt $\,\overline{\mbox{\tt t$\,\overline{\mbox{\tt he}}\,$}}\,\overline{\mbox{\tt $\,\overline{\mbox{\tt $\,\overline{\mbox{\tt p$\,\overline{\mbox{\tt ur}}\,$}}\,\overline{\mbox{\tt $\,\overline{\mbox{\tt $\,\overline{\mbox{\tt po}}\,$s}}\,$e}}\,$}}\,\overline{\mbox{\tt of}}\,$}}\,$}}\,$}}\,\overline{\mbox{\tt $\,\overline{\mbox{\tt $\,\overline{\mbox{\tt ma$\,\overline{\mbox{\tt in}}\,$}}\,\overline{\mbox{\tt ta$\,\overline{\mbox{\tt in}}\,$}}\,$}}\,\overline{\mbox{\tt $\,\overline{\mbox{\tt in}}\,$g}}\,$}}\,\overline{\mbox{\tt $\,\overline{\mbox{\tt $\,\overline{\mbox{\tt in}}\,\overline{\mbox{\tt t$\,\overline{\mbox{\tt er}}\,$}}\,$}}\,\overline{\mbox{\tt $\,\overline{\mbox{\tt n$\,\overline{\mbox{\tt a$\,\overline{\mbox{\tt t$\,\overline{\mbox{\tt i$\,\overline{\mbox{\tt on}}\,$}}\,$}}\,$}}\,$}}\,\overline{\mbox{\tt al}}\,$}}\,$}}\,\overline{\mbox{\tt $\,\overline{\mbox{\tt pe}}\,\overline{\mbox{\tt a$\,\overline{\mbox{\tt ce}}\,$}}\,$}}\,\overline{\mbox{\tt $\,\overline{\mbox{\tt an}}\,$d}}\,\overline{\mbox{\tt $\,\overline{\mbox{\tt p$\,\overline{\mbox{\tt ro}}\,$}}\,\overline{\mbox{\tt $\,\overline{\mbox{\tt mo}}\,$t}}\,\overline{\mbox{\tt $\,\overline{\mbox{\tt in}}\,$g}}\,$}}
\]
\[
\,\overline{\mbox{\tt t$\,\overline{\mbox{\tt he}}\,$}}\,\overline{\mbox{\tt $\,\overline{\mbox{\tt adv$\,\overline{\mbox{\tt a$\,\overline{\mbox{\tt n$\,\overline{\mbox{\tt ce}}\,$}}\,$}}\,$}}\,\overline{\mbox{\tt $\,\overline{\mbox{\tt $\,\overline{\mbox{\tt me}}\,$n}}\,$t}}\,$}}\,\overline{\mbox{\tt $\,\overline{\mbox{\tt of}}\,\overline{\mbox{\tt a$\,\overline{\mbox{\tt ll}}\,$}}\,$}}\,\overline{\mbox{\tt $\,\overline{\mbox{\tt pe}}\,\overline{\mbox{\tt op}}\,\overline{\mbox{\tt le}}\,$}}\,\overline{\mbox{\tt $\,\overline{\mbox{\tt $\,\overline{\mbox{\tt t$\,\overline{\mbox{\tt he}}\,$}}\,\overline{\mbox{\tt $\,\overline{\mbox{\tt $\,\overline{\mbox{\tt $\,\overline{\mbox{\tt un}}\,\overline{\mbox{\tt it}}\,$}}\,\overline{\mbox{\tt ed}}\,$}}\,\overline{\mbox{\tt $\,\overline{\mbox{\tt $\,\overline{\mbox{\tt st$\,\overline{\mbox{\tt at}}\,$}}\,$e}}\,$s}}\,$}}\,$}}\,\overline{\mbox{\tt $\,\overline{\mbox{\tt of}}\,\overline{\mbox{\tt a$\,\overline{\mbox{\tt me}}\,\overline{\mbox{\tt r$\,\overline{\mbox{\tt ic}}\,$}}\,$a}}\,$}}\,$}}\,\overline{\mbox{\tt $\,\overline{\mbox{\tt $\,\overline{\mbox{\tt jo}}\,\overline{\mbox{\tt in}}\,$}}\,\overline{\mbox{\tt ed}}\,$}}\,\overline{\mbox{\tt in}}
\]
\[
\overline{\mbox{\tt f$\,\overline{\mbox{\tt o$\,\overline{\mbox{\tt un}}\,$d}}\,$}}\,\overline{\mbox{\tt $\,\overline{\mbox{\tt in}}\,$g}}\,\overline{\mbox{\tt $\,\overline{\mbox{\tt t$\,\overline{\mbox{\tt he}}\,$}}\,\overline{\mbox{\tt $\,\overline{\mbox{\tt $\,\overline{\mbox{\tt $\,\overline{\mbox{\tt un}}\,\overline{\mbox{\tt it}}\,$}}\,\overline{\mbox{\tt ed}}\,$}}\,\overline{\mbox{\tt $\,\overline{\mbox{\tt n$\,\overline{\mbox{\tt a$\,\overline{\mbox{\tt t$\,\overline{\mbox{\tt i$\,\overline{\mbox{\tt on}}\,$}}\,$}}\,$}}\,$}}\,$s}}\,$}}\,$}}
\]

To judge the algorithm's performance, these hierarchies are compared to the
true segmentations.  Two measures are used, {\em recall} and {\em
  crossing-brackets}.  To define these, it is helpful to think of a
bracketing of a sentence $u = u_1\ldots u_l$ as a set of pairs $B(u) = \{
\langle i,j\rangle \}$ where $\langle i,j\rangle \in B(u)$ if a pair of
brackets exactly surrounds the subsequence $u_{i+1}\ldots u_{j}$.  Thus, the
bracketing [{\tt f}[{\tt o}[{\tt un}]{\tt d}]] has the bracket set

\[ \{ \langle 0,1 \rangle \langle 1,2 \rangle \langle 2,3 \rangle \langle
3,4 \rangle \langle 4,5 \rangle \langle 2,4 \rangle \langle 1,5 \rangle
\langle 0,5 \rangle \} \]

Then if $B_T(u)$ is the true segmentation of $u$, and $B_L(u)$ is the
bracketing of $u$ produced by the lexicon, the recall rate is defined

\[ \mbox{\em recall} = \frac{\sum_{u\in U} |B_T(u) \cap B_L(u)|}{\sum_{u\in U}
  |B_T(u)|}. \]

\noindent The recall rate is the proportion of the subsequences bracketed
in the true segmentation that are also bracketed at some level of the
algorithm's hierarchical representation of the input.  If the recall rate
is high, then it means the algorithm has learned most of the words in the
input, and that it properly parses the input into these words.  In the
example sentence, there is one recall error (the word {\tt founding} occurs
in the true segmentation but is not spanned by any parameter at any level
of the algorithm's hierarchy) for a recall rate of $\frac{25-1}{25}= 96\%$.

The crossing-brackets rate is the proportion of the subsequences bracketed
in the true segmentation that are crossed by some bracketed subsequence in
the algorithm's hierarchical representation.  It is defined by

\[ \mbox{\em crossing-bracket} = \frac{\sum_{u\in U} | \{ \langle i,j\rangle \in B_T(u)
  \ni \exists \langle k,l\rangle \in B_L(u) \ni k<i\wedge i<l<j\vee
  i<k<j \wedge l<j \} |}{\sum_{u\in U} |B_T(u)|}. \]

There are no crossing-brackets violations in the example sentence, so the
crossing-brackets rate is 0\%.  If the true segmentation had included a
bracket pair around {\tt unite} in {\tt united} there would be an error,
because the algorithm represents {\tt united} as {\tt unit} $\circ$ {\tt
  ed}, and the {\tt ed} crosses {\tt unite}.  If the crossing-brackets rate
is high, it means that the algorithm is making significant errors: it is
parsing the input in a way that is in conflict with the true segmentation.
The algorithm can trivially achieve a 0\% crossing-brackets rate by
representing each sentence as a sequence of terminals (imposing no
linguistic structure), but then the recall rate will be low.  A combination
of high recall rate and low crossing-brackets rate is the ideal situation.

In the sentence presented above, the algorithm's bracketing is constructed
by recursively expanding Viterbi representations.  Of course, the Viterbi
representation is only one of many possible representations for the input
and the parameters.  It would be possible (and perhaps desirable) to
compute recall and crossing-brackets as expected values over all possible
representations.  However, as the Viterbi representation tends to dominate
the total probability, for these tests it will be the only representation
considered.

\subsection{Segmentation Results}

The algorithm was run on the Brown corpus, producing a lexicon of 26,026
words (compression rate of 2.33 bits/char); some selections are presented
in figure~\ref{fig:dict2}.  Testing this lexicon on the input, the recall
rate is 90.5\% and the crossing-brackets rate is 1.7\%.  Run on the Chinese
corpus, the lexicon contains 57,885 words; the recall rate is 96.9\% and
the crossing-brackets rate is 1.3\%.  In both cases, almost all of the
recall errors are words that occur only once in the input, or several times
but always as part of the same larger phrase.  One of the reasons that
recall is higher on the Chinese corpus is that Chinese has fewer affixes
(like English's \word{-s}, \word{-ed} and \word{-ing}) that tend to
increase the size of the {\em hapax legomena} (the set of words that only
occur once).

Several examples of words that cause recall errors can be found in
figure~\ref{fig:dict2}.  For example, {\tt feasibility} is not bracketed in
{\tt feasibilityof}, {\tt diffusing} is not bracketed in {\tt
  primarilydiffusing}, and {\tt broiled} is not bracketed in {\tt
  charcoalbroiled} (parameters 26,002, 26,005 and 26,006).  Parameter
25,920 is included to provide an example of a crossing-brackets violation:
{\tt infiltratedwithneutrophils} is represented as {\tt infiltra} $\circ$
{\tt tedwith} $\circ$ {\tt neutrophils}.  Because {\tt infiltrated} is not
bracketed, the parameter causes two recall errors (one for each time it is
used) and because {\tt tedwith} crosses the true word {\tt infiltrated} the
parameter causes two crossing-brackets errors (one for each time it is
used).

\begin{figure}[p]
  \pageline \vspace{-.15in}
\begin{small}
\begin{center}
\begin{tabular}{rrrrll}
Rank & $-\log p_G(w)$ & $|w|_G$ & $c_G(w)$ & w & $\mbox{rep}(w)$\\ \hline 
    0&  4.589&  & 39820.24& {\tt s}& {\em terminal}\\
    1&  5.147& 16.661& 27042.71& {\tt [the]}&   {\tt [t[he]]}\\
    2&  5.155& 16.721& 26886.31& {\tt [and]}&   {\tt [[an]d]}\\
    3&  5.427&  & 22273.75& {\tt a}& {\em terminal}\\
    4&  6.171& 19.306& 13301.39& {\tt [of]}&   {\tt [of]}\\
    5&  6.180& 17.854& 13216.57& {\tt [in]}&   {\tt [in]}\\
    6&  6.593& 18.698& 9924.97& {\tt [to]}&   {\tt [to]}\\
    7&  7.079& 19.547& 7088.43& {\tt [that]}&   {\tt [[th][at]]}\\
    8&  7.322& 12.805& 5988.71& {\tt [is]}&   {\tt [is]}\\
  100& 10.123& 24.078&  859.41& {\tt [two]}&   {\tt [t[wo]]}\\
  101& 10.160& 22.040&  837.29& {\tt [even]}&   {\tt [e[ven]]}\\
  102& 10.161&  &  836.93& {\tt g}& {\em terminal}\\
  103& 10.222& 18.903&  802.52& {\tt [men]}&   {\tt [[me]n]}\\
  104& 10.277& 18.196&  772.19& {\tt [your]}&   {\tt [[you]r]}\\
  105& 10.280& 12.830&  770.85& {\tt [she]}&   {\tt [s[he]]}\\
  106& 10.282& 25.761&  769.83& {\tt [work]}&   {\tt [[wor]k]}\\
  107& 10.292& 15.832&  764.56& {\tt [hewas]}&   {\tt [[he][was]]}\\
  108& 10.295& 25.078&  762.46& {\tt [after]}&   {\tt [[aft][er]]}\\
 1000& 13.043& 23.794&  113.57& {\tt [drive]}&   {\tt [[dr][ive]]}\\
 1001& 13.043& 24.480&  113.56& {\tt [didnt]}&   {\tt [[did][nt]]}\\
 1002& 13.045& 27.501&  113.39& {\tt [performance]}&   {\tt [[perform][ance]]}\\
 1003& 13.046& 15.442&  113.33& {\tt [afterthe]}&   {\tt [[after][the]]}\\
 1004& 13.047& 23.689&  113.26& {\tt [mission]}&   {\tt [[miss][ion]]}\\
 1005& 13.047& 21.170&  113.25& {\tt [11]}&   {\tt [11]}\\
 1006& 13.048& 27.852&  113.17& {\tt [project]}&   {\tt [[pro][ject]]}\\
 1007& 13.048& 22.046&  113.15& {\tt [lie]}&   {\tt [l[ie]]}\\
 1008& 13.049& 16.026&  113.06& {\tt [outofthe]}&   {\tt [[outof][the]]}\\
10000& 16.063& 27.062&   13.99& {\tt [transmission]}&   {\tt [[trans][mission]]}\\
10001& 16.063& 27.063&   13.99& {\tt [corruption]}&   {\tt [[corrupt][ion]]}\\
10002& 16.063& 29.858&   13.99& {\tt [forthebenefitof]}&   {\tt [[forthe][benefit][of]]}\\
10003& 16.063& 19.948&   13.99& {\tt [stillhad]}&   {\tt [[still][had]]}\\
10004& 16.064& 24.526&   13.99& {\tt [tak]}&   {\tt [tak]}\\
10005& 16.064& 27.996&   13.99& {\tt [conservation]}&   {\tt [[conserv][ation]]}\\
10006& 16.064& 27.246&   13.99& {\tt [sermon]}&   {\tt [s[er][mon]]}\\
10007& 16.064& 22.338&   13.99& {\tt [ourcountry]}&   {\tt [[our][country]]}\\
10008& 16.064& 27.719&   13.99& {\tt [irrelevant]}&   {\tt [[ir][relevant]]}\\
22202& 17.870& 32.569&    4.00& \multicolumn{2}{l}{\tt [[massachusetts][instituteoftechnology]]}\\
25920& 18.870& 52.706&    2.00& \multicolumn{2}{l}{\tt [[infiltra][tedwith][neutrophils]]}\\
26000& 18.870& 43.904&    2.00& {\tt [pleuralbloodsupply]}&   {\tt [[pleural][blood][supply]]}\\
26001& 18.870& 41.349&    2.00& {\tt [anordinaryhappyfamily]}&   {\tt [[anordinary][happy][family]]}\\
26002& 18.870& 45.269&    2.00& {\tt [feasibilityof]}&   {\tt [f[eas][ibility][of]]}\\
26003& 18.870& 46.646&    2.00& \multicolumn{2}{l}{\tt [[lunar][brightness][distribution]]}\\
26004& 18.870& 43.008&    2.00& {\tt [primarilydiffusing]}&   {\tt [[primarily][diff][using]]}\\
26005& 18.870& 47.115&    2.00& {\tt [sodiumtripolyphosphate]}&   {\tt [[sodium][tri][polyphosphate]]}\\
26006& 18.870& 41.054&    2.00& {\tt [charcoalbroiled]}&   {\tt [[charcoal][broil][ed]]}\\
26007& 18.870& 41.171&    2.00& \multicolumn{2}{l}{\tt [[over][considerable][periodsoftime]]}\\
26008& 18.870& 42.300&    2.00& {\tt [per1000peryear]}&   {\tt [[per][1000][peryear]]}
\end{tabular}
\end{center}
\end{small}
\vspace{-.1in}
\caption{\label{fig:dict2}Some words from the lexicon with
  their representations, ranked by probability.}
\pageline
\end{figure}

\subsection{Discussion}

These results are very pleasing.  The algorithm discovers words in
unsegmented input and very reliably parses sentences into proper linguistic
structure (word recall rates are 90.5\% and 96.9.\%).  It would be
difficult to better these rates with any algorithm that does not include
words in the lexicon based on single occurrences (nothing precludes this
possibility).  At the same time, only rarely does the algorithm produce
analyses that are in conflict with what is known about the true linguistic
structure (word crossing-brackets rates are 1.7\% and 1.3\%).

Of course, the algorithm is producing far more structure than is tested by
checking word boundary conflicts.  Therefore the word {\em accuracy}
rate\footnote{The accuracy rate is defined by

\[ \mbox{\em accuracy} = \frac{\sum_{u\in U} |B_T(u) \cap B_L(u)|}{\sum_{u\in U}
  |B_L(u)|}. \]} (the proportion of bracket pairs produced by the algorithm
that are words as defined by the true bracketing) is generally
substantially lower than for algorithms that produce a single level of
structure.  One of the deficiencies of the segmentation tests is that they
look at only one facet of linguistic structure, namely that defined by
space placement in English and a standard dictionary in Chinese.  The
algorithm is given no credit for discovering units smaller than words (such
as {\em found} in {\em founding}, from the example in section~\ref{segex}),
or bigger than words (such as {\em unitedstatesofamerica} or {\em
  nationalfootballleague}).

\subsubsection{Comparisons with Other Results}

It is difficult to compare these results against others, because few
segmentation rates have been published for English, and most Chinese
segmentation algorithms start with dictionaries.  Furthermore, direct
comparison is impossible given that our algorithm produces hierarchical
segmentations whereas most other algorithms produce only a single level of
structure.

Olivier~\cite{Olivier68} presents an on-line word-learning algorithm and
applies it to 288,000 characters of unsegmented (spaceless), lower-case
English text taken from the nomination speeches of major-party presidential
nominees between 1928 (Al Smith) and 1960 (Richard Nixon).  The algorithm
achieves a peak word recall rate of about 80\%.  This result is the most
directly comparable to our Brown corpus tests.  The poorer recall rate
reflects the lengthy parameters learned to model regularities above the
word level.  Cartwright and Brent~\cite{Cartwright94}, testing several
word-learning algorithms on a very small (4000 phoneme) corpus of
phonemified English text,\footnote{Transcriptions of mothers' speech to
  children taken from the CHILDES database~\cite{MacWhinney85} and
  converted to phonemes in a manner that ensures each word is given a
  consistent transcription.  Spaces between words in the original text are
  used to define word boundaries in the phoneme sequences, but are removed
  in the evidence presented to the algorithm.} report a peak recall rate of
95.6\%, but this {\em drops} dramatically if the algorithm is given more
evidence, as the algorithm adds extralinguistic patterns to the lexicon.
They report in Brent and Cartwright~\cite{Brent96} substantially lower
recall rates (40\%-70\%) for similar algorithms tested on slightly
different data.  In contrast, our algorithm achieves a recall rate of only
65.5\% on the small sample used in their first tests (because it doesn't
learn words that only appear once) but this rate climbs to 96.5\% on a much
longer corpus of 34,438 utterances of motherese from the CHILDES database
transcribed in the same manner.  It seems therefore that our algorithm
performs substantially better.  Wolff~\cite{Wolff77,Wolff80} presents a
word-learning algorithm and applies it to English and pseudo-English text,
but does not provide results in a manner suitable for comparison; however,
experiments performed by Nevill-Manning~\cite{NevillManning96} indicate
that Wolff's algorithms are not competitive.  Finally,
Nevill-Manning~\cite{NevillManning96} applies his {\tt Sequitur} algorithm
to English text (with spaces) but reports results in a manner incomparable
with those presented here.  From the sample hierarchical structures he
provides it appears that his algorithm performs well, but has a lower
recall rate and a substantially higher rate of crossing-brackets.

There is a larger body of literature on the segmentation of Chinese (and
Japanese and other orthographically unsegmented languages).  Most of these
algorithms attack a slightly differently problem, starting with a lexicon
defined by hand-segmentations of text or man-made dictionaries.  However,
it is interesting to compare results.  Sproat {\em et al.}~\cite{Sproat94}
make the point (see also Luo and Roukos~\cite{Luo96}) that it is difficult
define ``true'' segmentations in Chinese: when people are asked to segment
sentences into words, their segmentations very often disagree.  This
reflects the fact that there are many levels of linguistic structure in a
sentence, and it can be difficult to define what a ``word'' is.  Many
segmentation algorithms (\cite{Luo96,Sproat94} and others) agree with human
segmenters at approximately the same rate as human segmenters agree with
one another (recall rates between 60\% and 90\%).  Our algorithm, in
contrast, has a recall rate of 96.9\%, substantially higher than any other
algorithm achieves (or could possibly achieve), because it produces
structure at multiple linguistic levels.
 
\subsubsection{Recall vs.\ Accuracy}

It is reasonable to ask whether our algorithm is in some sense cheating by
producing a hierarchical structure for each sentence.  Does this not make
it easy to achieve high recall rates?  Is there any information content to
the structure?

First of all, notice that the algorithm can not produce a bracketing that
brackets every subsequence of a sentence.  This is because the algorithm
outputs a tree-- its own brackets can not cross.  For a sentence of length
$n$, there can be at most $2n$ bracket pairs in the representation the
algorithm produces.  Yet there are $n(n-1)$ possible subsequences that
could be bracketed in the true segmentation.  Therefore the algorithm makes
a significant commitment in producing a representation: it is not in any
way the case that the algorithm can trivially raise the recall rate to 100\%
by reducing the accuracy rate.

Second, although there are applications where multilevel representations
are inappropriate (spell-checking, for example), there are many
applications that benefit substantially from them.  The most obvious case
is language acquisition, the central topic of this thesis.
Section~\ref{res:mean} will demonstrate how the compositional
representation aids the acquisition of word meanings.  Other examples
include document-indexing and retrieval.  Standard approaches to these
problems involve treating documents as a collection of features, where each
feature is a word that appears in the document.  It is well known that
words are not the ideal level of representation for this problem.  Often
performance is improved by removing affixes (converting \word{cars} to
\word{car}, for instance).  At the same time, performance can be improved
by combining words into bigger units (\word{national football league}).
The compositional framework offers the possibility that all parameters that
occur in the hierarchical representation of a document be treated as
features, whether they be above, at, or below the word level.  This adds
slightly to the number of features considered by the retrieval and indexing
algorithms, but such algorithms tend to be quite robust to the introduction
of superfluous features.

\section{Learning from Raw Speech}\label{results:speech}

Section~\ref{tla} argues that theories of language acquisition should
involve as few unjustified assumptions as possible, and be tested on input
similar to that children receive.  So far, however, all the experiments
that have been described treat the learning problem as one of learning from
text, not speech.  Learning from speech can be significantly more
challenging:

\begin{itemize}
\item Speech is continuous, rather than discrete.  Discretizing speech
  involves making choices, which introduces either errors or ambiguity.
\item In text, characters are given consistent representations.  In speech,
  sound units like phonemes are pronounced differently each time they are
  spoken, in a manner that is dependent on everything from speaker sex and
  age to blood-alcohol content.
\item In text, words generally receive consistent spellings.  In contrast,
  sounds in spoken words are dropped, added and changed in ways that depend
  on context and speech speed.
\end{itemize}

To demonstrate that our algorithm can also learn words from speech, it is
applied to a large, multi-speaker collection of continuous utterances.  A
two-part process is used.  First, the speech is transcribed (automatically)
into a phoneme sequence, and then the algorithm is applied as is to the
result.

\subsection{Input}

Two sources of speech are used, the Texas Instrument-MIT (TIMIT) database
(for training acoustic models) and the WSJ1 database (for testing).  These
are both large collections of digitized speech distributed by the
Linguistics Data Consortium.

The TIMIT collection is designed to be used for training speech
recognizers.  It consists of 6,300 utterances, each one a sentence read
aloud by one of 630 speakers of either sex from around the United States.
Each speaker reads 10 sentences.  Two are fixed ``calibration'' sentences,
five are ``phonetically compact'' sentences drawn from a set of 450
sentences designed for phonetic coverage, and three are ``phonetically
diverse'' sentences drawn from a set of 1,890 designed to add variety to
the collection.  We do not use the calibration sentences, leaving 5,040.
Of these, 3,696 are used for training and the remainder set aside.  Each
utterance in the TIMIT collection has been transcribed into a phoneme
sequence by phoneticians, with phoneme boundaries labeled in the acoustic
stream.

The WSJ1 collection is a large database of speech designed for experiments
and tests of continuous speech recognition systems.  It consists of 78,000
utterances totaling almost 73 hours of speech.  Of this, we use
approximately 68,000 utterances in our tests.  Each is a dictated sentence
from a Wall Street Journal article: 200 non-journalists read 150 sentences
each, another 25 read 1200 sentences each, and 20 journalists read 200
sentences and spontaneously composed 200 more.

\subsection{Method}

The HTK HMM toolkit developed by Young and Woodland was used to build a
triphone-based phoneme transcriber.  This is essentially an automatic
speech recognition device that outputs a sequence of phonemes rather than
the more traditional word sequence.\footnote{See Rabiner and
  Huang~\cite{Rabiner93} for an introduction to the methods of automatic
  speech recognition.  In a triphone-based speech recognizer, speech
  production is modeled as a three-stage process.  First a phoneme sequence
  is generated.  In our model phonemes are generated independently under a
  uniform distribution.  Each phoneme in the resulting sequence is further
  specialized by incorporating information about its two neighbors, forming
  a triphone (a context-dependent phoneme).  In our model neighbors are
  divided into eight classes (vowel, fricative, etc.), so one triphone is
  {\em vowel}-\uniJ-{\em vowel}.  In the second stage of production, each
  triphone independently generates a sequence of acoustic vectors.  In our
  case this process is modeled by a three-state HMM chain with a looping
  (variable-length) middle state.  Each state emits an acoustic vector
  under a mixture-of-gaussians distribution, where each acoustic-vector is
  an LPC-coded 40-vector consisting of an energy and 13 mel-frequency
  cepstral coefficients and their first and second time differences.  The
  final (deterministic) stage of speech production maps the resulting
  vector sequence to speech.  In this model of speech production there are
  free parameters in each of the many triphone acoustic models (the HMMs).
  These are estimated from speech using the Baum-Welch algorithm.}  The
transcriber does {\em not} incorporate a prior model of phoneme sequences,
as a normal speech recognition device would.  This is because the process
of learning a stochastic lexicon and grammar is that of learning a model of
phoneme sequences; incorporating a prior model into the phoneme recognition
device would defeat the purpose of the language acquisition experiment.

The transcriber uses a set of 48 phonemes.  The parameters in the acoustic
models for each triphone are trained on 3,696 utterances from the TIMIT
database, each of which has been pre-labeled with phoneme boundaries so
that supervised learning methods can be used.  Tests of the transcriber on
the TIMIT test data put phoneme recall at 55.5\% and phoneme accuracy at
68.7\%.  These numbers were computed by comparing the Viterbi analyses of
utterances against phoneticians' transcriptions.  It should be clear from
this performance level that the input to our algorithm will be very, very
noisy.  This is the consequence of not using a prior model over phonemes.
Ordinary speech recognition systems achieve substantially better rates by
building a prior model from hand-constructed dictionaries and word-sequence
models trained on text.  Some sentences with their ``true'' transcriptions
(as produced by phoneticians) and the output of the automatic transcriber
are presented in figure~\ref{fig:timit}.

\begin{figure}[tb]
\pageline
\def\ipa{\ipatenrm}
\begin{center}
\begin{tabular}{l}
\vspace{-.00in}Bricks are an alternative.\\
\vspace{-.02in}\mbox{\hspace{.1in} /\unib\unir\uniI\unik\unis\unit\unia\unir\unin\uniO\unil\unit\unir\unin\uniIX\unit\uniI\univ / {\em (phonetician)}}\\
\mbox{\hspace{.1in} /\unib\unir\uniIX\unik\uniz\unia\unir\uniE\unin\uniO\unil\unit\unir\uniIX\unin\uniIX\unit\uniI\univ / {\em (automatic)}}\\
\vspace{-.00in}Fat showed in loose rolls beneath the shirt.\\
\vspace{-.02in}\mbox{\hspace{.1in} /\unif\uniA\unit\uniS\unio\uniu\unid\unit\uniIX\unin\unil\uniu\unis\unir\unio\uniu\unil\uniz\unib\uniIX\unin\unii\uniT\uniIX\uniS\unir\unit / {\em (phonetician)}}\\
\mbox{\hspace{.1in} /\unif\uniA\unit\uniS\uniE\unid\unii\uniI\unin\unid\uniAX\uniD\uniAX\unil\uniIX\unis\uniw\unir\unil\unit\unis\unip\uniIX\unit\unin\unii\uniIX\uniT\uniD\uniIX\uniS\unir\unit / {\em (automatic)}}\\
\vspace{-.00in}It suffers from a lack of unity of purpose and respect for heroic leadership.\\
\vspace{-.02in}\mbox{\hspace{.1in} /\uniIX\unit\unis\uniAH\unif\unir\uniz\unif\unir\uniAX\unim\uniAX\unil\uniA\unik\uniIX\univ\uniy\uniu\unin\uniIX\unit\unii\uniAX\univ\unip\unir\unip\uniIX\unis\uniE\unin\unir\uniIX\unis\unip\uniE\unik\unit\unif\unir\uniH\unir\unio\uniu\uniIX\unik\unil\unii\unit\unir\uniS\uniI\unip / {\em (phonetician)}}\\
\mbox{\hspace{.1in} /\uniIX\unit\unis\uniT\uniAH\unip\unir\uniz\unif\unir\unin\unia\unil\uniA\unik\uniE\unid\unik\unii\uniIX\unin\uniI\unid\unis\uniEPI\unii\uniIX\unip\unir\unip\uniAH\unis\uniIX\unin\unir\uniIX\unis\unip\unib\uniA\unik\unit\unif\unir\unih\unir\uniA\unil\uniIX\unik\unil\unii\unir\uniS\uniA\unip / {\em (automatic)}}\\
\vspace{-.00in}His captain was thin and haggard and his beautiful boots were worn and shabby.\\
\vspace{-.02in}\mbox{\hspace{.1in} /\unih\uniI\uniz\unik\uniA\unip\unit\uniIX\unin\uniw\uniAX\unis\uniT\uniI\unin\uniA\unin\uniH\uniA\unig\unir\unid\uniIX\unin\uniI\uniz\unib\uniy\uniu\unit\uniu\unif\unil\unib\uniu\unit\unis\uniEPI\uniw\unir\uniw\uniO\unir\unin\uniIX\unin\uniEPI\uniS\uniA\unib\unii / {\em (phonetician)}}\\
\mbox{\hspace{.1in} /\unih\uniI\uniz\unik\unia\unit\uniAH\unin\uniw\uniAX\unis\unit\uniD\uniA\unin\uniAX\unin\unih\uniA\unig\uniI\unir\unid\uniE\unin\uniI\uniIX\uniz\unip\unib\uniy\uniu\unit\uniIX\unif\unil\unid\unib\unil\unio\uniu\unik\unit\uniz\uniEPI\uniw\unir\uniw\uniO\unir\unin\uniIX\uniG\uniS\uniA\unib\unii / {\em (automatic)}}\\
\\
\vspace{-.03in}The statute allows for a great deal of latitude agrees Arthur Christy the
first special prosecutor\\
\vspace{-.00in}appointed under the nineteen seventy eight law.\\
\vspace{-.02in}\mbox{\hspace{.1in} /\unin\unih\uniIX\unis\unit\uniA\unik\uniC\uniI\uniz\uniI\unil\unia\uniu\uniAX\uniz\unis\uniU\unid\unig\unir\unie\unii\unit\unii\unil\unio\uniu\uniz\unil\unia\unii\uniAH\unit\uniI\uniz\unih\unii\uniI\unih\uniIX\unig\unir\unii\uniz\uniIX\unir\uniT\uniI\uniZ\unih\uniI\unis\unit\unii\unih\uniIX\unis\unia\unir\unis\uniAH\uniS\unil\unid\unih\uniO\unis\uniEPI}\\
\mbox{\hspace{.1in} \ \unin\unit\uniC\uniIX\unit\unir\unim\unih\unil\unii\unit\uniI\uniD\uniAX\unin\uniIX\uniz\uniIX\unin\unia\unii\uniG\unit\unii\unin\unis\uniIX\unim\uniIX\unin\unii\unie\unii\uniIX\uniEPI\unil\unio\uniu\unin\unih\uniIX /}\\
\vspace{-.00in}In past investigations he notes the focus has been quite narrow.\\
\mbox{\hspace{.1in} /\unih\uniIX\unin\unid\unih\unia\uniu\unis\unid\uniz\uniIX\unin\uniz\uniE\uniz\uniIX\unig\unii\uniS\uniIX\uniz\unih\unii\unin\uniE\unil\uniAX\unis\unih\uniIX\unis\uniAH\unil\unid\unig\uniI\unis\unih\uniIX\uniz\unih\uniI\unin\unik\unil\unii\unit\unim\uniI\unir\unio\uniu\unil\uniu\unih\uniA /}\\
\vspace{-.00in}But in each case he suggests it's up to the counsel he certainly has the powers.\\
\mbox{\hspace{.1in} /\unit\uniA\unih\uniAH\unit\uniI\uniG\unii\unit\uniC\unit\unik\unii\unis\unih\unii\uniz\unii\unid\uniJ\uniE\unis\unid\uniz\unis\uniAX\unih\uniIX\unis\uniAH\uniz\unit\uniIX\uniD\uniIX\unik\uniE\unil\unis\uniEPI\unil\unih\unii\unis\uniE\uniz\uniAX\unil\unii\unih\uniE\uniz\uniEPI\uniAX\unih\uniE\unil\unir\uniu\unis\unin\unih\uniAX /}\\
\vspace{-.03in}The department said wages are rising an average one point two percent in
the first year of the\\
\vspace{-.00in}nineteen eighty six labor contracts.\\
\vspace{-.02in}\mbox{\hspace{.1in} /\unih\uniIX\unit\uniIX\unih\unia\unir\unim\uniA\unin\unis\uniIX\unid\unil\unii\unit\uniJ\uniI\uniz\unir\univ\unir\unia\unii\uniz\unii\uniG\uniI\unin\uniA\univ\uniIX\unid\unit\unim\uniw\uniE\unin\unih\uniAX\unil\unii\uniG\unid\unit\uniIX\unih\unir\unis\uniA\unin\unih}\\
\mbox{\hspace{.1in} $\,$\ \uniIX\unin\uniAX\unis\unir\uniS\unit\unii\unir\unih\uniAX\uniz\uniIX\unin\unia\unii\unit\unii\uniG\unii\unit\unii\unis\uniI\unik\unil\uniI\unih\unir\uniz\unih\uniO\unin\unit\unir\unia\uniu\unin\unid\unis\uniAX\unih /}
\end{tabular}
\end{center}
\caption{\label{fig:timit} Above, four utterances from the TIMIT corpus used
  to evaluate the performance of the automatic transcriber; both the
  phoneticians' transcriptions and the automatic transcriber's
  output are shown.  Below, four utterances from the WSJ1 corpus with the
  transcriber's output.  Note the extremely poor quality of this input to
  the learning algorithm.}  \pageline
\end{figure}

The automatic transcriber was run on each of the 68,000 utterances from the
WSJ1 corpus of continuous speech.  Only the maximum-likelihood (Viterbi)
phoneme sequence was recorded.  The resulting transcriptions are usually
unreadable, even by trained experts.  Figure~\ref{fig:timit} presents the
first four sentences from the WSJ1 corpus and their automatically generated
transcriptions.

\begin{figure}[p]
\pageline
\def\ipa{\ipaninerm}
\begin{small}
\begin{center}
\begin{tabular}{rrrrlll}
Rank & $-\log p_G(w)$ & $|w|_G$ & $c_G(w)$ & w & $\mbox{rep}(w)$ & Usage \\ \hline 
    0&  4.356&  & 137161.72& \unid& {\em terminal}\\
    1&  4.376&  & 135301.41& \unit& {\em terminal}\\
    2&  4.454&  & 128187.62& \uniIX& {\em terminal}\\
   80&  9.978& 14.735& 2785.07& [\unih\uniIX\uniz]&   [\unih\uniIX\uniz] & his\\
   81&  9.985& 15.799& 2772.66& [\uniD\unie\unii]&   [\uniD[\unie\unii]] & they\\
   82& 10.008& 13.516& 2728.53& [\uniIX\unis\unit]&   [\uniIX\unis\unit]\\
 1000& 13.568& 16.861&  231.30& [\unin\uniE\unim]&   [\unin\uniE\unim]\\
 1001& 13.570& 16.327&  231.03& [\unit\uniS\unin]&   [\unit\uniS\unin]\\
 1002& 13.570& 17.384&  231.00& [\uniy\uniu\unit\unii]&   [[\uniy\uniu]\unit\unii]\\
 9160& 18.829& 25.498&    6.03& [\unii\uniG\unib\uniI\uniz\unin]&   [[\unii\uniG][\unib\uniI\uniz]\unin]\\
 9161& 18.830& 25.870&    6.03& [\unid\unim\uniI\unin\uniI\uniz\unit]&
 [\unid\unim[\uniI\unin\uniI\uniz\unit]] & administration\\
 9162& 18.831& 25.752&    6.03&
 [\unip\unir\uniI\uniz\uniIX\unit\uniE\unin]&
 [[\unip\unir\uniI\uniz\uniIX]\unit[\uniE\unin]] & president\\
 9163& 18.833& 25.559&    6.02& [\uniE\unin\unid\unis\unip\unir]&   [[\uniE\unin]\unid\unis[\unip\unir]]\\
 9164& 18.837& 44.253&    6.00&
 [\unig\unio\uniu\unil\unid\unim\uniIX\unin\unis\uniA\unik\unis]&
 [[\unig\unio\uniu\unil]\unid[\unim\uniIX\unin]\unis[\uniA\unik\unis]] & Goldman-Sachs\\
 9165& 18.837& 33.683&    6.00& [\unik\unim\unip\uniS\uniu\unit\unir]&
 [[\unik\unim\unip][\uniS\uniu\unit]\unir] & computer\\
 9166& 18.837& 31.309&    6.00& [\unig\unia\univ\unir\unim\uniIX\unin]&
 [\unig\unia[\univ\unir\unim\uniIX\unin]] & government\\
 9167& 18.837& 31.549&    6.00&
 [\unio\uniu\unib\unil\uniz\uniAX\unih\uniu\unio\uniu]&
 [[\unio\uniu\unib\unil][\uniz\uniAX\unih\uniu\unio\uniu]] & double quote\\
 9168& 18.837& 31.174&    6.00&
 [\unim\uniIX\unin\uniIX\unis\unit\unir\unie\unii\uniS\uniIX\unin]&
 [[\unim\uniIX\unin]\uniIX[\unis\unit\unir\unie\unii\uniS\uniIX\unin]] & administration\\
 9169& 18.837& 23.988&    6.00& [\unit\uniJ\uniE\unir\uniIX\unin]&   [[\unit\uniJ\uniE]\unir[\uniIX\unin]]\\
 9170& 18.837& 30.343&    6.00&
 [\unih\uniAH\unib\unil\unih\uniAX\unih\uniw\unio\uniu]&
 [[\unih\uniAH\unib\unil][\unih\uniAX\unih\uniw\unio\uniu]] & double quote\\
 9171& 18.837& 29.909&    6.00& [\unis\uniAH\unim\unip\uniD\uniIX\uniG]&
 [\unis[\uniAH\unim\unip][\uniD\uniIX\uniG]] & something\\
 9172& 18.837& 32.469&    6.00& [\unip\unir\unip\unil\unio\uniu\uniz\unil]&
 [[\unip\unir][\unip\unil\unio\uniu]\uniz\unil] & proposal\\
 9173& 18.837& 30.133&    6.00& [\unib\unio\uniu\unis\unik\unig\unii]&
 [[\unib\unio\uniu][\unis\unik\unig]\unii]& (Ivan) Boesky\\
 9174& 18.838& 30.019&    6.00& [\unik\unig\uniE\unid\uniJ\uniIX\unil]&
 [[\unik\unig\uniE][\unid\uniJ\uniIX]\unil] & schedule\\
 9175& 18.838& 33.758&    6.00&
 [\unig\unio\uniu\unil\unid\unim\unia\unii\uniIX\unin\uniz]&
 [[\unig\unio\uniu\unil]\unid[\unim\unia\unii\uniIX\unin\uniz]] & Goldman-Sachs\\
 9176& 18.838& 29.464&    6.00&
 [\unik\uniO\unir\unip\unir\unie\unii\unit\uniI\unid]&
 [[\unik\uniO\unir\unip\unir][\unie\unii\unit\uniI\unid]] & incorporated\\
 9177& 18.838& 30.073&    5.99&
 [\unis\uniIX\unit\uniC\uniu\unie\unii\uniS\uniIX\unim]&
 [[\unis\uniIX\unit\uniC\uniu][\unie\unii\uniS\uniIX\unim]] & situation\\
 9178& 18.838& 30.214&    5.99& [\unik\uniAX\unim\unir\uniS\uniAX\unil]&
 [[\unik\uniAX\unim]\unir[\uniS\uniAX\unil]] & commercial\\
 9179& 18.838& 26.638&    5.99& [\uniz\unio\uniu\unig\unik\unis]&   [\uniz[\unio\uniu][\unig\unik\unis]]\\
 9180& 18.839& 31.360&    5.99&
 [\unii\uniIX\unin\unid\uniIX\uniz\unit\uniJ\unii]&
 [\unii[\uniIX\unin\unid][\uniIX\uniz\unit\uniJ]\unii]\\
 9181& 18.839& 28.854&    5.99& [\unil\uniA\uniz\unid\uniJ\uniI\unir]&
 [\unil\uniA[\uniz\unid\uniJ\uniI\unir]] & last year\\
 9182& 18.839& 28.147&    5.99& [\unih\unia\uniu\uniw\uniA\univ\unir]&
 [[\unih\unia\uniu\uniw][\uniA\univ\unir]] & however\\
 9183& 18.839& 28.110&    5.99& [\uniz\uniIX\unib\uniI\unil\uniAX\unit\unii]&   [[\uniz\uniIX\unib][\uniI\unil\uniAX\unit\unii]]\\
 9184& 18.840& 28.088&    5.99& [\uniI\unin\uniI\unid\uniI\uniS\uniI\unin]&
 [[\uniI\unin\uniI\unid\uniI\uniS]\uniI\unin] & in addition\\
 9185& 18.840& 24.205&    5.99& [\unir\unii\uniIX\unin\unid\uniJ\uniIX]&   [\unir[\unii\uniIX\unin][\unid\uniJ\uniIX]]\\
 9186& 18.840& 28.961&    5.99& [\unib\unii\unig\unid\unih\uniAH\unim]&
 [[\unib\unii\unig]\unid[\unih\uniAH\unim]] & become\\
 9187& 18.840& 30.456&    5.99& [\uniz\unih\uniE\unil\uniA\unim\unir]&   [[\uniz\unih\uniE][\unil\uniA\unim]\unir]\\
 9188& 18.840& 28.059&    5.99& [\unim\unia\unii\unir\unik\unig\uniI]&   [[\unim\unia\unii]\unir[\unik\unig\uniI]]\\
 9189& 18.841& 28.383&    5.98&
 [\unis\uniIX\unin\unil\unii\unim\uniIX\unin]&
 [\unis[\uniIX\unin\unil\unii][\unim\uniIX\unin]] & Solomon (Brothers)\\
 9190& 18.841& 29.154&    5.98&
 [\uniD\uniIX\unid\uniAH\univ\unir\unim\uniE\unin]&
 [[\uniD\uniIX]\unid[\uniAH\univ\unir\unim\uniE\unin]] & the government\\
 9191& 18.841& 28.658&    5.98& [\unip\unir\unia\unim\unil\uniAX\unim]&
 [[\unip\unir\unia]\unim[\unil\uniAX\unim]] & problem\\
 9192& 18.841& 27.380&    5.98& [\unid\uniJ\uniA\unin\unir\uniAX\unil]&
 [[\unid\uniJ][\uniA\unin][\unir\uniAX\unil]] & general\\
 9193& 18.841& 30.144&    5.98&
 [\unih\uniAX\unib\unil\uniz\uniAX\unih\uniw\unio\uniu]&
 [[\unih\uniAX\unib\unil\uniz\uniAX\unih][\uniw\unio\uniu]] & double quote\\
 9194& 18.841& 27.137&    5.98& [\unis\unit\unir\uniO\uniAX\uniG]&
 [[\unis\unit\unir\uniO\uniAX]\uniG] & strong\\
 9195& 18.842& 27.166&    5.98& [\unia\uniT\uniIX\unin\unii\uniz]&
 [\unia\uniT[\uniIX\unin\unii\uniz]] & Japanese\\
 9196& 18.843& 29.567&    5.98& [\uniIX\uniG\unik\unil\uniI\unit\uniI]&   [\uniIX[\uniG\unik\unil][\uniI\unit\uniI]]\\
 9197& 18.843& 29.517&    5.97& [\unil\uniA\unis\unit\uniS\unii\unir]&
 [[\unil\uniA\unis][\unit\uniS\unii]\unir] & last year\\
 9198& 18.843& 28.774&    5.97& [\unid\uniJ\uniA\unip\uniIX\unin\unii]&
 [[\unid\uniJ\uniA\unip][\uniIX\unin]\unii] & Japanese\\
 9199& 18.844& 24.464&    5.97& [\uniE\unir\unii\uniA\unil]&   [[\uniE\unir\unii]\uniA\unil]
\end{tabular}
\end{center}
\end{small}
\caption{\label{fig:speech}Some words from a lexicon learned from
  dictated Wall Street Journal articles.}  \pageline
\end{figure}
\label{grammar}

\subsection{Results}

The standard concatenative algorithm was run on the 68,000 phonemic
transcriptions from the WSJ1 corpus, separated at utterance boundaries.
The algorithm produces a lexicon of 9,624 words; excerpts are presented in
figure~\ref{fig:speech}.  Those words which are used consistently in the
representation of the input are labeled with the ``underlying words'' they
account for.  For example, parameter 9,164
(/\unig\unio\uniu\unil\unid\unim\uniIX\unin\unis\uniA\unik\unis /) is used
to represent spoken utterances about the Goldman-Sachs investment firm.  So
is parameter 9,175
(/\unig\unio\uniu\unil\unid\unim\unia\unii\uniIX\unin\uniz /), a slightly
different pronunciation of the same words.  Most of the longer parameters
near the end of the lexicon are used in a consistent manner; the list in
figure~\ref{fig:speech} reflects the financial nature of the speech.
Notice that many common phrases have several parameters devoted to them,
such as \word{Goldman-Sachs}, \word{Japanese}, \word{last year},
\word{administration} etc.  In some cases the pronunciations seem quite
strange.  For example, \word{double quote} (used by readers to refer to the
`` symbol) is captured by the parameter
/\unih\uniAX\unib\unil\uniz\uniAX\unih\uniw\unio\uniu/.  This reflects the
flaws of the speech recognition system-- it systematically mistranscribes
the sounds of \word{double quote}.  This is not a fundamental problem,
although it does make it difficult for us to interpret the lexicon.  All a
language learner needs is for the parameters of their lexicon to be used
{\em consistently}, so that meaning can be associated with sounds; internal
agreement with standard phonetic writing systems is irrelevant.

In many cases the compositional hierarchy is clearly performing as desired.
For example, parameter 9,182 (\word{however}
/\unih\unia\uniu\uniw\uniA\univ\unir /) is represented as \word{how}
/\unih\unia\uniu\uniw / $\circ$ \word{ever} /\uniA\univ\unir/.  See
similarly parameters 9,176 and 9,177 where morphological decomposition
takes place.  Also, in line with the discussion of
section~\ref{motiv:linguistics}, the algorithm compiles out word sequences
that have idiosyncratic pronunciations.  For example, parameter 9,181 is
\word{last year} /\unil\uniA\uniz\unid\uniJ\uniI\unir /.  Notice that the
underlying /\unis\unit\uniy/ sequence is pronounced /\uniz\unid\uniJ/.  In
English fast speech /\unit\uniy / is commonly pronounced /\unit\uniC /
(\word{want you} becomes \word{wantcha}).  And if a sound is pronounced
with vibrating vocal cords, the previous sound often assimilates that
property.  Thus, the transformation (for an unknown reason) of /\unit\uniC
/ to /\unid\uniJ/ also changes the /\unis / to a /\uniz /.
/\unil\uniA\uniz\unid\uniJ\uniI\unir / is therefore a natural pronunciation
of \word{last year}.  The algorithm has captured the fact that the two
words are pronounced differently together than separately by creating this
parameter.  If the algorithm had a mechanism for capturing sound changes
via perturbations, one would hope that this parameter would be represented
in terms of the two isolated words and a sound change.

At the same time, it is important to realize that the algorithm has not
learned enough to analyze any particular utterance well.  Parameters are
learned in cases where words or phrases are given consistent pronunciations
multiple times.  Since there is significant variation in word pronunciation
(or at least the transcriber's interpretation of word pronunciation),
infrequent words are not usually learned, and neither are many words with
lax vowels, which are transcribed inconsistently.

\subsection{Discussion}

In one sense, the dictionary presented in figure~\ref{fig:speech} is
extremely impressive.  It represents the first significant machine
acquisition of linguistic knowledge from raw speech, speech that is in many
ways much {\em more} complex than that children are exposed to.
Furthermore, this learning took place without access to the extralinguistic
environment.  This brings into question claims that language acquisition is
only possible because of special properties of mothers' speech and actions.

Nevertheless, this is only a very preliminary experiment, and suffers from
many deficiencies.  The acoustic models are trained using supervised
learning.  The compositional model has no means of representing sound
changes via perturbations.  In fact, the composition operator is quite
fundamentally flawed: if a word that ends with a given phoneme is composed
with a word that starts with the same phoneme, the result is a doubled
phoneme, even though such pairs are pronounced (and transcribed) as one.
In section~\ref{rep:speech} (see also de Marcken~\cite{deMarcken95d}) we
presented a more sophisticated composition and perturbation model that
incorporates significantly greater knowledge of phonology and phonetics,
and allows for sound changes.  Results of experiments with that model are
inconclusive: its computational burdens prevent it from being applied to
the large WSJ1 corpus, and on smaller tests we have performed there is not
enough data for substantial learning to take place.

Another significant flaw in the learning model is the use of Viterbi
transcriptions produced by the phoneme transcriber.  The automatic
transcriber assumes an uninformative, uniform language model.  From its
output our algorithm attempts to learn a more informative language model,
but the result is never used by the transcriber to improve the quality of
its output.  A conceptually and algorithmically small change that could
substantially improve the results of this experiment would be for the
transcriber to produce a phoneme lattice rather than a single sequence.  It
is not difficult to modify our algorithm to take such a lattice as input
(see section~\ref{concatmodel}).

\section{Learning Meanings}\label{res:mean}

This section reports some preliminary tests of the concatenative model
extended with the meaning perturbation operator.  The tests are completely
artificial in the sense that the meanings presented to the learning
algorithm are constructed from the orthography of sentences rather than
dervied from real situations.

\subsection{Input}

Evidence is constructed from the Nina portion of the CHILDES
database~\cite{MacWhinney85,Suppes73}.  This is a set of transcriptions of
interactions between a mother and a young child (Nina) over a multi-year
period.  Only the transcriptions of the mother's speech are used; these
amount to approximately 34,000 sentences of English text.  Each sentence is
converted to a phonemic form using a very simple text-to-phoneme converter.
This produces phoneme sequences that are not far removed from text; words
are pronounced consistently, for example.  These unsegmented phoneme
sequences are the sound-side of the input to the learning algorithm.

Meanings are constructed for each utterance by looking up each word in the
original (text) sentence in a small hand-constructed dictionary.  The
dictionary defines a set of sememes for each word.  The meaning of a
sentence is the union of the meanings of the words in the sentence.  The
dictionary has been constructed to test various properties of the
compositional framework.  Words related by simple morphological
transformations are given common sememe sets (usually a single sememe).
For example, the words \word{decorate}, \word{decorating},
\word{decoration} and \word{decorations} are defined to mean \sems{decor}.
Some function words (\word{a}, \word{an}, \word{the}, \word{of},
\word{this}, \word{that} and a few others) are assigned the empty meaning.
Some words with very different pronunciations are assigned the same
meaning: \word{OK} and \word{yes} both mean \sems{yes}.  Some words exhibit
simple compositional behavior: \word{nightgown} and \word{nightgowns} mean
\sems{night gown}; \word{unzip} means \sems{undo zip} and \word{unwrap}
means \sems{undo wrap}.  And finally, some words exhibit non-compositional
behavior: \word{blackboard} means \sems{black board blackboard};
\word{yesterday} means \sems{previous day}; \word{cranberry} and
\word{cranberries} mean \sems{red berry} and \word{strawberry} and
\word{strawberries} mean \sems{sweet berry}.  Figure~\ref{fig:seminput}
contains some sample utterances.

\begin{figure}[tb]
\pageline
\def\ipa{\ipatenrm}
\begin{center}
\begin{tabular}{ll}
\mbox{this is a book?} & \mbox{let's see if Linda can bring in a glass of something to drink.}\\
\mbox{/\uniD\uniI\unis\uniI\uniz\unie\unib\uniU\unik/} & \mbox{/\unil\uniE\unit\unis\unis\unii\uniI\unif\unil\unia\unii\unin\unid\uniAX\unik\uniA\unin\unib\unir\uniI\uniG\uniI\unin\unie\unig\unil\uniA\unis\uniAX\univ\unis\uniAH\unim\uniT\uniI\uniG\unit\uniu\unid\unir\uniI\uniG\unik/}\\
\sems{be book} & \{ \sem{let us see if linda can bring in glass}\\
               & \ \ \ \sem{something to drink} \} \\
\\
\mbox{what do you see in the book?} & \mbox{can you go see if there's a little cranberry juice left to drink?}\\
\mbox{/\uniw\unia\unit\unid\uniu\uniy\uniu\unis\unii\uniI\unin\uniD\uniAX\unib\uniU\unik/} & \mbox{/\unik\uniA\unin\uniy\uniu\unig\unio\uniu\unis\unii\uniI\unif\uniD\uniE\unir\uniz\unie\unil\uniI\unit\unit\uniAX\unil\unik\unir\uniA\unin\unib\unir\unir\unii\uniJ\uniu\unia\unii\unis\unil\uniE\unif\unit\unit\uniu\unid\unir\uniI\uniG\unik/}\\
\sems{what do you see in book} & \{ \sem{can you go see if there be little red berry}\\
                               & \ \ \ \sem{juice leave to drink} \}\\
\\
\mbox{how many rabbits?} & \mbox{I don't think we have anything else.}\\
\mbox{/\unih\unia\uniu\unim\uniE\unin\unii\unir\uniA\unib\unib\uniI\unit\unis/}& \mbox{/\unia\unii\unid\unio\uniu\unin\unit\uniT\uniI\uniG\unik\uniw\unii\unih\uniA\univ\uniE\unin\unii\uniT\uniI\uniG\uniE\unil\unis/}\\
\sems{how many rabbit} & \sems{i do not think we have anything else}
\end{tabular}
\end{center}

\caption{\label{fig:seminput} Six utterances as constructed from the Nina
  portion of the CHILDES database.  The left three are from the start of
  the corpus and the right three are from a bit later in Nina's life.  Each
  utterance is presented in three parts: first, the original text (this is
  not seen by the algorithm); then the phonemic form; then the meaning of
  the utterance, an unordered set of sememes.}  \pageline
\end{figure}

\subsection{Method}

Two experiments were performed, both over the same 10,000 utterance subset
of the 34,000 utterance corpus.  The data spanned the entire corpus, but
was filtered down to 10,000 utterances to reduce computation time.  In both
cases the basic concatenative algorithm was run for 10 iterations to
produce a seed dictionary for the meaning algorithm, which was run for an
additional 8 iterations.  This staged process was also designed to reduce
computation time.

In the first experiment, each utterance was paired with its meaning.  In
the second, three possible meanings were presented for each utterance,
weighted equally with $p(z|v) = 1$.  The meanings were taken from the
utterance and the two surrounding utterances.

After training, the original input was reparsed using the basic
concatenative algorithm: the dictionary contained meanings, but the
algorithm parsed the input on the basis of its sound only.  The Viterbi
representation of each utterance was used to construct a sememe set, and
this was compared against the ``true'' meaning of the utterance.

The description length of a sememe was set at 10 bits.

\subsection{Results}

Trained on single meanings, sememe accuracy was 97.6\%, sememe recall was
91.4\%.  Trained with the three ambiguous meanings, sememe accuracy was
96.5\%, sememe recall was 70.2\%.  For very similar results on a slightly
different data set, see de Marcken~\cite{deMarcken96b}.

Although it would have been valuable to do so, no experiments were
performed in which the algorithm was tested on different utterances than it
was trained on.  However, it was not the case that the meaning algorithm
created words bigger than those produced by the basic algorithm, so it is
not the case that the algorithm's good performance is due to an over-fitting
of the data.

\subsection{Discussion}

These results are very encouraging.  The algorithm very rarely learns the
wrong meaning for words (sememe accuracies of 97.6\% and 96.5\%), and
learns most word meanings (sememe recalls of 91.4\% and 70.2\%).  Some of
the recall errors are cases of words that only once in the training data.

However, the learning algorithm suffers when ambiguous meanings are
presented; notice the significantly lower recall.  The reason is that the
algorithm starts with ``empty'' meanings for each word.  It therefore
predicts that all utterances have the empty meaning, and when computing the
posterior probability of meanings, meanings which are simple (have few
sememes) get assigned a disproportionately high probability.  As a
consequence, the algorithm is excessively biased towards simple meanings,
and the confusion that results interferes with learning.

\chapter{Conclusions}\label{ch:conclusion}

This thesis has presented a broad computational theory of unsupervised
language acquisition, based on Bayesian inference with a prior defined in
terms of model size, and a common representation for grammars and evidence
that is both linguistically appropriate and statistically efficient.  It
has presented learning algorithms for several specific instantiations of
the theory, and tested these algorithms on complex text and speech signals.
The resulting grammars accord very well with known properties of the human
language processing mechanism.

This thesis represents a significant milestone for theories of language
acquisition, because it provides a concrete demonstration of how learning
can take place from evidence that is of comparable complexity to that
children receive.  Few other theories have been shown to produce
linguistically plausible grammars, and none from data that is unequivocally
available to children.  The experiments on learning words directly from
continuous speech and on learning to map from unsegmented character
sequences to representations of meaning are both firsts.

At the same time, the thesis has explained conditions that need to be met
for any theory of unsupervised language acquisition to converge to
linguistically plausible grammars.  These are conditions on the
relationship between linguistic mechanisms and statistical models.  Among
the most important is that grammars must be able to model patterns in the
input that arise from causes external to language, without sacrificing
linguistic structure.  Most other statistical theories of language
acquisition have failed because they have violated one or more of these
conditions.

It is interesting to look at why this work has succeeded whereas many
similar experiments have not.  The general learning framework is not new:
Bayesian inference, stochastic language models, and the minimum description
length principle are all standard tools in the machine learning community,
and have been applied by many to problems of language acquisition.  At the
same time, many of the specific types of language models discussed here are
also similar or even identical to those others have used.  Indeed, the
multigram model that is the foundation for all of the experiments described
here is a staple of the data compression and language modeling communities,
and has been applied to the problem of learning a lexicon many times over
the last thirty years.

Two innovations are key to the superior performance of our algorithms.  The
first is the compositional framework.  It provides a principled means for
the description length of a grammar to be computed, makes parameters
inexpensive, and biases the learner towards linguistically plausible
grammars.  Perhaps most fundamentally, it allows a grammar to capture
patterns at many different scales simultaneously, ensuring that linguistic
structure does not lose out in a statistical competition with other sources
of regularity in the input.  The second innovation is the type of learning
algorithm we use.  Unlike most other grammar-optimization procedures, our
algorithms do not directly manipulate, or even store, a representation of
the grammar.  Instead, they manipulate the ``content'' of the grammar--
information that determines how the grammar behaves, rather than how it
looks.  From this information an optimal representation can easily be
reconstructed.  This strategy avoids many of the local optima problems that
have traditionally plagued classes of grammars in which desired moves
require complex changes to representations.

\section*{Future Work}

The theory of language acquisition that has been presented here is very
general, and only a few instances have been explored in any depth.  Many
interesting ones remain open for further research.

The concatenative model, based on the multigram distribution, is simple but
weak.  It has no concept of type, and therefore can not capture patterns
over syntactic and semantic classes.  The simplest remedy to this problem
is to associate with each a parameter a class, as described in
section~\ref{rep:cfg}, but as discussed there this is a poor fix.  It makes
more sense in the compositional framework to model classes with features
that are inherited.  Each parameter introduces a unique feature and also
inherits the features of its components; perturbations alter the default
feature set of a parameter.  In this way, a phrase like \word{red apples}
has the features of \word{red} and \word{apples} and its own feature.  A
class is any set of objects with a common set of features.  Many
interesting statistical and algorithmic issues arise in such models.

Tests of the meaning perturbation operator have been completely artificial;
more interesting experiments would apply the algorithms to representations
of meaning that arise in real situations.  An obvious application is
machine translation.  Given a pair of translated documents, the methods
described in this thesis can be run to produce representations for each
document.  One of these (ambiguous) representations can be treated as the
meaning of the other, for purposes of learning a translation model.  The
fact that the framework explains some forms of non-compositional behavior
is very desirable for machine translation.

Perhaps the area that most deserves follow-up work is learning directly
from speech signals.  The experiments performed in this thesis are
promising but rudimentary, and only hint as to what is possible.  With
better acoustic models and models of sound change, and proper integration
of the language model with the acoustic model, results will no doubt
improve dramatically.  It may be that there are near-term limits on what is
learnable from speech alone and an intriguing possibility is to provide the
learning algorithm with textual transcripts as ``meanings''.  This extra
information may improve performance to the point that practical lexicons
for speech recognizers can be learned from transcribed speech.

A final area that warrants further research is the derivation of on-line
learning algorithms based on the ideas of this thesis.  The algorithms
described here make multiple passes over the input, which imposes limits on
the amount of evidence that can be used for learning, and makes it
difficult for the algorithms to adapt to non-stationary properties of the
data.

\appendix
\chapter{Phonemes Used in Transcriptions}\label{app:phonemes}

\def\ipa{\ipatenrm}

Sounds are transcribed in the text using the following set of symbols to
represent phonemes, taken from the International Phonetic Alphabet (IPA).
Phonemes and phoneme sequences are delimited by slash marks: the word
\word{canoe} might be transcribed /\unik\uniAX\unin\uniu/.

\begin{center}
\begin{tabular}{clcl}
Symbol & Example & Symbol & Example \\ \hline
\unib&\underline{b}ee&\unih&\underline{h}ay\\
\unip&\underline{p}ea&\uniH&a\underline{h}ead\\
\unid&\underline{d}ay&\uniI&b\underline{i}t\\
\unit&\underline{t}ea&\unii&b\underline{ee}t\\
\unig&\underline{g}ay&\uniU&b\underline{oo}k\\
\unik&\underline{k}ey&\uniu&b\underline{oo}t\\
\uniJ&\underline{j}oke&\uniE&b\underline{e}t\\
\uniC&\underline{ch}oke&\unie&b\underline{a}se\\
\unis&\underline{s}ea&\uniAH&b\underline{u}t\\
\uniS&\underline{sh}e&\unio&b\underline{o}ne\\
\uniz&\underline{z}one&\uniA&b\underline{a}t\\
\uniZ&a\underline{z}ure&\unia&b\underline{o}b\\
\unif&\underline{f}in&\uniO&b\underline{ou}ght\\
\univ&\underline{v}an&\uniIX&ros\underline{e}s\\
\uniT&\underline{th}in&\uniAX&\underline{a}bout\\
\uniD&\underline{th}en&\uniEPI&{\em silence}\\
\unim&\underline{m}om\\
\unin&\underline{n}oon\\
\uniG&si\underline{ng}\\
\unil&\underline{l}ay\\
\unir&\underline{r}ay\\
\uniw&\underline{w}ay\\
\uniy&\underline{y}acht
\end{tabular}
\end{center}

\bibliography{}

\begin{thebibliography}{100}

\bibitem{Anderson77}
J.~R. Anderson.
\newblock Induction of augmented transition networks.
\newblock {\em Cognitive Science}, 1:125--47, 1977.

\bibitem{Anderson81}
J.~R. Anderson.
\newblock A theory of language acquisition based on general learning
  principles.
\newblock In {\em Proc. of the 7th International Joint Conference on Artificial
  Intelligence}, pages 97--103, Vancouver, B.C., Canada, 1981.

\bibitem{Anderson88b}
S.~R. Anderson.
\newblock Morphology as a parsing problem.
\newblock {\em Linguistics}, 26:521--544, 1988.

\bibitem{Angluin78}
D.~Angluin.
\newblock On the complexity of minimum inference of regular sets.
\newblock {\em Information and Control}, 39:337--350, 1978.

\bibitem{Angluin80}
D.~Angluin.
\newblock Inductive inference of formal languages from positive data.
\newblock {\em Information and Control}, 45(2):117--135, May 1980.

\bibitem{Angluin82}
D.~Angluin.
\newblock Inference of reversible langauges.
\newblock {\em Journal of the Association for Computing Machinery},
  29:741--765, 1982.

\bibitem{Baayen96}
H.~Baayen and R.~Sproat.
\newblock Estimating lexical priors for low-frequency syncretic forms.
\newblock {\em Computational Linguistics}, 22(2):155--166, 1996.

\bibitem{Baker79}
J.~K. Baker.
\newblock Trainable grammars for speech recognition.
\newblock In {\em Proceedings of the 97th Meeting of the Acoustical Society of
  America}, pages 547--550, 1979.

\bibitem{Barton87b}
G.~E. Barton, R.~C. Berwick, and E.~S. Ristad.
\newblock {\em Computational Complexity and Natural Language}.
\newblock MIT Press, Cambridge, MA, 1987.

\bibitem{Baum89}
E.~B. Baum and D.~Haussler.
\newblock What size net gives valid generalization?
\newblock {\em Neural Computation}, 1(1):151--160, 1989.

\bibitem{Baum72}
L.~E. Baum.
\newblock An inequality and associated maximization technique in statistical
  estimation for probabilistic functions of {M}arkov processes.
\newblock {\em Inequalities}, 3:1--8, 1972.

\bibitem{Baum70}
L.~E. Baum, T.~Petrie, G.~Soules, and N.~Weiss.
\newblock A maximization technique occuring in the statistical analysis of
  probabilistic functions in {M}arkov chains.
\newblock {\em Annals of Mathematical Statistics}, 41:164--171, 1970.

\bibitem{Bell90}
T.~C. Bell, J.~G. Cleary, and I.~H. Witten.
\newblock {\em Text Compression}.
\newblock Prentice-Hall, Englewood Cliffs, New Jersey, 1990.

\bibitem{Bellman57}
R.~Bellman.
\newblock {\em Dynamic Programming}.
\newblock Princeton University Press, Princeton, New Jersey, 1957.

\bibitem{Berger96}
A.~L. Berger, S.~A. Della~Pietra, and V.~J. Della~Pietra.
\newblock A maximum entropy approach to natural language processing.
\newblock {\em Computational Linguistics}, 22(1):39--71, 1996.

\bibitem{Berwick85}
R.~C. Berwick.
\newblock {\em The Acquisition of Syntactic Knowledge}.
\newblock MIT Press, Cambridge, MA, 1985.

\bibitem{Berwick87b}
R.~C. Berwick and S.~Pilato.
\newblock Learning syntax by automata induction.
\newblock {\em Machine Learning}, 2:9--38, 1987.

\bibitem{Biermann72}
A.~W. Biermann and J.~W. Feldman.
\newblock A survey of results in grammatical inference.
\newblock In S.~Watanabe, editor, {\em Frontiers of Pattern Recognition}.
  Academic Press, 1972.

\bibitem{Bloomfield33}
L.~Bloomfield.
\newblock {\em Language}.
\newblock Holt, New York, 1933.

\bibitem{Blumer87}
A.~Blumer, A.~Ehrenfeucht, D.~Haussler, and M.~K. Warmuth.
\newblock {O}ccam's {R}azor.
\newblock {\em Information Processing Letters}, 24:377--380, 1987.

\bibitem{Breiman84}
L.~Breiman, J.~Friedman, R.~Olshen, and C.~Stone.
\newblock {\em Classification and Regression Trees}.
\newblock Probability Series. Wadsworth International Group, Belmont,
  California, 1984.

\bibitem{Brent93}
M.~R. Brent.
\newblock Untitled.
\newblock Unpublished ms., 1993.

\bibitem{Brent96}
M.~R. Brent and T.~A. Cartwright.
\newblock Distributional regularity and phonotactic constraints are useful for
  segmentation.
\newblock {\em Cognition}, 61, to appear.

\bibitem{Brent95}
M.~R. Brent, A.~Lundberg, and S.~Murthy.
\newblock Discovering morphemic suffixes: A case study in minimum description
  length induction.
\newblock In {\em Fifth International Workshop on AI and Statistics}, Ft.
  Lauderdale, Florida, 1995.

\bibitem{Briscoe92}
T.~Briscoe and N.~Waegner.
\newblock Robust stochastic parsing using the inside-outside algorithm.
\newblock In {\em Proc. of the AAAI Workshop on Probabilistic-Based Natural
  Language Processing Techniques}, pages 39--52, 1992.

\bibitem{Brown93}
P.~F. Brown, S.~A. Della~Pietra, V.~J. Della~Pietra, and R.~L. Mercer.
\newblock The mathematics of machine translation: Parameter estimation.
\newblock {\em Computational Linguistics}, 19(2):253--312, 1993.

\bibitem{Brown90}
P.~L. Brown, J.~Cocke, S.~A. Della~Pietra, V.~J. Della~Pietra, F.~Jelinek,
  J.~D. Lafferty, R.~L. Mercer, and P.~S. Rossin.
\newblock A statistical approach to machine translation.
\newblock {\em Computational Linguistics}, 16(2):79--86, 1990.

\bibitem{Brown92}
P.~L. Brown, S.~A. Della~Pietra, V.~J. Della~Pietra, J.~C. Lai, and R.~L.
  Mercer.
\newblock An estimate of an upper bound for the entropy of {E}nglish.
\newblock {\em Computational Linguistics}, 18(1):31--40, 1992.

\bibitem{Brown70}
R.~Brown and C.~Hanlon.
\newblock Derivational complexity and order of acquisition in child speech.
\newblock In J.~R. Hayes, editor, {\em Cognition and the Development of
  Language}. Wiley, New York, 1970.

\bibitem{Carroll95}
G.~Carroll.
\newblock {\em Learning Probabilistic Grammars for Language Modeling}.
\newblock PhD thesis, Brown University, Providence, Rhode Island, 1995.

\bibitem{Carroll92}
G.~Carroll and E.~Charniak.
\newblock Learning probabilistic dependency grammars from labeled text.
\newblock In {\em Working Notes, Fall Symposium Series, AAAI}, pages 25--31,
  1992.

\bibitem{Cartwright94}
T.~A. Cartwright and M.~R. Brent.
\newblock Segmenting speech without a lexicon: Evidence for a bootstrapping
  model of lexical acquisition.
\newblock In {\em Proc. of the 16th Annual Meeting of the Cognitive Science
  Society}, Hillsdale, New Jersey, 1994.

\bibitem{Chang92}
J.-S. Chang, S.-D. Chen, Y.~Zheng, Z.-Z. Liu, and S.-J. Ke.
\newblock Large-corpus-based methods for {C}hinese personal name recognition.
\newblock {\em Journal of Chinese Information Processing}, 6(3):7--15, 1992.

\bibitem{Chang95}
J.-S. Chang, Y.-C. Lin, and K.-Y. Su.
\newblock Automatic construction of a {C}hinese electronic dictionary.
\newblock In {\em Third Workshop on Very Large Corpora}, Cambridge,
  Massachusetts, 1995.

\bibitem{Chen95}
S.~F. Chen.
\newblock Bayesian grammar induction for language modeling.
\newblock In {\em Proc. 33rd Annual Meeting of the Association for
  Computational Linguistics}, pages 228--235, Cambridge, Massachusetts, 1995.

\bibitem{Chen96}
S.~F. Chen.
\newblock {\em Building Probabalistic Models for Natural Language}.
\newblock PhD thesis, Harvard University, Cambridge, Massachusetts, 1996.

\bibitem{Chomsky51}
N.~A. Chomsky.
\newblock Morphophonemics of modern {H}ebrew.
\newblock Master's thesis, University of Pennsylvania, 1951.

\bibitem{Chomsky55}
N.~A. Chomsky.
\newblock {\em The Logical Structure of Linguistic Theory}.
\newblock Plenum Press, New York, 1955.

\bibitem{Chomsky61}
N.~A. Chomsky.
\newblock On the notion ``rule of grammar''.
\newblock In R.~Jakobson, editor, {\em Proceedings of the Symposia in Applied
  Mathematics, Volume VII}, New York, 1961.

\bibitem{Chomsky65}
N.~A. Chomsky.
\newblock {\em Aspects of The Theory of Syntax}.
\newblock MIT Press, Cambridge, MA, 1965.

\bibitem{Chomsky86}
N.~A. Chomsky.
\newblock {\em Knowledge of Language: Its Nature, Origin, and Use}.
\newblock Praeger, New York, 1986.

\bibitem{Cleary84}
J.~G. Cleary and I.~H. Witten.
\newblock Data compression using adaptive coding and partial string matching.
\newblock {\em IEEE Transactions on Communications}, 32(4):396--402, 1984.

\bibitem{Cook76}
C.~M. Cook, A.~Rosenfeld, and A.~R. Aronson.
\newblock Grammatical inference by hill climbing.
\newblock {\em Information Sciences}, 10:59--80, 1976.

\bibitem{Cover50}
T.~M. Cover and R.~C. King.
\newblock A convergent gambling estimate of the entropy of {E}nglish.
\newblock {\em IEEE Transactions on Information Theory}, 24(4):413--421, 1950.

\bibitem{Csiszar84}
I.~Csiszar and G.~Tusn\'{a}dy.
\newblock Information geometry and alternating minimization procedures.
\newblock {\em Statistics and Decisions}, Supplemental Issue 1:205--237, 1984.

\bibitem{Cutler94}
A.~Cutler.
\newblock Segmentation problems, rhythmic solutions.
\newblock {\em Lingua}, 92(1--4), 1994.

\bibitem{deMarcken94b}
C.~de~Marcken.
\newblock The acquisition of a lexicon from paired phoneme sequences and
  semantic representations.
\newblock In {\em International Colloquium on Grammatical Inference}, pages
  66--77, Alicante, Spain, 1994.

\bibitem{deMarcken95b}
C.~de~Marcken.
\newblock Lexical heads, phrase structure and the induction of grammar.
\newblock In {\em Third Workshop on Very Large Corpora}, Cambridge,
  Massachusetts, 1995.

\bibitem{deMarcken95d}
C.~de~Marcken.
\newblock The unsupervised acquisition of a lexicon from continuous speech.
\newblock Memo 1558, MIT Artificial Intelligence Lab., Cambridge,
  Massachusetts, 1995.

\bibitem{deMarcken96b}
C.~de~Marcken.
\newblock Linguistic structure as composition and perturbation.
\newblock In {\em Proc. 34th Annual Meeting of the Association for
  Computational Linguistics}, pages 335--341, Santa Cruz, California, 1996.

\bibitem{Deligne95}
S.~Deligne and F.~Bimbot.
\newblock Language modeling by variable length sequences: Theoretical
  formulation and evaluation of multigrams.
\newblock In {\em Proceedings of the International Conference on Speech and
  Signal Processing}, volume~1, pages 169--172, 1995.

\bibitem{DellaPietra94}
S.~Della~Pietra, V.~Della~Pietra, J.~Gillett, J.~Lafferty, H.~Printz, and
  L.~Ure\u{s}.
\newblock Inference and estimation of a long-range trigram model.
\newblock In {\em International Colloquium on Grammatical Inference}, pages
  78--92, Alicante, Spain, 1994.

\bibitem{DellaPietra95}
S.~Della~Pietra, V.~Della~Pietra, and J.~Lafferty.
\newblock Inducing features of random fields.
\newblock Technical Report CMU-CS-95-144, Carnegie Mellon University,
  Pittsburgh, Pennsylvania, May 1995.

\bibitem{Dempster77}
A.~P. Dempster, N.~M. Liard, and D.~B. Rubin.
\newblock Maximum liklihood from incomplete data via the {EM} algorithm.
\newblock {\em Journal of the Royal Statistical Society}, B(39):1--38, 1977.

\bibitem{Dresher90}
B.~E. Dresher and J.~D. Kaye.
\newblock A computational learning model for metrical phonology.
\newblock {\em Cognition}, 34:137--195, 1990.

\bibitem{Ehrenfeucht89}
A.~Ehrenfeucht, D.~Haussler, M.~Kearns, and L.~Valiant.
\newblock A general lower bound on the number of examples needed for learning.
\newblock {\em Information and Control}, 82(3):247--251, 1989.

\bibitem{Ellison92}
T.~M. Ellison.
\newblock {\em The Machine Learning of Phonological Structure}.
\newblock PhD thesis, University of Western Australia, 1992.

\bibitem{Ellison94}
T.~M. Ellison.
\newblock The iterative learning of phonological rules.
\newblock {\em Computational Linguistics}, 20(3), 1994.

\bibitem{Francis82}
W.~N. Francis and H.~Kucera.
\newblock {\em Frequency Analysis of {E}nglish usage: Lexicon and Grammar}.
\newblock Houghton-Mifflin, Boston, 1982.

\bibitem{Geman92}
S.~Geman, E.~Bienenstock, and R.~Doursat.
\newblock Neural networks and the bias/variance dilemma.
\newblock {\em Neural Computation}, 4:1--58, 1992.

\bibitem{Gibson94}
E.~Gibson and K.~Wexler.
\newblock Triggers.
\newblock {\em Linguistic Inquiry}, 25:407--454, 1994.

\bibitem{Gleitman90}
L.~Gleitman.
\newblock The structural sources of verb meanings.
\newblock {\em Language Acquisition}, 1(1):3--55, 1990.

\bibitem{Gold67}
E.~M. Gold.
\newblock Language identification in the limit.
\newblock {\em Information and Control}, 10:447--474, 1967.

\bibitem{Gold78}
E.~M. Gold.
\newblock Complexity of automaton identification from given data.
\newblock {\em Information and Control}, 37:302--320, 1978.

\bibitem{Goodluck91}
H.~Goodluck.
\newblock {\em Language Acquisition}.
\newblock Blackwell Publishers, Cambridge, MA, 1991.

\bibitem{Halle61}
M.~Halle.
\newblock On the role of simplicity in linguistic descriptions.
\newblock In R.~Jakobson, editor, {\em Proceedings of the Symposia in Applied
  Mathematics, Volume VII}, New York, 1961.

\bibitem{Halle83b}
M.~Halle.
\newblock On distinctive features and their articulatory implementation.
\newblock {\em Natural Language and Linguistic Theory}, 1:91--105, 1983.

\bibitem{Harris93}
J.~W. Harris.
\newblock Integrity of prosodic constituents and the domain of syllabification
  rules in {S}panish and {C}atalan.
\newblock In K.~Hale and S.~J. Keyser, editors, {\em The View from Building 20:
  Essays in Linguistics in Honor of {S}ylvain {B}romberger}. MIT Press,
  Cambridge, MA, 1993.

\bibitem{Huffman52}
D.~A. Huffman.
\newblock A method for the construction of minimum-redundancy codes.
\newblock {\em Proceedings of the IRE}, 40(9):1098--1101, 1952.

\bibitem{Jelinek90}
F.~Jelinek, J.~D. Lafferty, and R.~L. Mercer.
\newblock Basic methods of probabilistic context free grammars.
\newblock In P.~Laface and R.~DeMori, editors, {\em Speech Recognition and
  Understanding}, pages 345--360. Springer-Verlag, New York, NY, 1990.

\bibitem{Joshi75}
A.~K. Joshi.
\newblock Tree adjunct grammars.
\newblock {\em Journal of Computer and System Sciences}, 10:136--163, 1975.

\bibitem{Jusczyk93}
P.~W. Jusczyk.
\newblock Discovering sound patterns in the native language.
\newblock In {\em Proc. of the 15th Annual Meeting of the Cognitive Science
  Society}, pages 49--60, 1993.

\bibitem{Jusczyk94}
P.~W. Jusczyk.
\newblock Infants speech perception and the development of the mental lexicon.
\newblock In J.~C. Goodman and H.~C. Nusbaum, editors, {\em The Development of
  Speech Perception}. MIT Press, Cambridge, MA, 1994.

\bibitem{Kanazawa94}
M.~Kanazawa.
\newblock {\em Learnability Classes of Categorial Grammars}.
\newblock PhD thesis, Stanford University, Stanford, California, 1994.

\bibitem{Kazman94}
R.~Kazman.
\newblock Simulating the child's acquisition of the lexicon and syntax--
  experiences with {B}abel.
\newblock {\em Machine Learning}, 16:87--120, 1994.

\bibitem{Kearns94}
M.~J. Kearns and U.~V. Vazirani.
\newblock {\em An Introduction to Computational Learning Theory}.
\newblock MIT Press, Cambridge, MA, 1994.

\bibitem{Keeping62}
E.~S. Keeping.
\newblock {\em Introduction to Statistical Inference}.
\newblock Van Nostrand, Princeton, New Jersey, 1962.

\bibitem{Kenstowicz94}
M.~Kenstowicz.
\newblock {\em Phonology in Generative Grammar}.
\newblock Blackwell Publishers, Cambridge, MA, 1994.

\bibitem{Klatt92}
D.~H. Klatt.
\newblock Review of selected models of speech perception.
\newblock In W.~Marslen-Wilson, editor, {\em Lexical Representation and
  Process}. MIT Press, Cambridge, MA, 1992.

\bibitem{Kolmogorov65}
A.~N. Kolmogorov.
\newblock Three approaches to the quantitative definition of information.
\newblock {\em Problems of Information Transmission}, 1(1):1--7, 1965.

\bibitem{Koskenniemi83}
K.~Koskenniemi.
\newblock {\em Two-Level Morphology: A General Computational Model for
  Word-Form Recognition and Production}.
\newblock PhD thesis, University of Helsinki, Helsinki, Finland, 1983.

\bibitem{Kupiec92b}
J.~Kupiec.
\newblock Robust part-of-speech tagging using a hidden {M}arkov model.
\newblock {\em Computer Speech and Language}, 6:225--242, 1992.

\bibitem{Lari90}
K.~Lari and S.~J. Young.
\newblock The estimation of stochastic context-free grammars using the
  inside-outside algorithm.
\newblock {\em Computer Speech and Language}, 4:35--56, 1990.

\bibitem{Levelt91}
W.~J.~M. Levelt.
\newblock {\em Speaking}.
\newblock MIT Press, Cambridge, MA, 1991.

\bibitem{Lewis75}
D.~Lewis.
\newblock Languages and language.
\newblock In K.~Gunderson, editor, {\em Language, Mind and Knowledge}.
  University of Minnesota Press, Minneapolis, 1975.

\bibitem{Li91}
M.~Li and P.~Vitanyi.
\newblock Inductive reasoning.
\newblock In E.~S. Ristad, editor, {\em Language Computations: DIMACS Series
  vol. 17}. American Mathematical Society, 1991.

\bibitem{Li93}
M.~Li and P.~Vitanyi.
\newblock {\em An introduction to {K}olmogorov complexity and its
  applications}.
\newblock Springer-Verlag, New York, NY, 1993.

\bibitem{Lieven94}
E.~V.~M. Lieven.
\newblock Crosslinguistic and crosscultural aspects of language addressed to
  children.
\newblock In C.~Gallaway and B.~J. Richards, editors, {\em Input and
  Interaction in Language Acquisition}, pages 56--73. Cambridge University
  Press, New York, NY, 1994.

\bibitem{Luo96}
X.~Luo and S.~Roukos.
\newblock An iterative algorithm to build {C}hinese language models.
\newblock In {\em Proc. 34th Annual Meeting of the Association for
  Computational Linguistics}, pages 139--143, Santa Cruz, California, 1996.

\bibitem{MacWhinney85}
B.~Mac\-Whinney and C.~Snow.
\newblock The child language data exchange system.
\newblock {\em Journal of Child Language}, 12:271--296, 1985.

\bibitem{MacWhinney78}
B.~MacWhinney.
\newblock Conditions on acquisition models.
\newblock In {\em Proceedings of the Association for Computing Machinery},
  pages 421--426, 1978.

\bibitem{Marcus93}
G.~F. Marcus.
\newblock Negative evidence in language acquisition.
\newblock {\em Cognition}, 46:53--85, 1993.

\bibitem{Markov13}
A.~A. Markov.
\newblock Example of a statistical investigation of the text of ``{E}ugene
  {O}negin'' illustrating the dependence between samples in chain.
\newblock {\em Bulletin de l'Academie Imperiale des Sciences de St.
  Petersbourg}, pages 153--162, 1913.

\bibitem{Marr82}
D.~Marr.
\newblock {\em Vision}.
\newblock W. H. Freeman and Company, San Francisco, 1982.

\bibitem{Miller84}
V.~S. Miller and M.~N. Wegman.
\newblock Variations on a theme by {Z}iv and {L}empel.
\newblock In A.~Apostolico and Z.~Galil, editors, {\em Combinatorial Algorithms
  on Words}, pages 131--140. Springer-Verlag, New York, NY, 1984.

\bibitem{Moffat90}
A.~Moffat.
\newblock Implementing the {PPM} data compression scheme.
\newblock {\em IEEE Transactions on Communications}, 38(11):1917--1921, 1990.

\bibitem{Murphy95}
P.~M. Murphy.
\newblock An empirical analysis of the benefit of decision tree size biases as
  a function of concept distribution.
\newblock Technical Report 95-29, Department of Information and Computer
  Science, University of California, Irvine, 1995.

\bibitem{Murphy94}
P.~M. Murphy and M.~J. Pazzani.
\newblock Exploring the decision forest: an empirical investigation of
  {O}ccam's razor in decision-tree induction.
\newblock {\em Journal of Artificial Intelligence Research}, 1:257--275, 1994.

\bibitem{NevillManning96}
C.~G. Nevill-Manning.
\newblock {\em Inferring Sequential Structure}.
\newblock PhD thesis, Department of Computer Science, University of Waikato,
  New Zealand, 1996.

\bibitem{Niyogi95}
P.~Niyogi.
\newblock {\em The Informational Complexity of Learning from Examples}.
\newblock PhD thesis, Massachusetts Institute of Technology, Cambridge,
  Massachusetts, 1995.

\bibitem{Niyogi94}
P.~Niyogi and R.~C. Berwick.
\newblock A {M}arkov language learning model for finite parameter spaces.
\newblock In {\em Proc. 32nd Annual Meeting of the Association for
  Computational Linguistics}, pages 171--180, Las Cruces, New Mexico, 1994.

\bibitem{Olivier68}
D.~C. Olivier.
\newblock {\em Stochastic Grammars and Language Acquisition Mechanisms}.
\newblock PhD thesis, Harvard University, Cambridge, Massachusetts, 1968.

\bibitem{Pasco76}
R.~C. Pasco.
\newblock {\em Source Coding Algorithms for Fast Data Compression}.
\newblock PhD thesis, Stanford University, Stanford, California, 1976.

\bibitem{Pereira92}
F.~Pereira and Y.~Schabes.
\newblock Inside-outside reestimation from partially bracketed corpora.
\newblock In {\em Proc. 30th Annual Meeting of the Association for
  Computational Linguistics}, pages 128--135, Berkeley, California, 1992.

\bibitem{Pinker94}
S.~Pinker.
\newblock {\em The Language Instinct}.
\newblock William Morrow and Company, New York, 1994.

\bibitem{Pitt89}
L.~Pitt.
\newblock Inductive inference, {DFA}s, and computational complexity.
\newblock Technical Report UIUCDCS-R-89-1530, University of Illinois at
  Urbana-Champaign, Urbana, Illinois, 1989.

\bibitem{Rabiner89}
L.~R. Rabiner.
\newblock A tutorial on hidden markov models and selected applications in
  speech recognition.
\newblock {\em Proceedings of the IEEE}, 77(2):257--286, 1989.

\bibitem{Rabiner93}
L.~R. Rabiner and B.-H. Juang.
\newblock {\em Fundamentals of Speech Recognition}.
\newblock Prentice Hall, Englewood Cliffs, NJ, 1993.

\bibitem{Rayner88}
M.~Rayner, A.~sa~Hugosson, and G.~Hagert.
\newblock Using a logic grammar to learn a lexicon.
\newblock Technical Report R88001, Swedish Institute of Computer Science, 1988.

\bibitem{Rissanen76}
J.~Rissanen.
\newblock Generalized {K}raft inequality and arithmetic coding.
\newblock {\em IBM Journal of Research and Development}, 20, 1976.

\bibitem{Rissanen78}
J.~Rissanen.
\newblock Modeling by shortest data description.
\newblock {\em Automatica}, 14:465--471, 1978.

\bibitem{Rissanen83}
J.~Rissanen.
\newblock A universal data compression system.
\newblock {\em IEEE Transactions on Information Theory}, 29(5):656--664, 1983.

\bibitem{Rissanen89}
J.~Rissanen.
\newblock {\em Stochastic Complexity in Statistical Inquiry}.
\newblock World Scientific, Singapore, 1989.

\bibitem{Rissanen91}
J.~Rissanen and E.~S. Ristad.
\newblock Language acquisition in the {MDL} framework.
\newblock In E.~S. Ristad, editor, {\em Language Computations: DIMACS Series
  vol. 17}. American Mathematical Society, 1991.

\bibitem{Ristad94}
E.~S. Ristad.
\newblock {\em The Language Complexity Game}.
\newblock MIT Press, Cambridge, MA, 1994.

\bibitem{Ristad95}
E.~S. Ristad and R.~G. Thomas.
\newblock New techniques for context modeling.
\newblock In {\em Proc. 33rd Annual Meeting of the Association for
  Computational Linguistics}, Cambridge, Massachusetts, 1995.

\bibitem{Sachs72}
J.~Sachs and M.~Johnson.
\newblock Language development in a hearing child of deaf parents.
\newblock In {\em International Symposium on First Language Acquisition},
  Florence, Italy, 1972.

\bibitem{Sakakibara92}
Y.~Sakakibara.
\newblock Efficient learning of context-free grammars from positive structural
  examples.
\newblock {\em Information and Control}, 97(1):23--60, Mar. 1992.

\bibitem{Schaffer94}
C.~Schaffer.
\newblock A conservation law for generalization performance.
\newblock In {\em Proc. of the 1994 International Conference on Machine
  Learning}, 1994.

\bibitem{DiSciullo87}
A.-M.~D. Sciullo and E.~Williams.
\newblock {\em On the Definition of Word}.
\newblock MIT Press, Cambridge, MA, 1987.

\bibitem{Selfridge81}
M.~Selfridge.
\newblock A computer model of child language acquisition.
\newblock In {\em Proc. of the 7th International Joint Conference on Artificial
  Intelligence}, pages 92--96, Vancouver, B.C., Canada, 1981.

\bibitem{Shannon48}
C.~E. Shannon.
\newblock A mathematical theory of communication.
\newblock {\em Bell System Technical Journal}, 27:623--656, 1948.

\bibitem{Shannon51}
C.~E. Shannon.
\newblock Prediction and entropy of printed {E}nglish.
\newblock {\em Bell System Technical Journal}, 30:50--64, 1951.

\bibitem{Shinohara90}
T.~Shinohara.
\newblock Inductive inference from positive data is powerful.
\newblock In {\em 1990 Workshop on Computational Learning Theory}, pages
  97--110, Los Altos, CA, 1990. Morgan Kaufmann.

\bibitem{Siklossy72}
L.~Sikl\'{o}ssy.
\newblock Natural language learning by computer.
\newblock In H.~A. Simon and L.~Sikl\'{o}ssy, editors, {\em Representation and
  Meaning}, pages 289--327. Prentice Hall, Englewood Cliffs, NJ, 1972.

\bibitem{Siskind92}
J.~M. Siskind.
\newblock Naive physics, event perception, lexical semantics, and language
  acquisition.
\newblock PhD thesis TR-1456, MIT Artificial Intelligence Lab., 1992.

\bibitem{Siskind93b}
J.~M. Siskind.
\newblock Lexical acquisition as constraint satisfaction.
\newblock Technical Report IRCS-93-41, University of Pennsylvania Institute for
  Research in Cognitive Science, Philadelphia, Pennsylvania, 1993.

\bibitem{Siskind94}
J.~M. Siskind.
\newblock Lexical acquisition in the presence of noise and homonymy.
\newblock In {\em Proc. of the American Association for Artificial
  Intelligence}, Seattle, Washington, 1994.

\bibitem{Sleator91}
D.~D. Sleator and D.~Temperley.
\newblock Parsing {E}nglish with a link grammar.
\newblock Technical Report CMU-CS-91-196, Carnegie Mellon University,
  Pittsburgh, Pennsylvania, 1991.

\bibitem{Snow76}
C.~E. Snow, A.~Arlmann-Rupp, Y.~Hassin, J.~Jobse, J.~Joosten, and J.~Vorster.
\newblock Mothers' speech in three social classes.
\newblock {\em Journal of Psycholinguistic Research}, 5:1--20, 1976.

\bibitem{Sokolov94}
J.~L. Sokolov and C.~E. Snow.
\newblock The changing role of negative evidence in theories of language
  development.
\newblock In C.~Gallaway and B.~J. Richards, editors, {\em Input and
  Interaction in Language Acquisition}, pages 38--55. Cambridge University
  Press, New York, NY, 1994.

\bibitem{Solomonoff59}
R.~J. Solomonoff.
\newblock A new method for discovering the grammars of phrase structure
  languages.
\newblock {\em Information Processing}, pages 258--290, 1959.

\bibitem{Solomonoff60}
R.~J. Solomonoff.
\newblock The mechanization of linguistic learning.
\newblock In {\em Proceedings of the 2nd International Conference on
  Cybernetics}, pages 180--193, 1960.

\bibitem{Solomonoff64}
R.~J. Solomonoff.
\newblock A formal theory of inductive inference.
\newblock {\em Information and Control}, 7:224--254, 1964.

\bibitem{Spencer91}
A.~Spencer.
\newblock {\em Morphological Theory}.
\newblock Blackwell Publishers, Cambridge, MA, 1991.

\bibitem{Sproat94}
R.~Sproat, N.~Chang, C.~Shih, and W.~Gale.
\newblock A stochastic finite-state word-segmentation algorithm for {C}hinese.
\newblock In {\em Proc. 32nd Annual Meeting of the Association for
  Computational Linguistics}, pages 66--73, Las Cruces, New Mexico, 1994.

\bibitem{Stolcke94}
A.~Stolcke.
\newblock {\em Bayesian Learning of Probabilistic Language Models}.
\newblock PhD thesis, University of California at Berkeley, Berkeley, CA, 1994.

\bibitem{Suppes73}
P.~Suppes.
\newblock The semantics of children's language.
\newblock {\em American Psychologist}, 1973.

\bibitem{Teahan96}
W.~J. Teahan and J.~G. Cleary.
\newblock The entropy of {E}nglish using {PPM}-based techniques.
\newblock In {\em Proceedings of the 1996 Data Compression Conference}, Salt
  Lake City, Utah, 1996.

\bibitem{Tishby94}
N.~Tishby and A.~Gorin.
\newblock Algebraic learning of statistical associations for language
  acquisition.
\newblock {\em Computer Speech and Language}, 8:51--78, 1994.

\bibitem{Valiant84}
L.~G. Valiant.
\newblock A theory of the learnable.
\newblock {\em Comm. of the ACM}, 27(11):1134--1142, 1984.

\bibitem{Vapnik82}
V.~N. Vapnik.
\newblock {\em Estimation of Dependences Based on Empirical Data}.
\newblock Springer-Verlag, New York, NY, 1982.

\bibitem{Vapnik71}
V.~N. Vapnik and A.~Y. Chervonenkis.
\newblock On the uniform convergence of relative frequencies of events to their
  probabilities.
\newblock {\em Theory of Probability and its Applications}, 16(2):264--280,
  1971.

\bibitem{Humboldt1836}
W.~von Humboldt.
\newblock {\em \"{U}ber die Verschiedenheit des menschlichen Sprachbaues}.
\newblock Berlin, 1836.

\bibitem{Wang92}
L.-J. Wang, W.-C. Li, and C.-H. Chang.
\newblock Recognizing unregistered names for {M}andarin word identification.
\newblock In {\em COLING 92: International Conference on Computational
  Linguistics}, pages 1239--1243, Nantes, France, 1992.

\bibitem{Webb96}
G.~I. Webb.
\newblock Further experimental evidence against the utility of {O}ccam's
  {R}azor.
\newblock {\em Journal of Artificial Intelligence Research}, 4:397--417, 1996.

\bibitem{Welch84}
T.~Welch.
\newblock A technique for high-performance data compression.
\newblock {\em IEEE COMPUTER}, pages 8--19, June 1984.

\bibitem{Wexler80}
K.~Wexler and P.~Culicover.
\newblock {\em Formal Principles of Language Acquisition}.
\newblock Cambridge University Press, New York, NY, 1980.

\bibitem{Wolff77}
J.~G. Wolff.
\newblock The discovery of segments in natural language.
\newblock {\em British Journal of Psychology}, 68:97--106, 1977.

\bibitem{Wolff80}
J.~G. Wolff.
\newblock Language acquisition and the discovery of phrase structure.
\newblock {\em Language and Speech}, 23(3):255--269, 1980.

\bibitem{Wolff82}
J.~G. Wolff.
\newblock Language acquisition, data compression and generalization.
\newblock {\em Language and Communication}, 2(1):57--89, 1982.

\bibitem{Wolpert95}
D.~H. Wolpert.
\newblock Off-training set error and {\em a priori} distinctions between
  learning algorithms.
\newblock Technical Report 95-01-003, Santa Fe Institute, 1995.

\bibitem{Wu93}
Z.~Wu and G.~Tseng.
\newblock {C}hinese text segmentation for text retrieval: achievements and
  problems.
\newblock {\em Journal of the American Society for Information Science},
  44(9):532--542, 1993.

\bibitem{Zipf49}
G.~Zipf.
\newblock {\em Human Behavior and the Principle of Least Effort}.
\newblock Addison-Wesley, Reading, MA, 1949.

\bibitem{Ziv77}
J.~Ziv and A.~Lempel.
\newblock A universal algorithm for sequential data compression.
\newblock {\em IEEE Transactions on Information Theory}, 23(3):337--343, 1977.

\bibitem{Ziv78}
J.~Ziv and A.~Lempel.
\newblock Compression of individual sequences by variable rate coding.
\newblock {\em IEEE Transactions on Information Theory}, 24(5):530--536, 1978.

\end{thebibliography}
\bibliographystyle{abbrv}
\end{document}